\documentclass{article}

    \usepackage[final]{timeseries_workshop}

\usepackage[utf8]{inputenc} % allow utf-8 input
\usepackage[T1]{fontenc}    % use 8-bit T1 fonts
\usepackage{hyperref}       % hyperlinks
\usepackage{url}            % simple URL typesetting
\usepackage{booktabs}       % professional-quality tables
\usepackage{amsfonts}       % blackboard math symbols
\usepackage{nicefrac}       % compact symbols for 1/2, etc.
\usepackage{microtype}      % microtypography
\usepackage{xcolor}         % colors
\usepackage{graphicx}
\usepackage{subcaption}
\usepackage{amsmath}
\usepackage{wrapfig}
\DeclareGraphicsExtensions{.pdf,.png,.jpg}

\newcommand{\threeplots}[3]{%
  \begin{minipage}[b]{0.32\linewidth}\centering
    \includegraphics[width=\linewidth]{#1}
  \end{minipage}\hfill
  \begin{minipage}[b]{0.32\linewidth}\centering
    \includegraphics[width=\linewidth]{#2}
  \end{minipage}\hfill
  \begin{minipage}[b]{0.32\linewidth}\centering
    \includegraphics[width=\linewidth]{#3}
  \end{minipage}%
}

\title{On the Internal Semantics of Time-Series Foundation Models}

\author{%
  Atharva Pandey\thanks{Equal contribution.} \\
  Kairosity \\
  \texttt{atpan@kairosity.ai} \\
  \And
  Abhilash Neog\footnotemark[1] \\
  Virginia Tech \\
  \texttt{abhilash22@vt.edu} \\
  \And
  Gautam Jajoo \\
  Kairosity \\
  \texttt{jajoo@kairosity.ai} \\
}

\begin{document}

\maketitle

\begin{abstract}
Time-series Foundation Models (TSFMs) have recently emerged as a universal paradigm for learning across diverse temporal domains. However, despite their empirical success, the internal mechanisms by which these models represent fundamental time-series concepts remain poorly understood. In this work, we undertake a systematic investigation of concept interpretability in TSFMs. Specifically, we examine: (i) which layers encode which concepts, (ii) whether concept parameters are linearly recoverable, (iii) how representations evolve in terms of concept disentanglement and abstraction across model depth, and (iv) how models process compositions of concepts. We systematically probe these questions using layer-wise analyses, linear recoverability tests, and representation similarity measures, providing a structured account of TSFM semantics. The resulting insights show that early layers mainly capture local, time-domain patterns (e.g., AR(1), level shifts, trends), while deeper layers encode dispersion and change-time signals, with spectral and warping factors remaining the hardest to recover linearly. In compositional settings, however, probe performance degrades, revealing interference between concepts. This highlights that while atomic concepts are reliably localized, composition remains a challenge, underscoring a key limitation in current TSFMs' ability to represent interacting temporal phenomena.
\end{abstract}

\section{Introduction}

Foundation models have recently been extended to time series, where large-scale pretraining over heterogeneous temporal data yields strong zero/few-shot performance in forecasting and classification across healthcare, finance, climate, and energy \citep{timesfm,chronos,moment,moirai,timegpt}. Yet, unlike language and vision, our understanding of what these models encode internally remains limited. Interpretability in NLP and CV has shown that probing methods like linear and structural probes as well as representational similarity can localize information across layers and provide insight into model organization \citep{alain2016linear,hewitt2019structural,kornblith2019cka}. For TSFMs, early studies such as \citep{wilinski2024tsfm_interventions} reveal block-like layer similarity and the success of latent interventions, underscoring the value of probing. Complementary instance-level explanations in time series, e.g., saliency, attribution, and shapelets, offer rationales for individual predictions but do not illuminate model-wide semantics \citep{ismail2020benchmark,grabocka2014shapelets}.

This gap motivates a systematic investigation into how TSFMs internally represent fundamental time-series phenomena. We study \emph{concept interpretability} in TSFMs across seven canonical concepts that span stochastic, structural, and spectral behavior: \textit{AR1}, \textit{Level Shift}, \textit{Random Walk}, \textit{Spectral}, \textit{Time Warped}, \textit{Trend}, and \textit{Variance Shift}. Our analysis is guided by four central questions: \textbf{RQ1} - Where do concepts localize across layers? \textbf{RQ2} - Are concept parameters linearly recoverable from intermediate embeddings? \textbf{RQ3} - How do representations evolve in terms of disentanglement and abstraction with depth? and \textbf{RQ4} - How do models represent compositions of concepts?

To address these questions, we employ a probing-based interpretability framework. Methodologically, we adapt established tools - linear probes, structural probes, and CKA - tailored to quantify concept presence and parameter recoverability \citep{alain2016linear,hewitt2019structural,kornblith2019cka}. While these diagnostics are widely used in other domains, their systematic application to a controlled, diverse suite of time-series concepts offers new insights into the inductive biases and limitations of modern TSFMs \citep{timesfm,chronos,moment,moirai,timegpt,wilinski2024tsfm_interventions}.

\textbf{Contributions.}
(i) A \textit{concept-centric probing framework} for TSFMs covering seven canonical time-series concepts;
(ii) \textit{diagnostic tasks} for assessing concept localization, parameter recoverability, and compositional interaction;
(iii) \textit{empirical insights} revealing inductive biases and failure modes, which may potentially inform architectural choices, training curricula, and evaluation protocols.
\section{Methods}

\textbf{Layer-wise Concept Probing.} \textit{RQ1} and \textit{RQ2} examine which concepts are encoded across layers and whether their parameters are linearly recoverable. We investigate this by analyzing the latent representations of each layer using linear probes - a methodology widely used in language models to reveal the emergence of syntax and semantics at specific depths. For time series, this allows us to pinpoint where autoregressive structure, spectral frequency, or trend parameters become accessible. Given a synthetic dataset $\mathbf{X} \in \mathbb{R}^{S \times V}$ with generative parameters $\theta$ (e.g., AR coefficient, trend slope, frequency amplitude), a TSFM with $L$ layers produces hidden states $\mathbf{H}^{(l)} = f^{(l)}(\mathbf{X}) \in \mathbb{R}^{S \times d}$, which are pooled into $\mathbf{z}^{(l)} = \text{Pool}(\mathbf{H}^{(l)}) \in \mathbb{R}^{d}$. A linear probe then predicts parameters as $\hat{\theta}^{(l)} = \mathbf{W}^{(l)} \mathbf{z}^{(l)} + \mathbf{b}^{(l)}$. Performance is measured by mean squared error,
$\mathcal{L}^{(l)} = \tfrac{1}{N} \sum_{i=1}^{N} \|\theta_i - \hat{\theta}^{(l)}_i\|^2,
$
quantifying parameter recoverability across depth.

\textbf{Concept Representation.} \textit{RQ3} examines how representations evolve across depth - whether they become more abstract or more disentangled. In computer vision, representational similarity analyses reveal progressive shifts from low-level edges to object-level semantics. We adopt a similar lens for TSFMs, asking whether distinct time-series concepts occupy separable or overlapping regions in embedding space, and how this organization changes across layers. To quantify representational similarity across concepts and layers, we compute centered kernel alignment (CKA) between embedding sets $\mathbf{H}^{(l_1)}$ and $\mathbf{H}^{(l_2)}$:
$
\text{CKA}(\mathbf{H}^{(l_1)}, \mathbf{H}^{(l_2)}) = 
\frac{\| \mathbf{H}^{(l_1)\top} \mathbf{H}^{(l_2)} \|_F^2}
{\| \mathbf{H}^{(l_1)\top} \mathbf{H}^{(l_1)} \|_F \, \| \mathbf{H}^{(l_2)\top} \mathbf{H}^{(l_2)} \|_F}.
$
Additionally, we visualize embeddings via PCA, UMAP, and t-SNE \citep{jolliffe2002pca, mcinnes2018umap,maaten2008tsne} applied to pooled vectors $\mathbf{z}^{(l)}$, allowing inspection of cluster structure and concept separation.

\textbf{Concept Composition.} \textit{RQ4} examines how TSFMs handle compositions and whether concept-specific information transfers to their mixtures. We adopt a two-step \emph{probe-transfer} protocol: (i) train layer-wise linear probes on \emph{atomic} data for each concept $C_j$ to predict its parameters $\theta_j$ (backbone frozen); (ii) evaluate these frozen probes for $C_1$ and $C_2$ on \emph{composite} series to assess whether the original parameters remain linearly recoverable. We report per-layer MSE on composite data.

We study two families of compositions: \emph{structured} (segment-wise interleaving with continuity preservation) and \emph{functional} (additive mixing, optionally with per-series normalization and mixing coefficients $\alpha$). Full construction details, masks, and sampling ranges are provided in Appendix~\ref{app:dataset}.

\section{Results and Discussions}

% Helper: only defined if not already present elsewhere
\providecommand{\threeplots}[3]{%
  \begin{minipage}[b]{0.32\linewidth}\centering
    \includegraphics[width=\linewidth]{#1}
  \end{minipage}\hfill
  \begin{minipage}[b]{0.32\linewidth}\centering
    \includegraphics[width=\linewidth]{#2}
  \end{minipage}\hfill
  \begin{minipage}[b]{0.32\linewidth}\centering
    \includegraphics[width=\linewidth]{#3}
  \end{minipage}%
}

% \begin{wrapfigure}{r}{0.485\linewidth} % {placement}[r=right] {width}
%   \centering
%   \vspace{-1em} % Optional: Adjusts vertical spacing at the top
  
%   % --- First plot ---
%   \includegraphics[width=0.9\linewidth]{figures/LinearProbeMSE/chronos_layerwise_eval_mse.png}
%   \vspace{0.25em}
%   \small Chronos: probe MSE vs. layer
  
%   \vspace{1.5em} % Adds space between the two plots
  
%   % --- Second plot ---
%   \includegraphics[width=0.9\linewidth]{figures/LinearProbeMSE/moment_layerwise_eval_mse.png}
%   \vspace{0.25em}
%   \small MOMENT: probe MSE vs. layer
  
%   % --- Optional: Add a real caption ---
%   \caption{Layer-wise probe MSE for Chronos and MOMENT.}
%   \label{fig:wrap-probe-mse}
% \end{wrapfigure}

\begin{figure}[t]
  \centering

  % ---------------- Top row: probe MSE vs layers (all concepts per model) ----------------

  \vspace{0.8em}

  % ---------------- Bottom row: UMAP snapshots at early/mid/late layers ----------------
  \threeplots{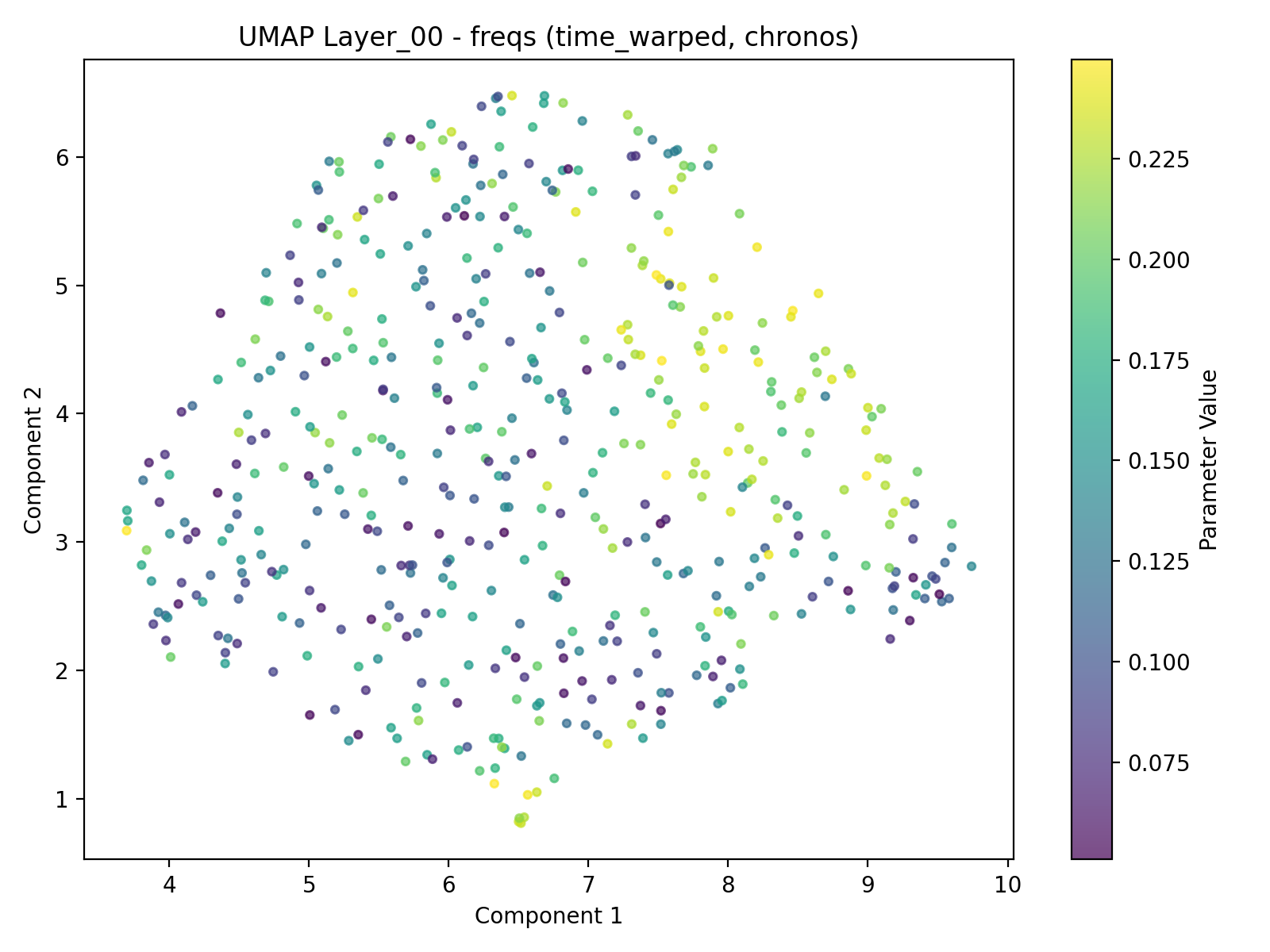}%
             {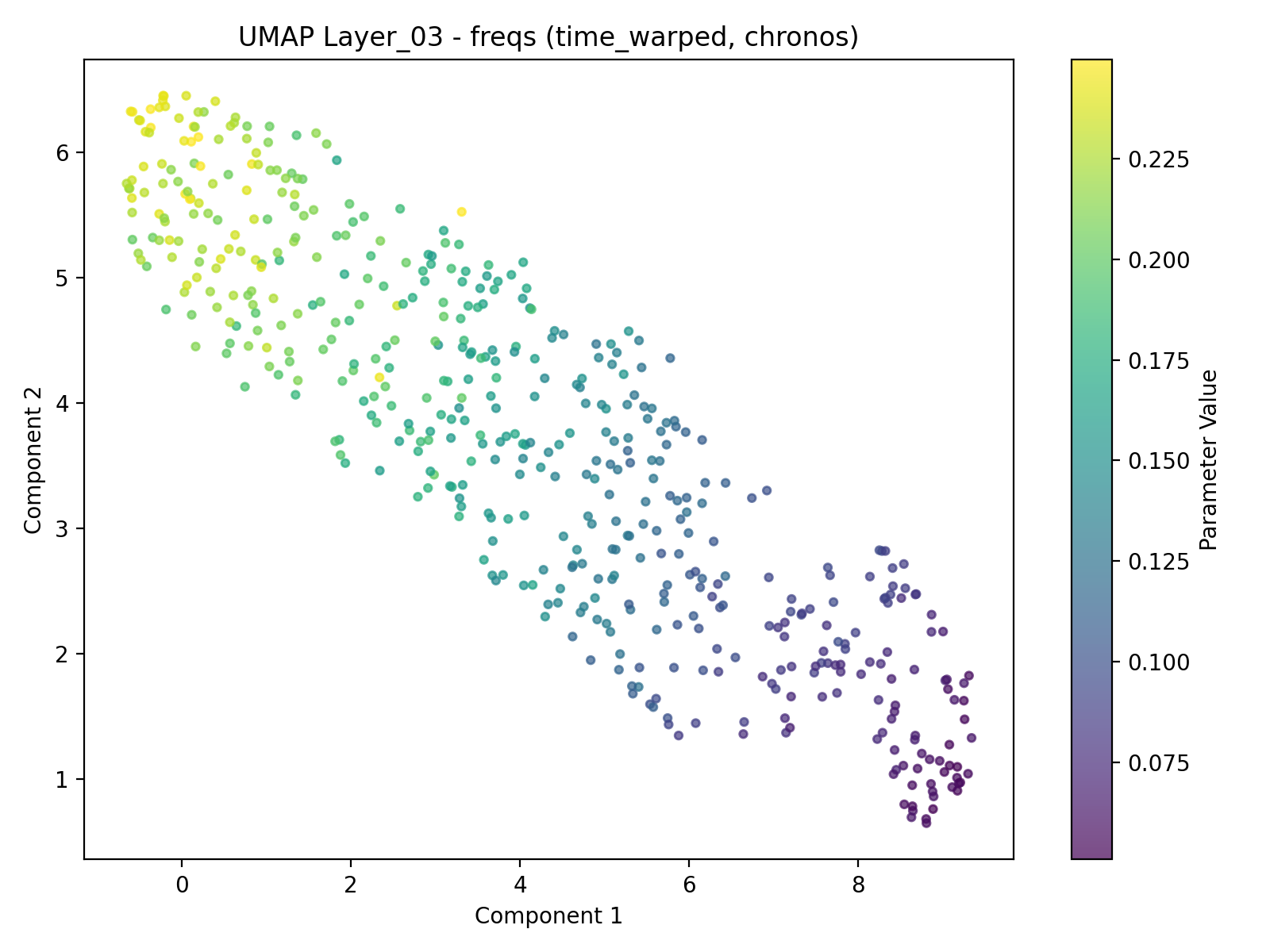}%
             {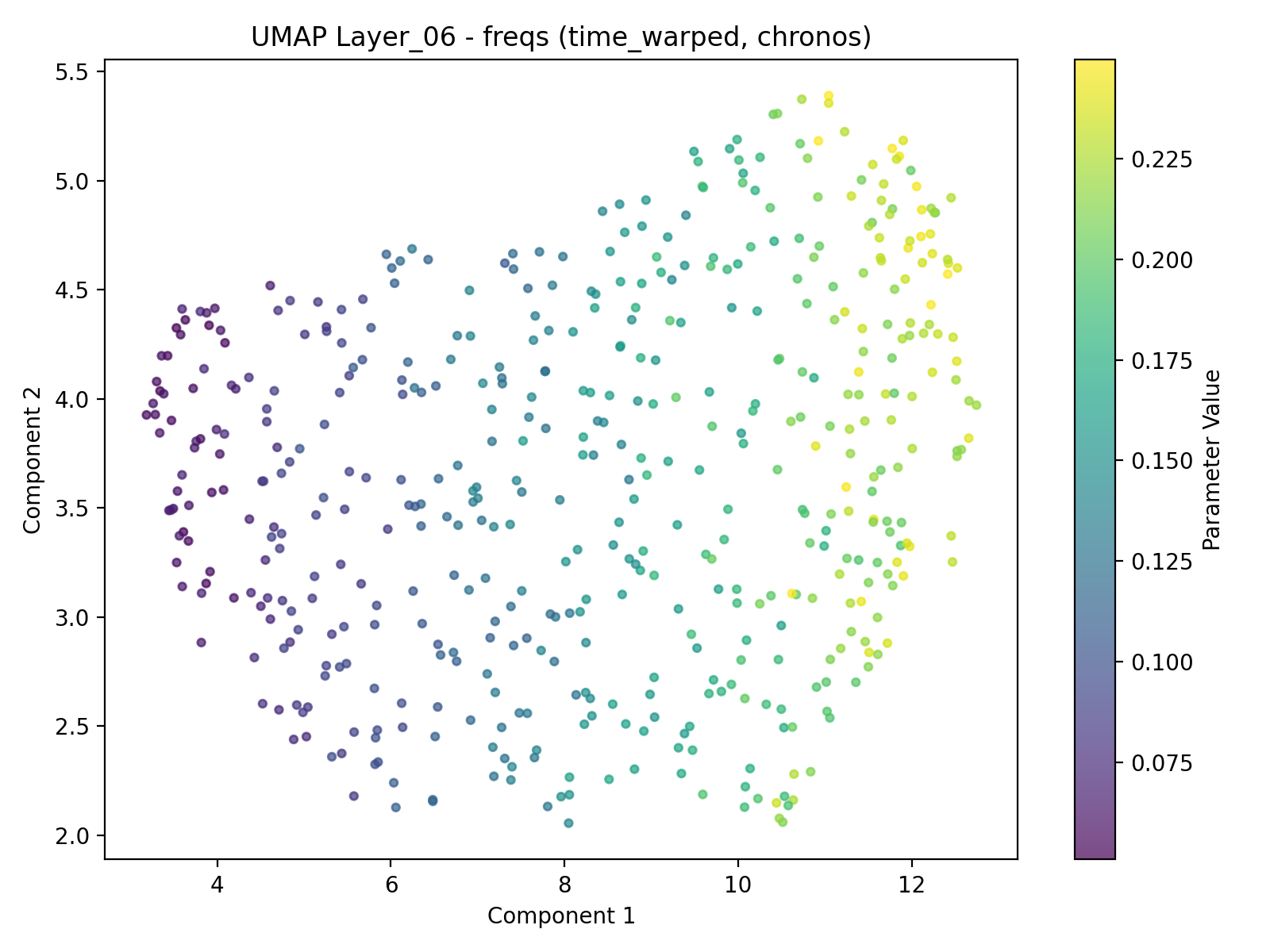}

  \caption{UMAP of pooled embeddings at early, mid, and late layers, time-warp concept.}
  \label{fig:layerwise_summary}
  \vspace{-15pt}
\end{figure}

\textbf{\textit{RQ1 \& RQ2}: \textit{Which layers encode which concepts, and are parameters linearly recoverable?}}
\vspace{-2pt}

\textbf{UMAP-probe alignment.} Examining UMAP embeddings of layer-wise latent representations reveals how the model organizes conceptual information internally. When representations are compact and well-ordered, linear probes can recover concept parameters with low error, suggesting the model forms localized embeddings for those concepts. Further, when a parameter varies smoothly along the UMAP manifold, probe accuracy improves even more - indicating that the model has not only captured the concept but also encoded a meaningful, controllable parameterization. Such alignment could be particularly useful for applications that steer activations conditionally. Through our experiments, we observe that structural and time-domain concepts such as AR(1) coefficient, trend slope, and level shifts, tend to be well-captured with lower probe errors. In contrast, spectral and time-warping concepts exhibit fragmented or tangled UMAP structures and higher probe errors, reflecting non-linear entanglement that resists simple linear decoding (see Fig.\ref{fig:layerwise_summary} and Appendix\ref{app:layerwise-dimred}).

% \begin{wrapfigure}{r}[h!] % [h!] means "place it here"
%   \centering
%   \resizebox{0.8\linewidth}{!}{%
%   \begin{minipage}[b]{0.49\linewidth}
%     \centering
%     \includegraphics[width=\linewidth]{figures/LinearProbeMSE/chronos_layerwise_eval_mse.png}
%     % \vspace{0.25em}
%     % \small Chronos: 
%   \end{minipage}
%   \hfill % Pushes the two minipages apart
%   \begin{minipage}[b]{0.49\linewidth}
%     \centering
%     \includegraphics[width=\linewidth]{figures/LinearProbeMSE/moment_layerwise_eval_mse.png}
%     % \small MOMENT: probe MSE vs. layer
%   \end{minipage}
%   }
%   \vspace{-6pt}
%   \caption{Layer-wise probe (y-axis MSE; x-axis layers) for Chronos (left) and MOMENT (right)}
%   \label{fig:side-by-side-mse}
%   \vspace{-15pt}
% \end{wrapfigure}
\begin{wrapfigure}{r}{0.689\linewidth}
\vspace{-12pt}
  % \centering
  \resizebox{\linewidth}{!}{% % Use \linewidth (100% of the wrapfigure)
    \begin{minipage}[b]{0.43\linewidth}
      \centering
      \includegraphics[width=\linewidth]{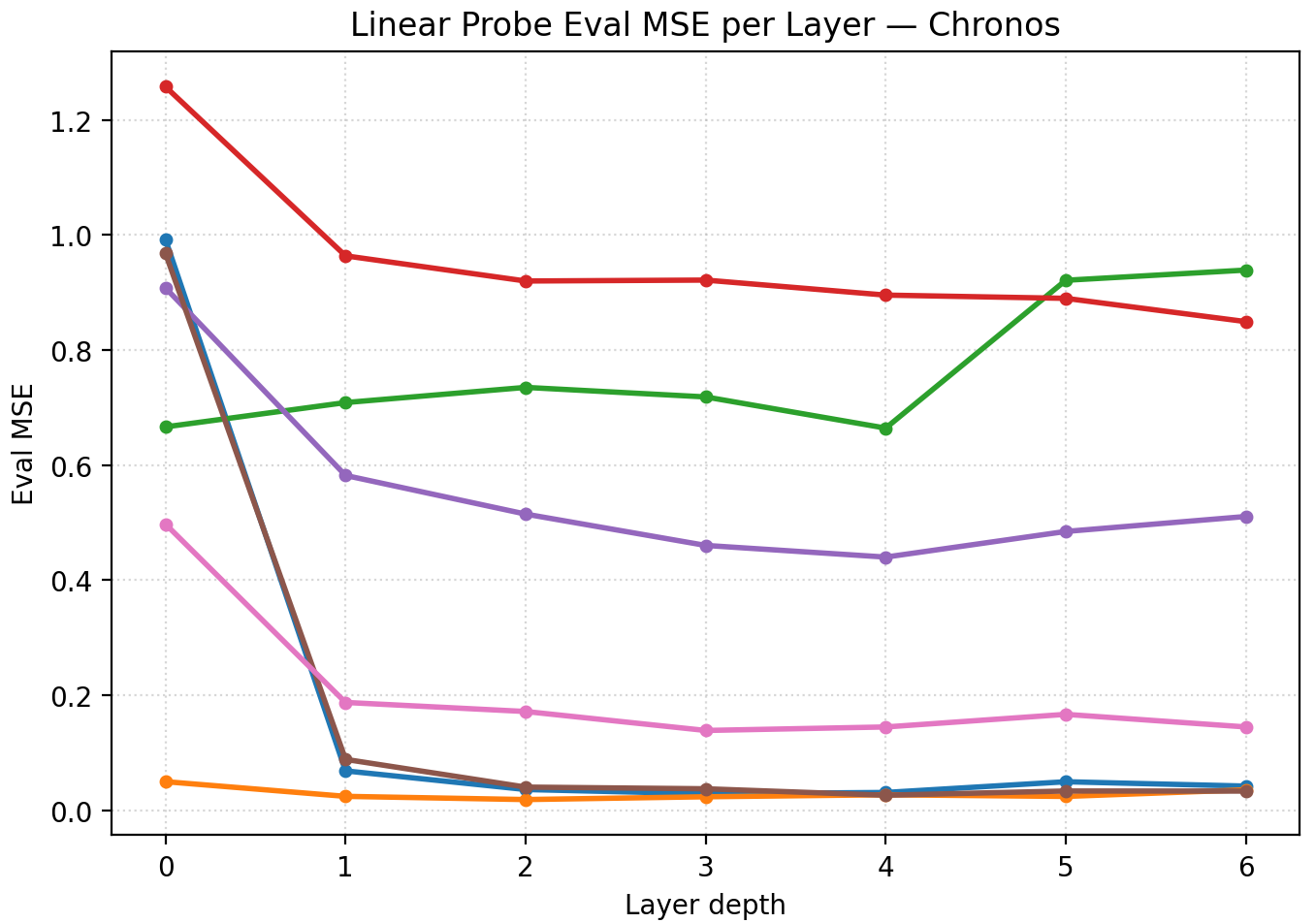}
    \end{minipage}
    \hfill % Pushes the two minipages apart
    \begin{minipage}[b]{0.557\linewidth}
      % \centering
      \includegraphics[width=\linewidth]{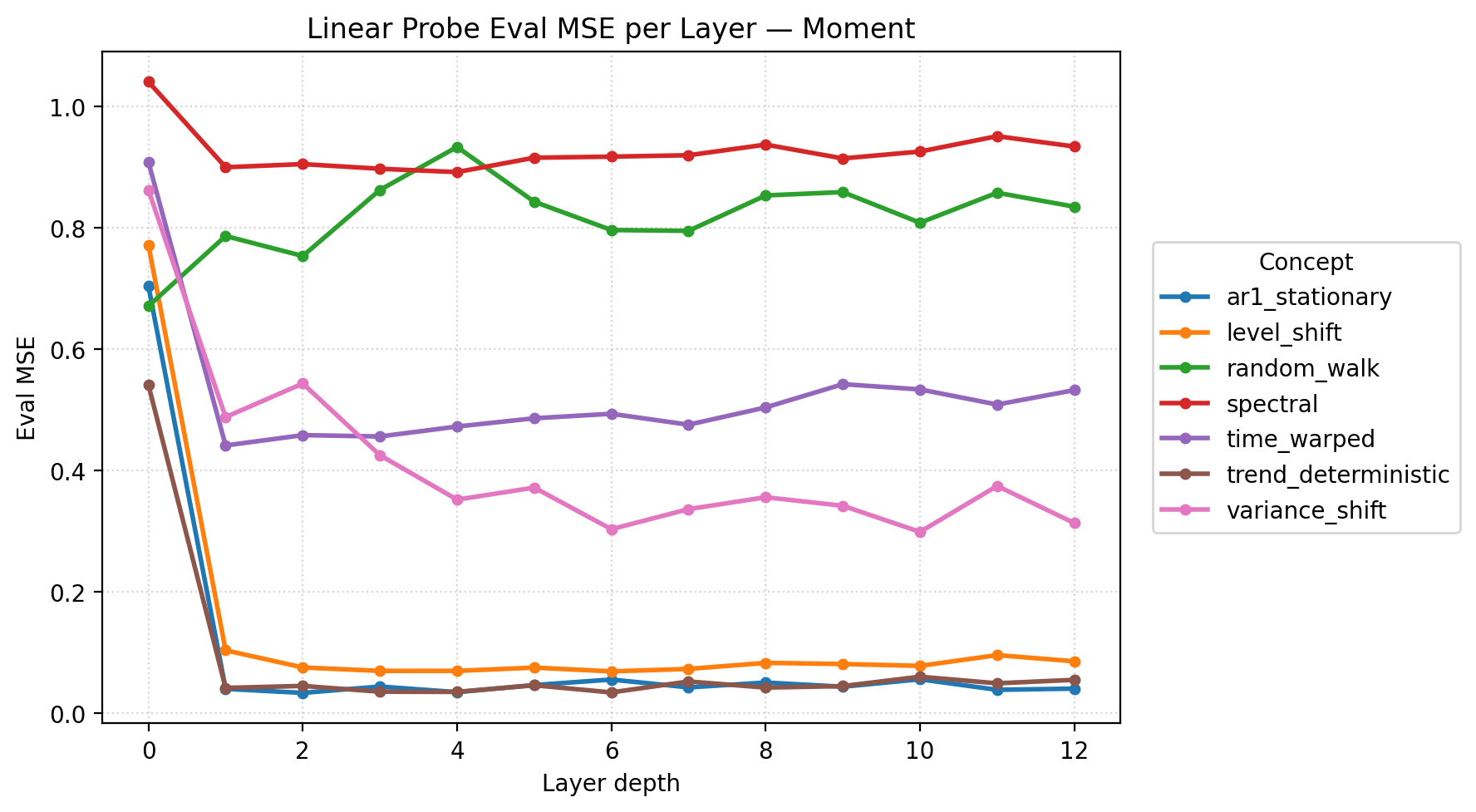}
    \end{minipage}
  }
  \vspace{-12pt} % Small space adjustment
  
  % --- Replaced \caption with plain text ---
  \footnotesize 
  % \centering 
  \caption{Layer-wise probe (y-axis MSE; x-axis layers) for Chronos (left) and MOMENT (right). Each curve represents a concept}
  \vspace{-12pt}
  % \label{} cannot be used here
\end{wrapfigure}

\textbf{Model comparison and depth.} Across identical experimental setups, \textsc{Chronos} consistently produces better-organized UMAP representations that are more linearly recoverable than those of \textsc{Moment} for all evaluated concepts. Most concepts are captured early-typically by the second transformer layer-after which performance plateaus. In contrast, representations of dispersion and change-point phenomena (e.g., variance shifts) continue to improve with depth, becoming more localized in the later layers (Fig.~\ref{fig:layerwise_summary}). 

\textit{Simple structural and time-domain concepts emerge early in well-organized, linearly recoverable representations, while complex or change-sensitive patterns gradually refine in deeper layers, reflecting a layered hierarchy of concept learning.}

\begin{figure*}[htbp]
    \centering
    \vspace{-4pt}
     \resizebox{\linewidth}{!}{
    \begin{subfigure}{0.48\textwidth}
        \includegraphics[width=\linewidth]{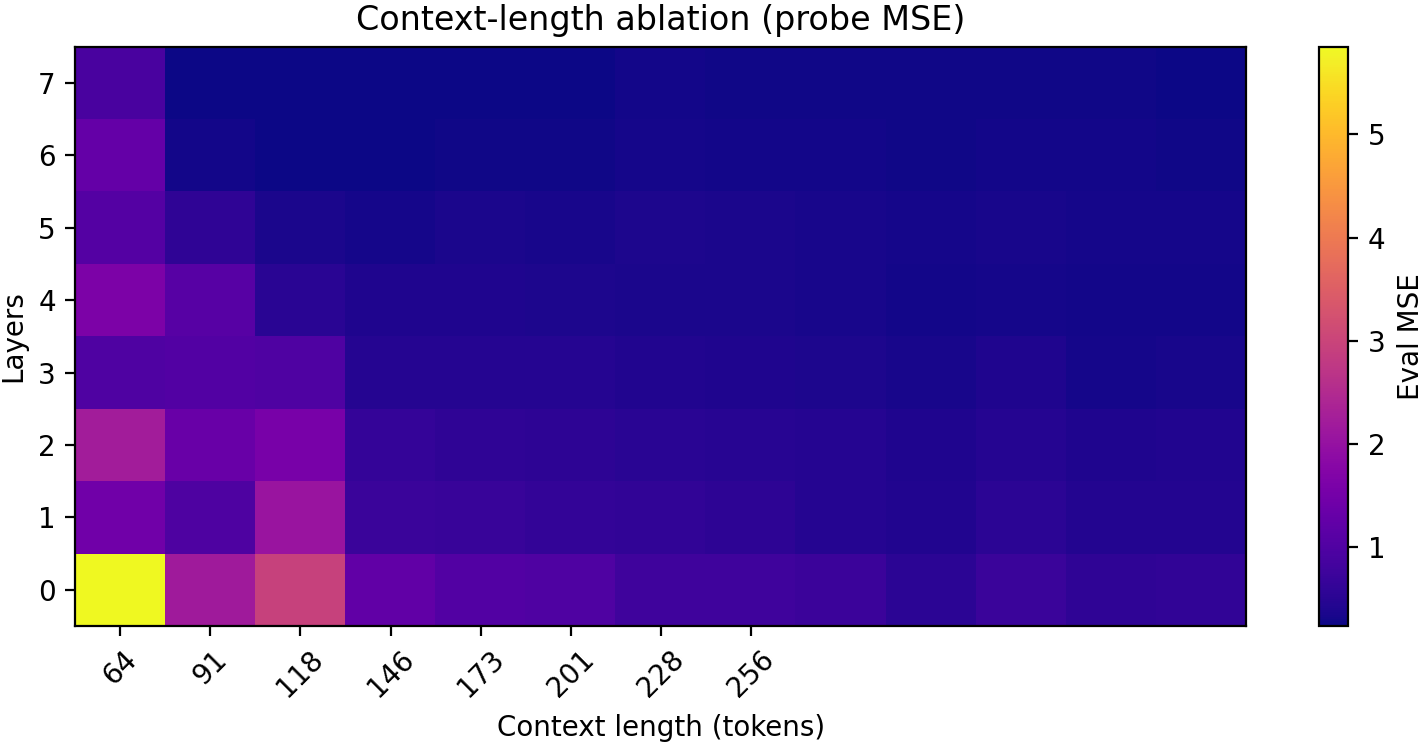}
        \caption{Time Warped Concept}
        \label{fig:moment_context_length_ablation_heatmap_time_warped}
    \end{subfigure}
    \begin{subfigure}{0.48\textwidth}
        \includegraphics[width=\linewidth]{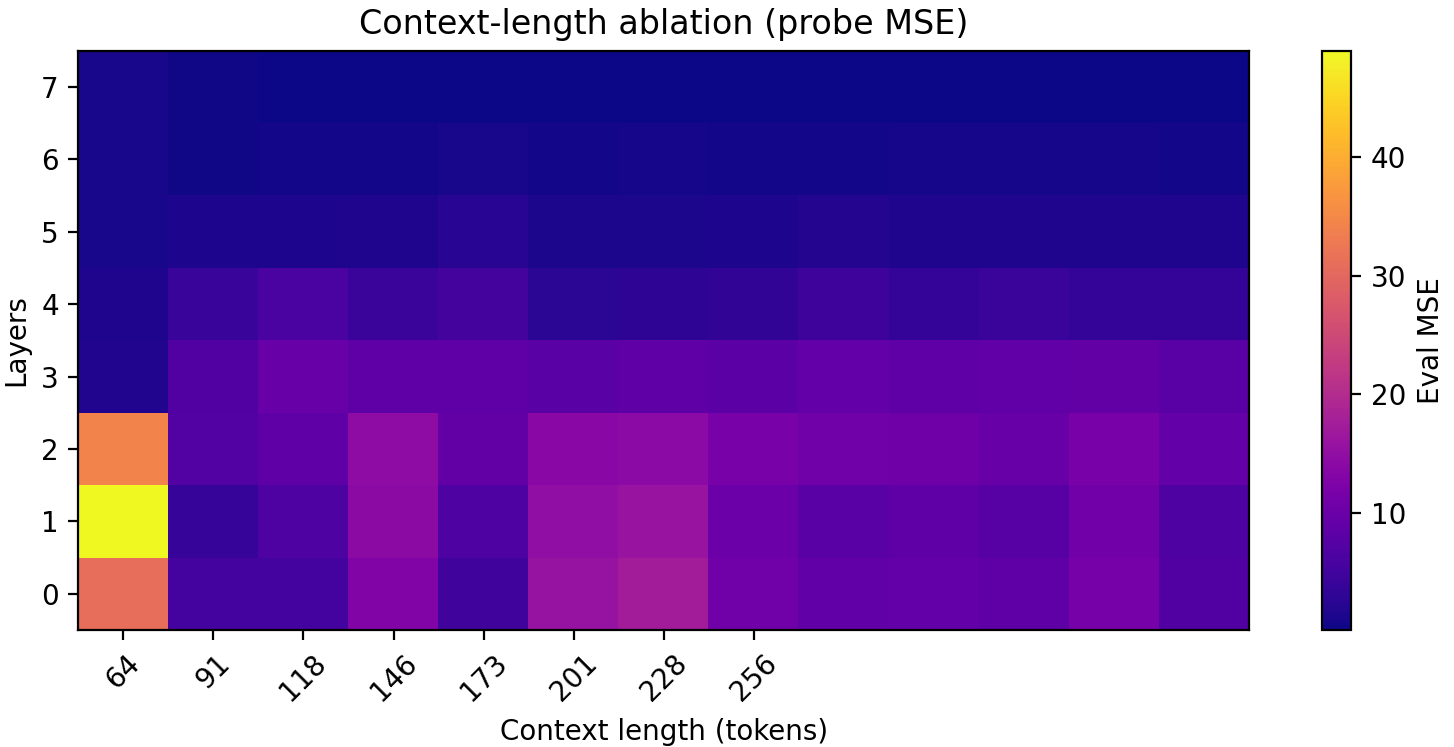}
        \caption{Variance Shift Concept}
        \label{fig:moment_variance_shift_context_length_ablation_heatmap}
    \end{subfigure}}
    \vspace{-15pt}
    \caption{Context Length ablations on MOMENT}
    \vspace{-7pt}
\end{figure*}

\textit{\textbf{RQ3: How do representations evolve with depth?}}
\vspace{-2pt}

UMAP snapshots (see Figures in the Appendix G, for e.g. Figure \ref{app:fig15} and Figure \ref{app:fig16}) reveal increasing cluster separation from early to late layers. \textit{Early layers reflect locally volatile structure; mid layers show partial disentanglement; late layers consolidate concept-level separation while compressing intra-concept variance.} This pattern aligns with the drop in probe MSE after layer 1, indicating a shift from generic to concept-aligned features. 

We further probe each layer's reliance on temporal context by cropping inputs to multiple fractions (25-100\%), extracting pooled embeddings, and evaluating the pretrained per-layer linear probes on the target concept. From Figure \ref{fig:moment_context_length_ablation_heatmap_time_warped} and Figure \ref{fig:moment_variance_shift_context_length_ablation_heatmap} we can see how MSE changes with available history; \textit{deeper layers improve as context grows (encoding longer-range dynamics) compared to relatively less improvement in early layers (capturing short, local structure).}

\textit{\textbf{RQ4: How are concept compositions represented?}}
\vspace{-2pt}

TSFMs can effectively encode atomic time-series concepts, but real-world data often involves compositions of multiple concepts. To study the behavior of TSFMs under composite concepts, we conduct two complementary experiments: \textbf{(a)} \textbf{Vector Arithmetic} - Inspired by word embedding compositionality, we test whether TSFM embeddings exhibit similar additive properties. Specifically, we evaluate whether the element-wise sum of atomic concept embeddings ($\mathbf{emb_{1}} + \mathbf{emb_{2}}$) approximates the embedding of their composite concept ($\mathbf{emb_{3}}$) using cosine similarity and relative distance metrics across model layers. \textbf{(b)} \textbf{Temporal Alignment Analysis} - Since time-series have inherent temporal structure, we test compositional stability across different sequence lengths (32, 64, 128, 256 timesteps). This assesses whether compositional relationships hold consistently across temporal horizons or are sensitive to sequence length. 

\begin{wrapfigure}{r}{0.6\textwidth}
    \vspace{-20pt}
    \centering
    \includegraphics[width=0.6\textwidth]{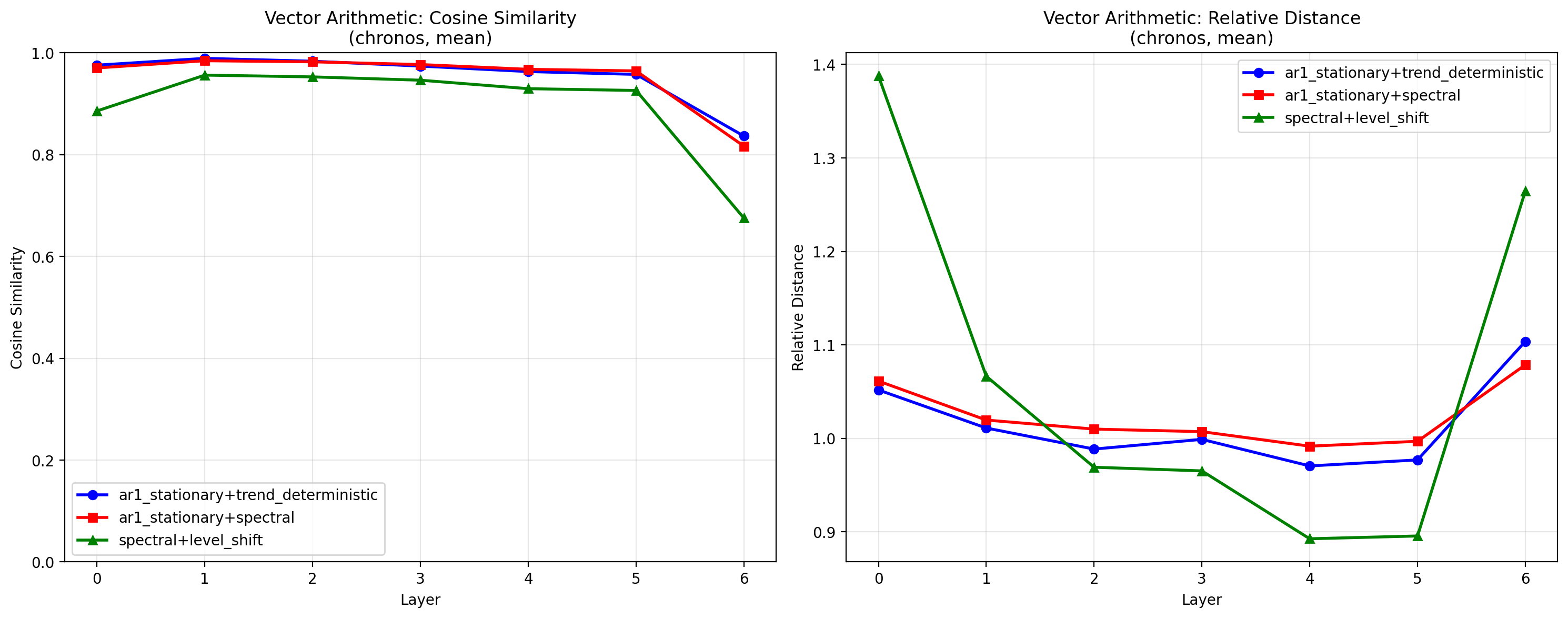}
    \caption{\small Vector arithmetic experiments with \textsc{Chronos}. 
    Atomic embeddings combine nearly linearly 
    ($\mathbf{emb_{1}} + \mathbf{emb_{2}} \approx \mathbf{emb_{3}}$), 
    except for temporally disparate concept pairs.}
    \label{fig:vec_arithmetic}
    \vspace{-12pt}
\end{wrapfigure}

Figure \ref{fig:vec_arithmetic} \textbf{reveals} strong compositional properties in TSFMs, with cosine similarities approaching 1.0 across most layers, indicating that atomic concept embeddings combine nearly linearly ($\mathbf{emb_{1}} + \mathbf{emb_{2}} \approx \mathbf{emb_{3}}$). Performance degradation in initial and final layers suggests that early representations lack full compositional structure, while deeper layers specialize in task-specific features that deviate from additive composition. The anomalous behavior of \textit{spectral+level shift}, which shows substantially higher relative distances, indicates non-linear interactions between concepts with fundamentally different temporal characteristics-frequency-domain properties versus abrupt discontinuities. \textit{Overall, TSFMs learn compositional representations similar to word embeddings for most concept pairs, with notable exceptions requiring more sophisticated composition mechanisms for temporally disparate concepts.}

\begin{figure}[htbp]
    \centering
    \threeplots{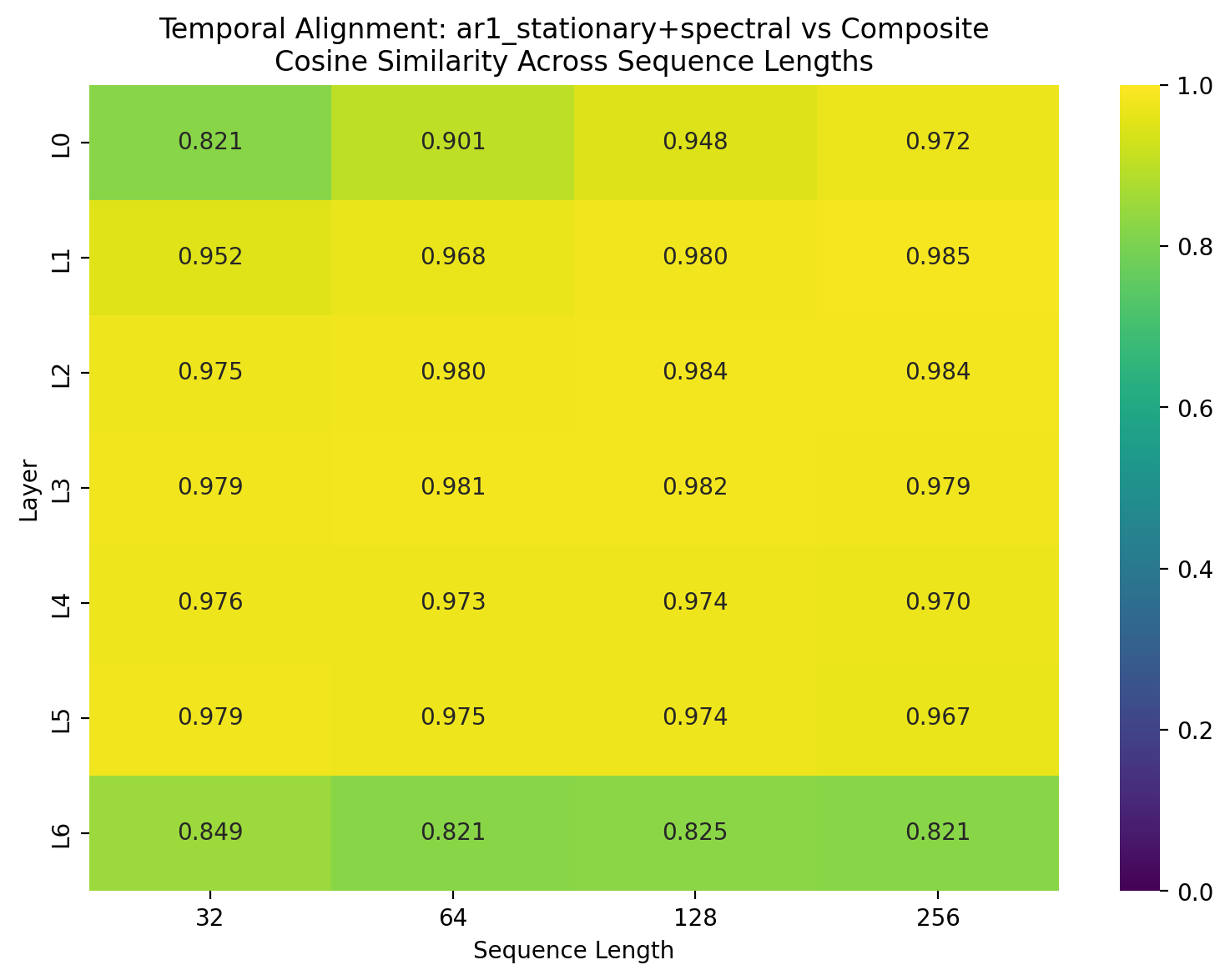}%
               {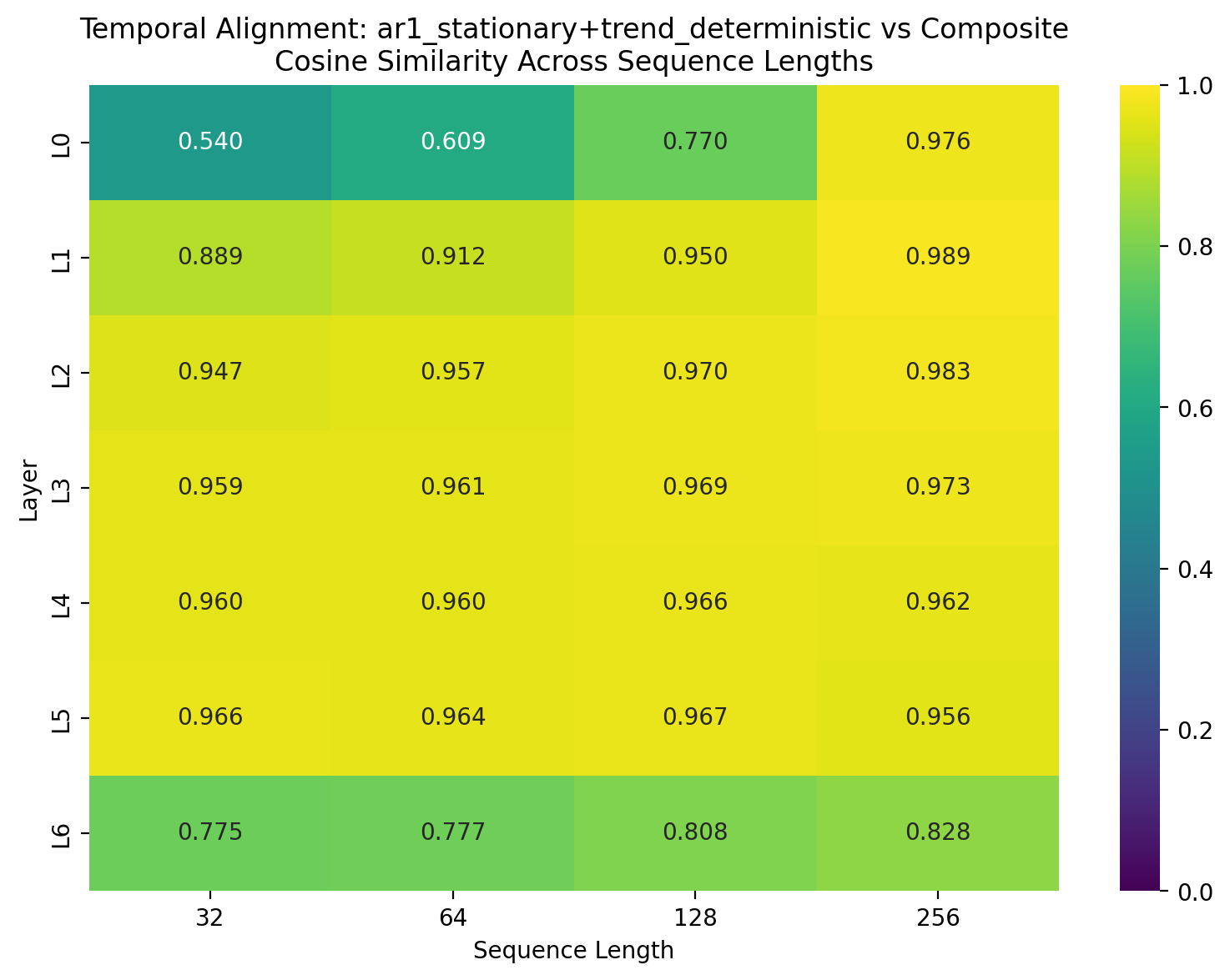}%
               {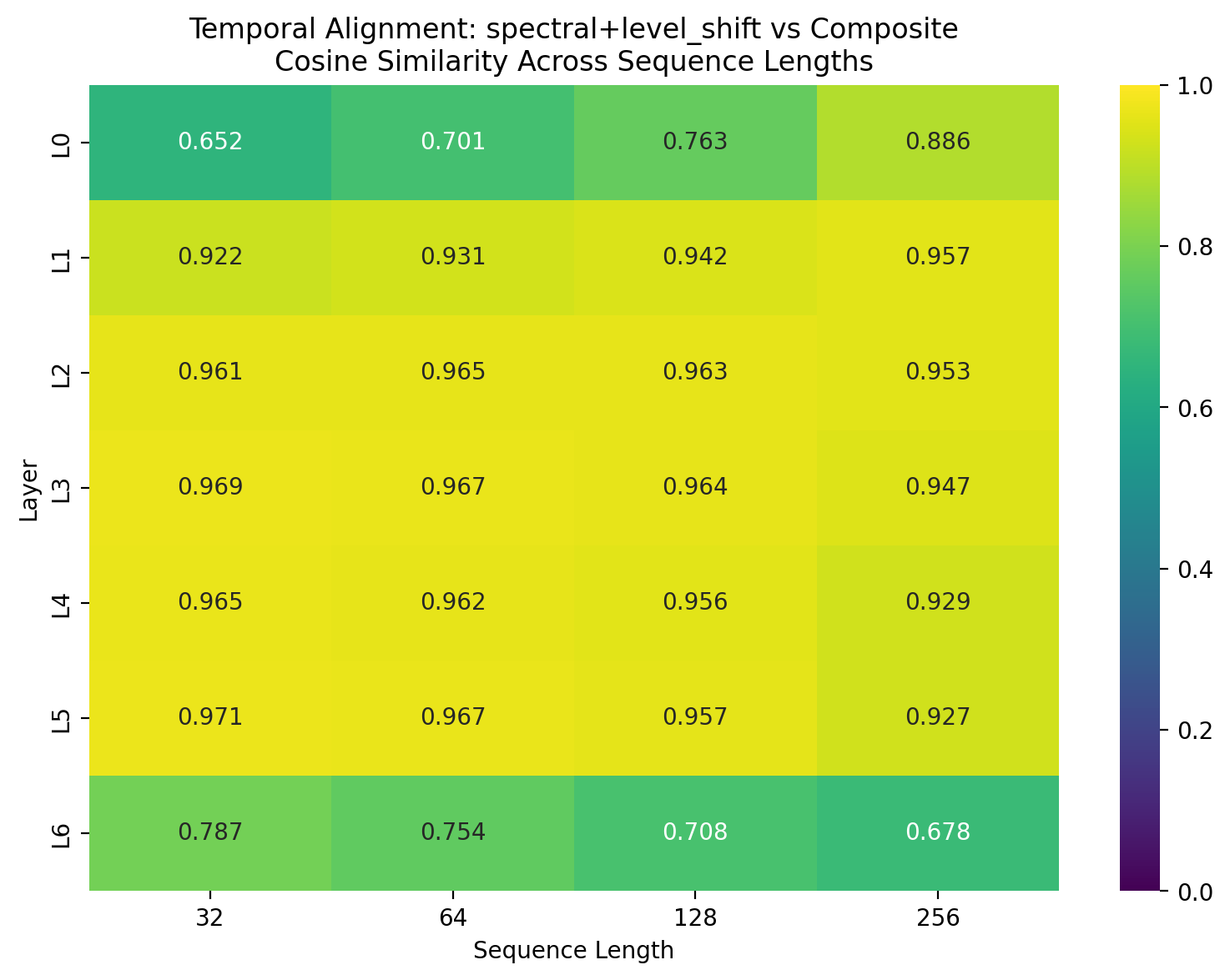}
    \caption{Chronos – Temporal alignment experiments. We show stability of compositional relationships across multiple atomic-concept pairs.}
    \label{fig:chronos_temporal_alignment}
\end{figure}

The temporal alignment analysis results (see Figure \ref{fig:chronos_temporal_alignment}) \textbf{demonstrate} robust compositional stability across sequence lengths, with consistently high similarities throughout most layers and temporal horizons. Reduced similarity at shorter sequences (32–128) in the initial and final layers suggests that compositional understanding requires sufficient temporal context to emerge and stabilize. \textit{The uniformly high performance at longer sequences confirms that TSFMs' compositional properties are temporally robust once adequate context is provided.} Please refer to Appendix \ref{app:compositeresults} for add. results.

\section{Conclusion and Future Work}

We present a probe-based analysis of time-series foundation models across seven canonical concepts. Early layers expose local, time domain structure (AR(1), level shift, trend), deeper layers specialize in dispersion and change-point signals. Spectral and time-warping factors are the least linearly accessible. TSFMs also exhibit strong linear compositional properties across most layers and concept pairs. Future works could explore additional TSFMs, multivariate and irregular datasets; adopt controlled-capacity non-linear or causal probes; architectures and objectives that better linearize phase and time-warping, and non-linear conditional steering.

\bibliographystyle{plainnat} % or abbrvnat if you prefer compact author names
\bibliography{references}
\newpage
\appendix
% \section{Appendix}
\section{Related Works}\label{app:related-works}

\paragraph{Time-series foundation models.}
Recent TSFMs demonstrate strong zero/few-shot performance via large-scale pretraining and task-agnostic architectures. Representative families include TimesFM (decoder-only with patched attention), Chronos (tokenized values with T5-style training), MOMENT (open models and the Time Series Pile), Moirai (masked-encoder universal forecaster), and TimeGPT (closed-source API). These works establish the empirical promise of TSFMs but do not characterize concept-level internal semantics. \citep{timesfm,chronos,moment,moirai,timegpt}

\paragraph{Probing and representational similarity.}
Linear probes and related diagnostic tools are widely used to localize information across layers in deep networks, originating with linear classifier probes and extended by structural probes in NLP to test linear recoverability of syntax. Centered Kernel Alignment (CKA) is commonly used to compare layer representations within and across models due to its invariances and robustness relative to earlier CCA-style measures. Our study adapts these established tools to TSFMs and focuses them on time-series concepts and parameters. \citep{alain2016linear,hewitt2019structural,kornblith2019cka}

\paragraph{Interpreting TSFMs and time-series models.}
Closest to our work, Wili{\'n}ski et al.\ analyze internal redundancy and concept steering in TSFMs, reporting block-like layer similarity and latent-space interventions; we complement this by centering \emph{concept parameters}, layer-wise recoverability, and controlled compositions. Broader interpretability for time series has emphasized saliency/attribution and shapelet-based explanations; these provide instance-level rationales, whereas our focus is on \emph{representation-level} concept encoding across depth. \citep{wilinski2024tsfm_interventions,ismail2020benchmark,grabocka2014shapelets}

% \section{Experimental Setup}
% \subsection{Datasets}
% \major{1) describe the synthetic dataset generation - 2) what are the type of concepts considered. Refer the reader to the appendix for more details, 3) describe the compositional datasets. Refer the reader to the appendix for more details}
% \subsection{Models and Implementation Details}
% \major{1) Describe the Chronos and MOMENT models - variants used. 2) Talk about linear regression models used for linear probing}

% \major{implementation details - 1) explain how the experiments were conducted - linear probing (regression models are trained on normalized features from each layer which are pooled). 2) Hardware details}

\section{Experimental Setup}

\textbf{Datasets.}
We evaluate seven synthetic concepts: AR(1), Level Shift, Random Walk, Spectral (sum of sinusoids), Time-Warped Sinusoid, Deterministic Trend, and Variance Shift. Generation procedures, parameter ranges, and normalization rules follow Appendix D (Dataset Generation and Description). We additionally construct compositional datasets by pairing two base concepts.

\textbf{Models.}
We use publicly released checkpoints of two time-series foundation models: Chronos and MOMENT since both are T-5 like models transformer architecture. Model weights are frozen and no finetuning is performed. 

\textbf{Evaluation and reporting.}
We use an 80/20 train/validation split for each concept and composition. Metric is mean squared error (MSE) for parameter recovery.

\label{app:experimentalsetup}

\section{Dimensionality Reduction Techniques}\label{app:dimred}
\paragraph{Principal Component Analysis (PCA).} Given pooled representations $\{\mathbf{z}^{(l)}_i\}_{i=1}^N$, we compute the empirical covariance matrix
\[
\mathbf{\Sigma}^{(l)} = \frac{1}{N} \sum_{i=1}^N \left(\mathbf{z}^{(l)}_i - \bar{\mathbf{z}}^{(l)}\right) \left(\mathbf{z}^{(l)}_i - \bar{\mathbf{z}}^{(l)}\right)^\top.
\]
Eigen-decomposition yields orthogonal axes capturing the largest variance directions:
\[
\mathbf{\Sigma}^{(l)} \mathbf{u}_k = \lambda_k \mathbf{u}_k, \quad \lambda_1 \ge \lambda_2 \ge \dots.
\]
These principal axes reveal which parameters dominate the representation space and whether layers compress or expand information.

\paragraph{t-SNE.} To assess local neighborhoods, we apply t-distributed Stochastic Neighbor Embedding (t-SNE), which constructs pairwise similarities in high- and low-dimensional spaces. For two points $\mathbf{z}_i, \mathbf{z}_j$, their similarity in the original space is
\[
p_{ij} = \frac{\exp(-\|\mathbf{z}_i - \mathbf{z}_j\|^2 / 2\sigma_i^2)}{\sum_{k \ne i} \exp(-\|\mathbf{z}_i - \mathbf{z}_k\|^2 / 2\sigma_i^2)},
\]
while in 2D space the similarity is
\[
q_{ij} = \frac{\left(1 + \|\mathbf{y}_i - \mathbf{y}_j\|^2\right)^{-1}}{\sum_{k \ne l} \left(1 + \|\mathbf{y}_k - \mathbf{y}_l\|^2\right)^{-1}}.
\]
t-SNE minimizes the Kullback–Leibler divergence:
\[
\text{KL}(P \| Q) = \sum_{i \ne j} p_{ij} \log \frac{p_{ij}}{q_{ij}}.
\]
This highlights fine-grained clusters and separability of parameter values.

\paragraph{UMAP.} Uniform Manifold Approximation and Projection seeks a balance between local and global structure. It constructs a weighted $k$-nearest-neighbor graph and optimizes a low-dimensional embedding $\{\mathbf{y}_i\}$ by minimizing the cross-entropy between high- and low-dimensional fuzzy simplicial sets:
\[
\mathcal{L}_{\text{UMAP}} = \sum_{(i,j)} w_{ij} \log \sigma(\|\mathbf{y}_i - \mathbf{y}_j\|) 
    + (1 - w_{ij}) \log (1 - \sigma(\|\mathbf{y}_i - \mathbf{y}_j\|)),
\]
where $\sigma$ is a differentiable approximation of a step function. UMAP can reveal concept families and hierarchical relationships (e.g., stationary vs. nonstationary).

These projections provide intuition about the embedding geometry—global variance (PCA), local clusters (t-SNE), and local-global trade-offs (UMAP)—which the linear probes then quantify.
\newpage
\section{Synthetic Datasets}\label{app:dataset}
% Section: Dataset Generation and Description

This section summarizes the synthetic time--series concepts used in our experiments, their generating equations, and key parameters. Unless noted, \(\varepsilon_t\) denotes i.i.d. Gaussian noise.

\subsection{AR(1) (Stationary)}
\begin{align}
  x_t &= \phi\, x_{t-1} + \varepsilon_t, \quad |\phi| < 1,\\
  \varepsilon_t &\sim \mathcal{N}(0, \sigma^2), \quad x_0 \text{ drawn from the stationary distribution.}
\end{align}
Parameters: autoregressive coefficient \(\phi\) (sampled from an interval), innovation std \(\sigma\). Default normalization: per-series z-score.

\subsection{Level Shift}
\begin{align}
  x_t &= \eta_t + \Delta\, \mathbf{1}\{t \geq \tau\}, \quad \eta_t \sim \mathcal{N}(0, \text{noise\_std}^2).
\end{align}
Parameters: signed shift magnitude \(\Delta\), changepoint \(\tau\), noise std. Default normalization: none (scale encodes the signal).

\subsection{Random Walk (With Drift)}
\begin{align}
  x_t &= x_{t-1} + \mu + \varepsilon_t,\\
  \varepsilon_t &\sim \mathcal{N}(0, \sigma^2).
\end{align}
Parameters: drift \(\mu\), innovation std. Default normalization: none.

\subsection{Spectral (Sum of Sinusoids)}
\begin{align}
  x_t &= \sum_{j=1}^{k} a_j \sin\big(2\pi f_j t + \phi_j\big) + \varepsilon_t, \quad 0 < f_j < 0.5.
\end{align}
Parameters: number of components \(k \in \{1,\dots,k_{\max}\}\); amplitudes \(a_j\); frequencies \(f_j\) sampled from \([\text{freq\_low},\text{freq\_high}]\); phases \(\phi_j \sim \text{Uniform}(0,2\pi)\); noise std. Default normalization: per-series z-score.

\subsection{Time-Warped Sinusoid}
Generate a base sinusoid \(b_t = \sin(2\pi f t + \phi)\), draw positive steps from a Gamma distribution, form a monotone cumulative warp \(u\) rescaled to \([0, T-1]\), then reinterpolate back to the regular grid:
\begin{align}
  x_t = \operatorname{interp}(t, u, b) + \varepsilon_t.
\end{align}
Parameters: base frequency \(f\), phase \(\phi\), warp strength, noise std. Default normalization: per-series z-score.

\subsection{Deterministic Trend}
\begin{align}
  x_t &= \beta\, t + \varepsilon_t, \quad \varepsilon_t \sim \mathcal{N}(0, \text{noise\_std}^2).
\end{align}
Parameters: slope \(\beta\), noise std. Default normalization: per-series z-score.

\subsection{Variance Shift}
\begin{align}
  x_t &\sim \begin{cases}
    \mathcal{N}(0, \sigma_1^2), & t < \tau,\\
    \mathcal{N}(0, \sigma_2^2), & t \geq \tau.
  \end{cases}
\end{align}
Parameters: changepoint \(\tau\), standard deviations \(\sigma_1, \sigma_2\). Default normalization: none.

\paragraph{Notes on Normalization} Concepts where magnitude/level is the signal (e.g., level or variance shift, random walk) use no normalization by default; others use per-series z-scoring. See the code reference (\texttt{concepts\_dataset.py}) for full details and sampling ranges.

\subsection{Time-series Concepts}
% \major{Describe time-series generation using concept parameters -- use the table from TimeFrechet}
\begin{figure*}[htbp]
    \centering
    % First row
    \begin{subfigure}{0.32\textwidth}
        \includegraphics[width=\linewidth]{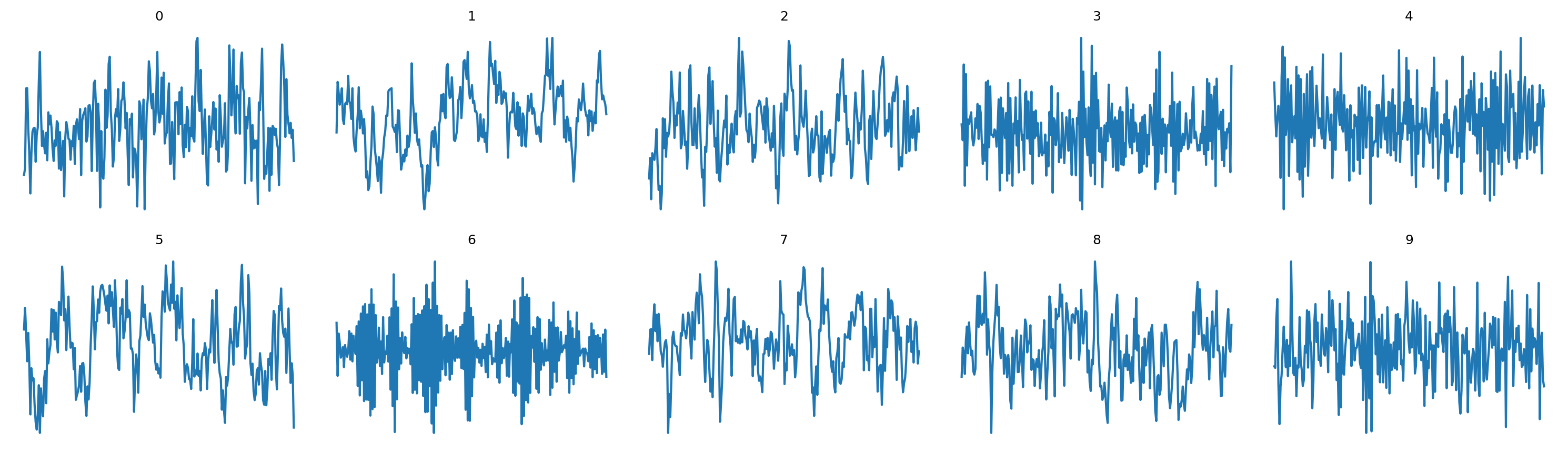}
        \caption{AR1}
        \label{fig:ar1}
    \end{subfigure}
    \begin{subfigure}{0.32\textwidth}
        \includegraphics[width=\linewidth]{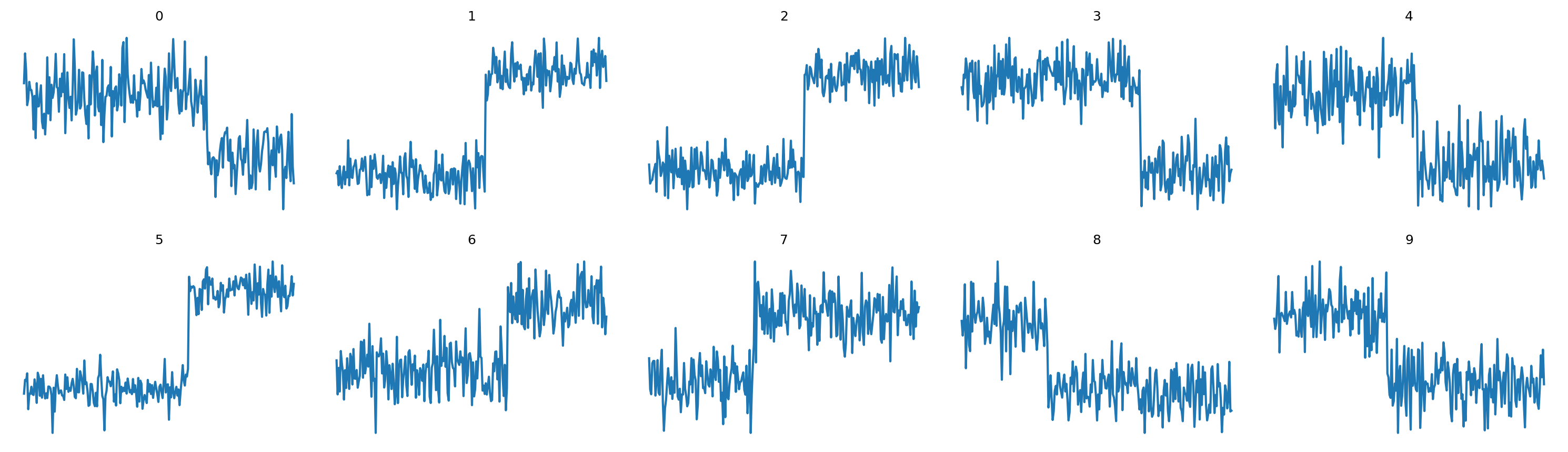}
        \caption{Level Shift}
        \label{fig:level_shift}
    \end{subfigure}
    \begin{subfigure}{0.32\textwidth}
        \includegraphics[width=\linewidth]{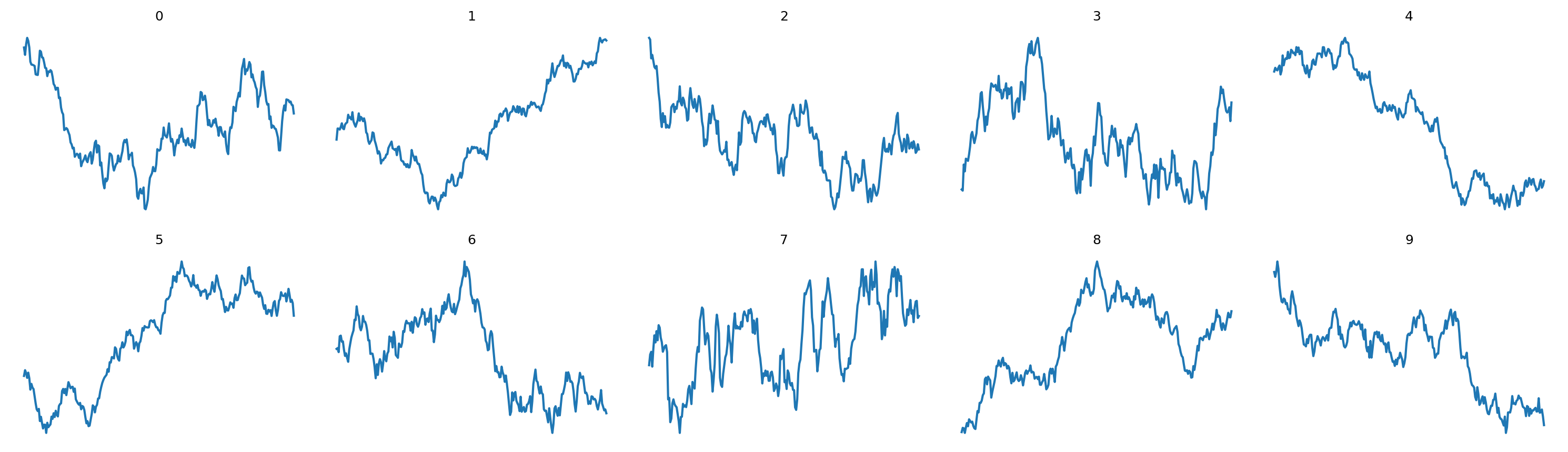}
        \caption{Random Walk}
        \label{fig:random_walk}
    \end{subfigure}
    
    % Second row
    \begin{subfigure}{0.32\textwidth}
        \includegraphics[width=\linewidth]{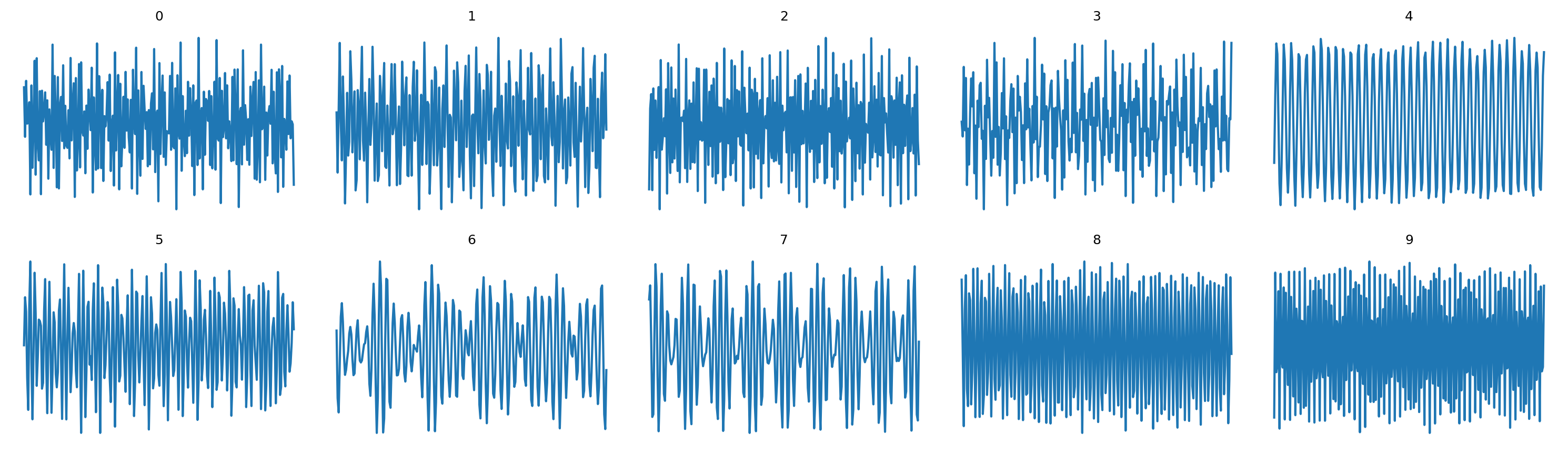}
        \caption{Spectral}
        \label{fig:spectral}
    \end{subfigure}
    \begin{subfigure}{0.32\textwidth}
        \includegraphics[width=\linewidth]{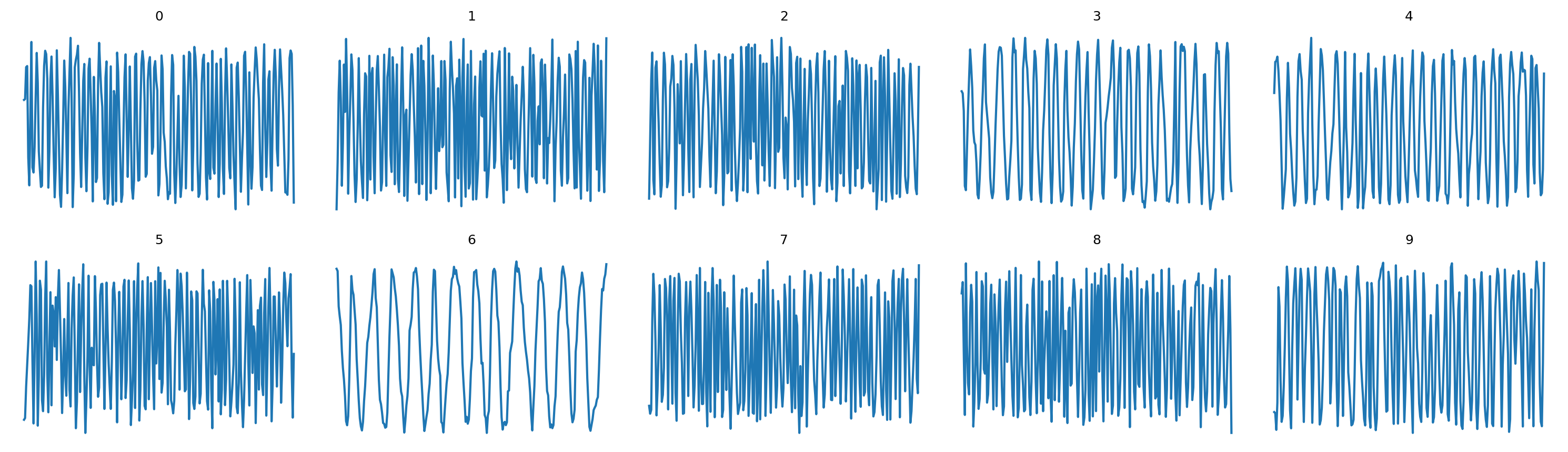}
        \caption{Time warped}
        \label{fig:time_warped}
    \end{subfigure}
    \begin{subfigure}{0.32\textwidth}
        \includegraphics[width=\linewidth]{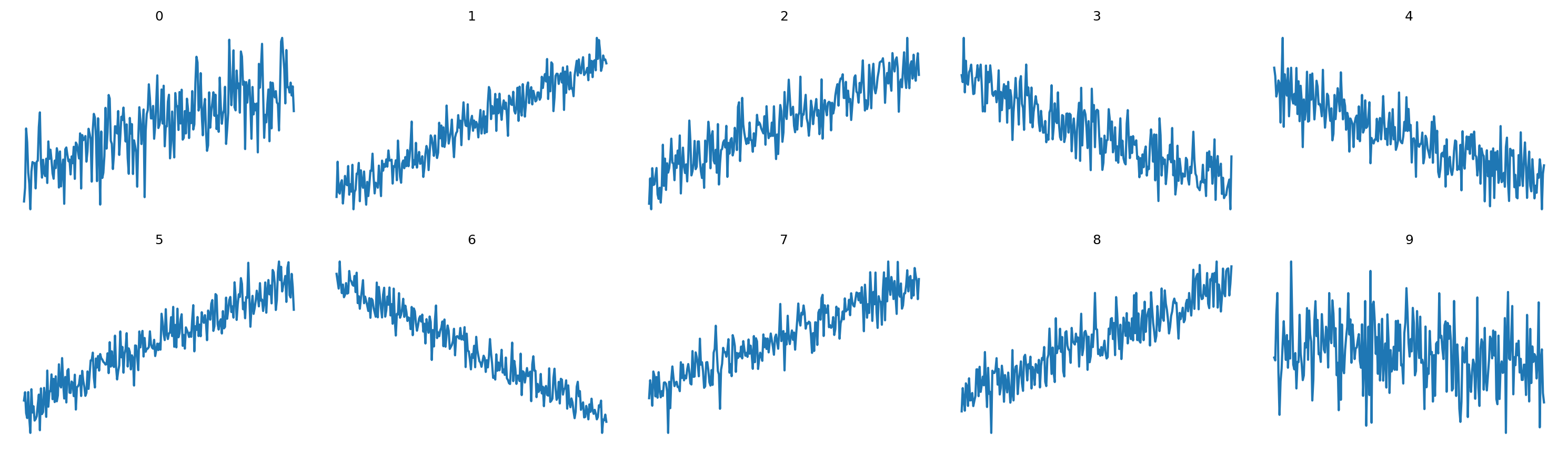}
        \caption{Trend}
        \label{fig:trend_deterministic}
    \end{subfigure}

    % Third row
    \begin{center}
    \begin{subfigure}{0.32\textwidth}
        \includegraphics[width=\linewidth]{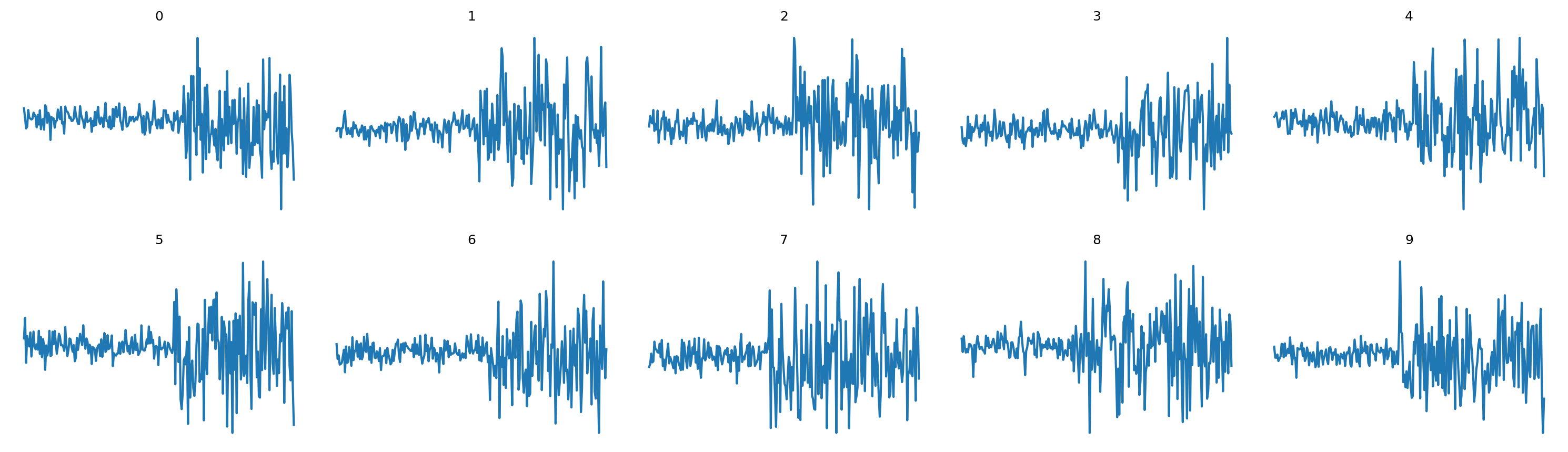}
        \caption{Variance Shift}
        \label{fig:variance_shift}
    \end{subfigure}
    \end{center}
    \caption{Visualization of the synthetic time-series samples generated}
    \label{fig:synth_data}
\end{figure*}
\newpage
\subsection{Time series composition}
\label{ts_composition}
Let $\mathcal{T}1 = \{T_1^{(i)}\}{i=1}^N$ and $\mathcal{T}2 = \{T_2^{(i)}\}{i=1}^N$ be two sets of time series generated from concepts $C_1$ and $C_2$ respectively, where each $T_j^{(i)} \in \mathbb{R}^T$.

\textbf{Structured Composition}. Temporal interleaving with continuity constraints, preserving local concept characteristics in different time segments.

For each sample $i$, we generate breakpoints $a_i, b_i$ where:
$$a_i \sim \mathcal{U}(\lfloor \alpha_{\text{low}} \cdot T \rfloor, \lfloor \alpha_{\text{high}} \cdot T \rfloor)$$
$$b_i \sim \mathcal{U}(\lfloor \beta_{\text{low}} \cdot T \rfloor, \lfloor \beta_{\text{high}} \cdot T \rfloor)$$
with constraints $0 \leq a_i < b_i \leq T$ and default ranges $\alpha_{\text{low}} = 0.2, \alpha_{\text{high}} = 0.4, \beta_{\text{low}} = 0.6, \beta_{\text{high}} = 0.8$.
The structured compositional series $X_{\text{struct}}^{(i)}$ is defined as:
$$X_{\text{struct}}^{(i)}[t] = \begin{cases}
T_1^{(i)}[t] & \text{if } t < a_i \\
T_2^{(i)}[t] + \delta_1^{(i)} & \text{if } a_i \leq t < b_i \\
T_1^{(i)}[t] + \delta_2^{(i)} & \text{if } t \geq b_i
\end{cases}$$
where the continuity offsets are:
$$\delta_1^{(i)} = T_1^{(i)}[a_i] - T_2^{(i)}[a_i]$$
$$\delta_2^{(i)} = T_2^{(i)}[b_i] - T_1^{(i)}[b_i] + \delta_1^{(i)}$$
The corresponding mask $M^{(i)} \in \{0,1\}^T$ indicates the source concept:
$$M^{(i)}[t] = \begin{cases}
0 & \text{if } t < a_i \text{ or } t > b_i \text{ (from } C_1\text{)} \\
1 & \text{if } a_i \leq t \leq b_i \text{ (from } C_2\text{)}
\end{cases}$$

\textbf{Functional Composition}. Elementwise addition creating global interaction between concepts throughout the entire time series
Both approaches generate datasets containing the composed series $X$, original component series $T_1, T_2$, and metadata preserving the generative parameters from both source concepts.

For functional composition, we first optionally normalize each time series:
% $$\tilde{T}j^{(i)} = \begin{cases}
% \frac{T_j^{(i)} - \mu_j^{(i)}}{\sigma_j^{(i)}} & \text{if normalize\each = True} \\
% T_j^{(i)} & \text{otherwise}
% \end{cases}$$
$$
\tilde{T}_j^{(i)} = 
\begin{cases}
\frac{T_j^{(i)} - \mu_j^{(i)}}{\sigma_j^{(i)}} & \text{if normalize = True} \\
T_j^{(i)} & \text{otherwise}
\end{cases}
$$

where $\mu_j^{(i)} = \frac{1}{T}\sum_{t=1}^T T_j^{(i)}[t]$ and $\sigma_j^{(i)} = \sqrt{\frac{1}{T}\sum_{t=1}^T (T_j^{(i)}[t] - \mu_j^{(i)})^2}$.
The functional compositional series is then:
$$X_{\text{func}}^{(i)} = \tilde{T}1^{(i)} + \tilde{T}_2^{(i)}$$

\newpage
\section{Layer Representation Similarity}\label{app:layer-rep-similarity}
\begin{figure*}[htbp]
    \centering
    % First row
    \begin{subfigure}{0.48\textwidth}
        \includegraphics[width=\linewidth]{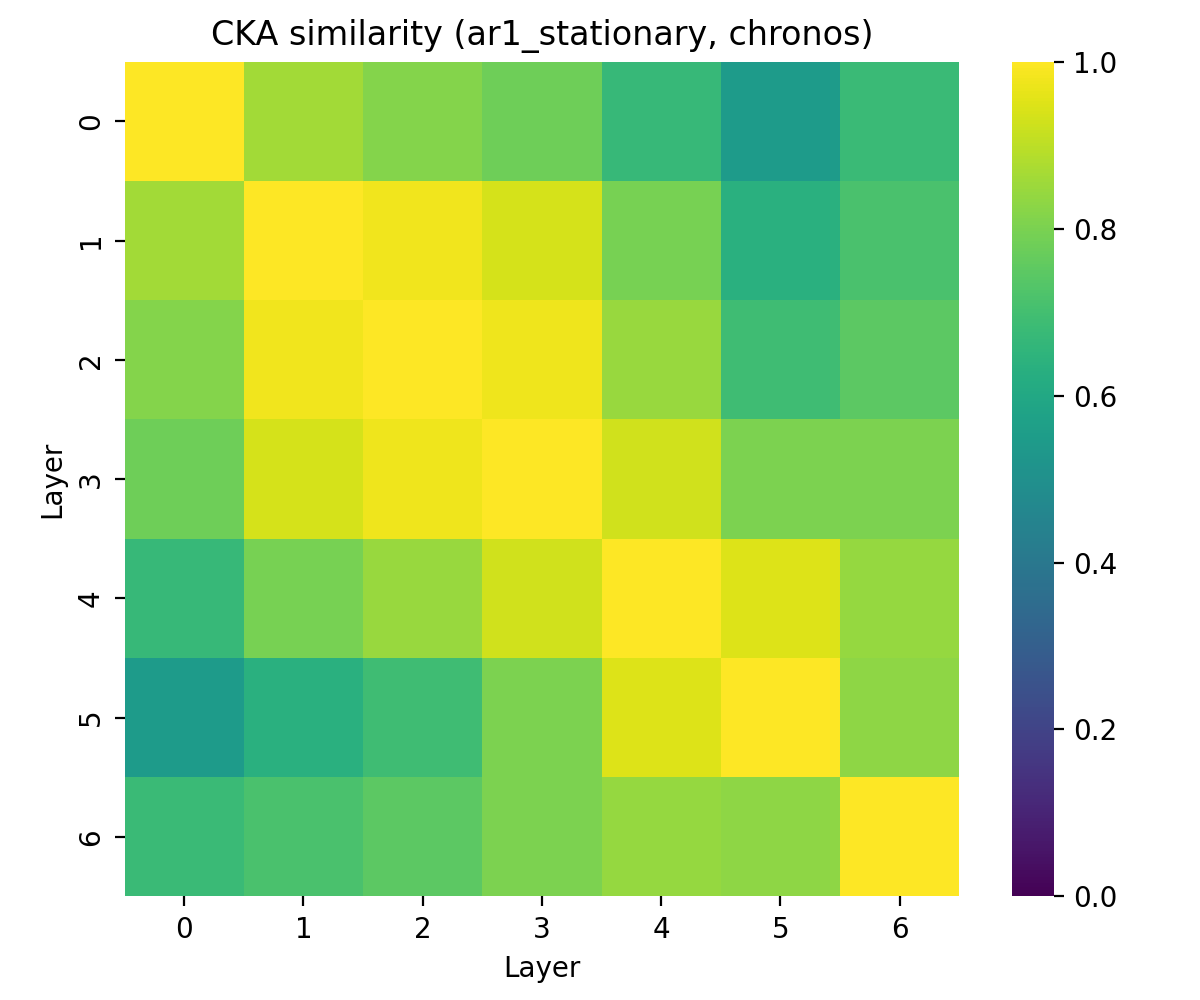}
        \caption{AR1}
        \label{fig:chronos_ck_ar1_stationary}
    \end{subfigure}
    \begin{subfigure}{0.48\textwidth}
        \includegraphics[width=\linewidth]{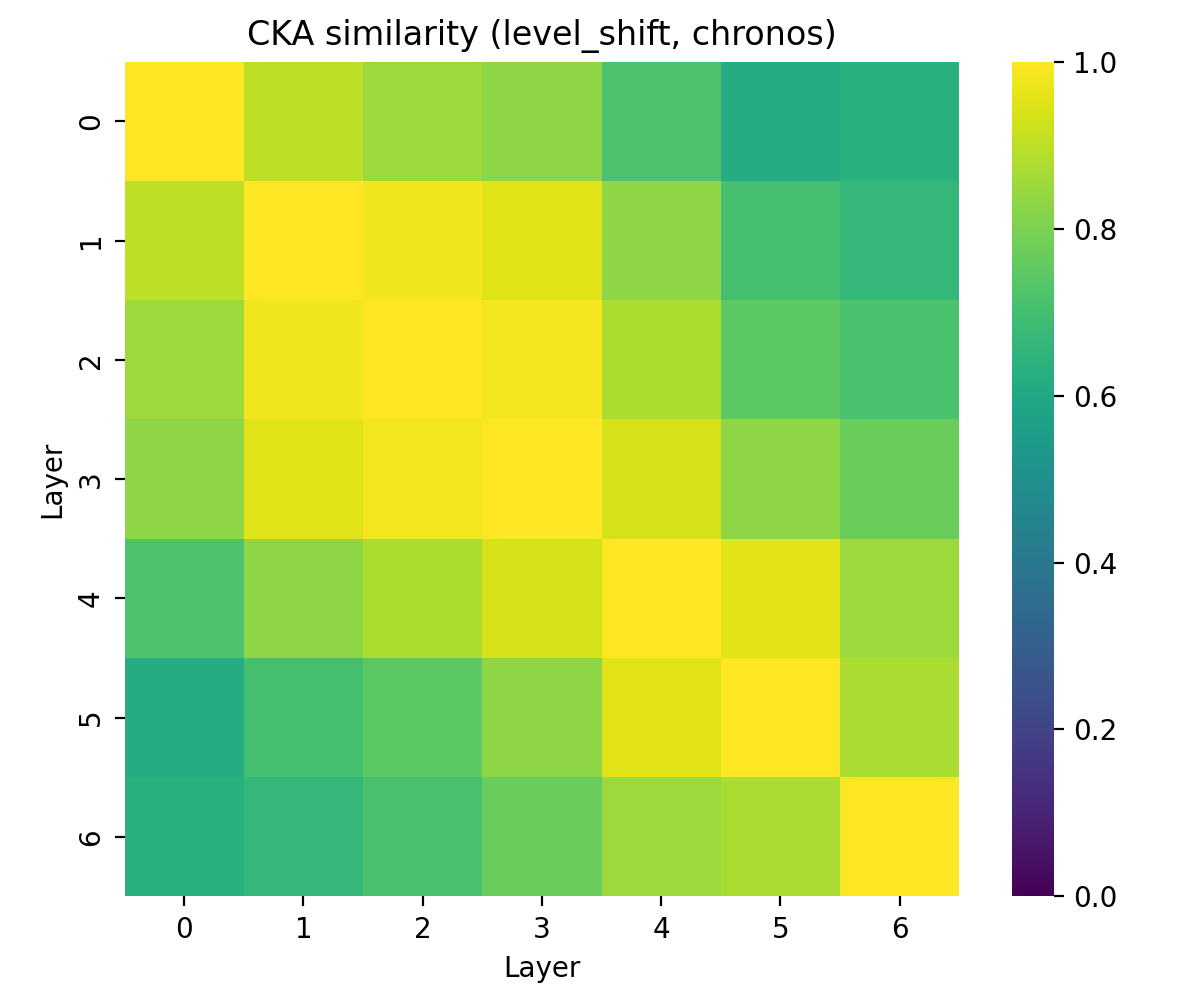}
        \caption{Level Shift}
        \label{fig:chronos_ck_level_shift}
    \end{subfigure}

    % Second row
    \begin{subfigure}{0.48\textwidth}
        \includegraphics[width=\linewidth]{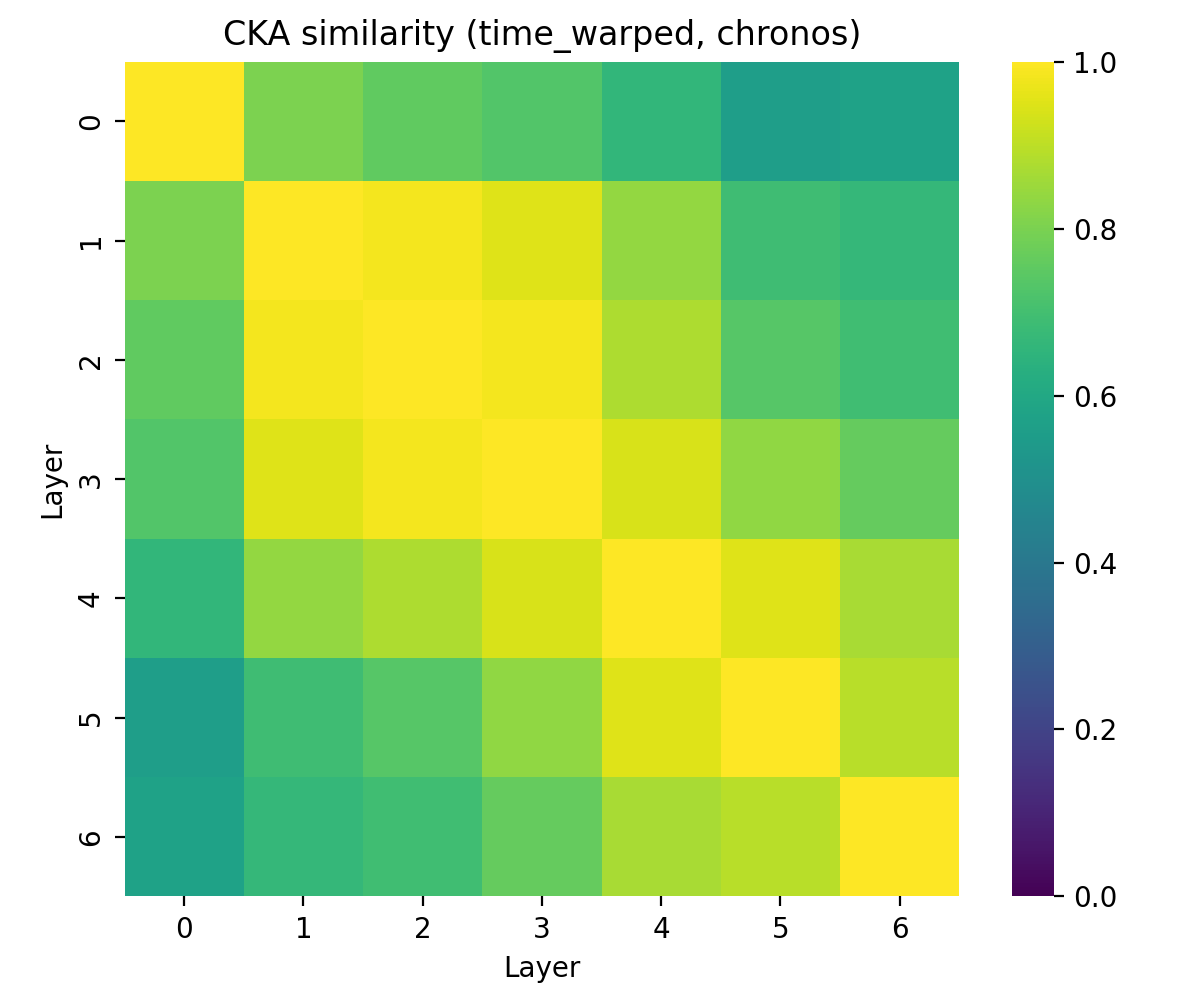}
        \caption{Time warped}
        \label{fig:chronos_ck_time_warped}
    \end{subfigure}
    \begin{subfigure}{0.48\textwidth}
        \includegraphics[width=\linewidth]{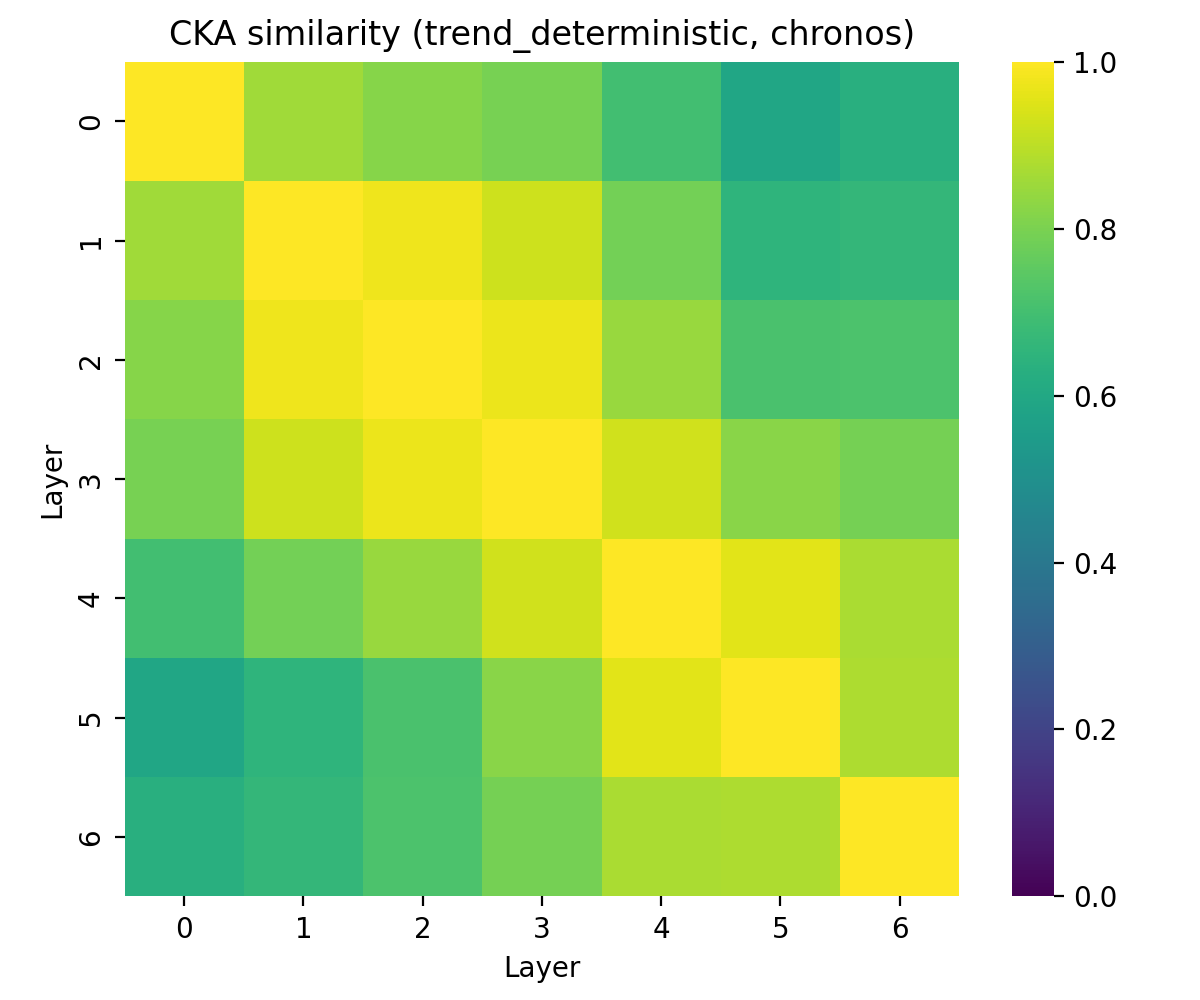}
        \caption{Trend}
        \label{fig:chronos_ck_trend_deterministic}
    \end{subfigure}

    % % Third row
    % % \begin{center}
    % \begin{subfigure}{0.32\textwidth}
    %     \includegraphics[width=\linewidth]{figures/chronos_ck_variance_shift.png}
    %     \caption{Variance Shift}
    %     \label{fig:chronos_ck_variance_shift}
    % \end{subfigure}
    % \begin{subfigure}{0.32\textwidth}
    %     \includegraphics[width=\linewidth]{figures/chronos_cpc_cka_similarity.png}
    %     \caption{Changepoint piecewise}
    %     \label{fig:chronos_cpc_cka_similarity}
    % \end{subfigure}
    
    % \end{center}
    \caption{CKA Similarity among layers of Chronos TSFM}
    \label{fig:tsfm_results}
\end{figure*}

\begin{figure*}[htbp]
    \centering
    % First row
    \begin{subfigure}{0.48\textwidth}
        \includegraphics[width=\linewidth]{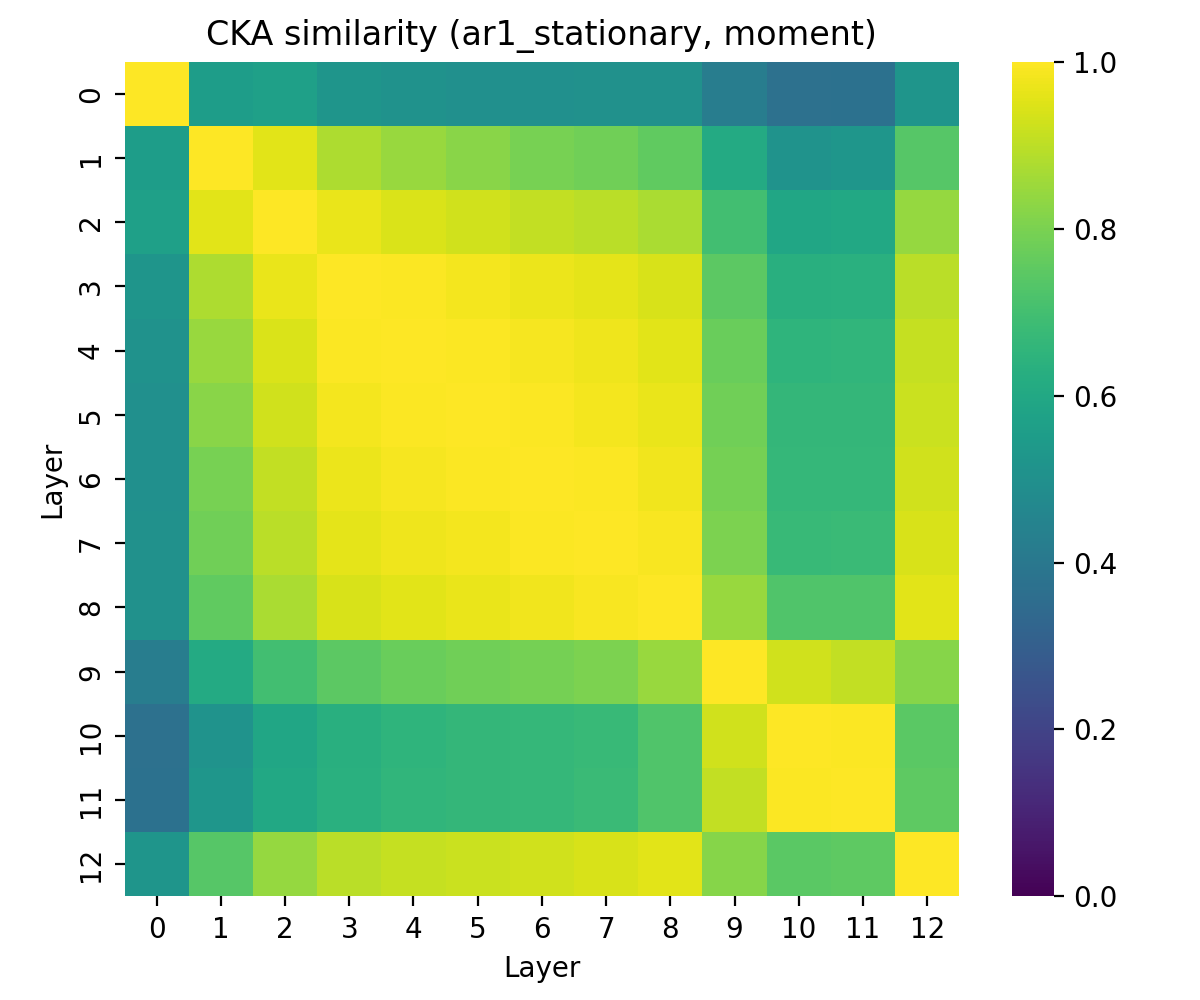}
        \caption{AR1}
        \label{fig:moment_ck_ar1_stationary}
    \end{subfigure}
    \begin{subfigure}{0.48\textwidth}
        \includegraphics[width=\linewidth]{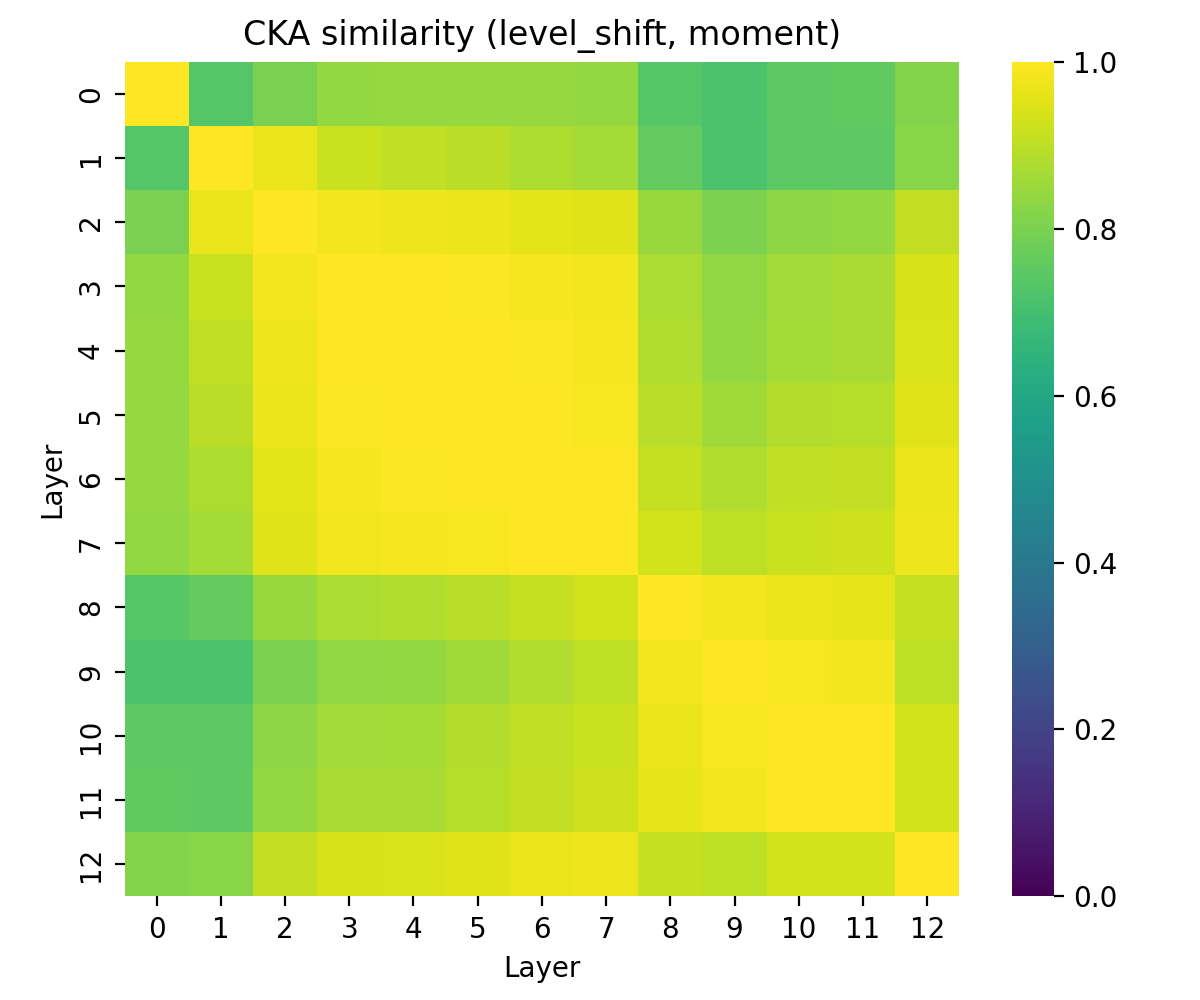}
        \caption{Level Shift}
        \label{fig:moment_ck_level_shift}
    \end{subfigure}
    % \begin{subfigure}{0.32\textwidth}
    %     \includegraphics[width=\linewidth]{figures/moment_ck_random_walk.png}
    %     \caption{Random Walk}
    %     \label{fig:moment_ck_random_walk}
    % \end{subfigure}
    
    % Second row
    % \begin{subfigure}{0.32\textwidth}
    %     \includegraphics[width=\linewidth]{figures/moment_ck_spectral.png}
    %     \caption{Spectral}
    %     \label{fig:moment_ck_spectral}
    % \end{subfigure}
    \begin{subfigure}{0.48\textwidth}
        \includegraphics[width=\linewidth]{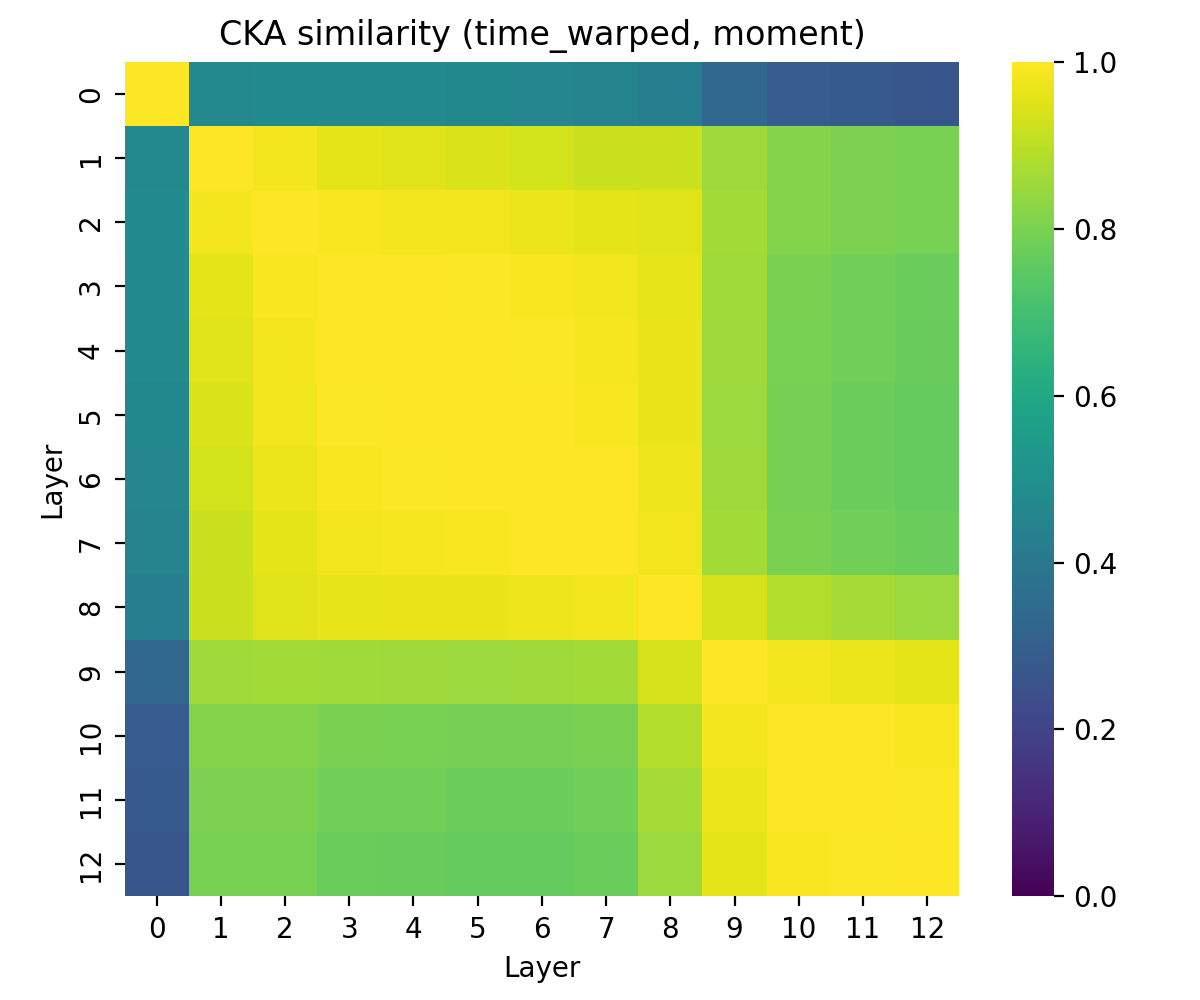}
        \caption{Time warped}
        \label{fig:moment_ck_time_warped}
    \end{subfigure}
    \begin{subfigure}{0.48\textwidth}
        \includegraphics[width=\linewidth]{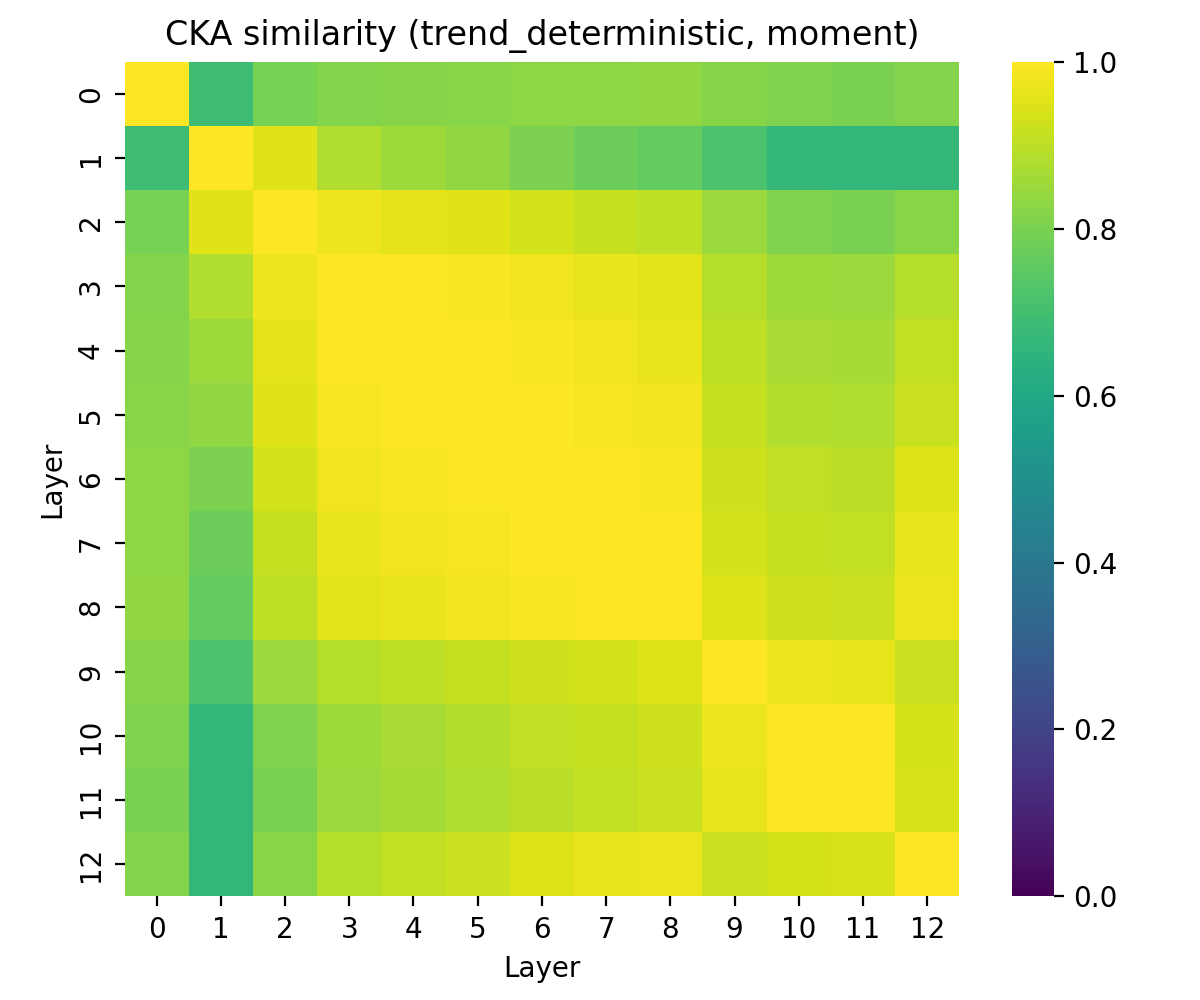}
        \caption{Trend}
        \label{fig:moment_ck_trend_deterministic}
    \end{subfigure}

    % Third row
    % \begin{center}
    % \begin{subfigure}{0.32\textwidth}
    %     \includegraphics[width=\linewidth]{figures/moment_ck_variance_shift.png}
    %     \caption{Variance Shift}
    %     \label{fig:moment_ck_variance_shift}
    % \end{subfigure}
    % \begin{subfigure}{0.32\textwidth}
    %     \includegraphics[width=\linewidth]{figures/moment_cpc_cka_similarity.png}
    %     \caption{Changepoint piecewise}
    %     \label{fig:moment_cpc_cka_similarity}
    % \end{subfigure}
    
    % \end{center}
    \caption{CKA Similarity among layers of MOMENT TSFM}
    \label{fig:tsfm_results}
\end{figure*}

\newpage

\section{Linear Probe Loss}\label{app:linearprobeloss}

\begin{figure*}[htbp]
    \centering
    % First row
    \begin{subfigure}{0.48\textwidth}
        \includegraphics[width=\linewidth]{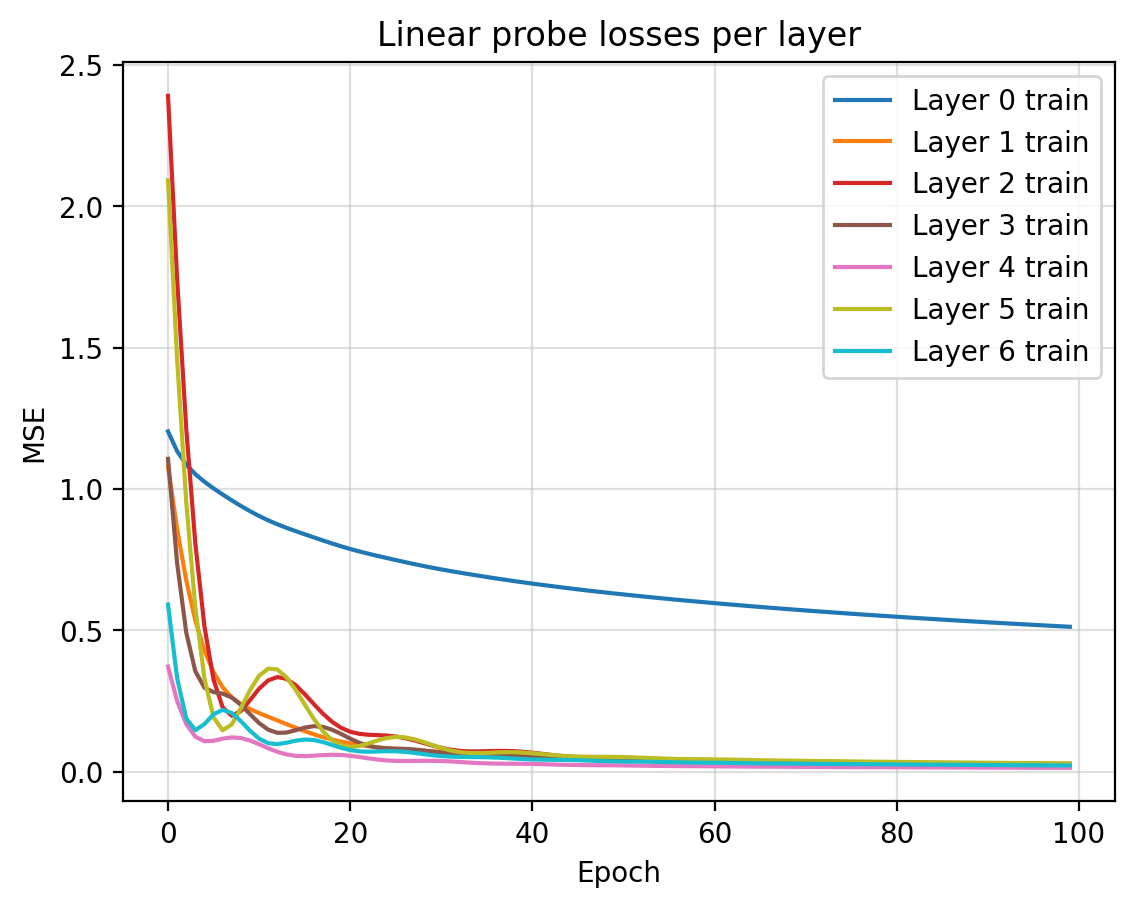}
        \caption{AR1}
        \label{fig:chronos_linear_probe_loss_ar1}
    \end{subfigure}
    \begin{subfigure}{0.48\textwidth}
        \includegraphics[width=\linewidth]{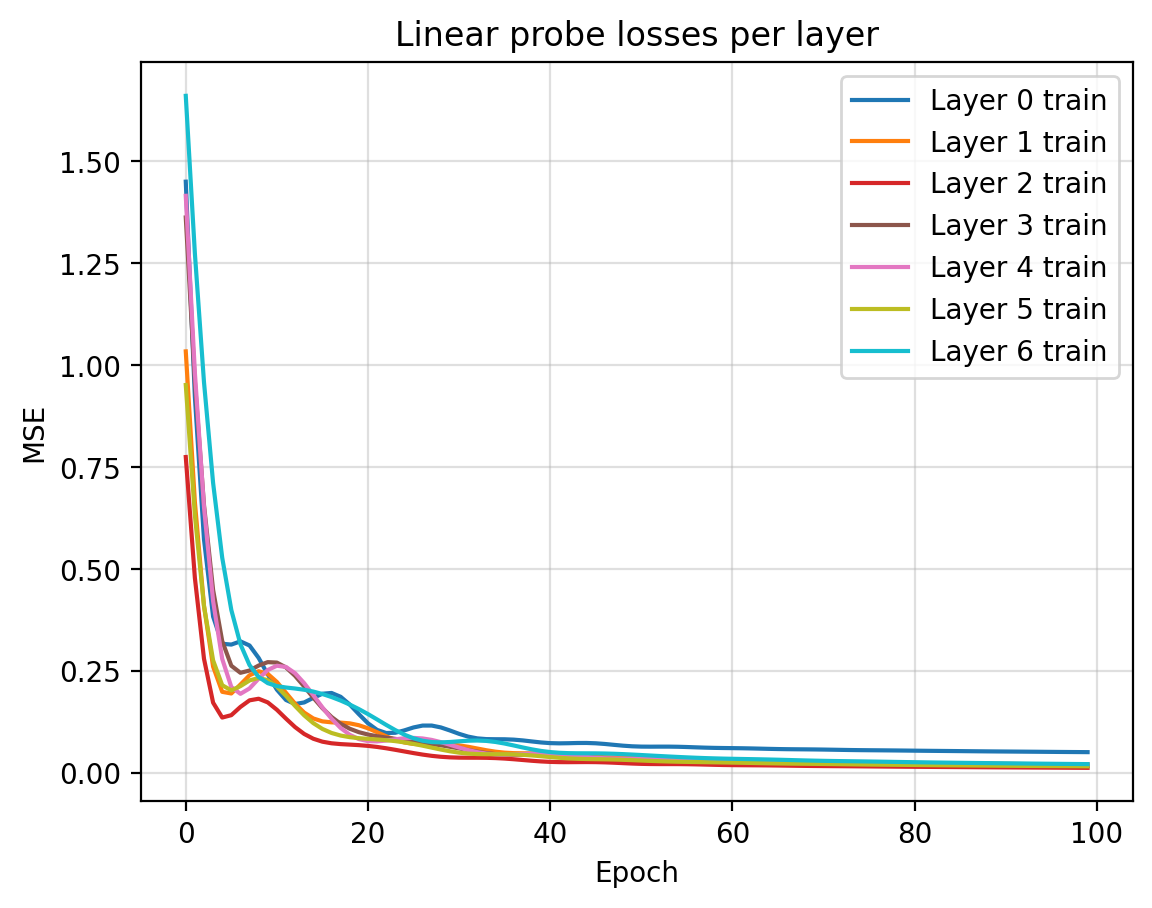}
        \caption{Level Shift}
        \label{fig:chronos_linear_probe_loss_level_shift}
    \end{subfigure}
    % \begin{subfigure}{0.32\textwidth}
    %     \includegraphics[width=\linewidth]{figures/linear_probe_loss/chronos_linear_probe_loss_random_walk.png}
    %     \caption{Random Walk}
    %     \label{fig:chronos_linear_probe_loss_random_walk}
    % \end{subfigure}
    
    % Second row
    % \begin{subfigure}{0.32\textwidth}
    %     \includegraphics[width=\linewidth]{figures/linear_probe_loss/chronos_linear_probe_loss_spectral.png}
    %     \caption{Spectral}
    %     \label{fig:chronos_linear_probe_loss_spectral}
    % \end{subfigure}
    \begin{subfigure}{0.48\textwidth}
        \includegraphics[width=\linewidth]{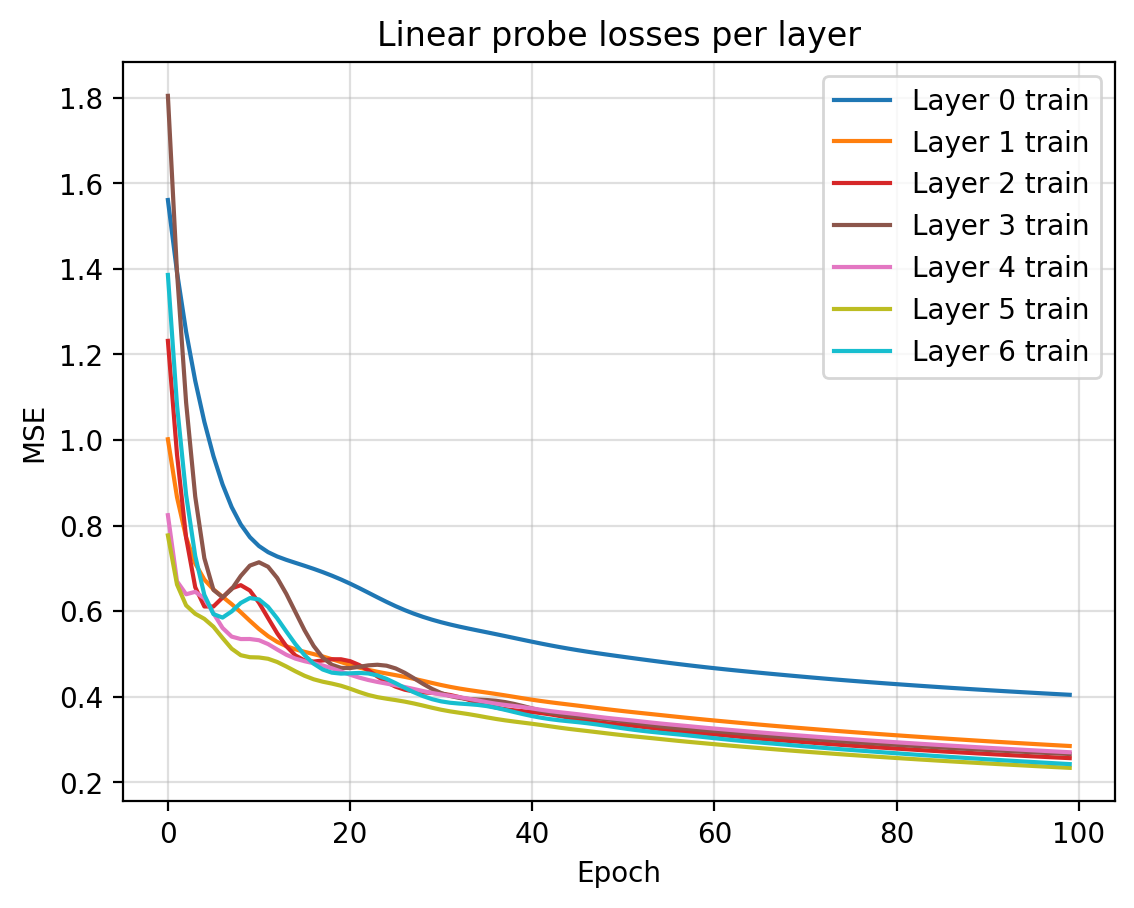}
        \caption{Time warped}
        \label{fig:chronos_linear_probe_loss_time_warped}
    \end{subfigure}
    \begin{subfigure}{0.48\textwidth}
        \includegraphics[width=\linewidth]{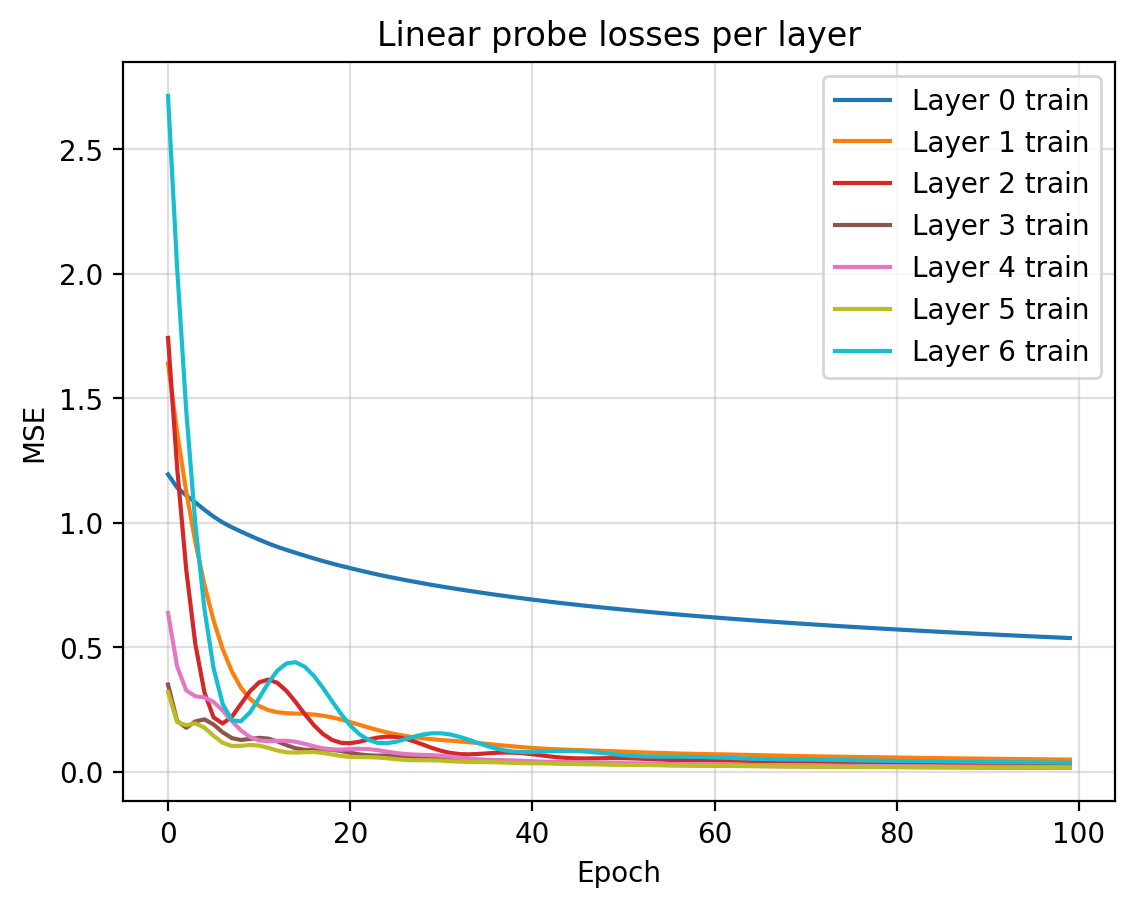}
        \caption{Trend}
    \label{fig:chronos_linear_probe_loss_trend_deterministic}
    \end{subfigure}

    % Third row
    % \begin{center}
    % \begin{subfigure}{0.32\textwidth}
    %     \includegraphics[width=\linewidth]{figures/linear_probe_loss/chronos_linear_probe_loss_variance_shift.png}
    %     \caption{Variance Shift}
    %     \label{fig:chronos_linear_probe_loss_variance_shift}
    % \end{subfigure}
    % \end{center}
    \caption{Layer-wise Loss in Chronos TSFM}
    \label{fig:tsfm_results}
\end{figure*}
\newpage

\begin{figure*}[htbp]
    \centering
    % First row
    \begin{subfigure}{0.48\textwidth}
        \includegraphics[width=\linewidth]{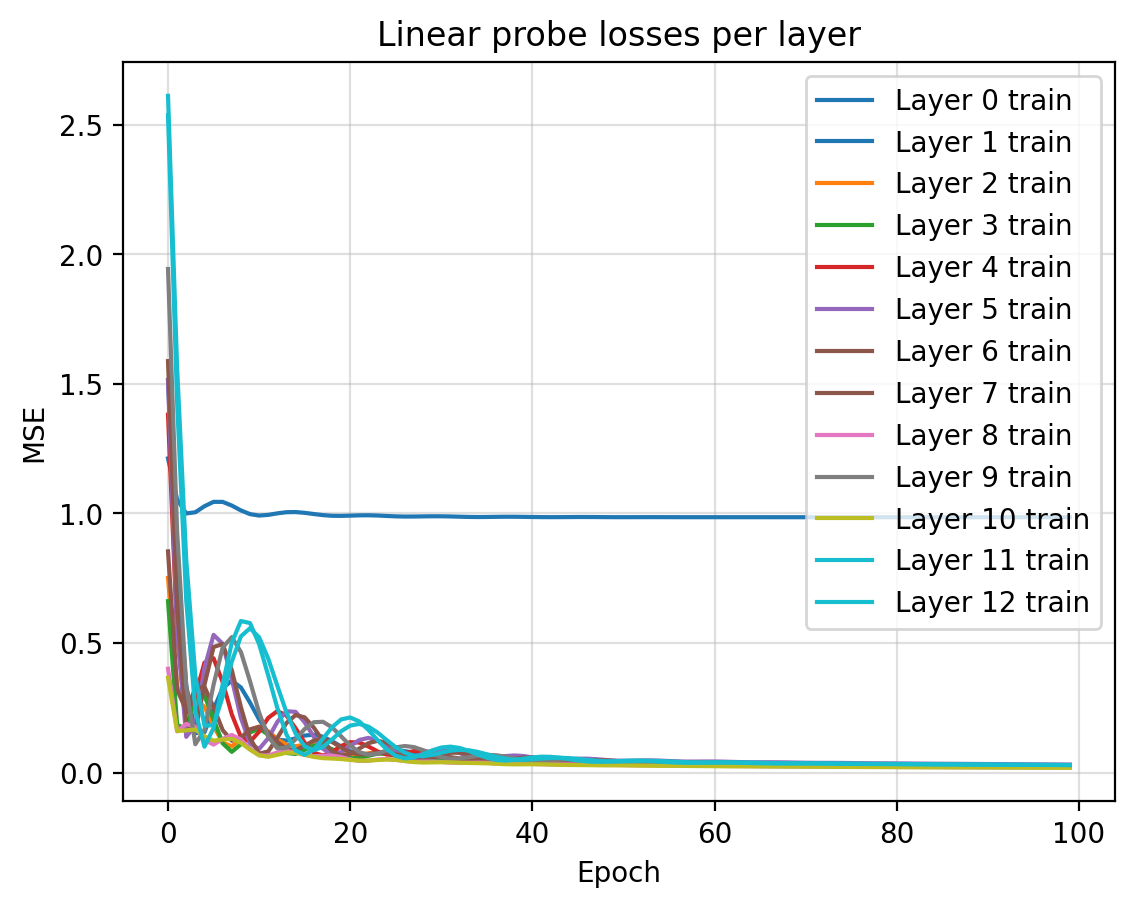}
        \caption{AR1}
        \label{fig:moment_linear_probe_ar1}
    \end{subfigure}
    \begin{subfigure}{0.48\textwidth}
        \includegraphics[width=\linewidth]{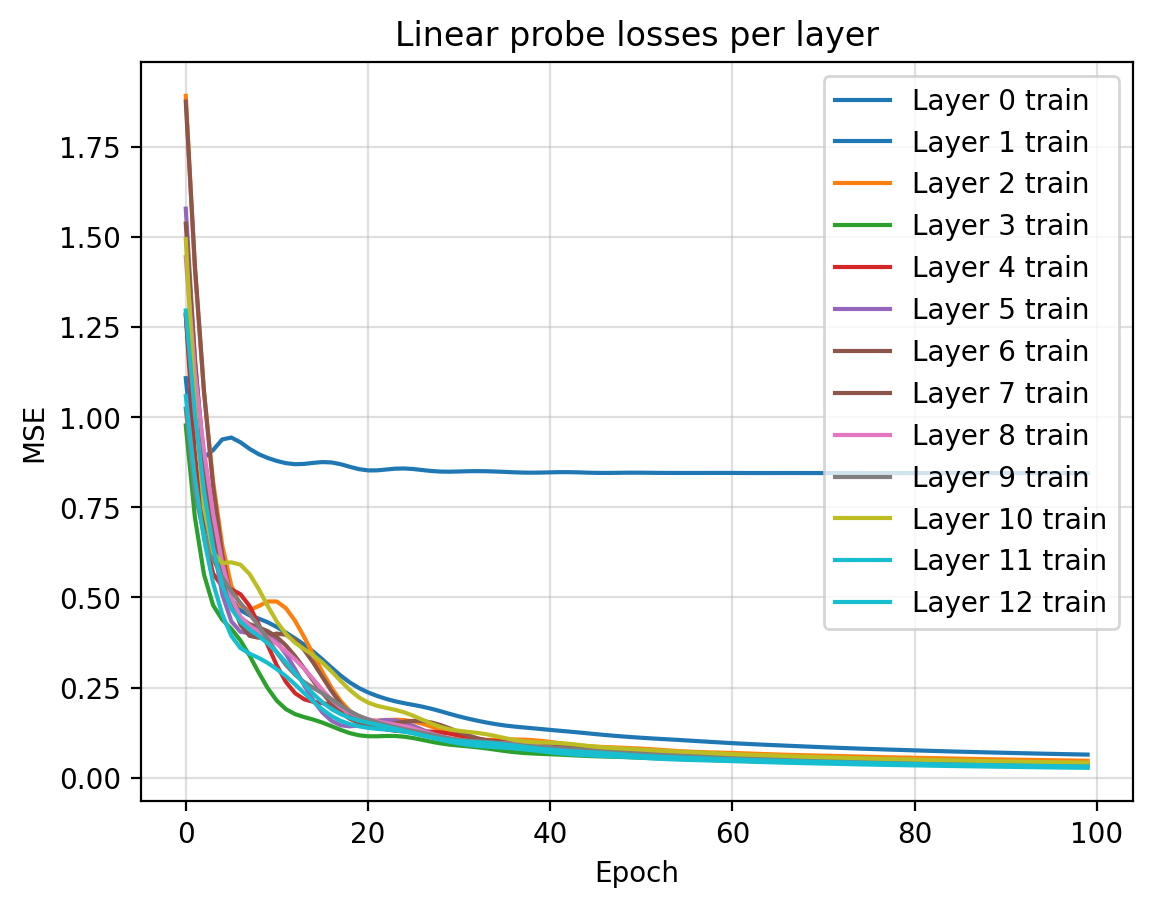}
        \caption{Level Shift}
        \label{fig:moment_linear_probe_level_shift}
    \end{subfigure}
    % \begin{subfigure}{0.32\textwidth}
    %     \includegraphics[width=\linewidth]{figures/linear_probe_loss/moment_linear_probe_random_walk.png}
    %     \caption{Random Walk}
    %     \label{fig:moment_linear_probe_random_walk}
    % \end{subfigure}
    
    % Second row
    % \begin{subfigure}{0.32\textwidth}
    %     \includegraphics[width=\linewidth]{figures/linear_probe_loss/moment_linear_probe_spectral.png}
    %     \caption{Spectral}
    %     \label{fig:moment_linear_probe_spectral}
    % \end{subfigure}
    \begin{subfigure}{0.48\textwidth}
        \includegraphics[width=\linewidth]{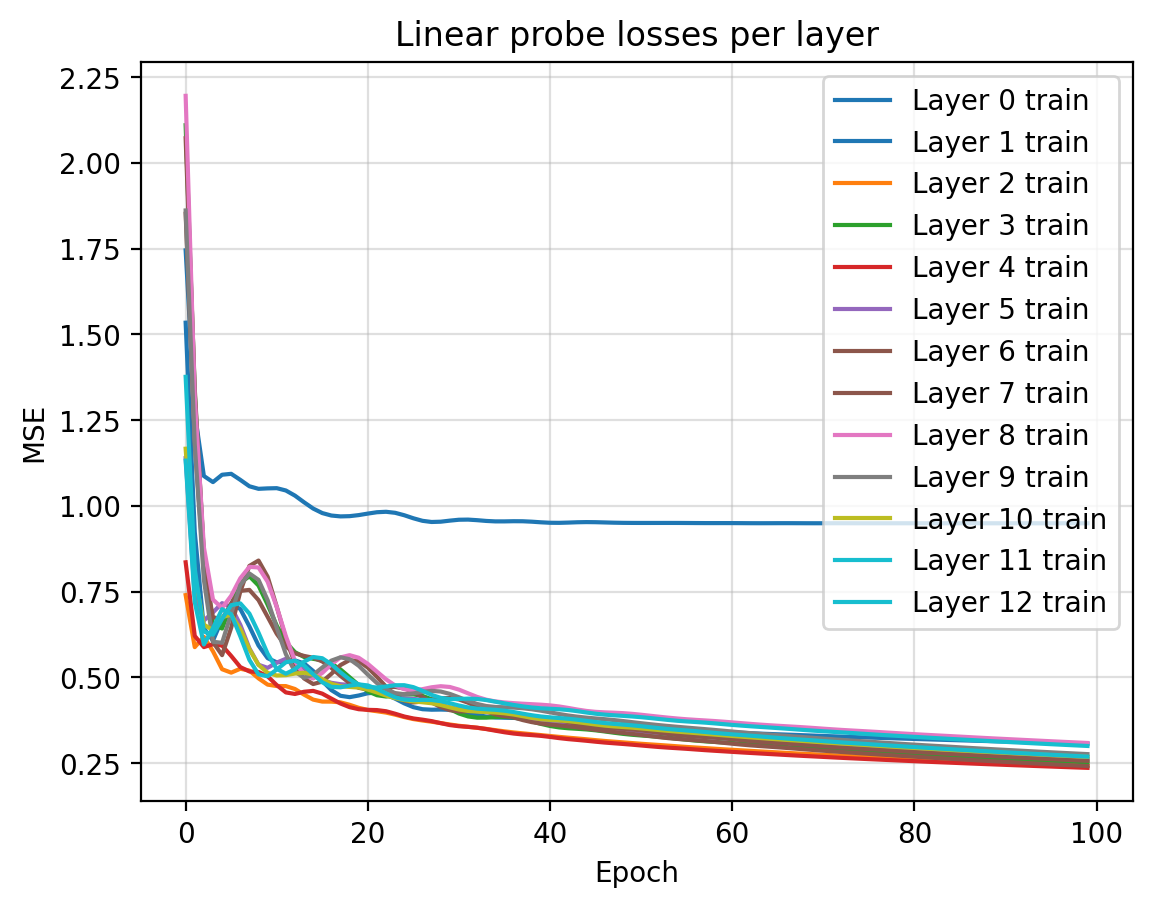}
        \caption{Time warped}
        \label{fig:moment_linear_probe_time_warped}
    \end{subfigure}
    \begin{subfigure}{0.48\textwidth}
        \includegraphics[width=\linewidth]{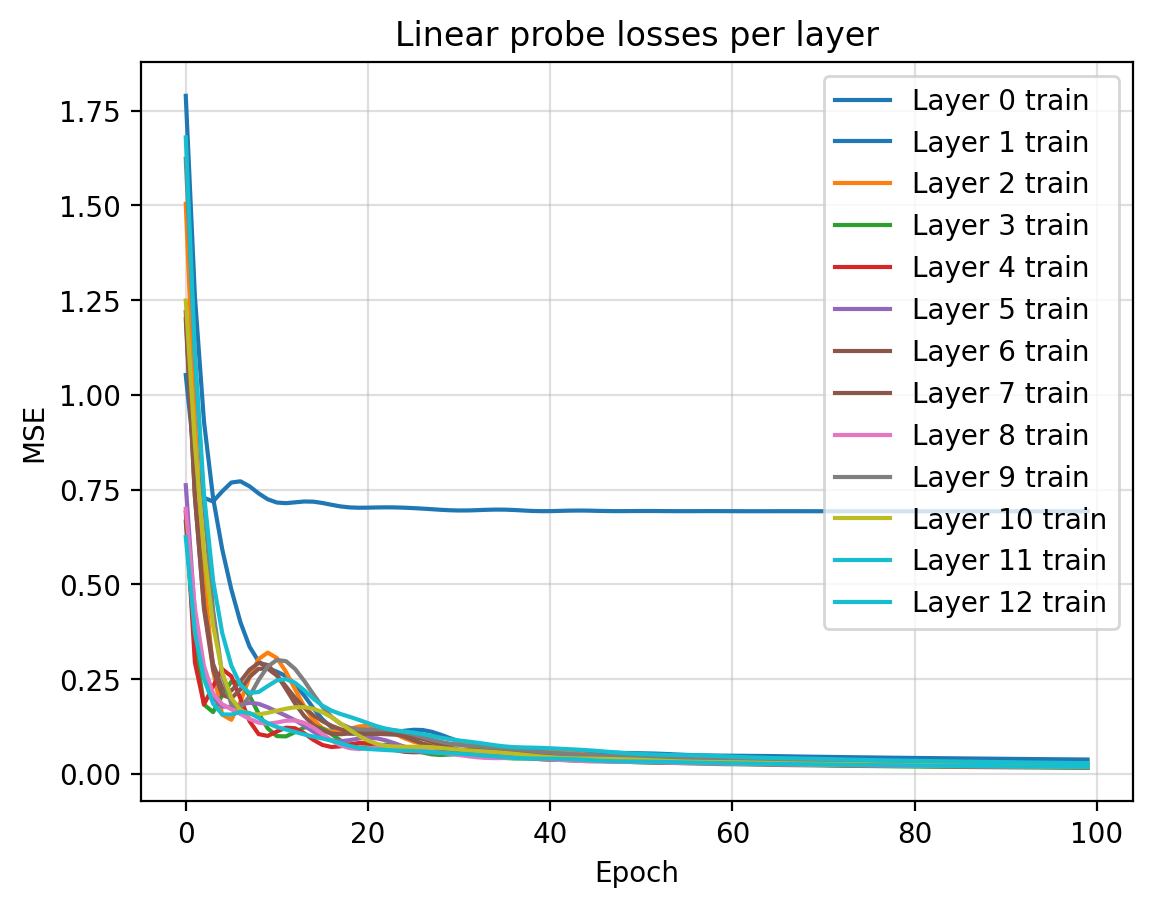}
        \caption{Trend}
    \label{fig:moment_linear_probe_trend_deterministic}
    \end{subfigure}

    % Third row
    % \begin{center}
    % \begin{subfigure}{0.32\textwidth}
    %     \includegraphics[width=\linewidth]{figures/linear_probe_loss/moment_linear_probe_variance_shift.png}
    %     \caption{Variance Shift}
    %     \label{fig:moment_linear_probe_variance_shift}
    % \end{subfigure}
    % \end{center}
    \caption{Layer-wise Loss in MOMENT TSFM}
    \label{fig:tsfm_results}
\end{figure*}
\newpage
% Section: Results --- Layerwise Plots
\section{ Layerwise Respresentation Visualization}\label{app:layerwise-dimred}

This section summarizes layerwise embeddings visualized via PCA, t-SNE, and UMAP for each concept and model. We show triplets of layers per method.

% Helper: a 3-wide figure row macro (natbib-safe: no captions here)
% \newcommand{\threeplots}[3]{%
%   \begin{minipage}[b]{0.32\linewidth}\centering
%     \includegraphics[width=\linewidth]{#1}
%   \end{minipage}\hfill
%   \begin{minipage}[b]{0.32\linewidth}\centering
%     \includegraphics[width=\linewidth]{#2}
%   \end{minipage}\hfill
%   \begin{minipage}[b]{0.32\linewidth}\centering
%     \includegraphics[width=\linewidth]{#3}
%   \end{minipage}%
% }

% ---------------- AR(1) ----------------
\subsection{AR(1) (Stationary)}

\paragraph{Moment (parameter: $\phi$).}
% \begin{figure}[t]
%   \centering
%   \threeplots{figs/ar1_moment/ar1/moment_ar1_pca_Layer_00_phis.png}%
%              {figs/ar1_moment/ar1/moment_ar1_pca_Layer_06_phis.png}%
%              {figs/ar1_moment/ar1/moment_ar1_pca_Layer_12_phis.png}
%   \caption{AR(1) --- Moment --- PCA (Layers 00/06/12)}
% \end{figure}

% \begin{figure}[t]
%   \centering
%   \threeplots{figs/ar1_moment/ar1/moment_ar1_tsne_Layer_00_phis.png}%
%              {figs/ar1_moment/ar1/moment_ar1_tsne_Layer_06_phis.png}%
%              {figs/ar1_moment/ar1/moment_ar1_tsne_Layer_12_phis.png}
%   \caption{AR(1) --- Moment --- t-SNE (Layers 00/06/12)}
% \end{figure}

\begin{figure}[t]
  \centering
  \threeplots{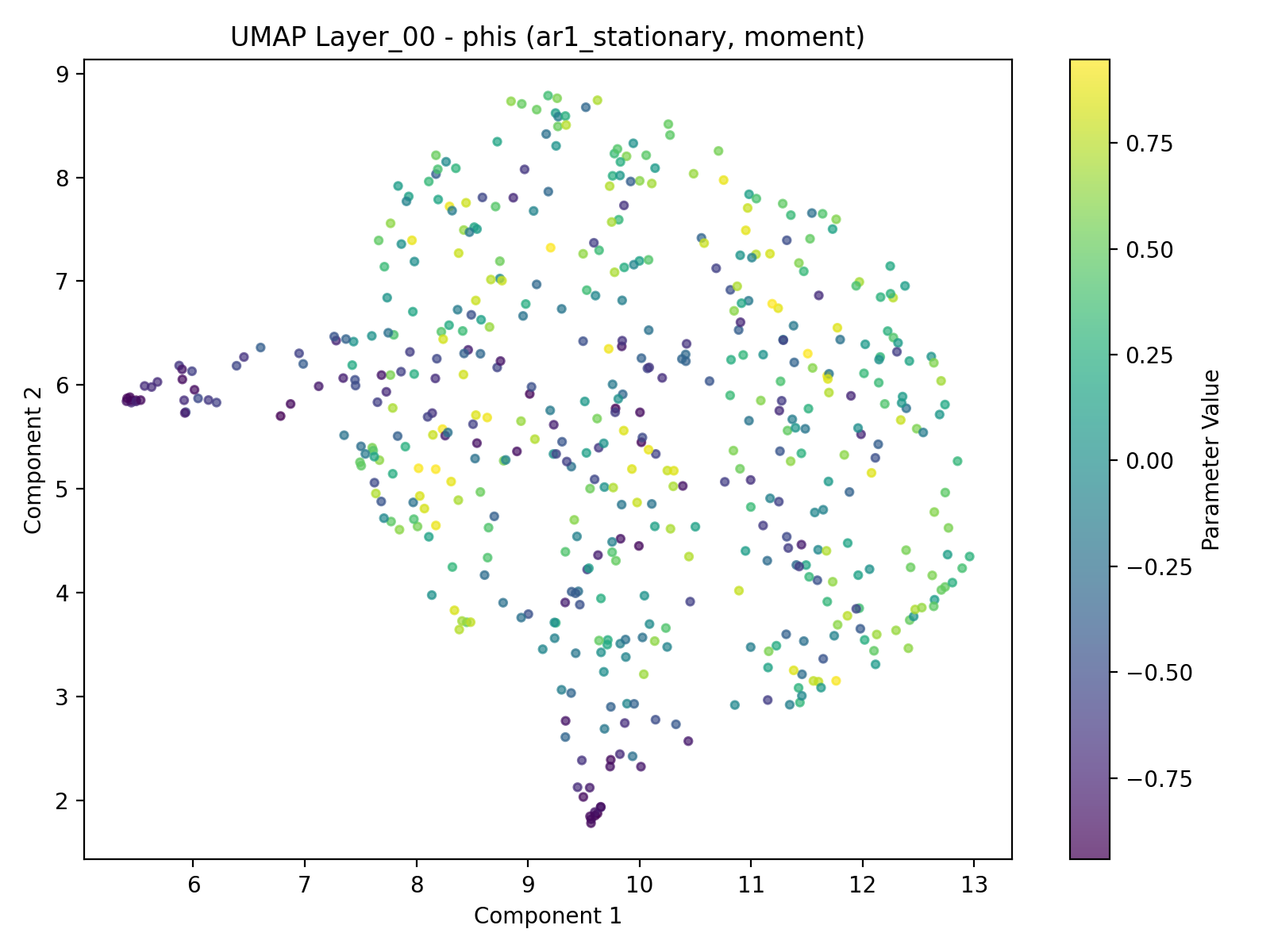}%
             {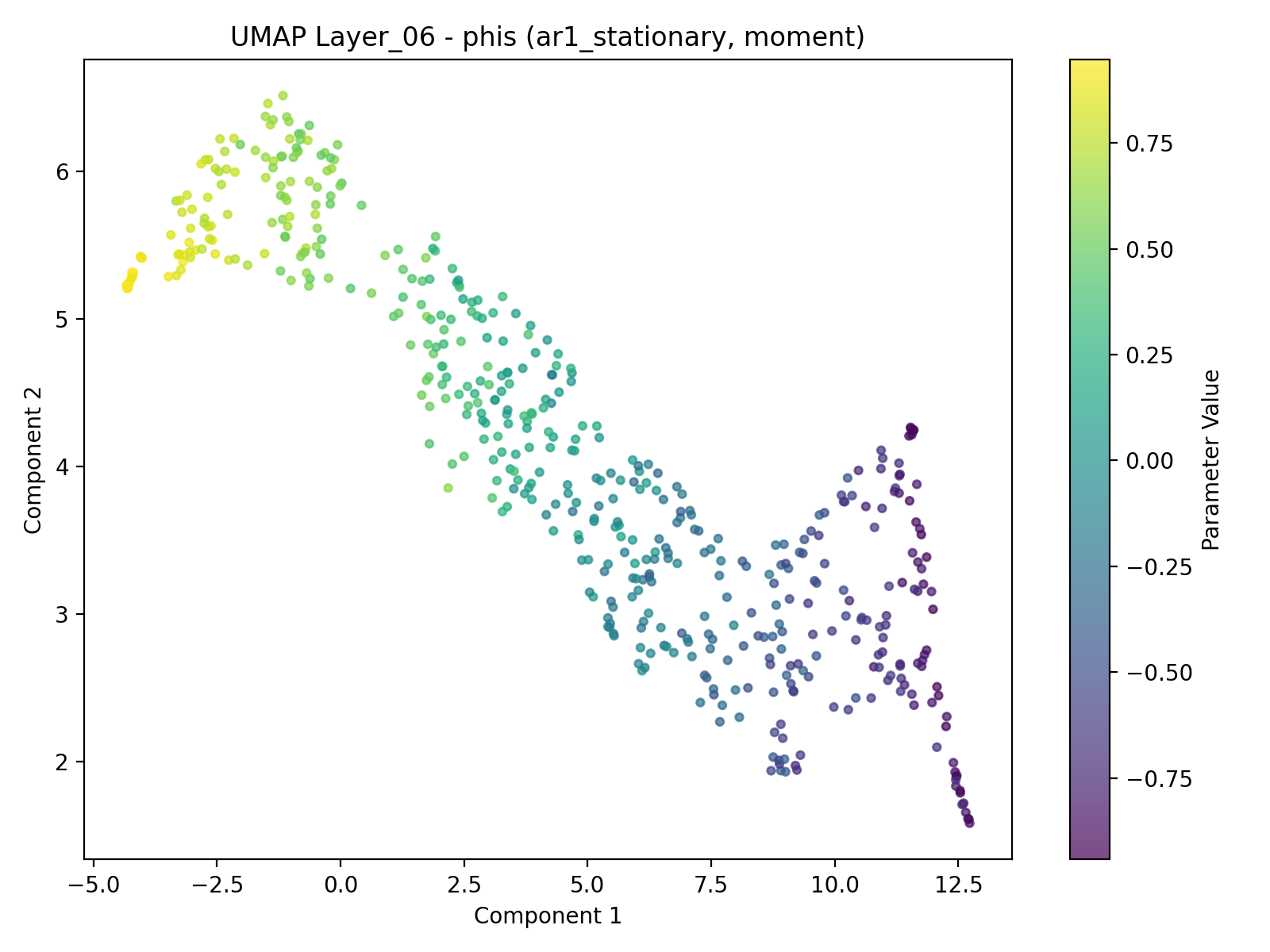}%
             {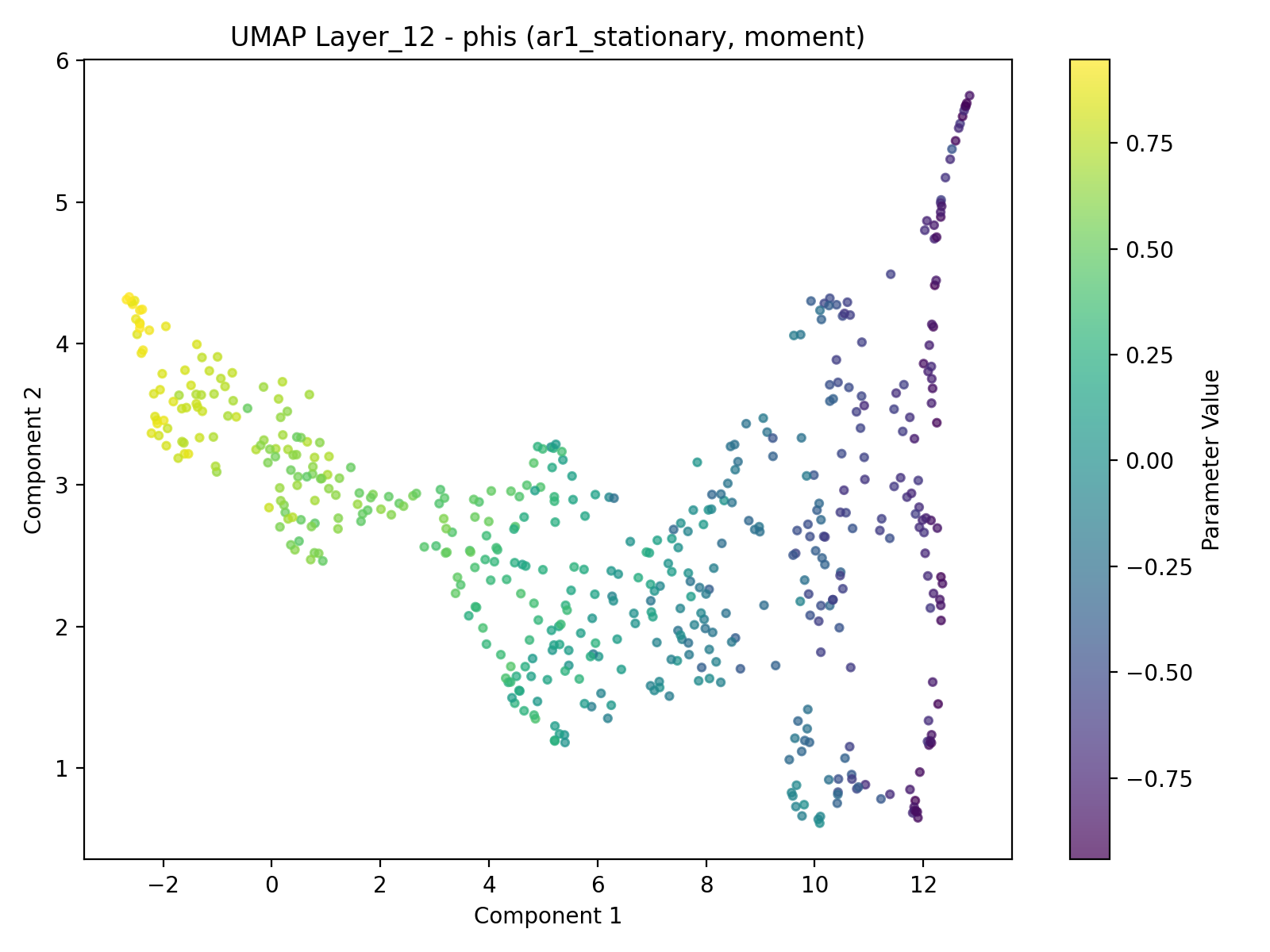}
  \caption{AR(1) --- Moment --- UMAP (Layers 00/06/12)}
\end{figure}
\newpage
\paragraph{Chronos (parameter: $\phi$).}
% \begin{figure}[t]
%   \centering
%   \threeplots{figs/ar1_stationary_chronos/ar1_stationary/chronos_ar1_pca_Layer_00_phis.png}%
%              {figs/ar1_stationary_chronos/ar1_stationary/chronos_ar1_pca_Layer_03_phis.png}%
%              {figs/ar1_stationary_chronos/ar1_stationary/chronos_ar1_pca_Layer_06_phis.png}
%   \caption{AR(1) --- Chronos --- PCA (Layers 00/03/06)}
% \end{figure}

% \begin{figure}[t]
%   \centering
%   \threeplots{figs/ar1_stationary_chronos/ar1_stationary/chronos_ar1_tsne_Layer_00_phis.png}%
%              {figs/ar1_stationary_chronos/ar1_stationary/chronos_ar1_tsne_Layer_03_phis.png}%
%              {figs/ar1_stationary_chronos/ar1_stationary/chronos_ar1_tsne_Layer_06_phis.png}
%   \caption{AR(1) --- Chronos --- t-SNE (Layers 00/03/06)}
% \end{figure}

\begin{figure}[t]
  \centering
  \threeplots{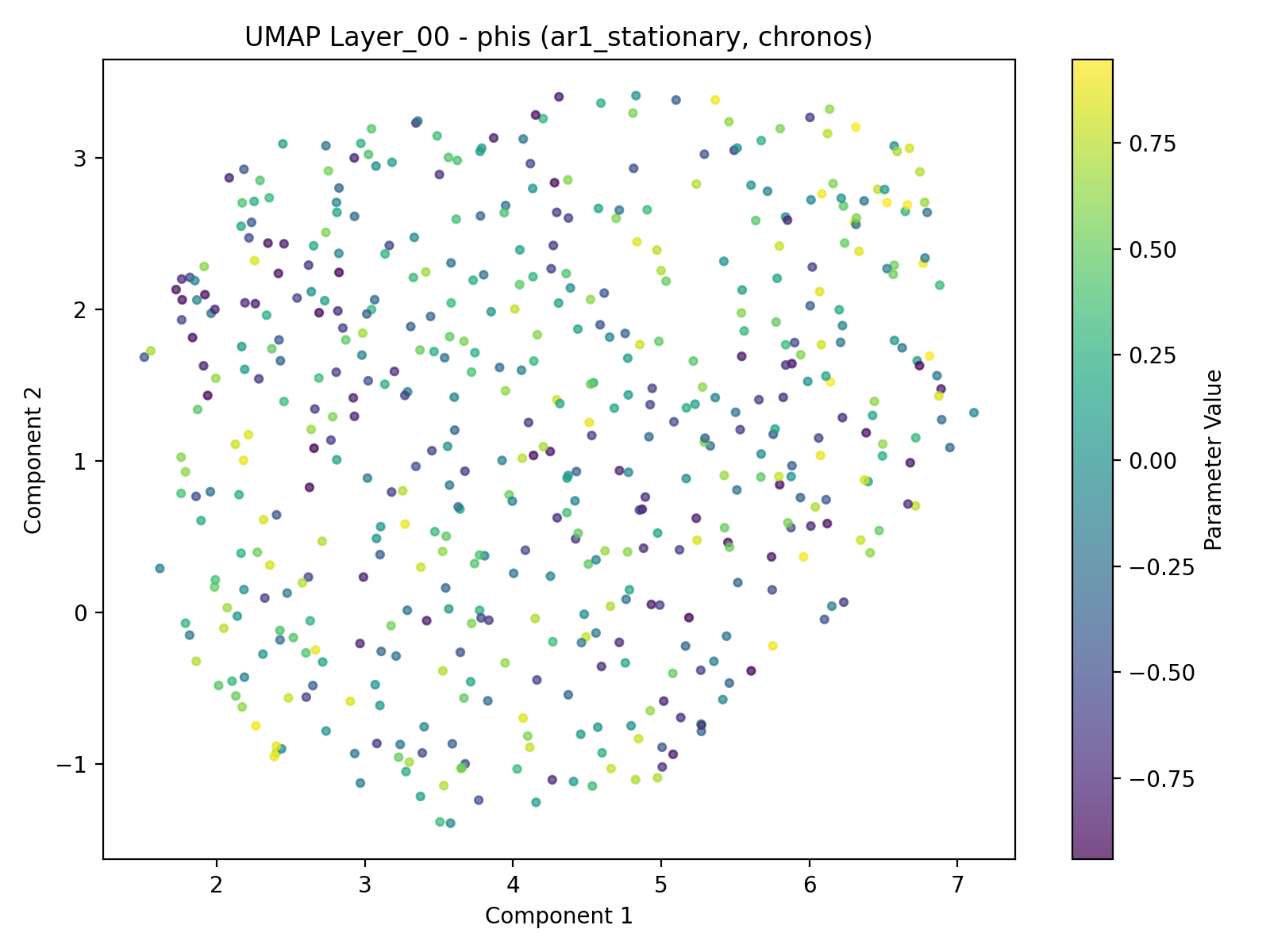}%
             {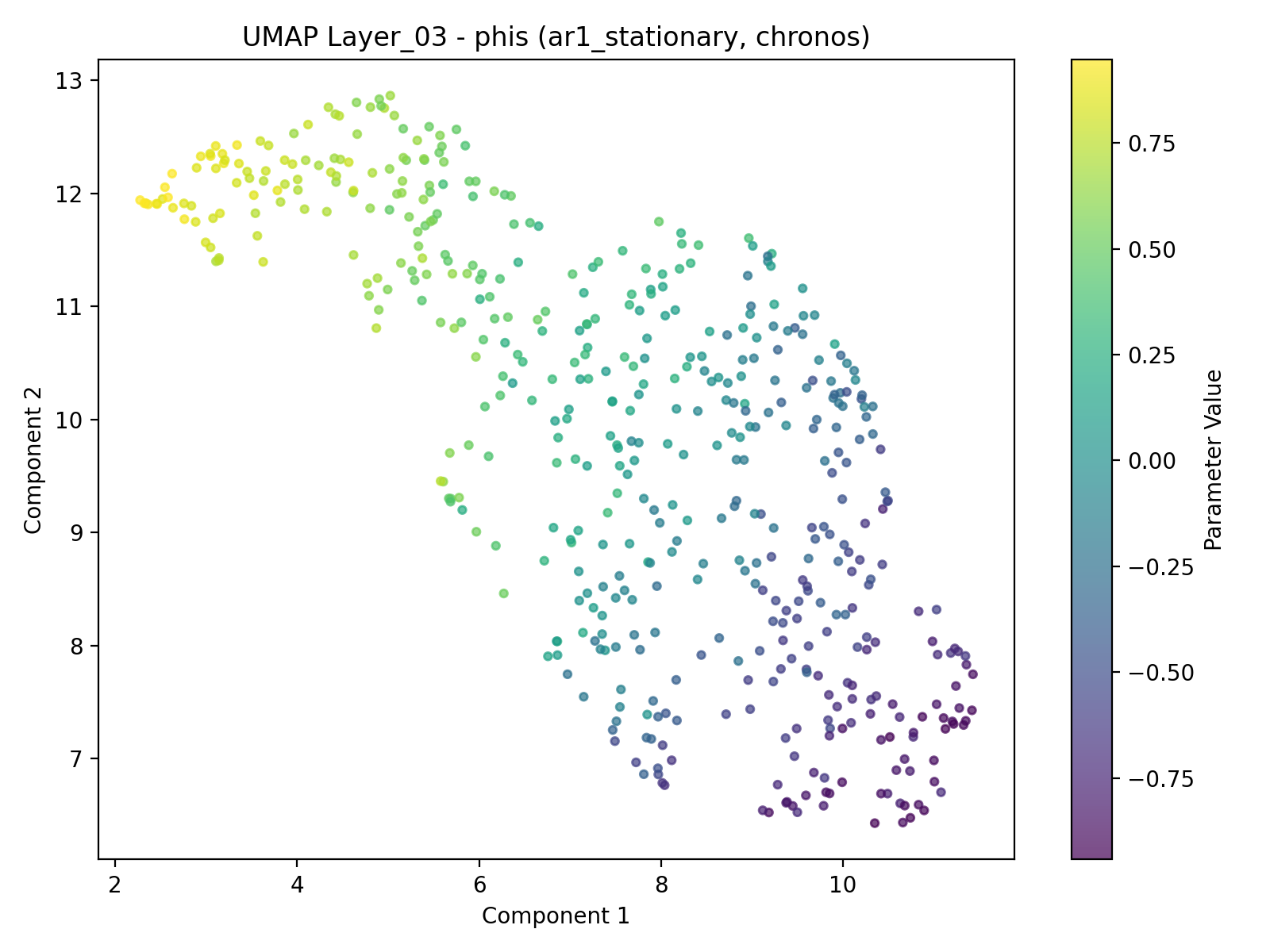}%
             {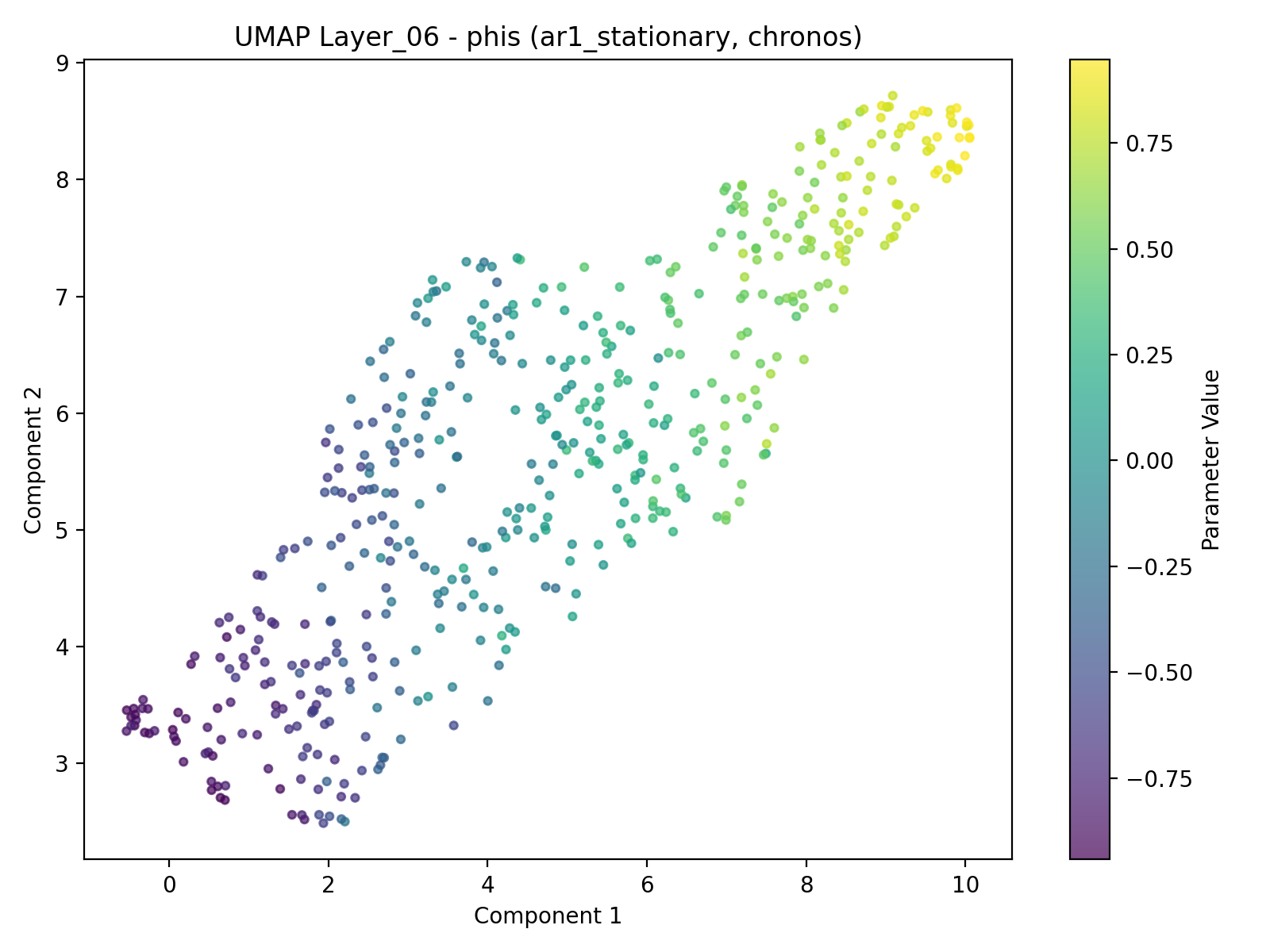}
  \caption{AR(1) --- Chronos --- UMAP (Layers 00/03/06)}
\end{figure}
\newpage
% ---------------- Level Shift ----------------
\subsection{Level Shift}

\paragraph{Moment (parameters: shift, $\tau$).}
\begin{figure}[t]
  \centering
  \threeplots{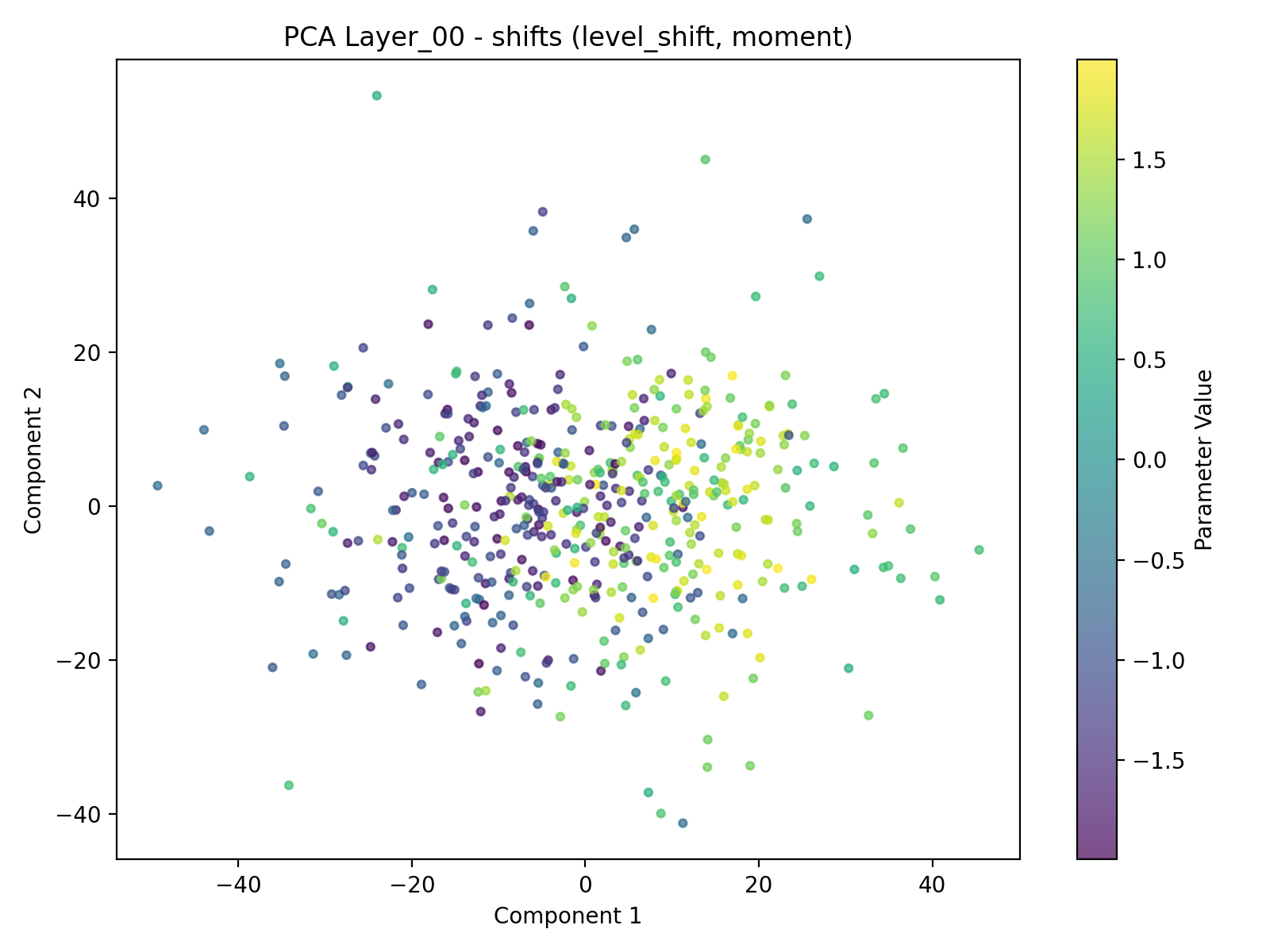}%
             {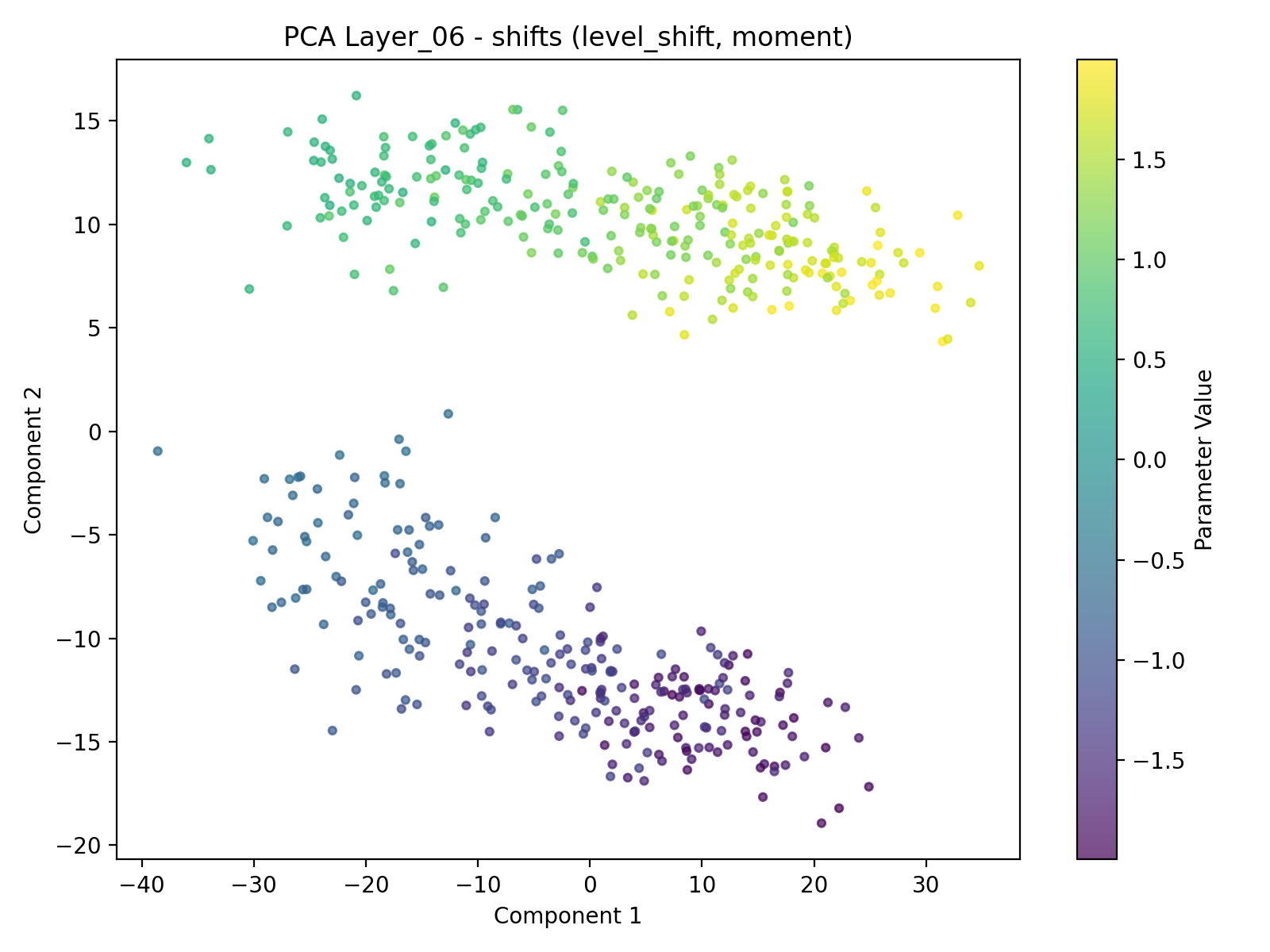}%
             {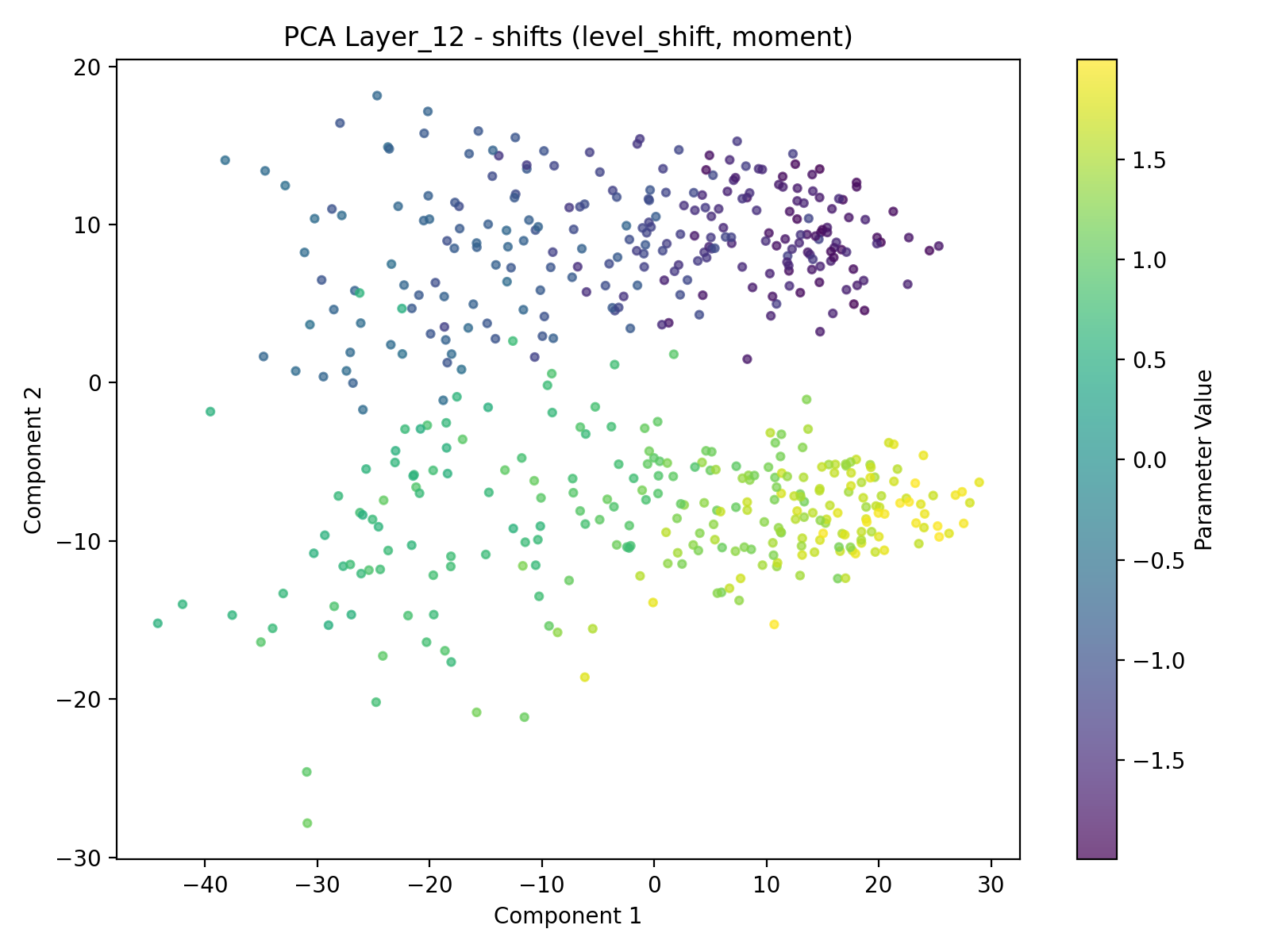}
  \caption{Level Shift --- Moment --- Shift --- PCA (Layers 00/06/12)}
  \label{app:fig15}
\end{figure}

\begin{figure}[t]
  \centering
  \threeplots{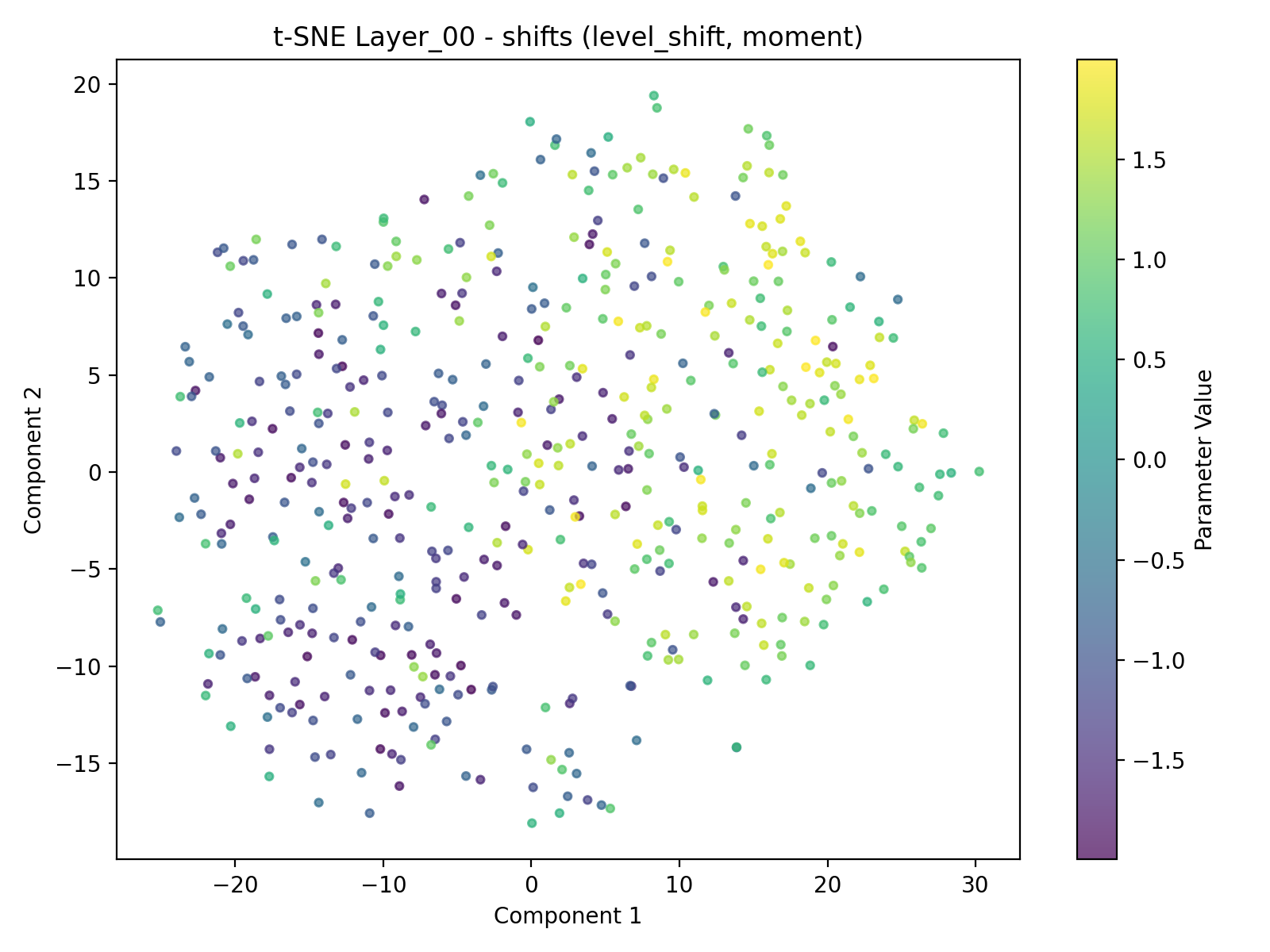}%
             {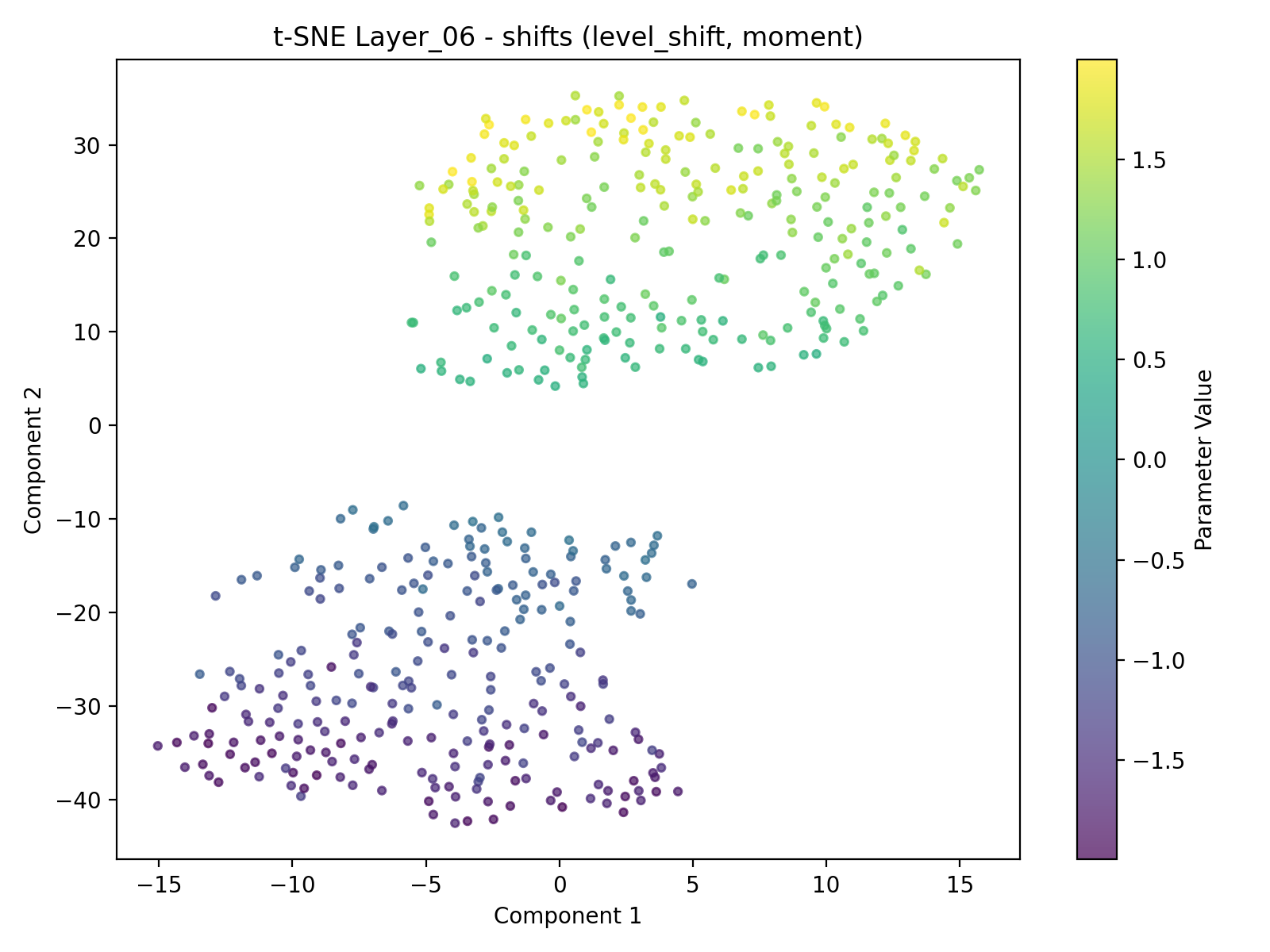}%
             {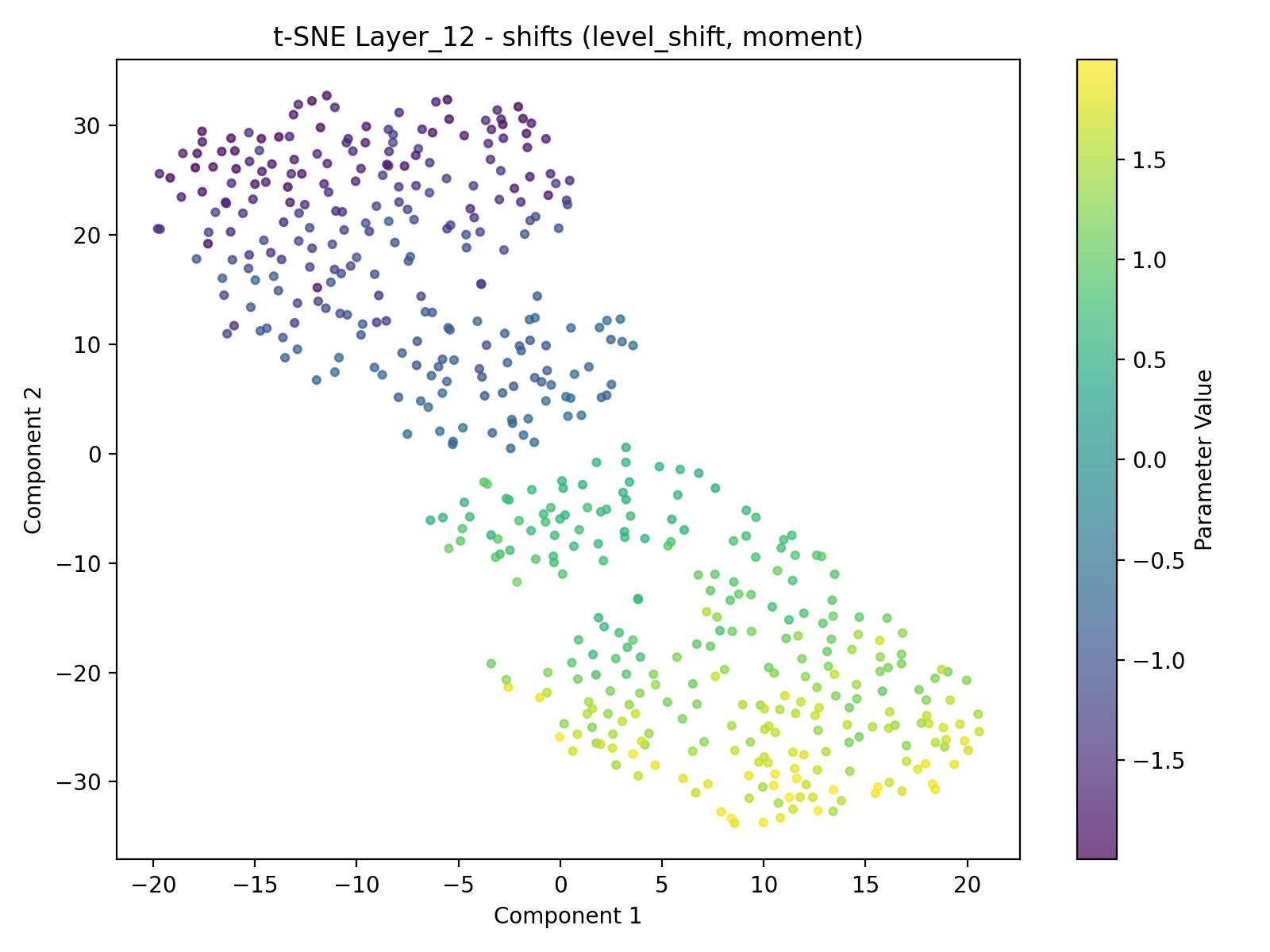}
  \caption{Level Shift --- Moment --- Shift --- t-SNE (Layers 00/06/12)}
  \label{app:fig16}
\end{figure}

\begin{figure}[t]
  \centering
  \threeplots{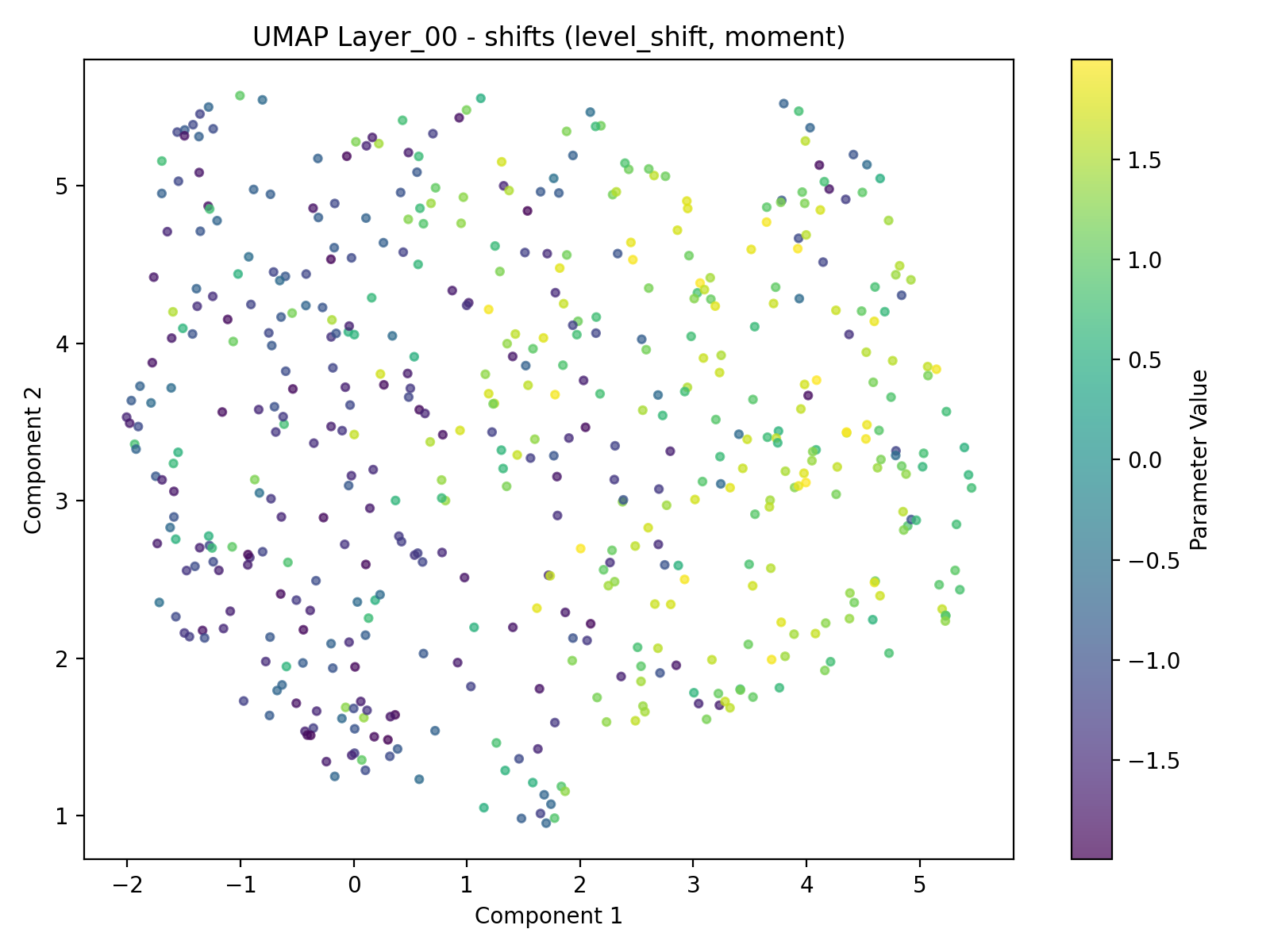}%
             {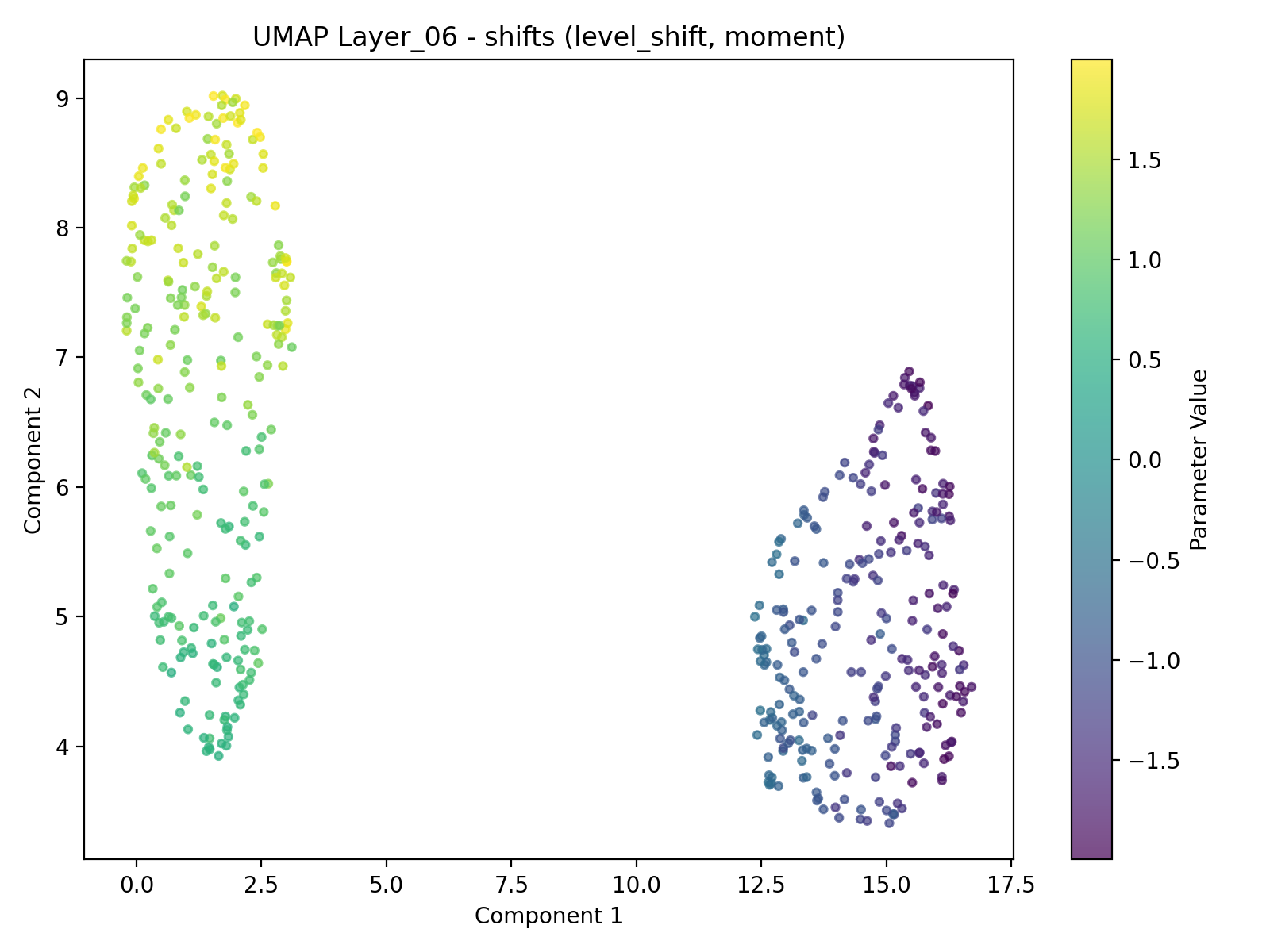}%
             {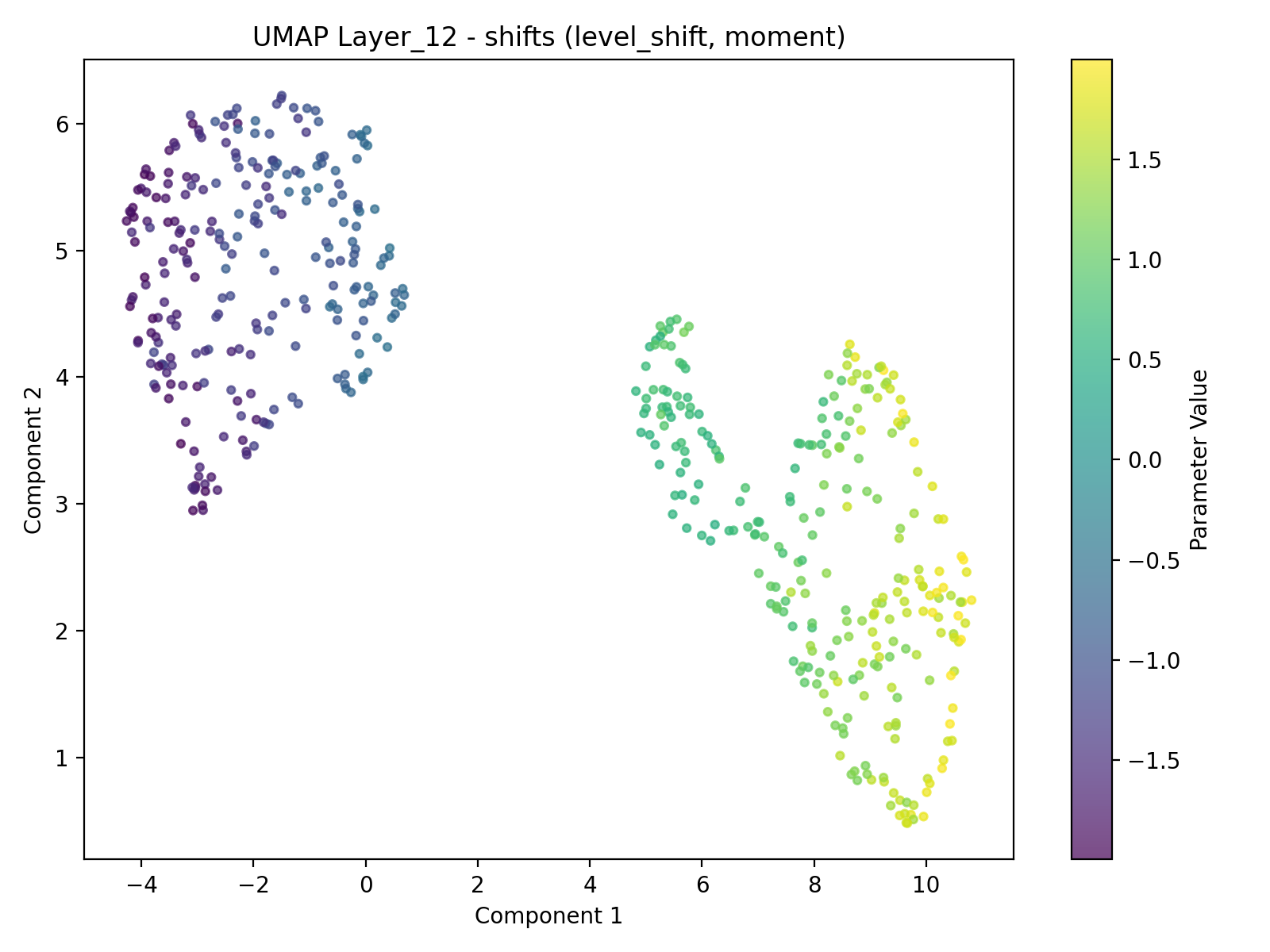}
  \caption{Level Shift --- Moment --- Shift --- UMAP (Layers 00/06/12)}
\end{figure}

\begin{figure}[t]
  \centering
  \threeplots{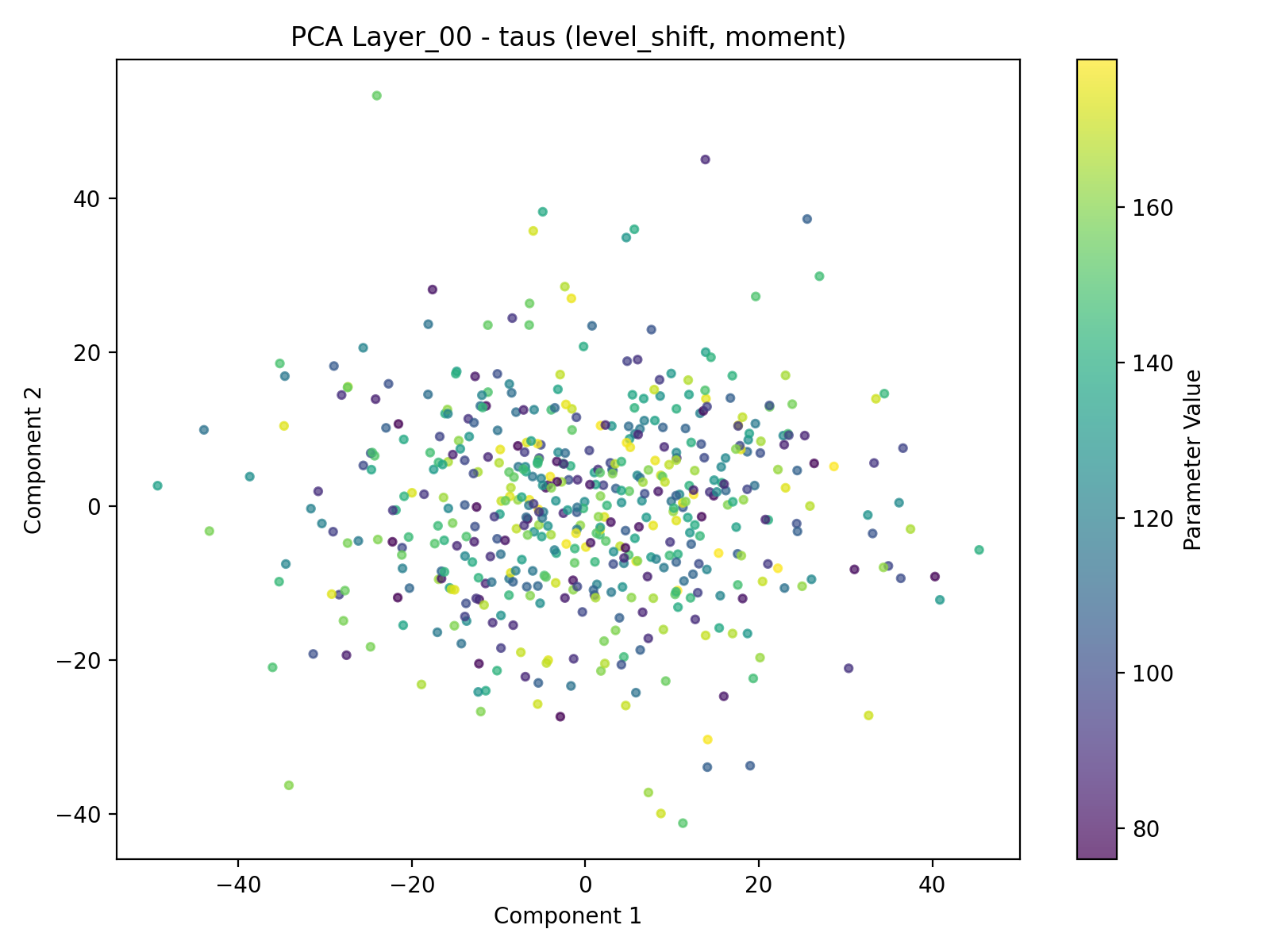}%
             {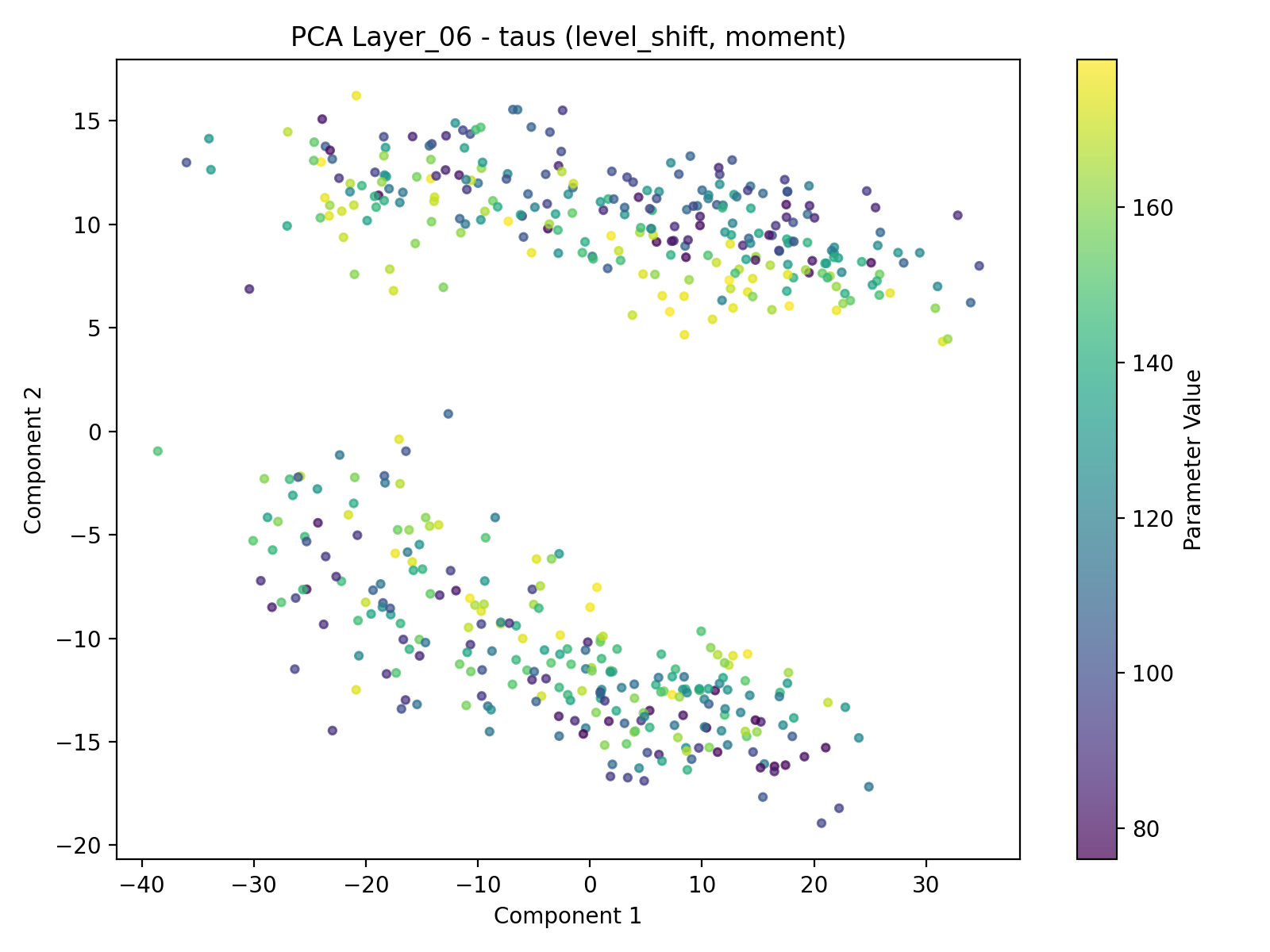}%
             {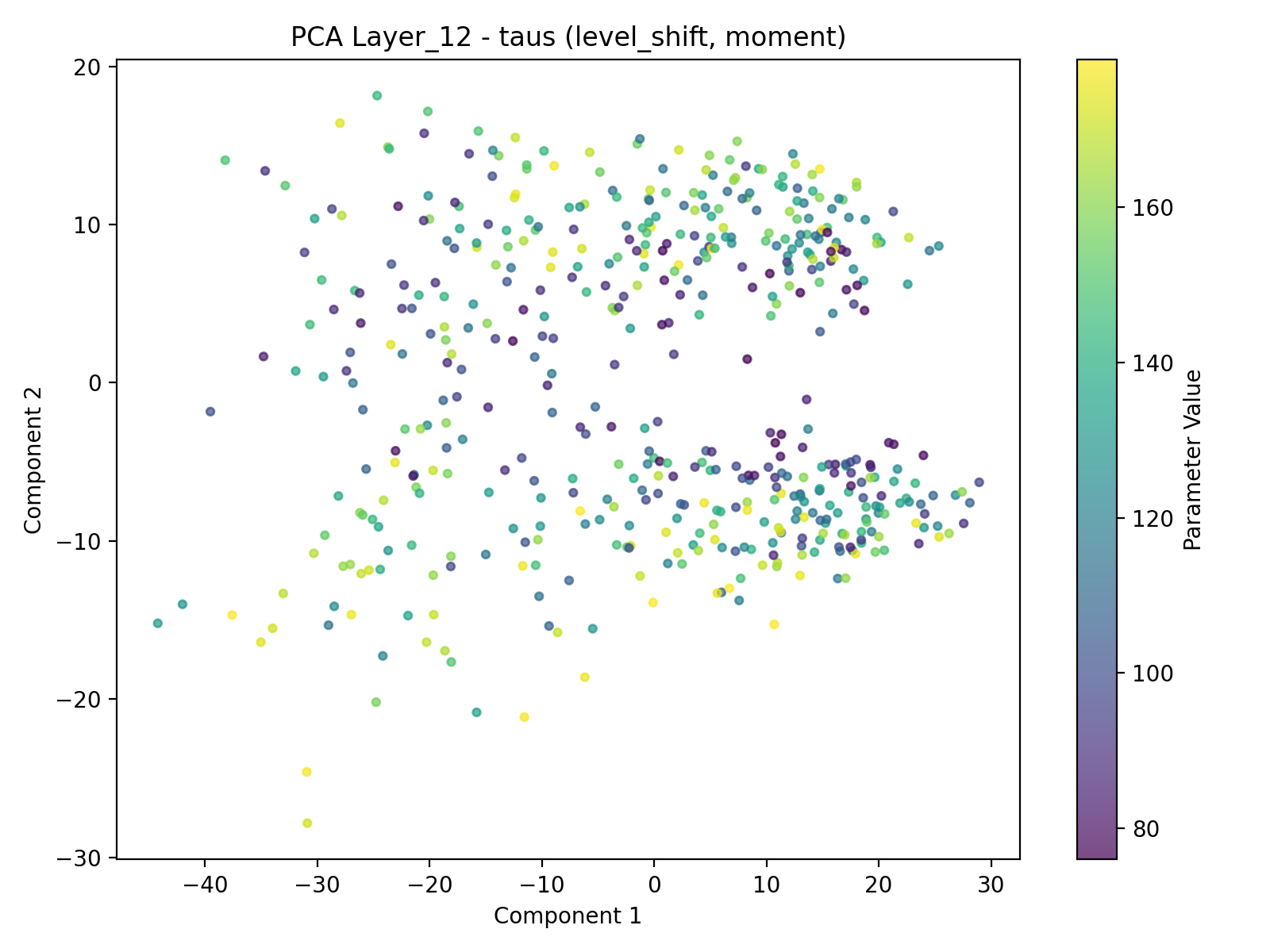}
  \caption{Level Shift --- Moment --- $\tau$ --- PCA (Layers 00/06/12)}
\end{figure}

\begin{figure}[t]
  \centering
  \threeplots{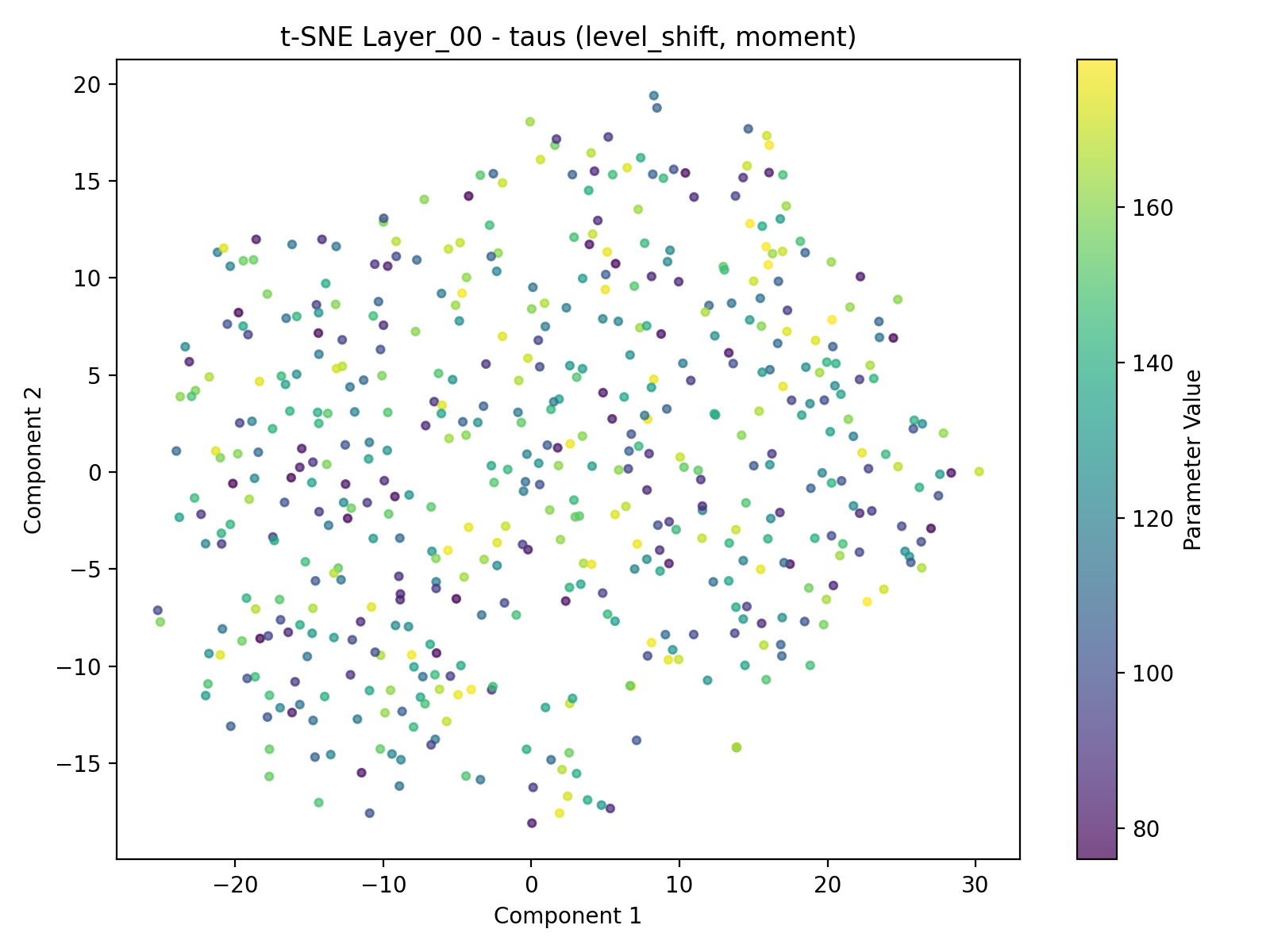}%
             {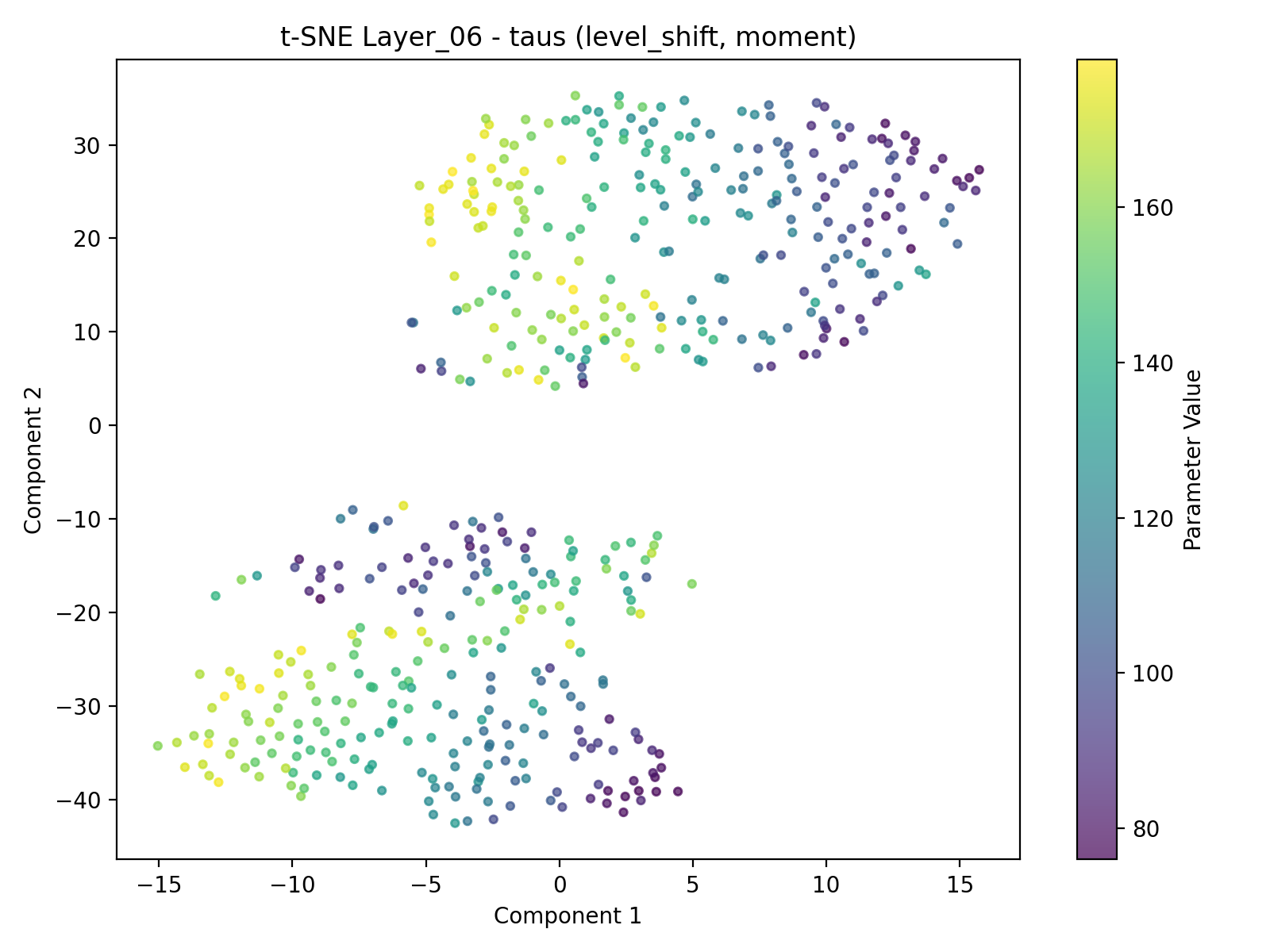}%
             {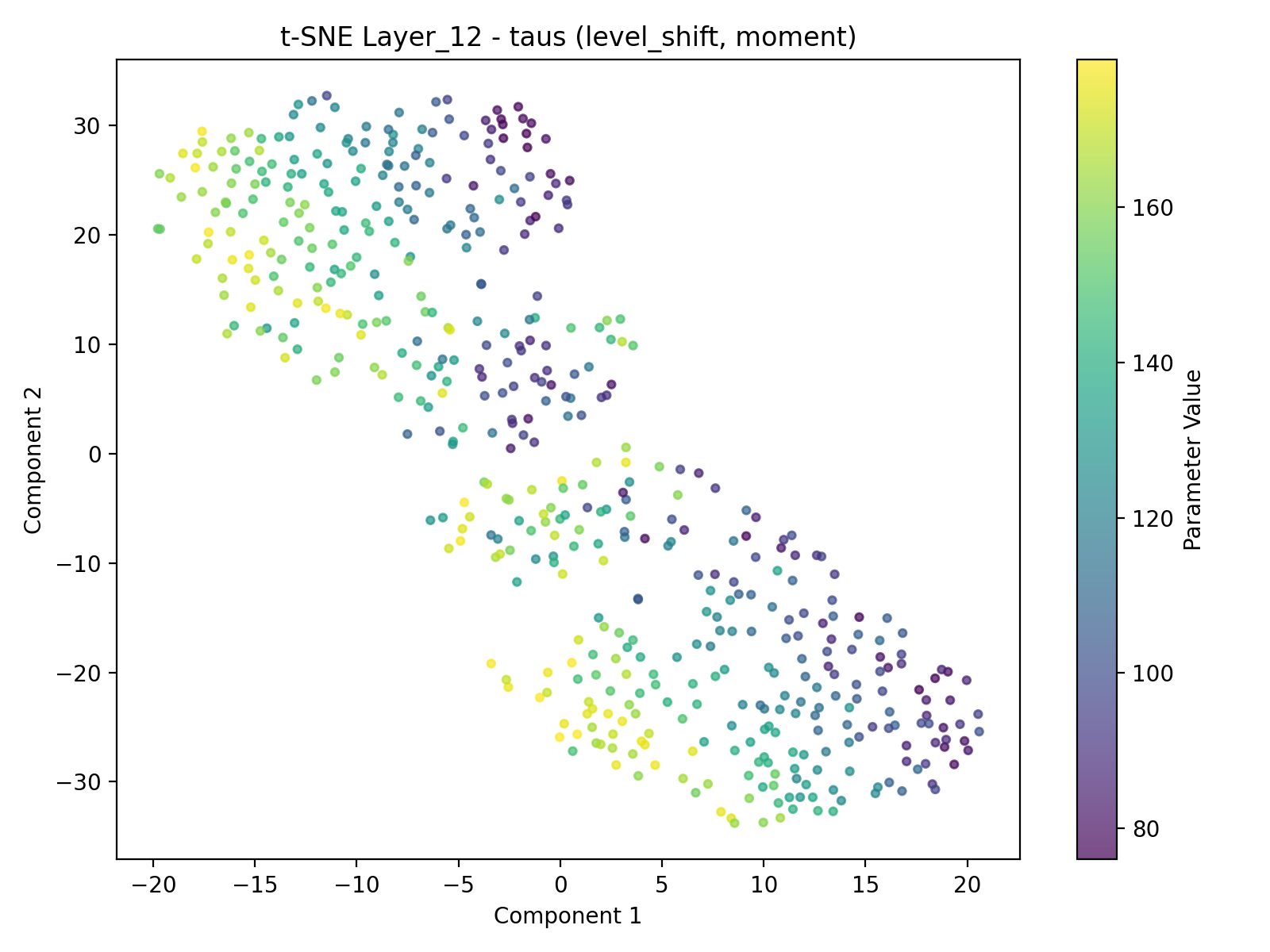}
  \caption{Level Shift --- Moment --- $\tau$ --- t-SNE (Layers 00/06/12)}
\end{figure}

\begin{figure}[t]
  \centering
  \threeplots{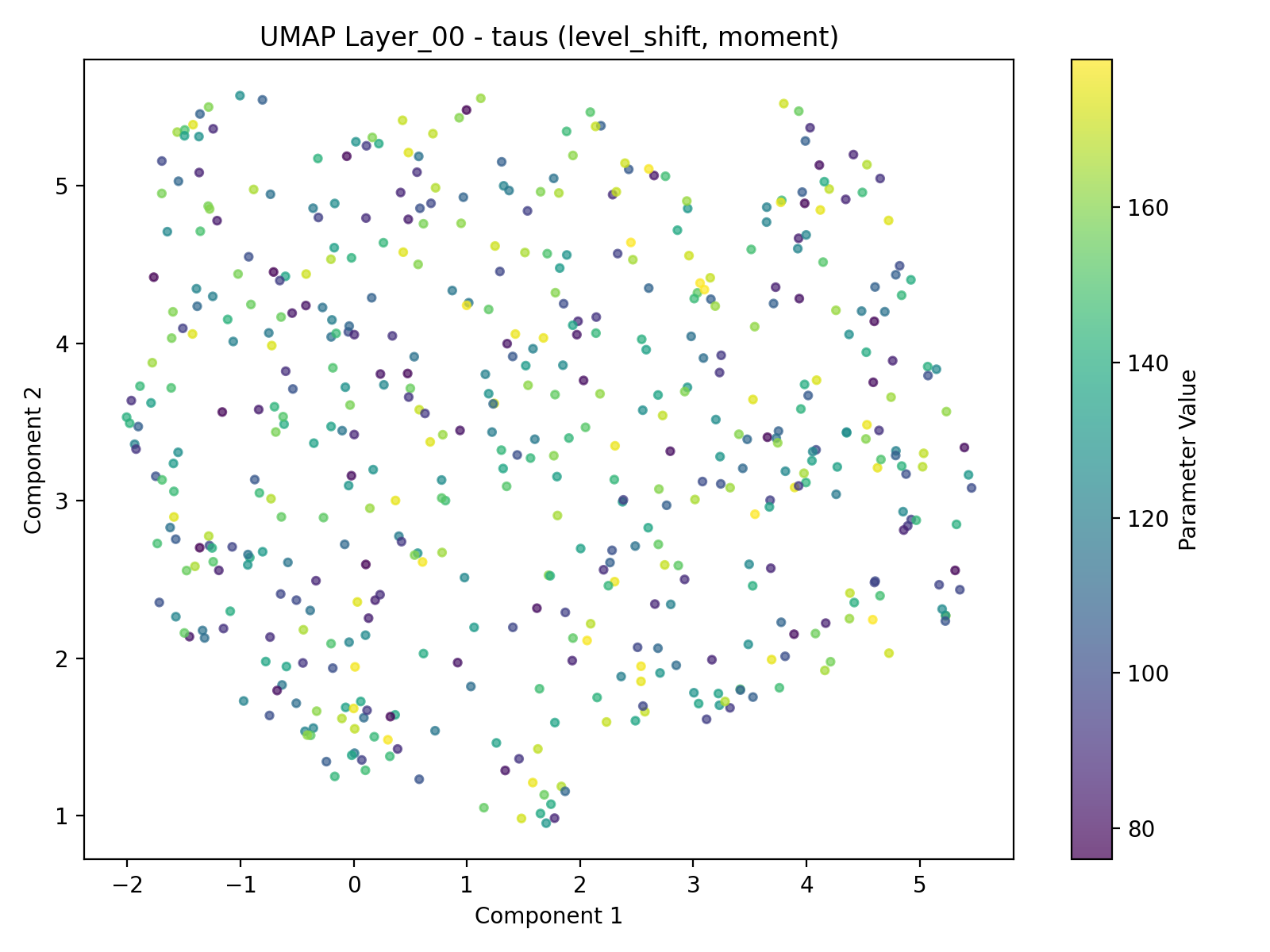}%
             {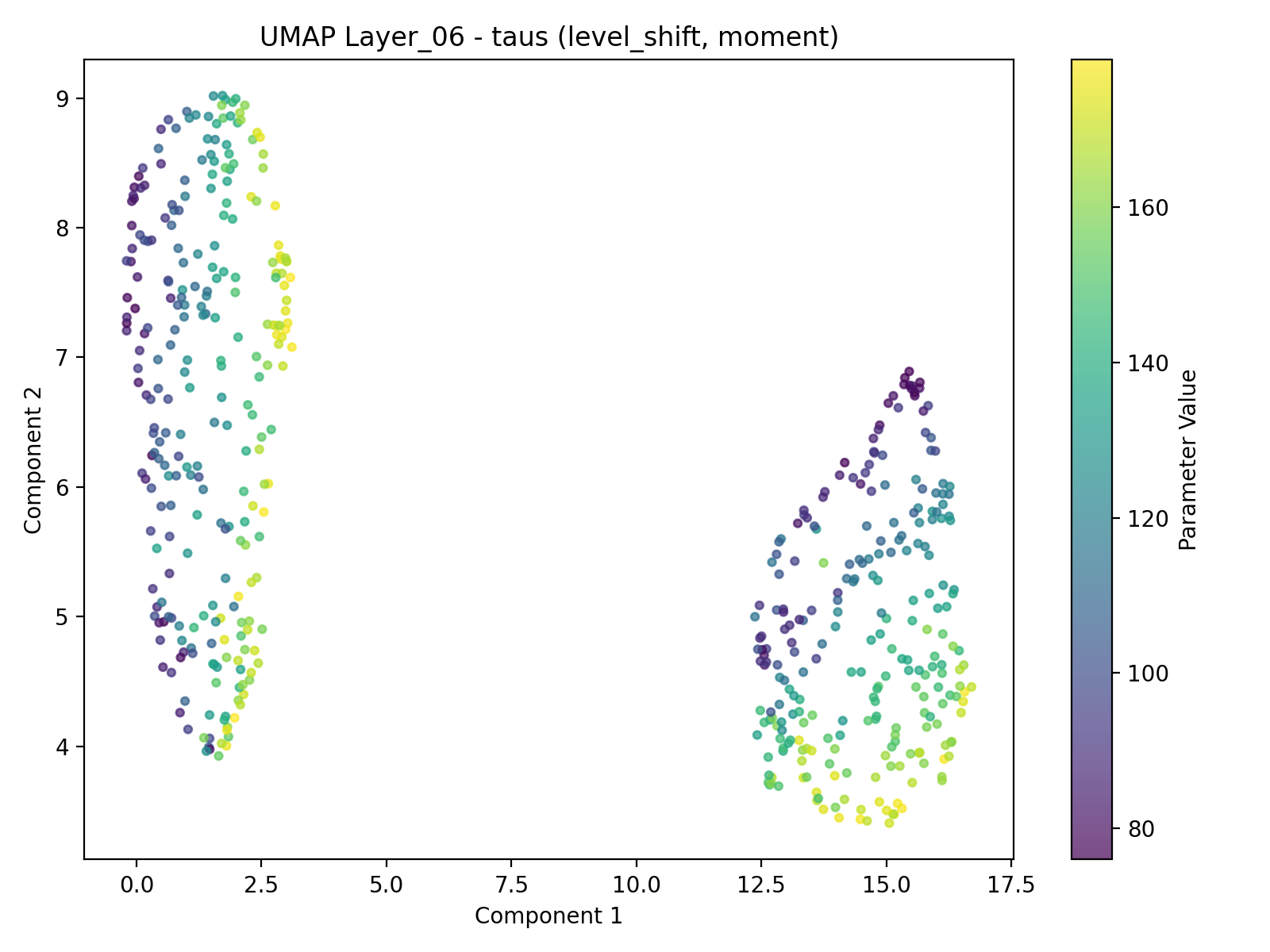}%
             {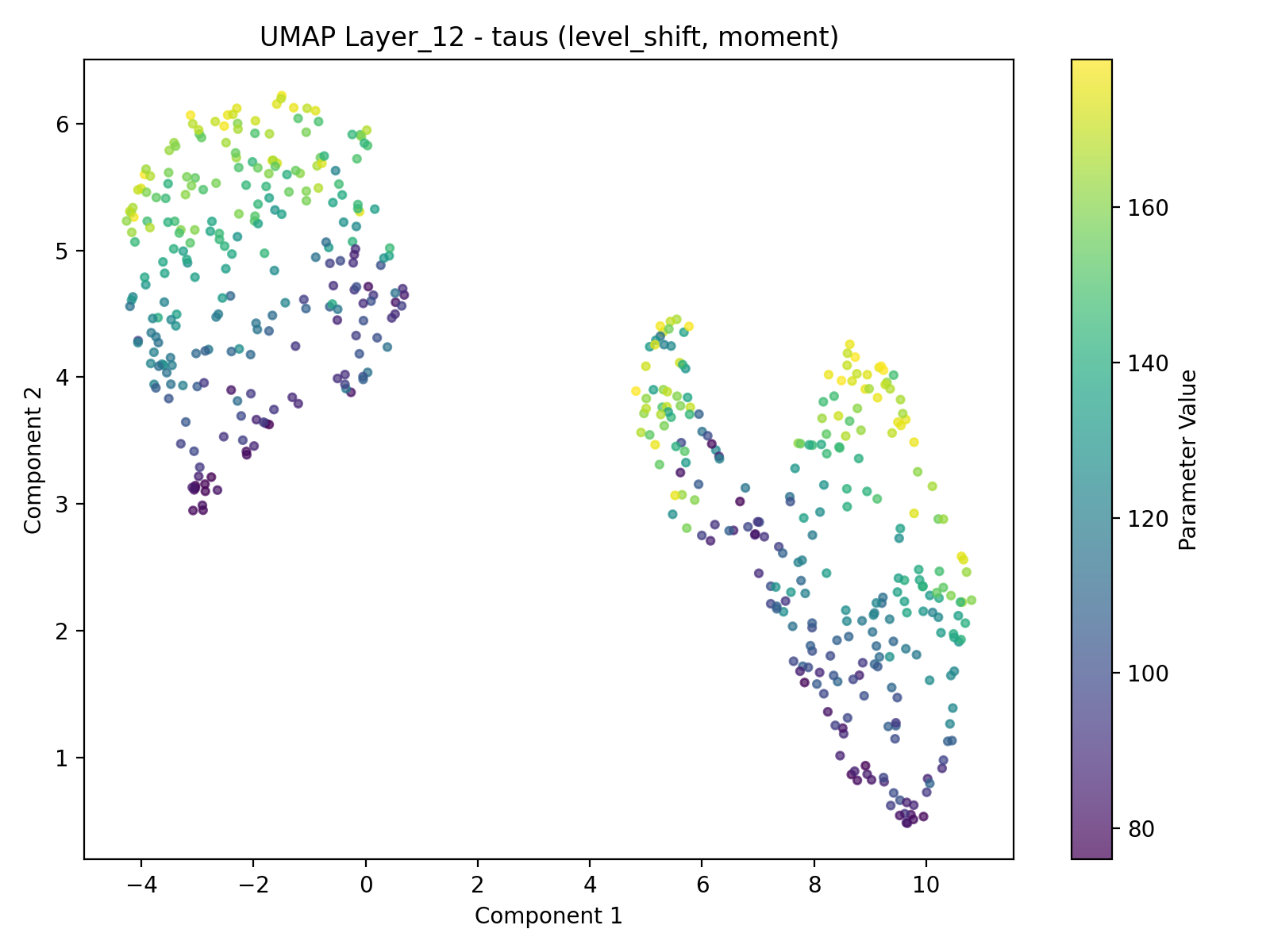}
  \caption{Level Shift --- Moment --- $\tau$ --- UMAP (Layers 00/06/12)}
\end{figure}
\newpage
\paragraph{Chronos (parameters: shift, $\tau$).}
\begin{figure}[t]
  \centering
  \threeplots{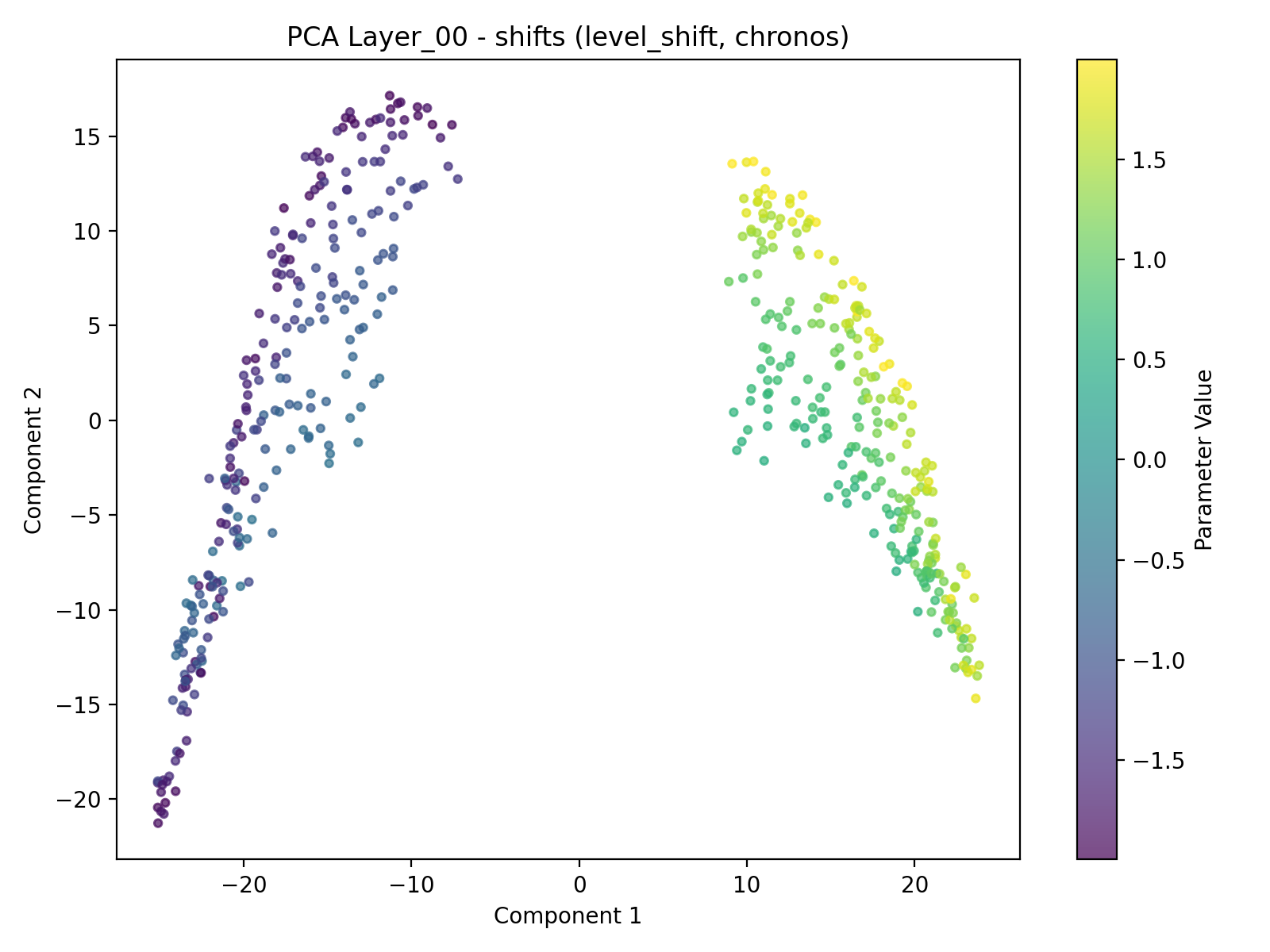}%
             {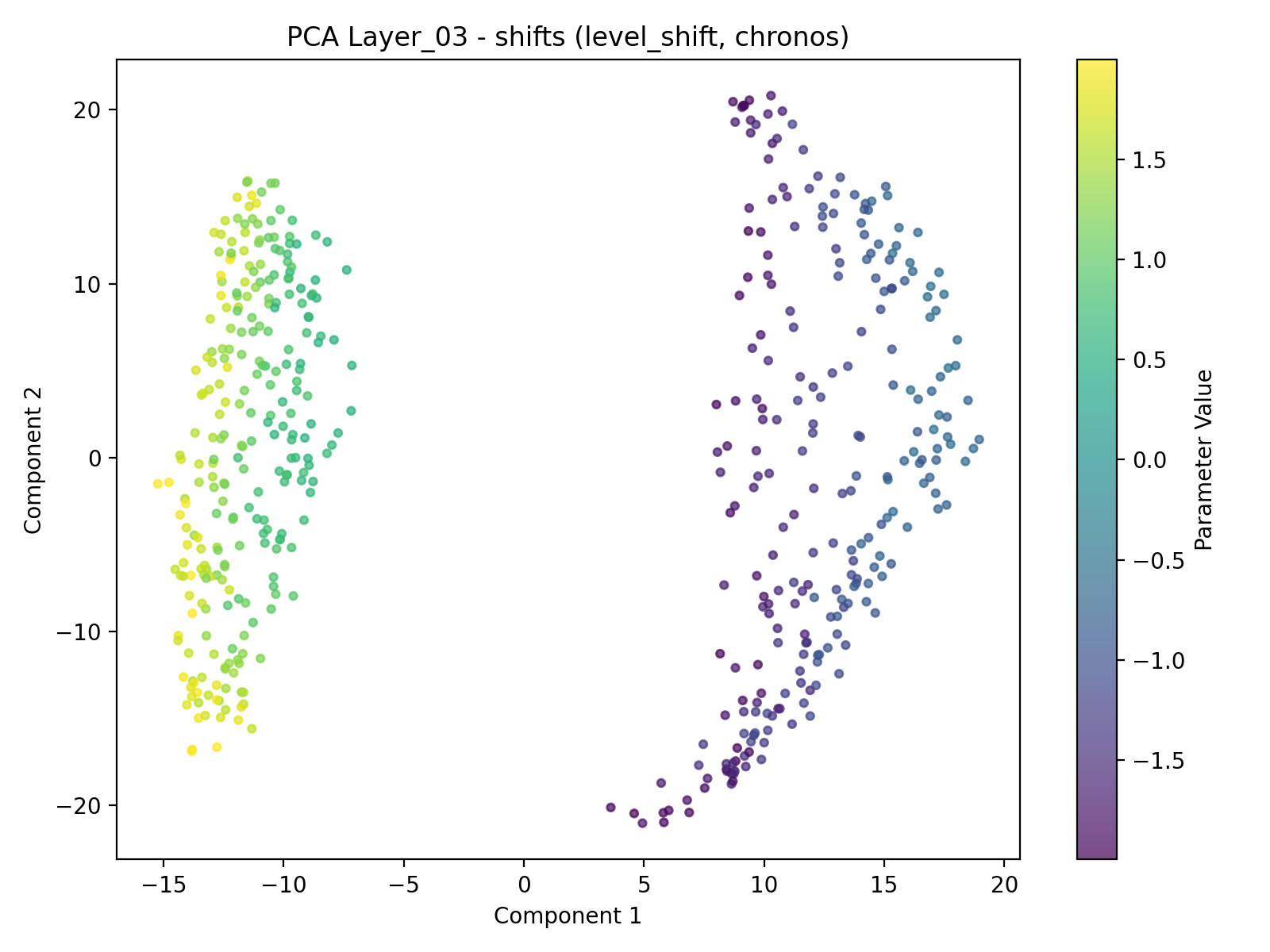}%
             {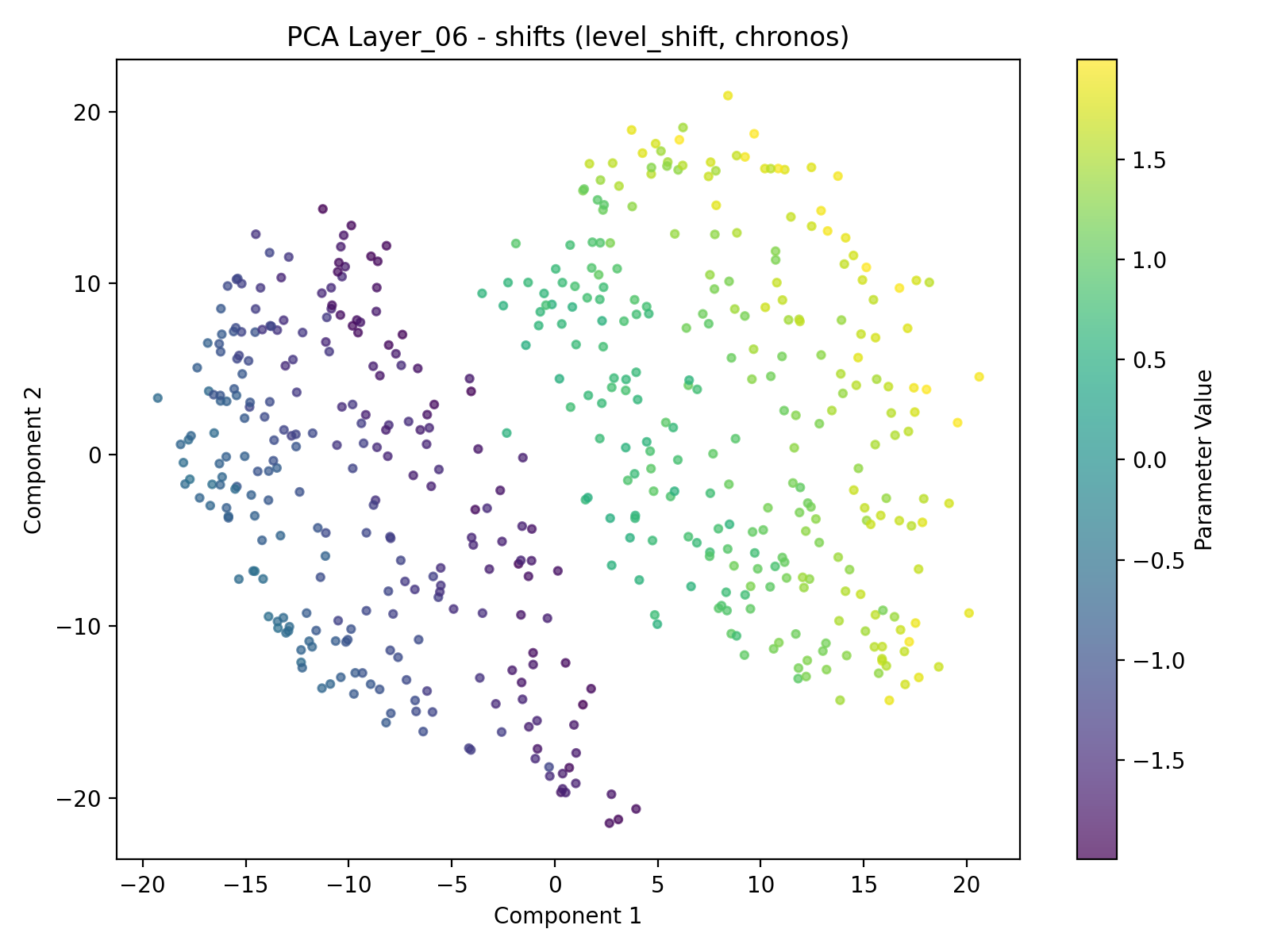}
  \caption{Level Shift --- Chronos --- Shift --- PCA (Layers 00/03/06)}
\end{figure}

\begin{figure}[t]
  \centering
  \threeplots{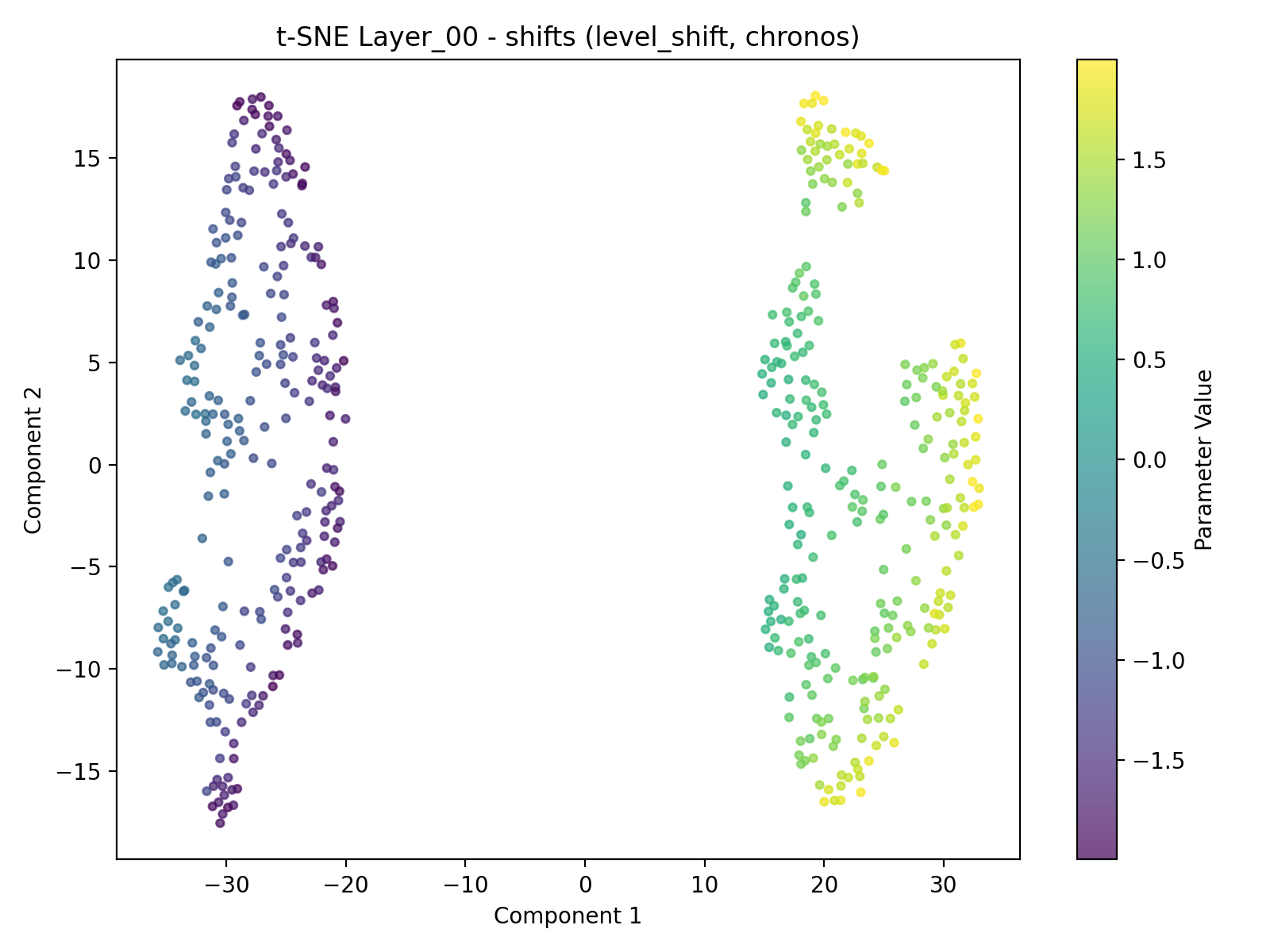}%
             {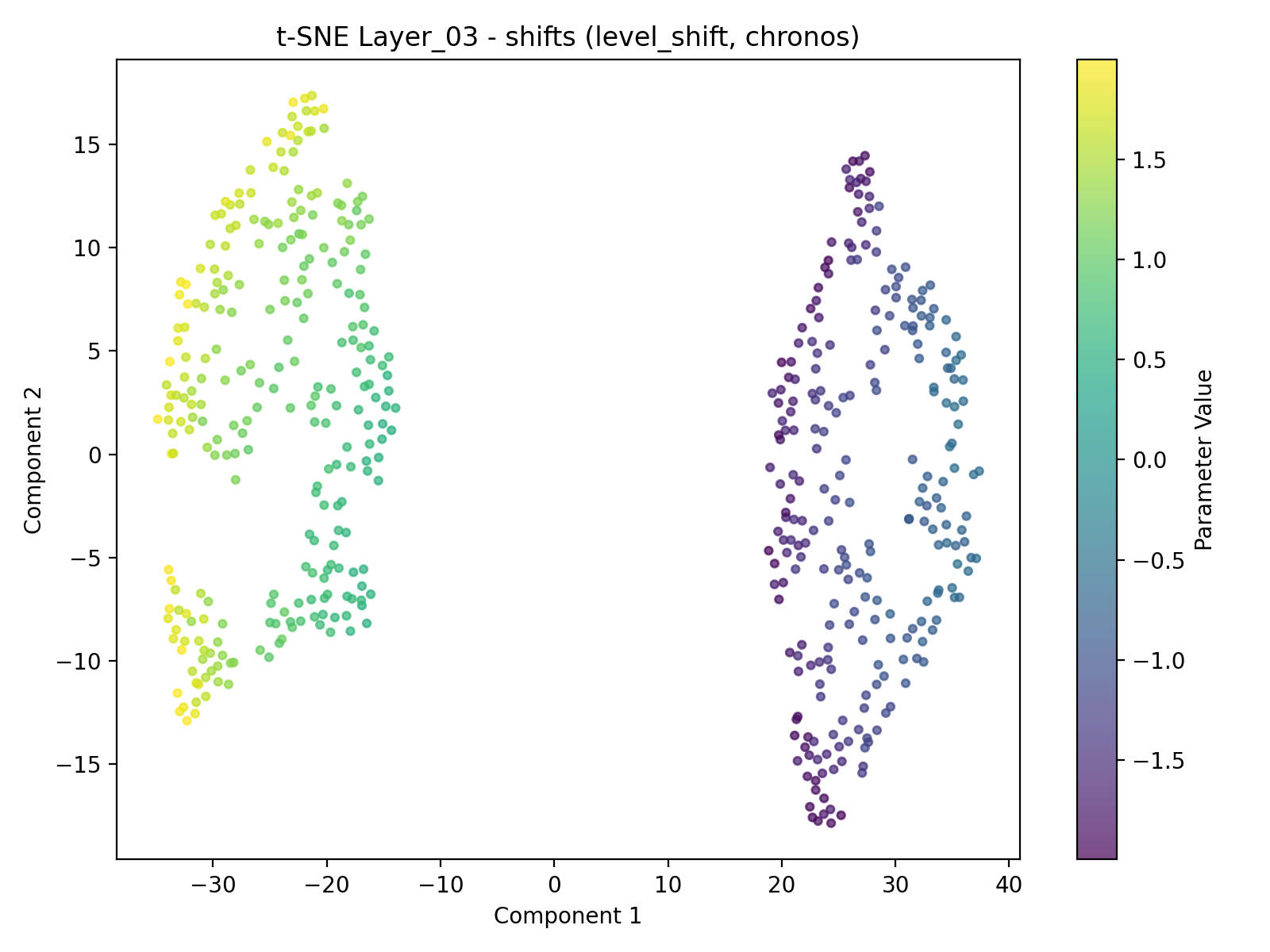}%
             {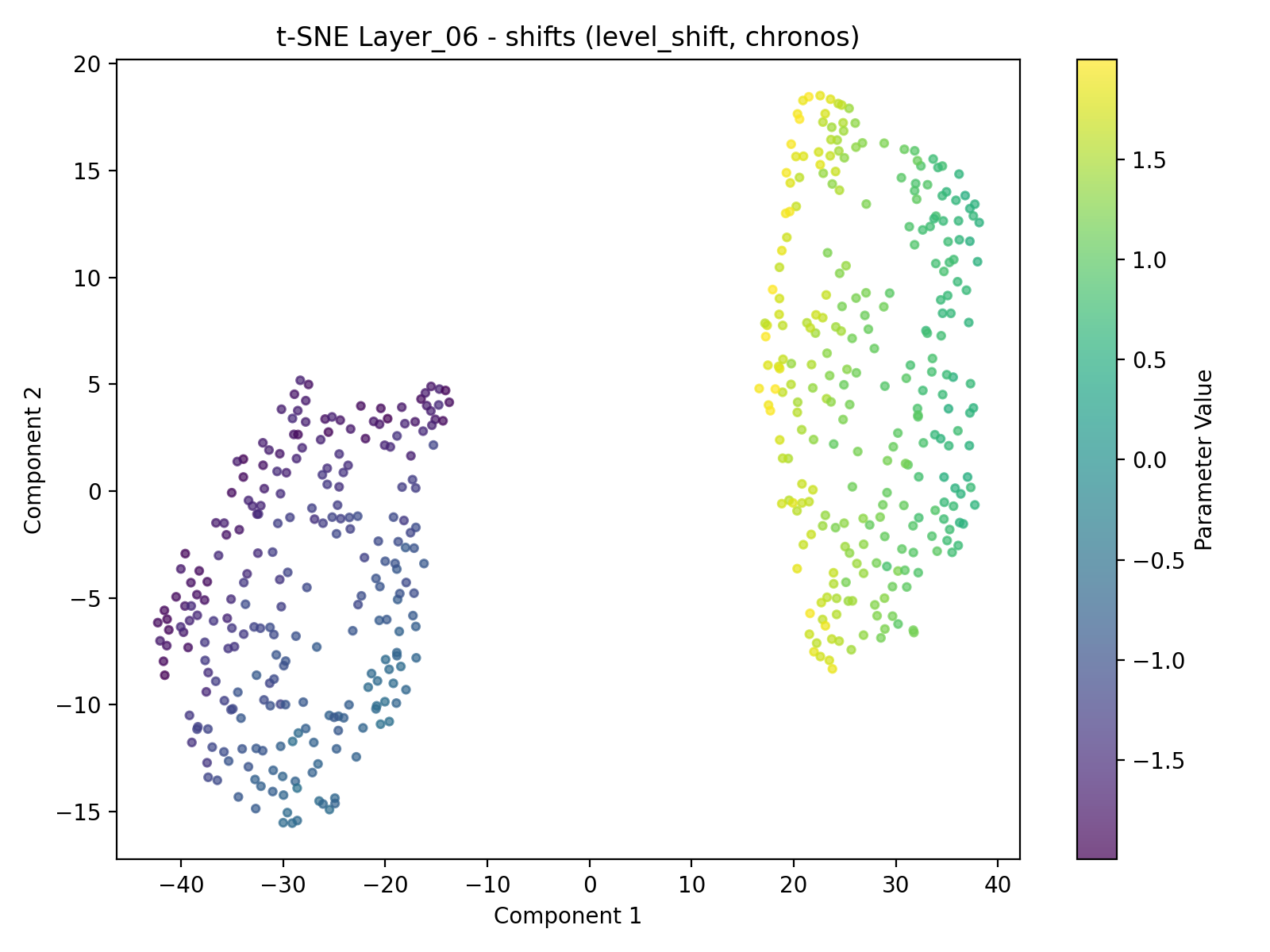}
  \caption{Level Shift --- Chronos --- Shift --- t-SNE (Layers 00/03/06)}
\end{figure}

\begin{figure}[t]
  \centering
  \threeplots{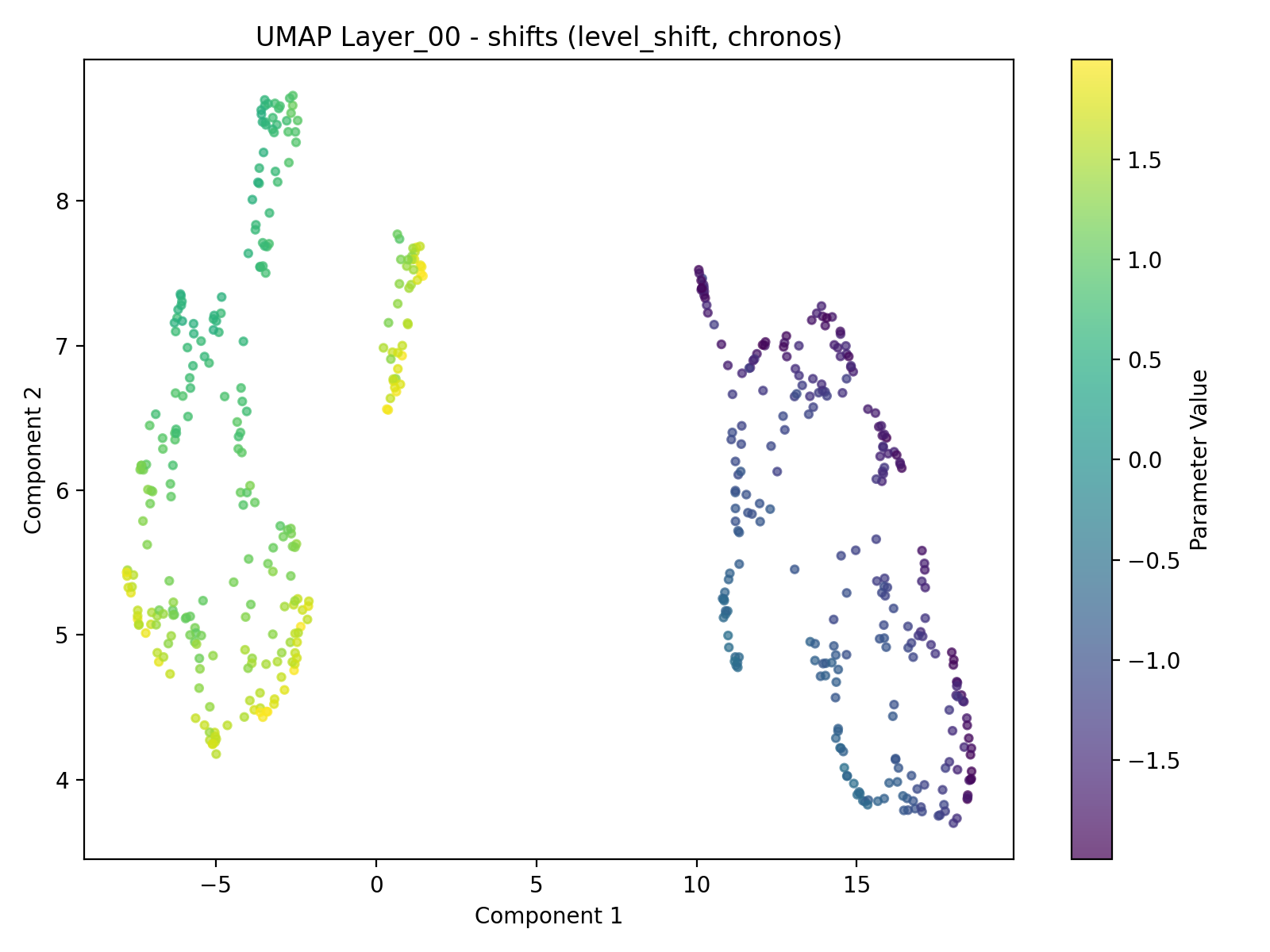}%
             {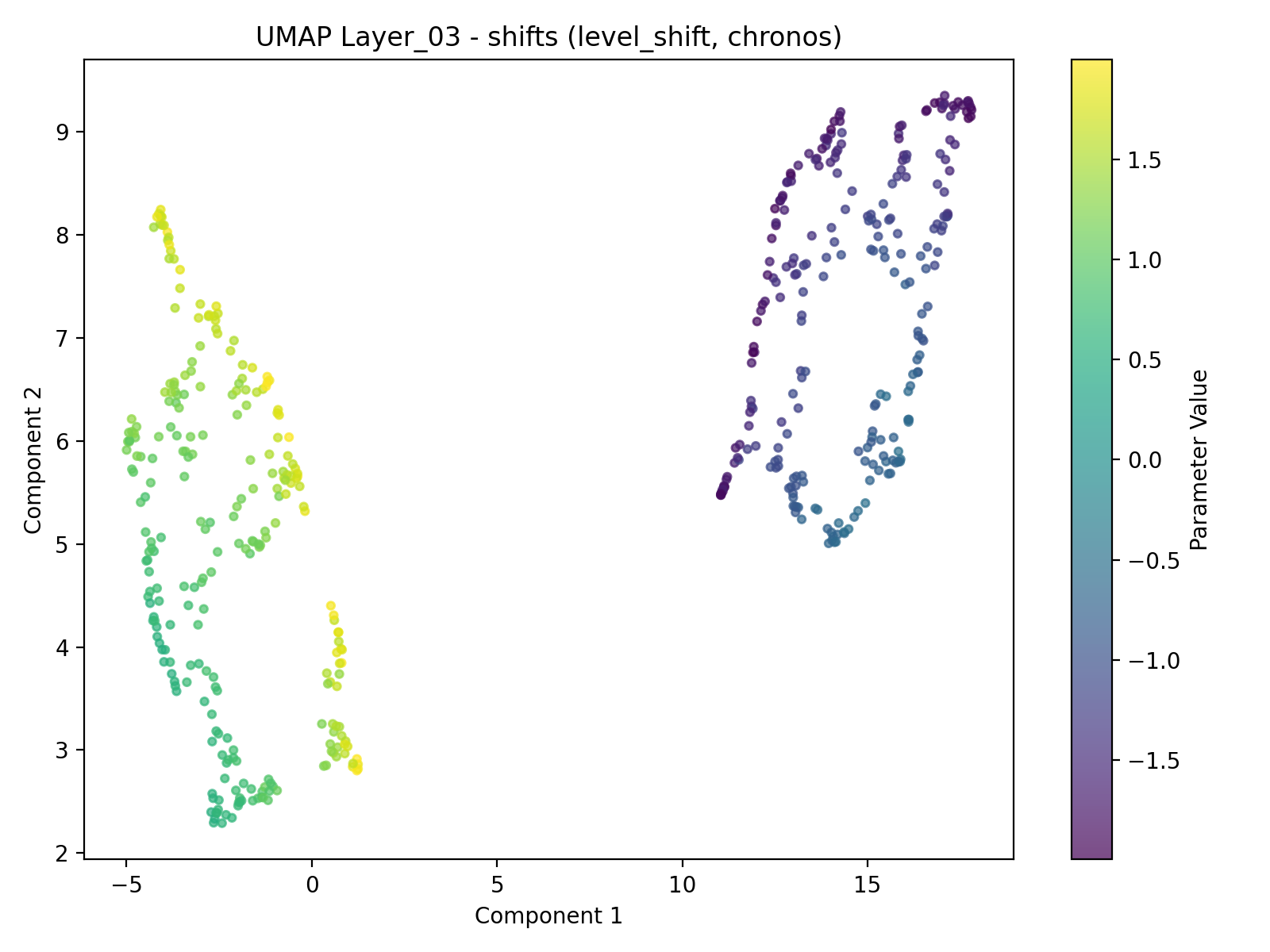}%
             {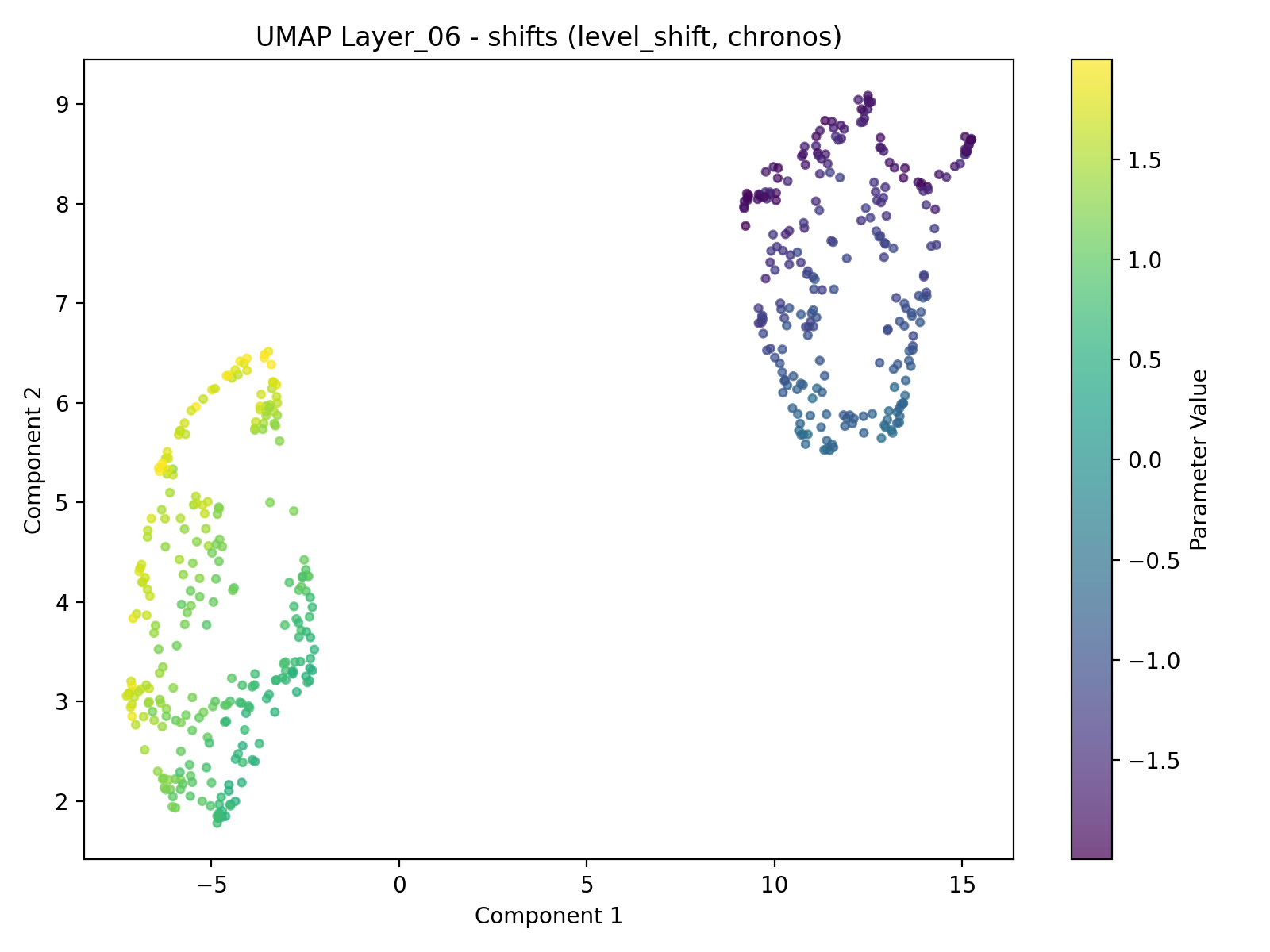}
  \caption{Level Shift --- Chronos --- Shift --- UMAP (Layers 00/03/06)}
\end{figure}

\begin{figure}[t]
  \centering
  \threeplots{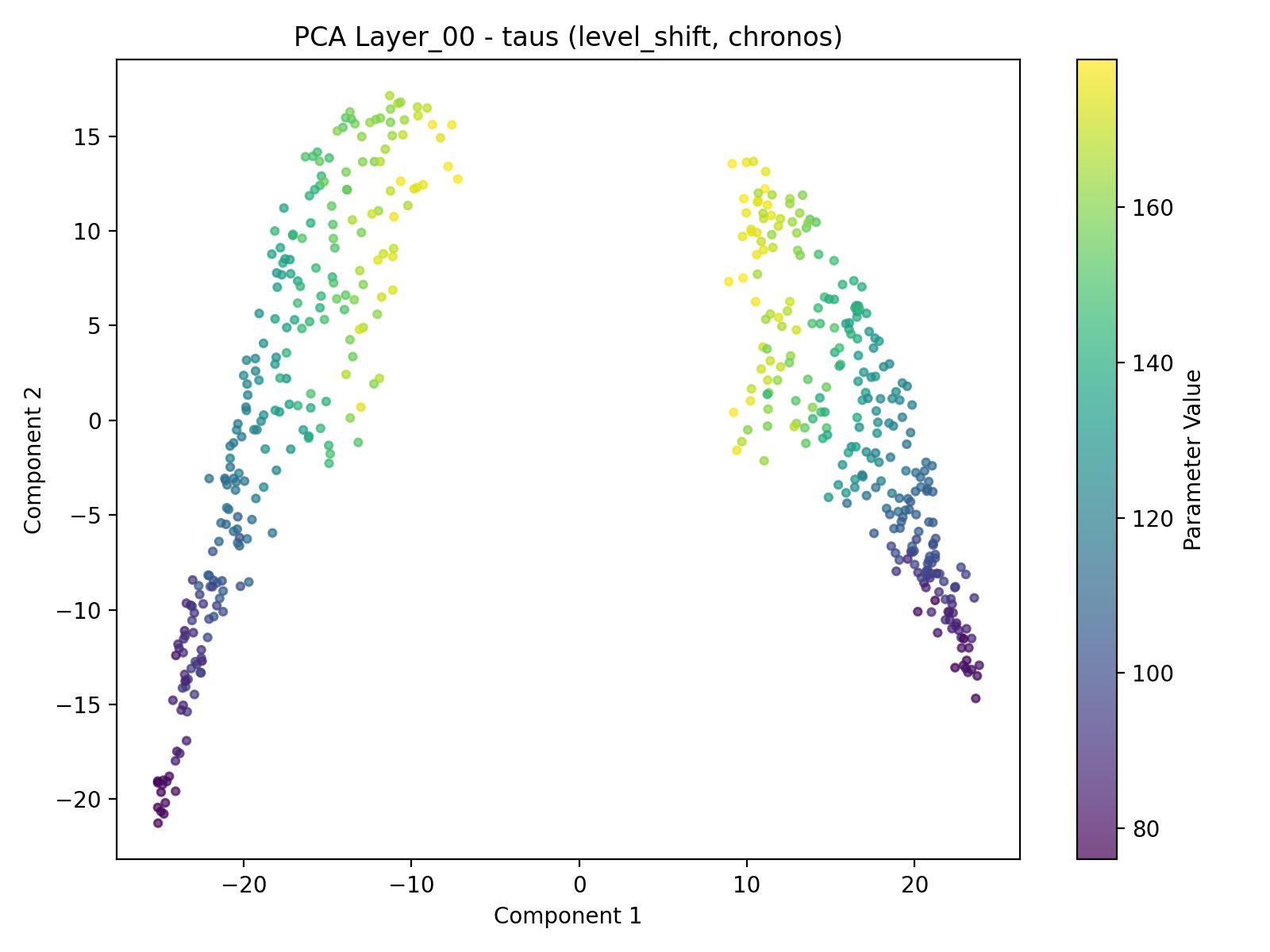}%
             {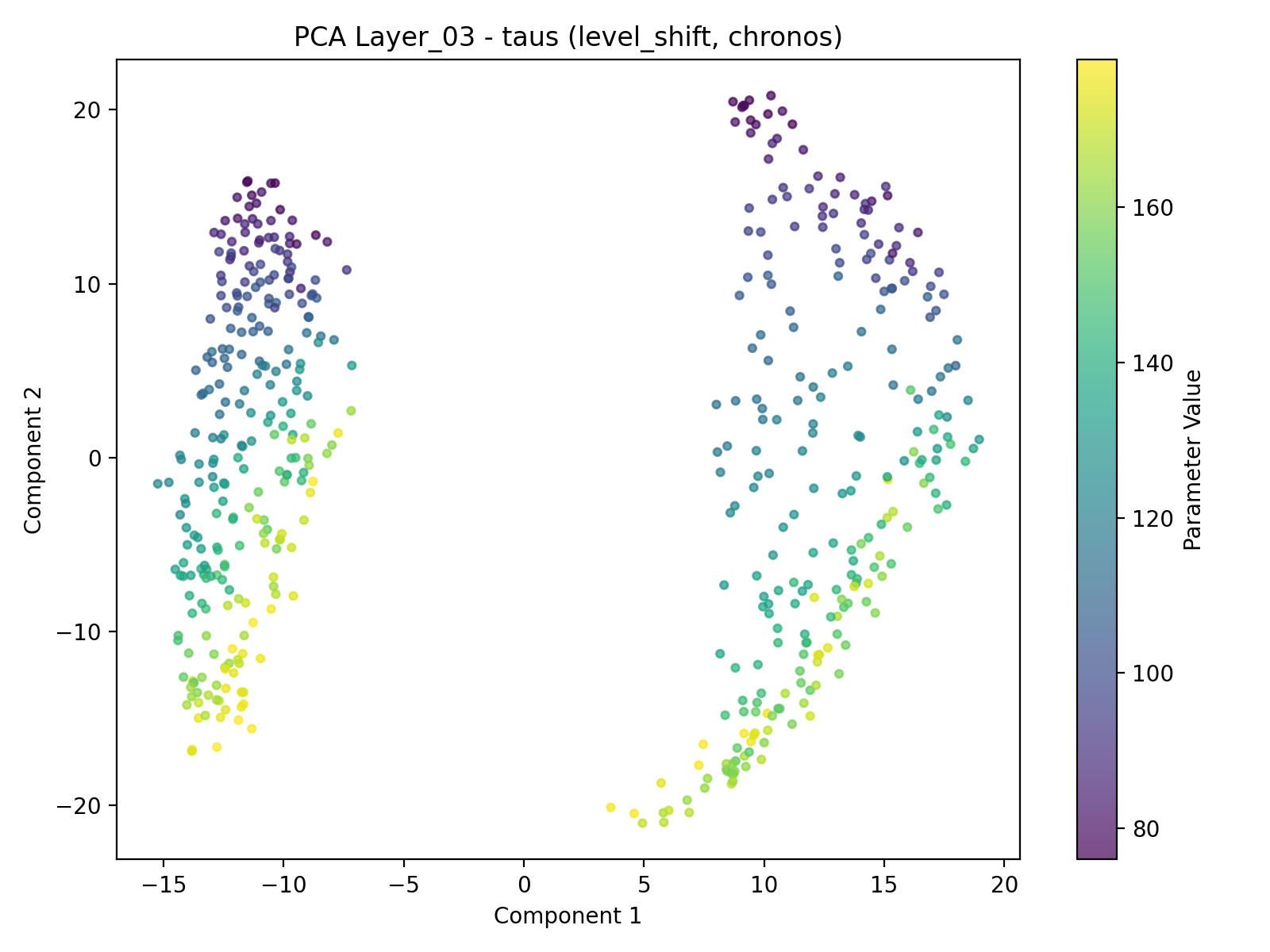}%
             {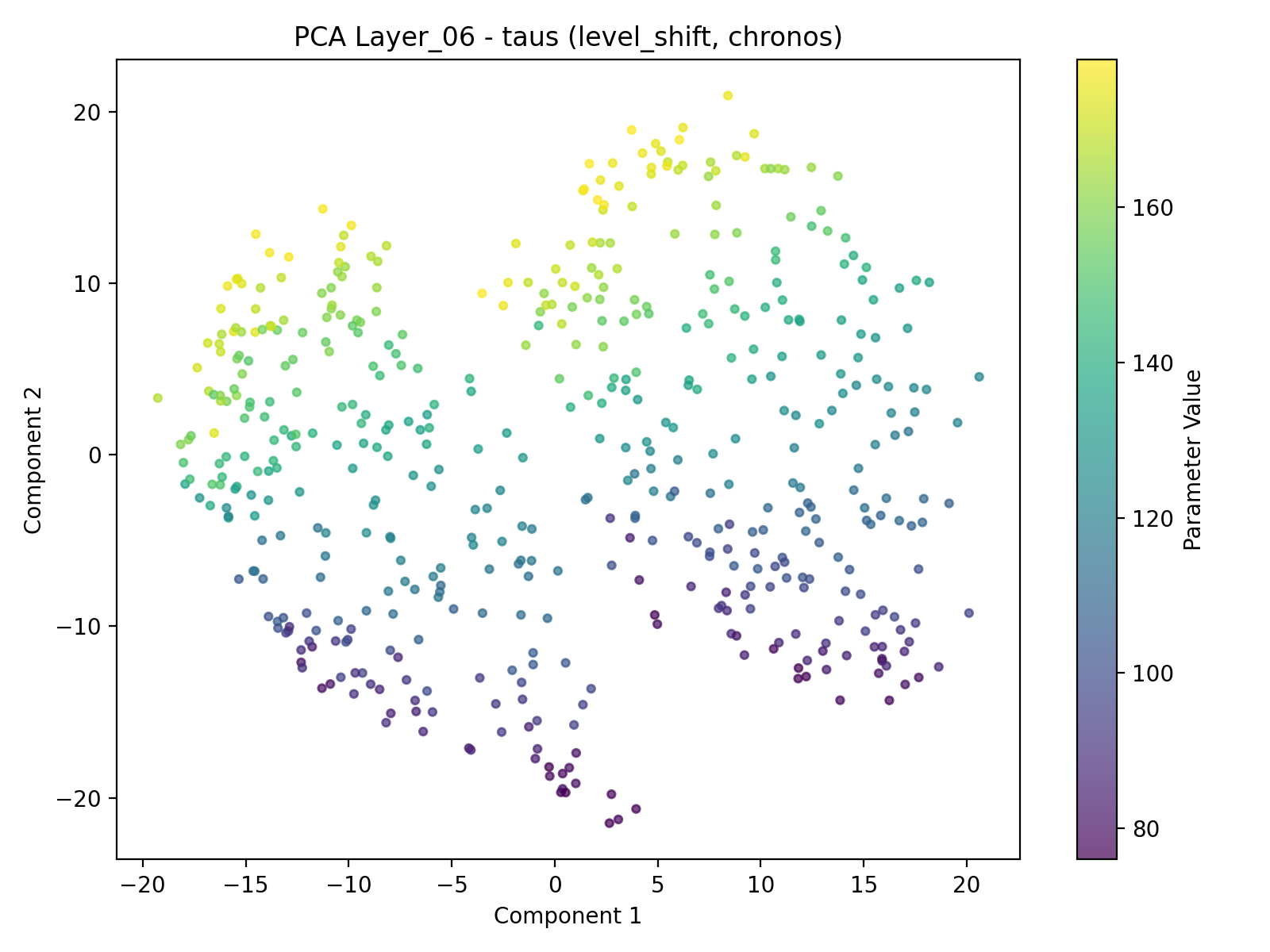}
  \caption{Level Shift --- Chronos --- $\tau$ --- PCA (Layers 00/03/06)}
\end{figure}

\begin{figure}[t]
  \centering
  \threeplots{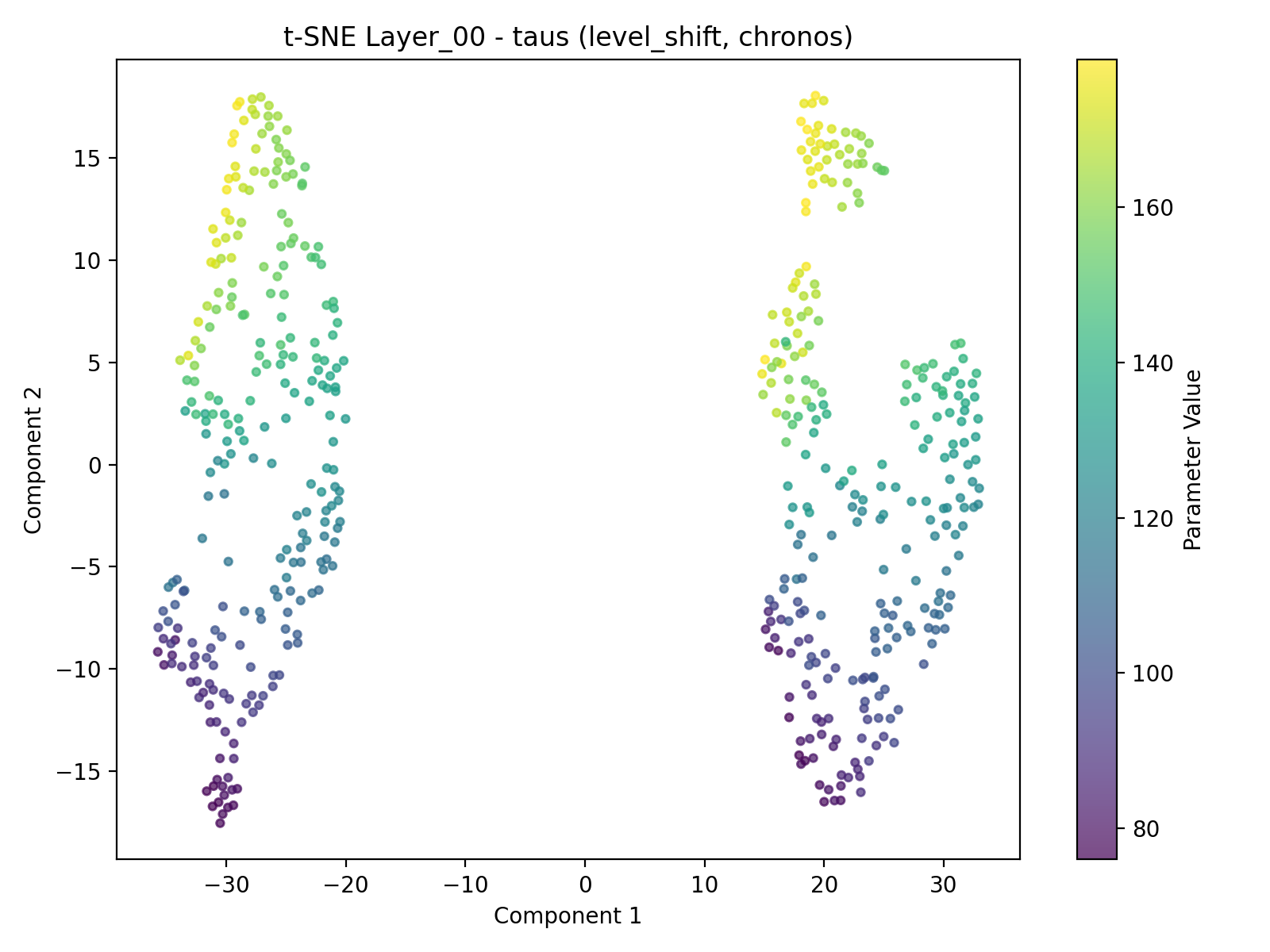}%
             {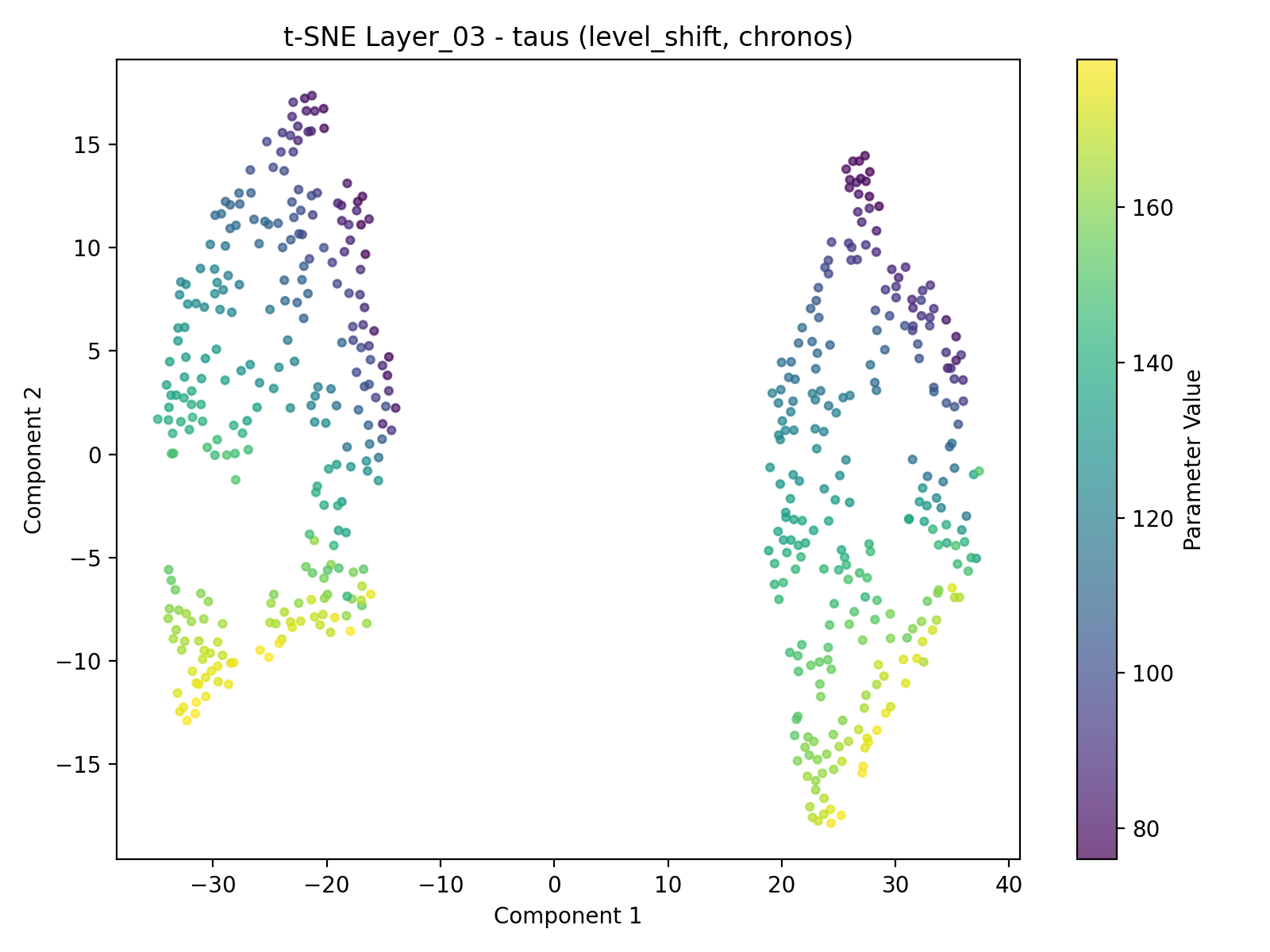}%
             {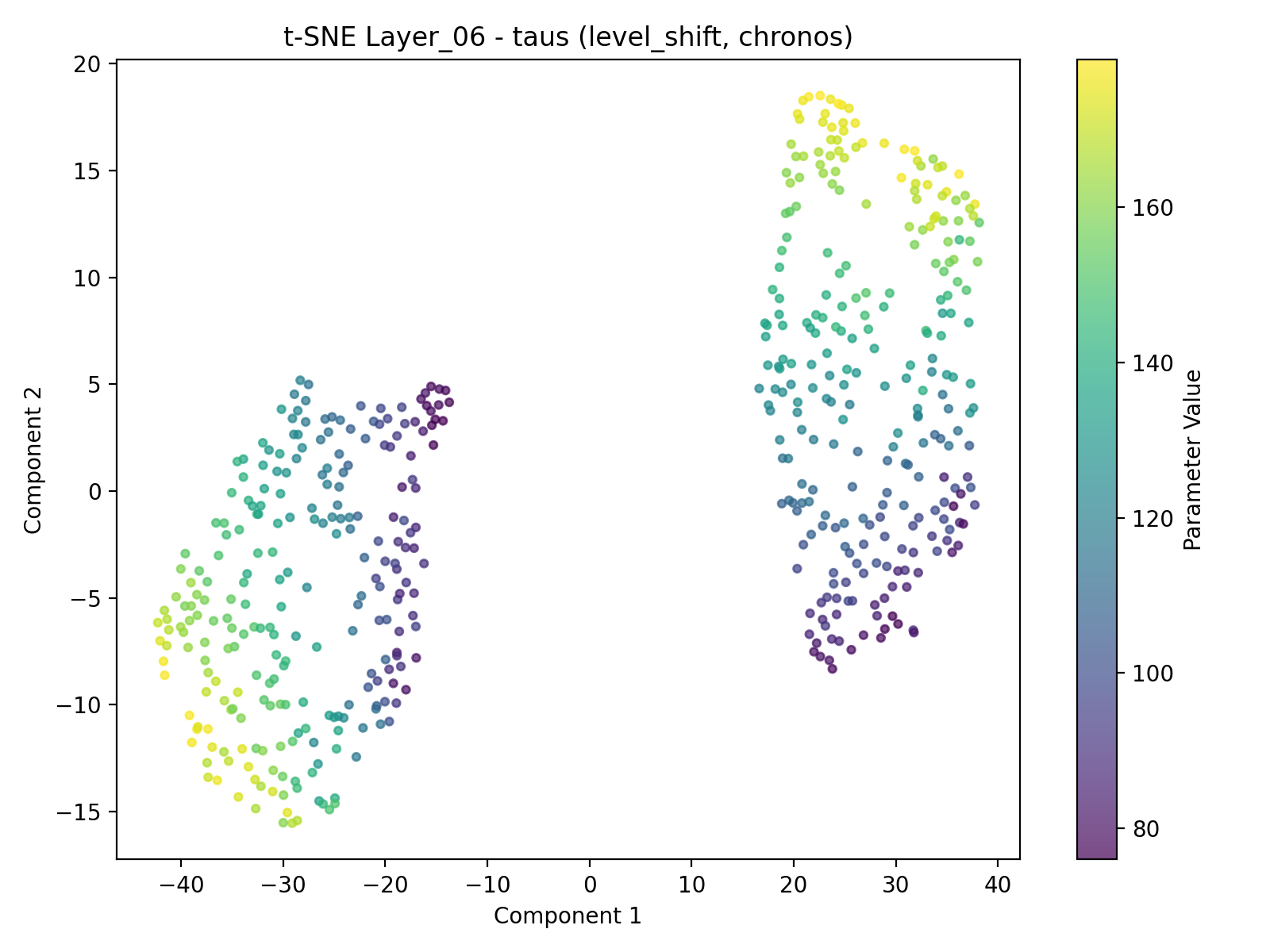}
  \caption{Level Shift --- Chronos --- $\tau$ --- t-SNE (Layers 00/03/06)}
\end{figure}

\begin{figure}[t]
  \centering
  \threeplots{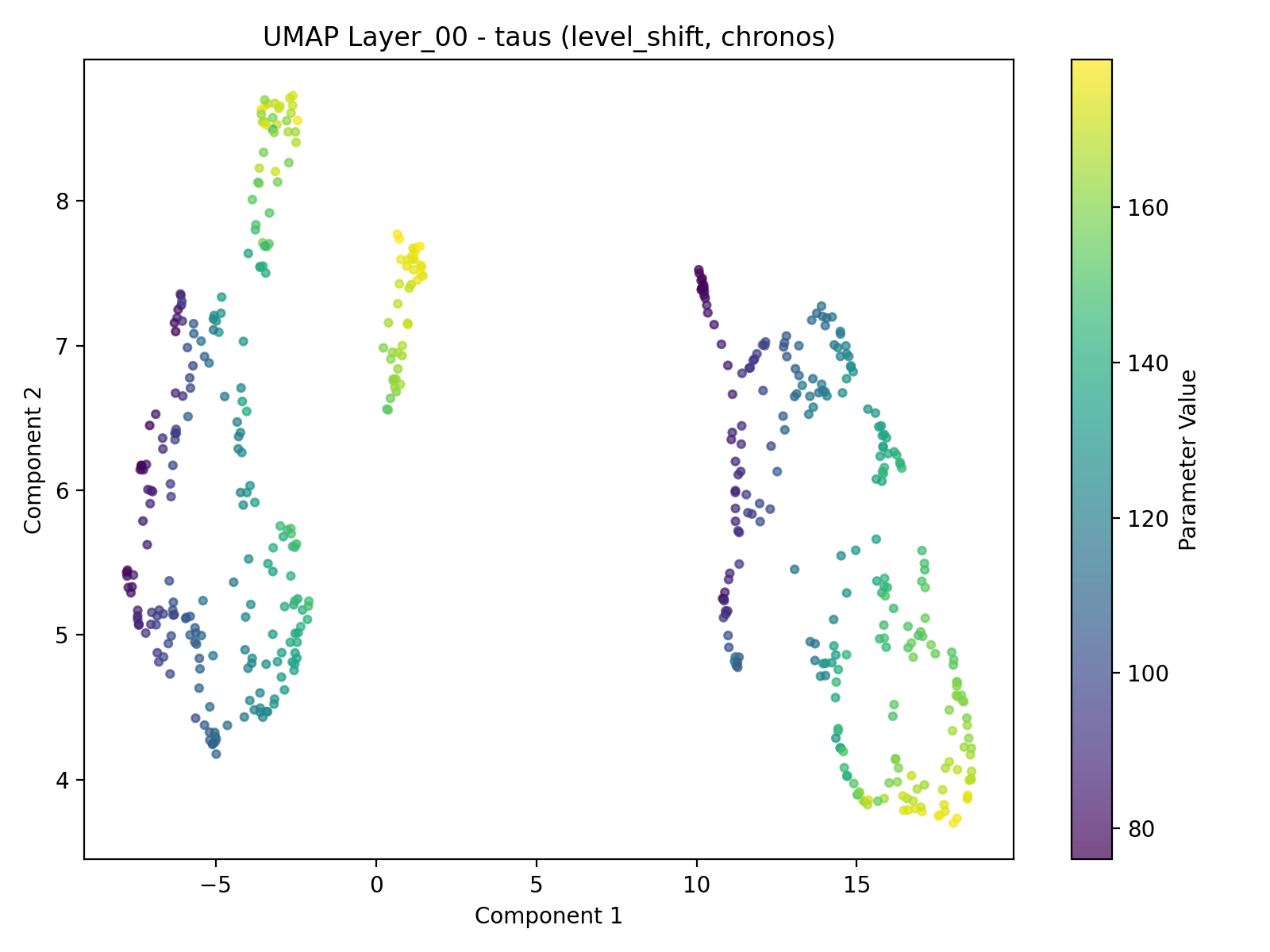}%
             {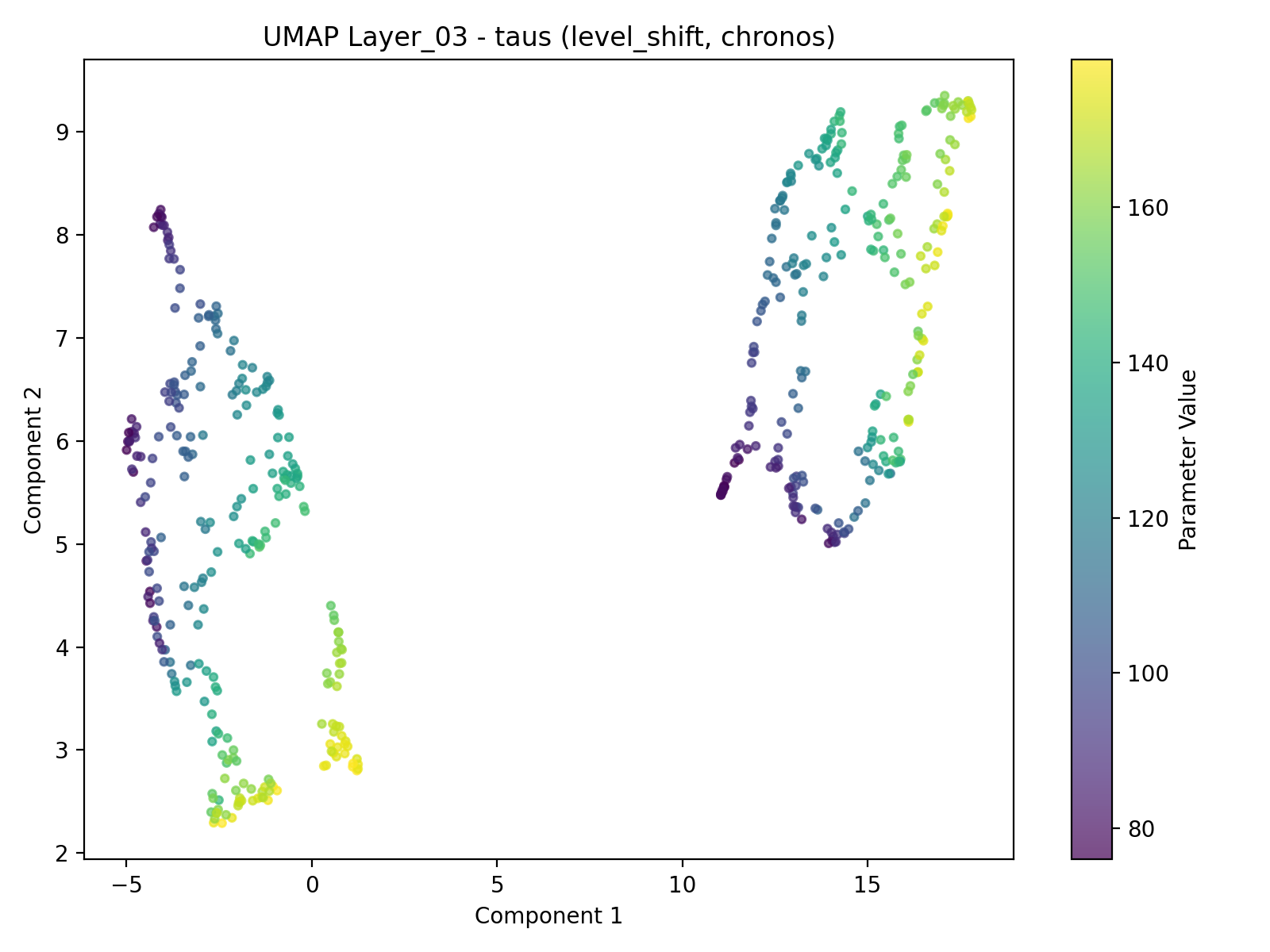}%
             {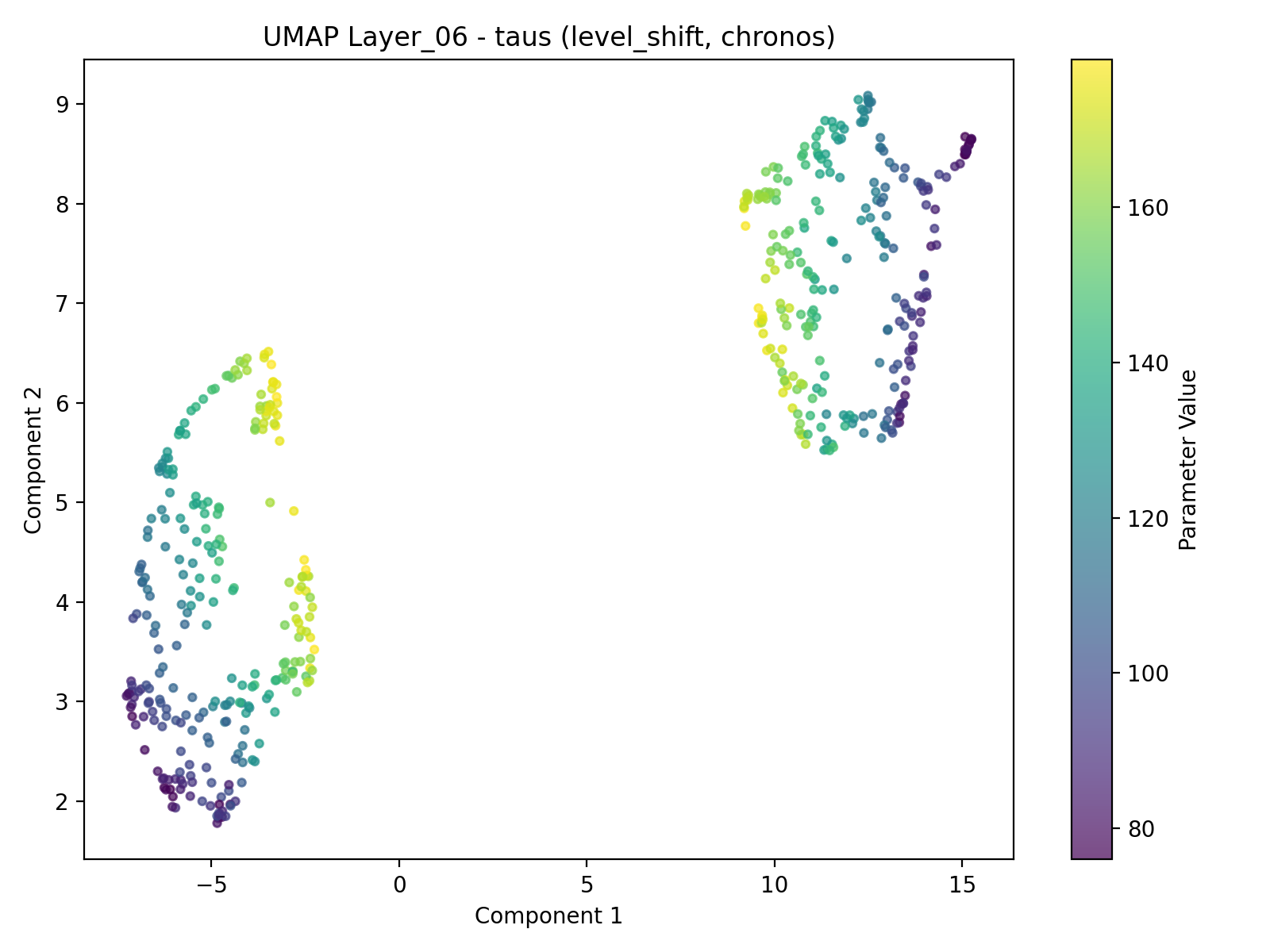}
  \caption{Level Shift --- Chronos --- $\tau$ --- UMAP (Layers 00/03/06)}
\end{figure}
\newpage
% ---------------- Random Walk ----------------
\subsection{Random Walk}

\paragraph{Chronos (parameter: drift).}
% \begin{figure}[t]
%   \centering
%   \threeplots{figs/random_walk_chronos/random_walk/chronos_random_walk_pca_Layer_00_drifts.png}%
%              {figs/random_walk_chronos/random_walk/chronos_random_walk_pca_Layer_03_drifts.png}%
%              {figs/random_walk_chronos/random_walk/chronos_random_walk_pca_Layer_06_drifts.png}
%   \caption{Random Walk --- Chronos --- PCA (Layers 00/03/06)}
% \end{figure}

% \begin{figure}[t]
%   \centering
%   \threeplots{figs/random_walk_chronos/random_walk/chronos_random_walk_tsne_Layer_00_drifts.png}%
%              {figs/random_walk_chronos/random_walk/chronos_random_walk_tsne_Layer_03_drifts.png}%
%              {figs/random_walk_chronos/random_walk/chronos_random_walk_tsne_Layer_06_drifts.png}
%   \caption{Random Walk --- Chronos --- t-SNE (Layers 00/03/06)}
% \end{figure}

\begin{figure}[t]
  \centering
  \threeplots{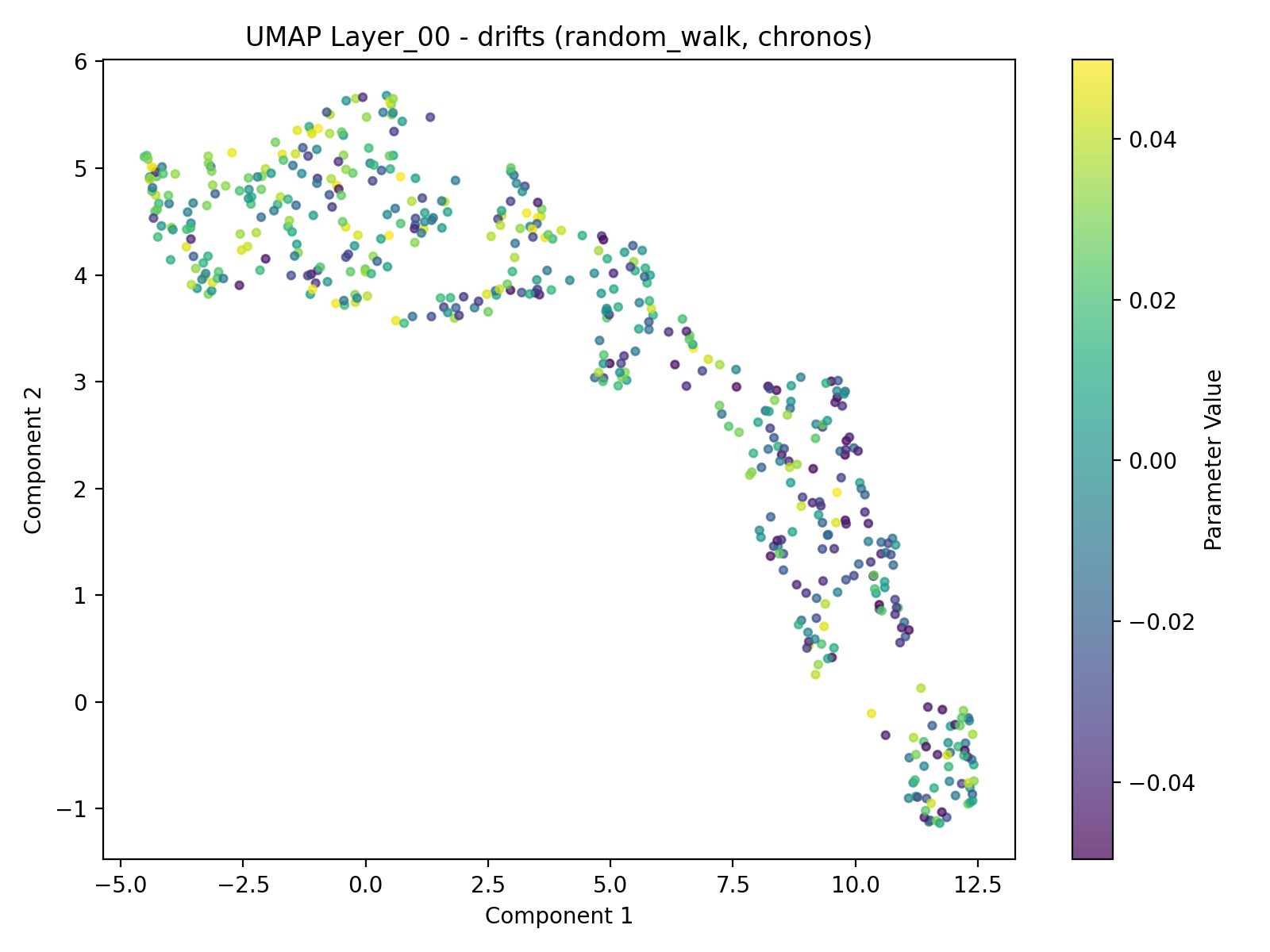}%
             {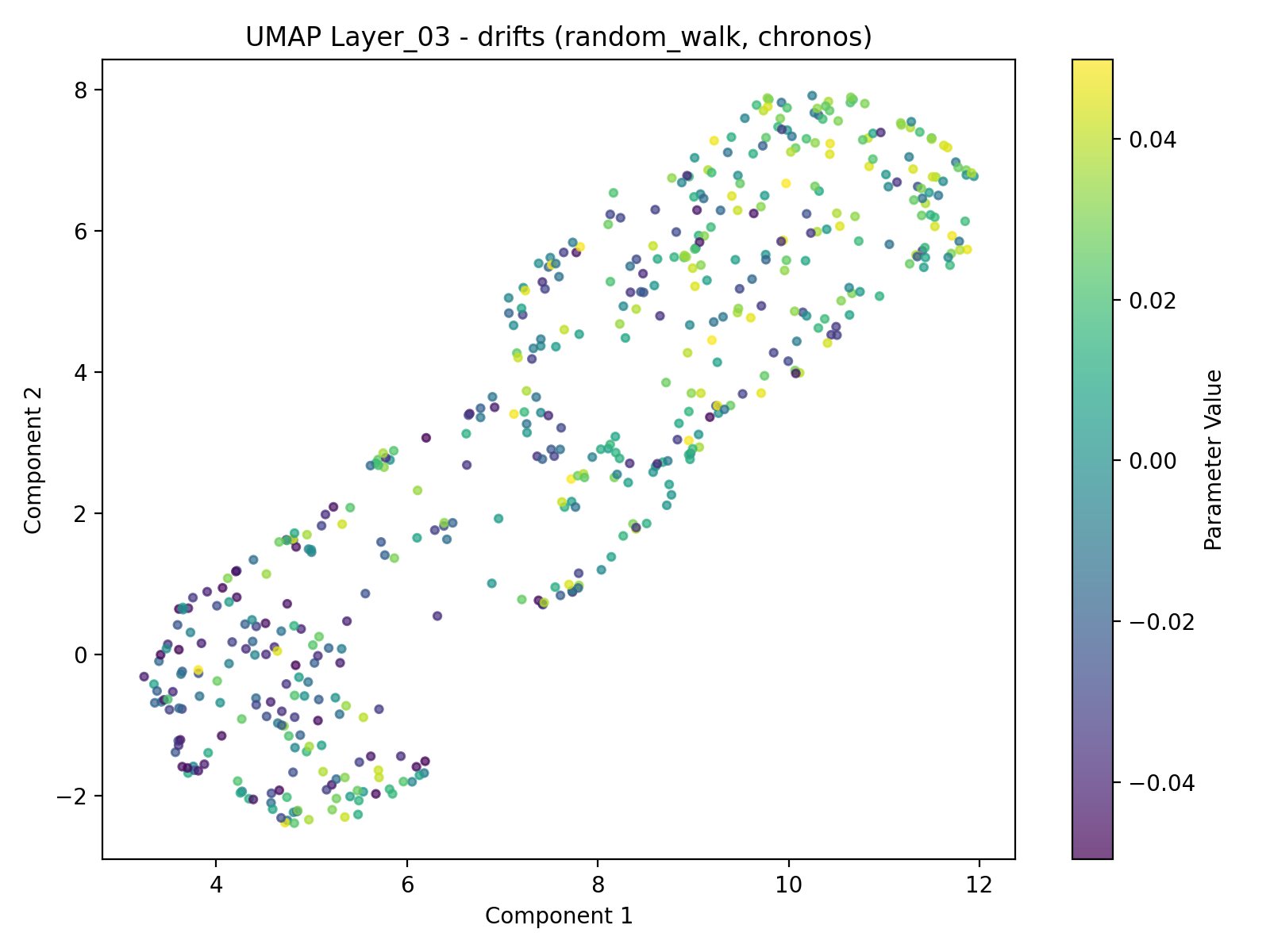}%
             {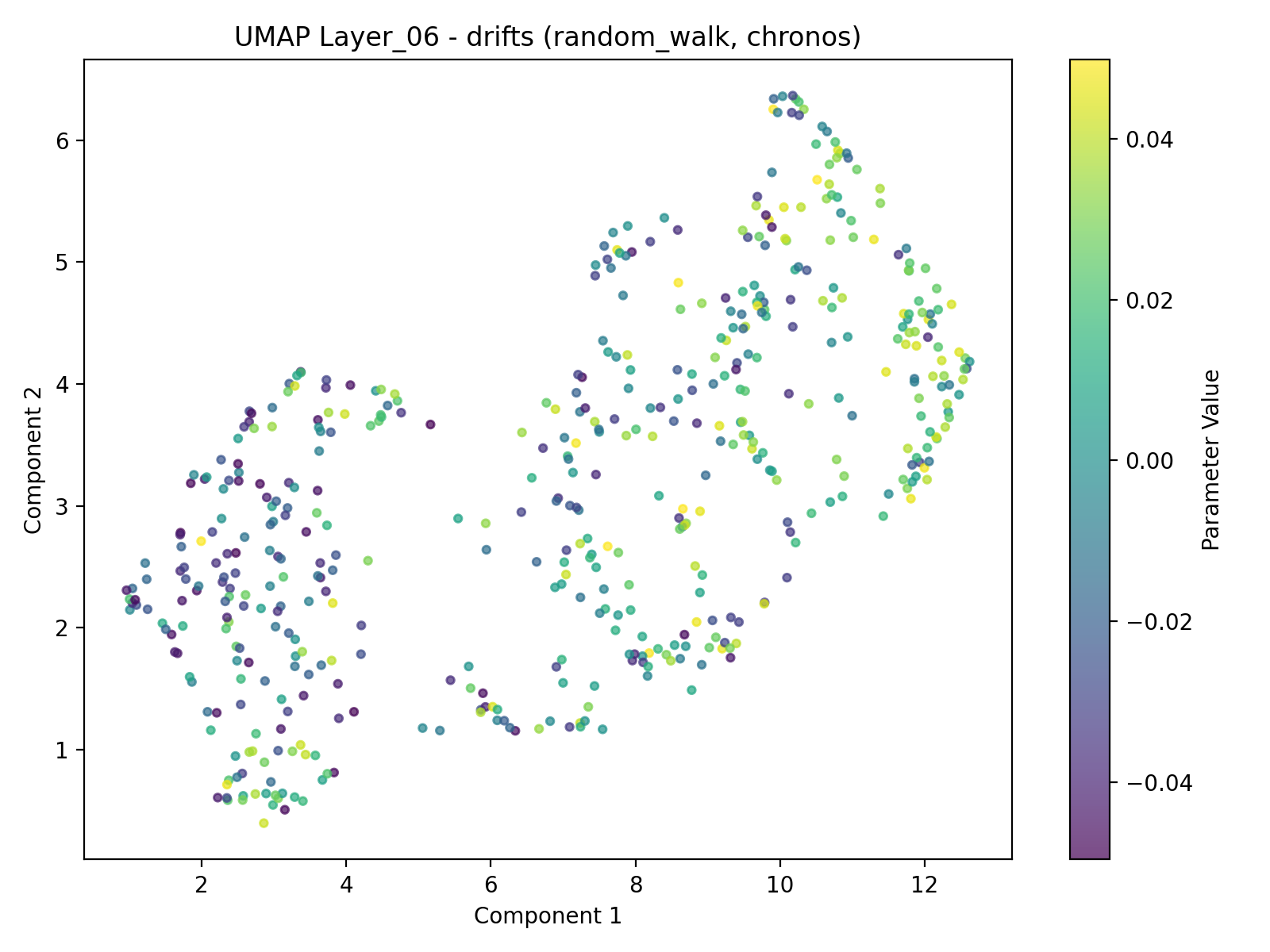}
  \caption{Random Walk --- Chronos --- UMAP (Layers 00/03/06)}
\end{figure}
\newpage
\paragraph{Moment (parameter: drift).}
% \begin{figure}[t]
%   \centering
%   \threeplots{figs/random_walk_moment/random_walk/moment_random_walk_pca_Layer_00_drifts.png}%
%              {figs/random_walk_moment/random_walk/moment_random_walk_pca_Layer_06_drifts.png}%
%              {figs/random_walk_moment/random_walk/moment_random_walk_pca_Layer_12_drifts.png}
%   \caption{Random Walk --- Moment --- PCA (Layers 00/06/12)}
% \end{figure}

% \begin{figure}[t]
%   \centering
%   \threeplots{figs/random_walk_moment/random_walk/moment_random_walk_tsne_Layer_00_drifts.png}%
%              {figs/random_walk_moment/random_walk/moment_random_walk_tsne_Layer_06_drifts.png}%
%              {figs/random_walk_moment/random_walk/moment_random_walk_tsne_Layer_12_drifts.png}
%   \caption{Random Walk --- Moment --- t-SNE (Layers 00/06/12)}
% \end{figure}

\begin{figure}[t]
  \centering
  \threeplots{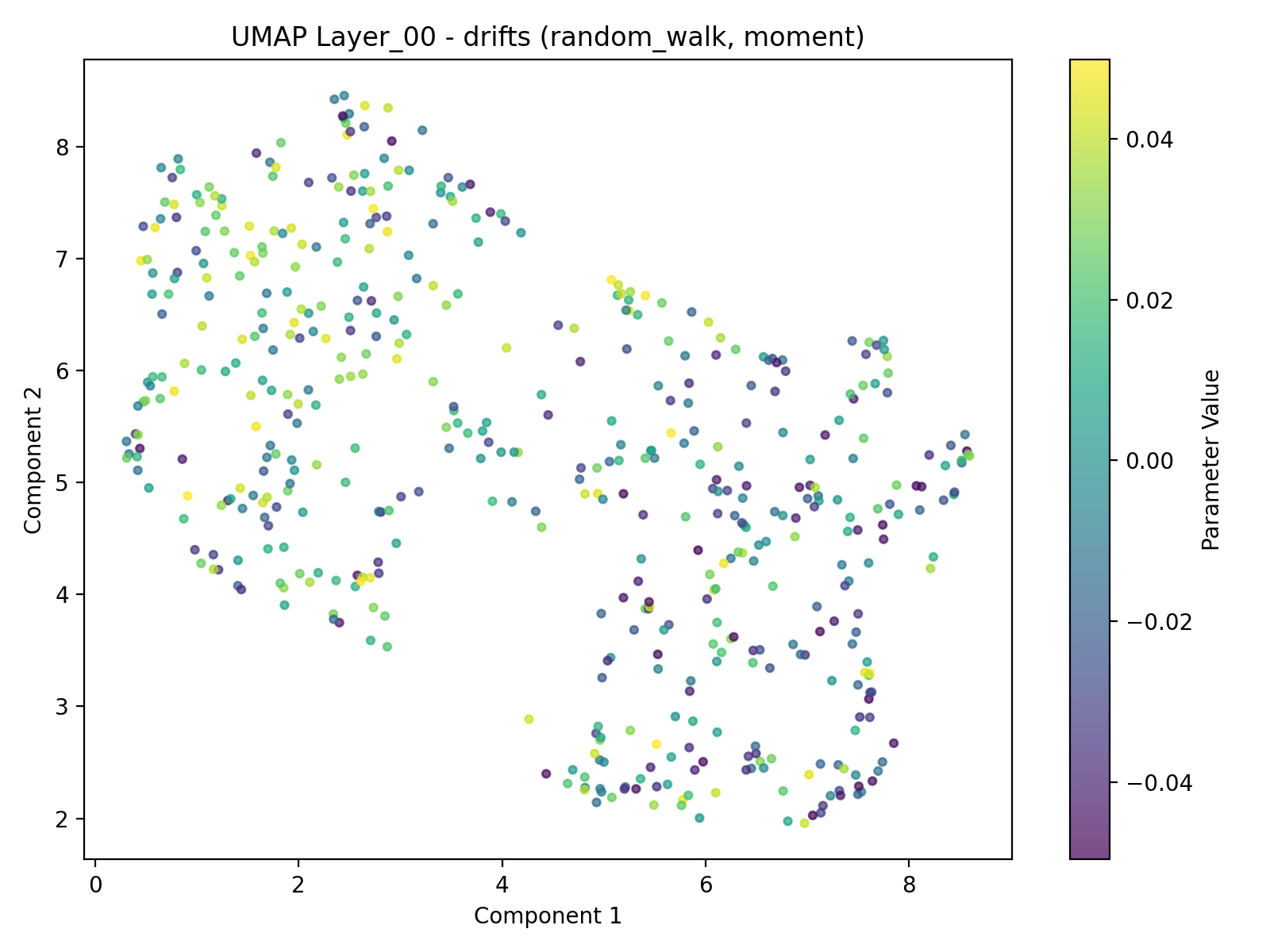}%
             {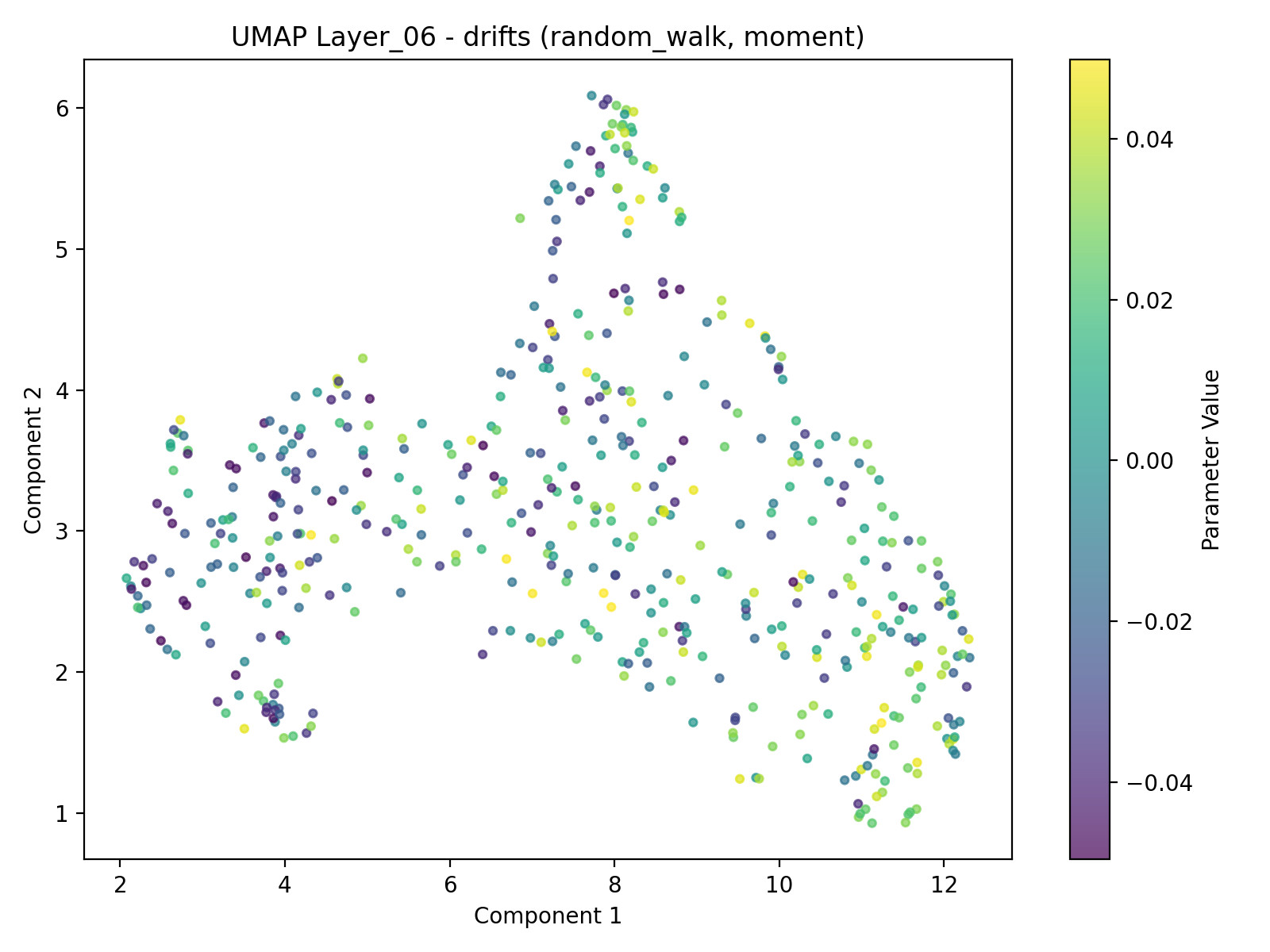}%
             {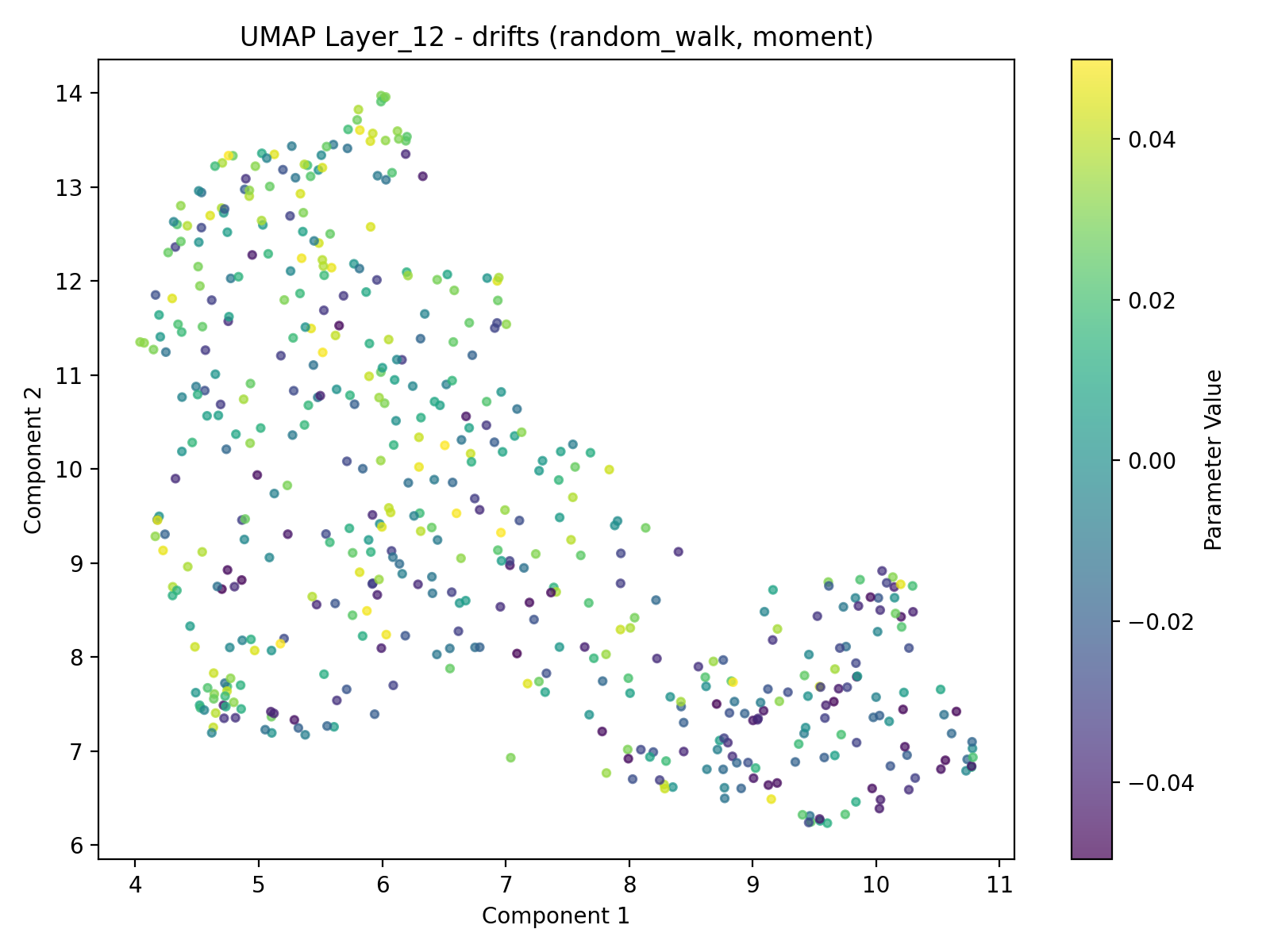}
  \caption{Random Walk --- Moment --- UMAP (Layers 00/06/12)}
\end{figure}
\newpage
% ---------------- Spectral ----------------
\subsection{Spectral (Sum of Sinusoids)}

\paragraph{Chronos ( frequency).}
% \begin{figure}[t]
%   \centering
%   \threeplots{figs/spectral_chronos/spectral/chronos_spectral_pca_Layer_00_amplitude.png}%
%              {figs/spectral_chronos/spectral/chronos_spectral_pca_Layer_03_amplitude.png}%
%              {figs/spectral_chronos/spectral/chronos_spectral_pca_Layer_06_amplitude.png}
%   \caption{Spectral --- Chronos --- Amplitude --- PCA (Layers 00/03/06)}
% \end{figure}

% \begin{figure}[t]
%   \centering
%   \threeplots{figs/spectral_chronos/spectral/chronos_spectral_tsne_Layer_00_amplitude.png}%
%              {figs/spectral_chronos/spectral/chronos_spectral_tsne_Layer_03_amplitude.png}%
%              {figs/spectral_chronos/spectral/chronos_spectral_tsne_Layer_06_amplitude.png}
%   \caption{Spectral --- Chronos --- Amplitude --- t-SNE (Layers 00/03/06)}
% \end{figure}

% \begin{figure}[t]
%   \centering
%   \threeplots{figs/spectral_chronos/spectral/chronos_spectral_umap_Layer_00_amplitude.png}%
%              {figs/spectral_chronos/spectral/chronos_spectral_umap_Layer_03_amplitude.png}%
%              {figs/spectral_chronos/spectral/chronos_spectral_umap_Layer_06_amplitude.png}
%   \caption{Spectral --- Chronos --- Amplitude --- UMAP (Layers 00/03/06)}
% \end{figure}

\begin{figure}[t]
  \centering
  \threeplots{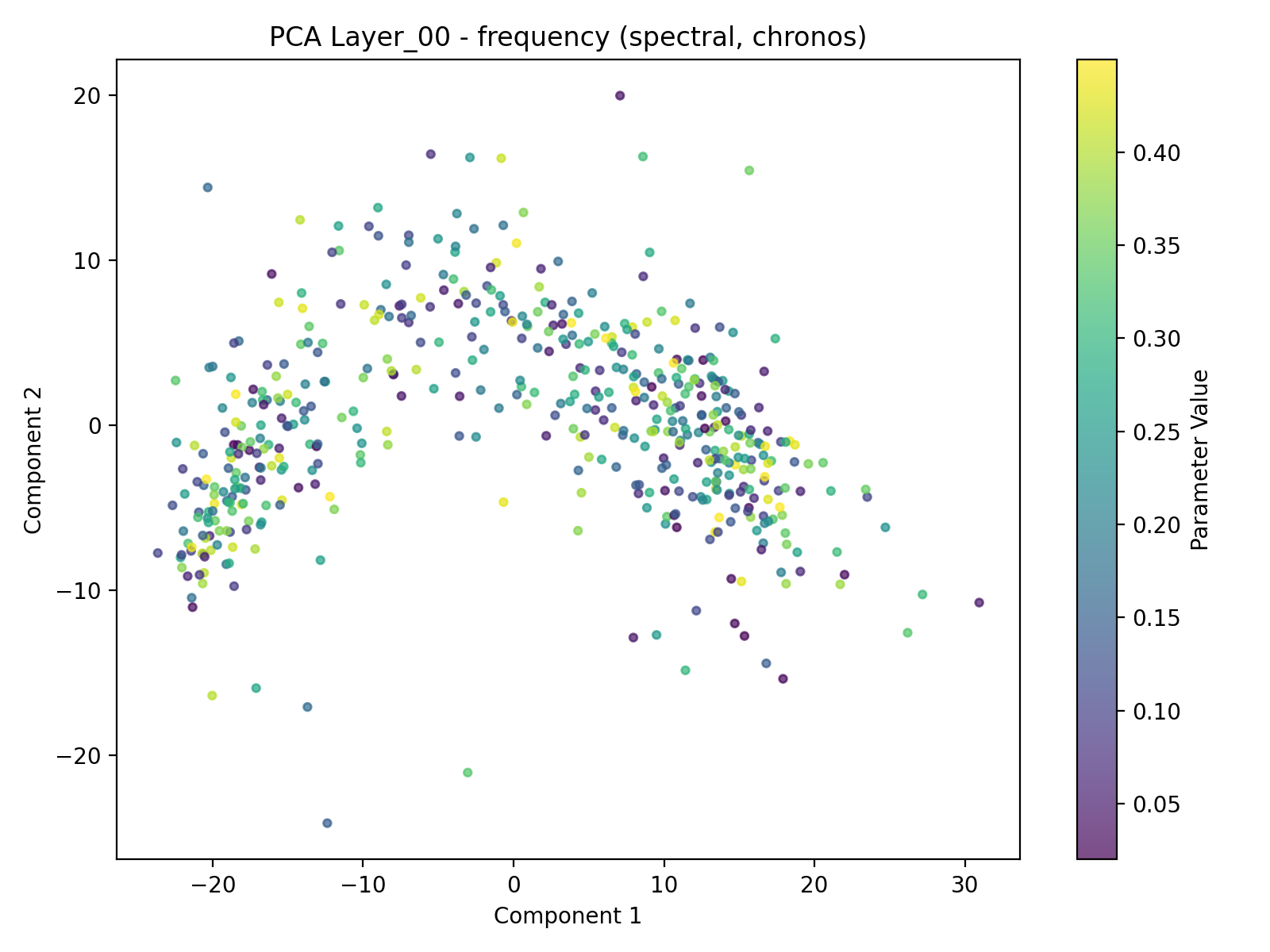}%
             {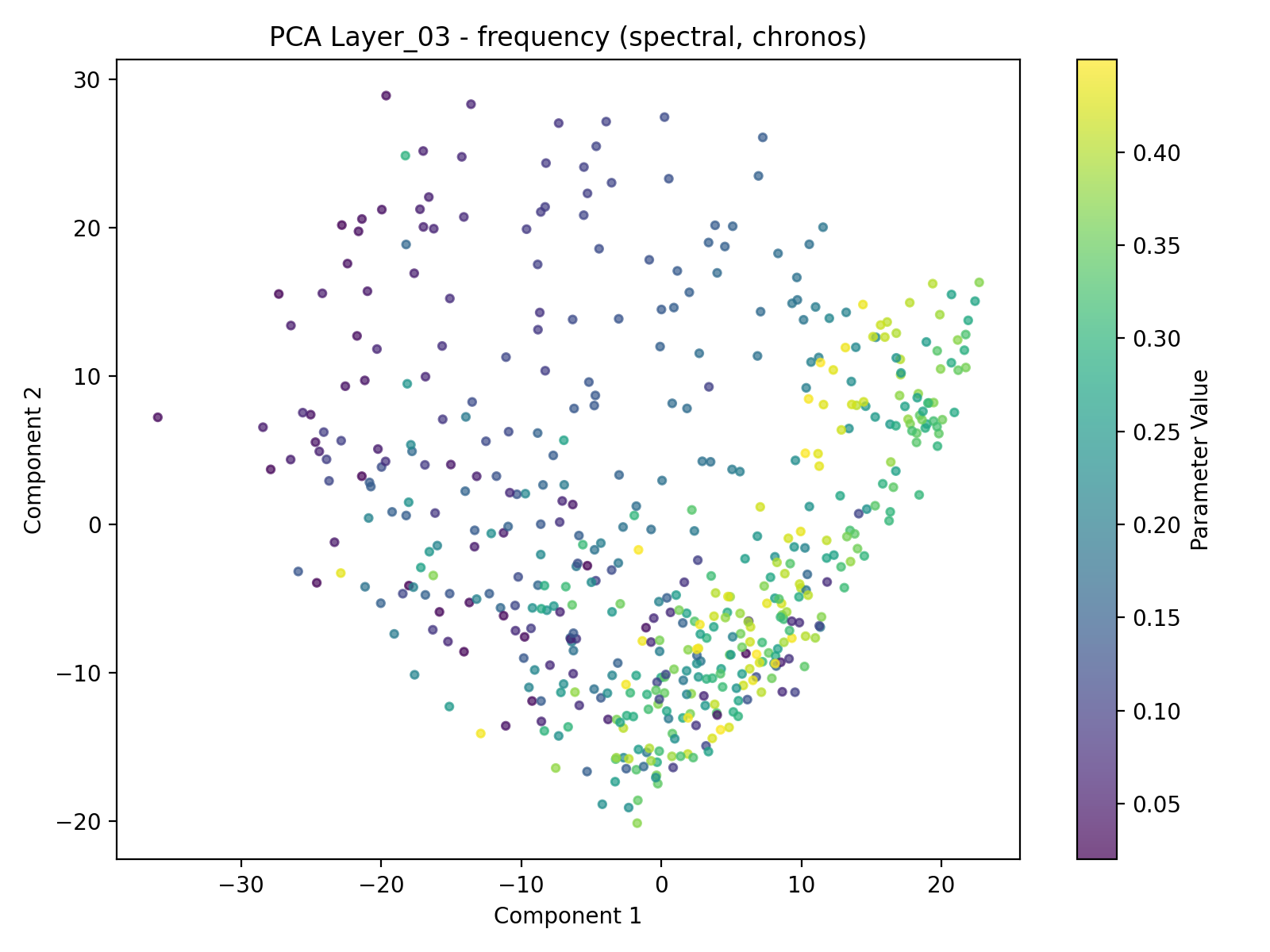}%
             {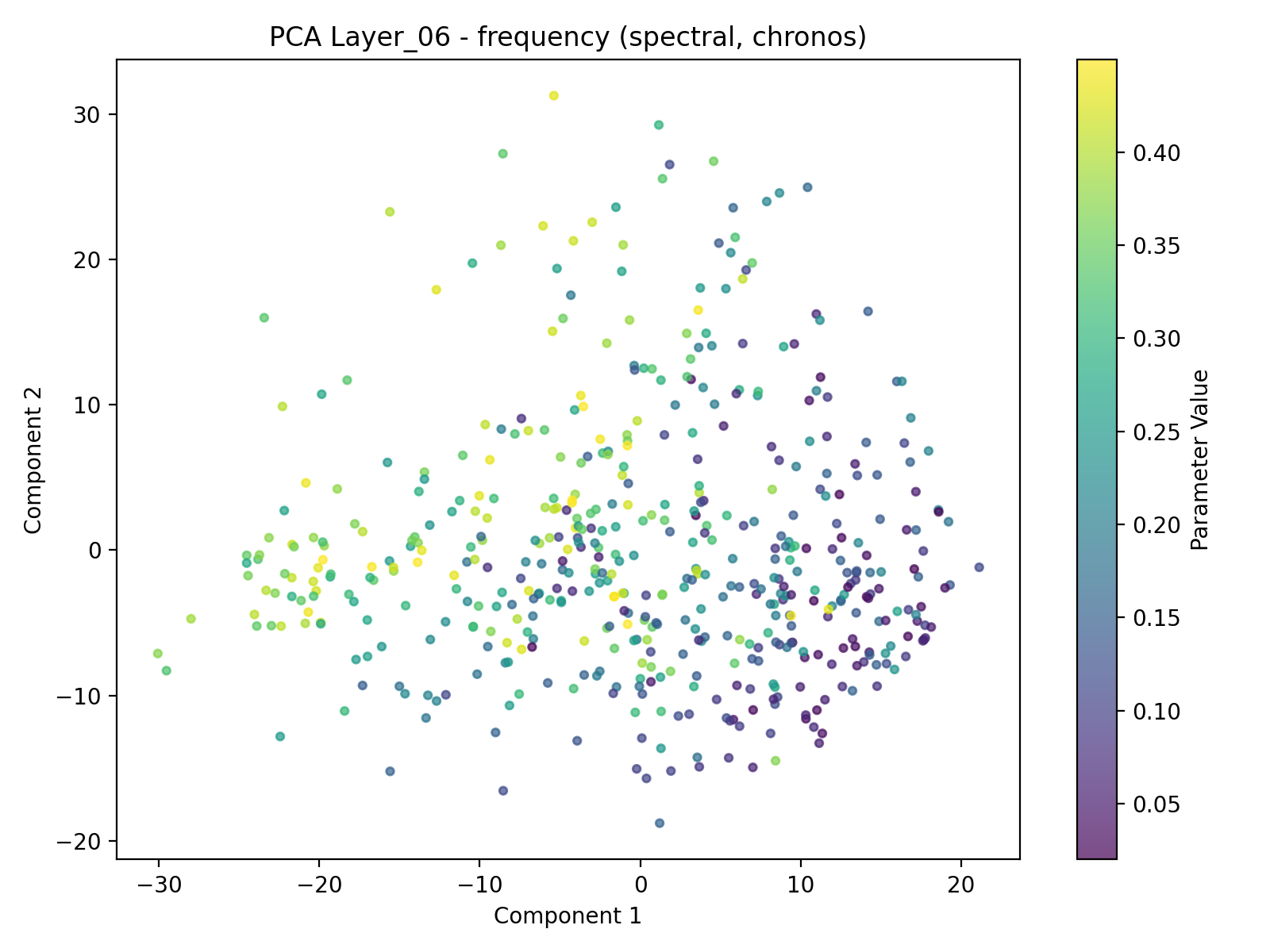}
  \caption{Spectral --- Chronos --- Frequency --- PCA (Layers 00/03/06)}
\end{figure}

\begin{figure}[t]
  \centering
  \threeplots{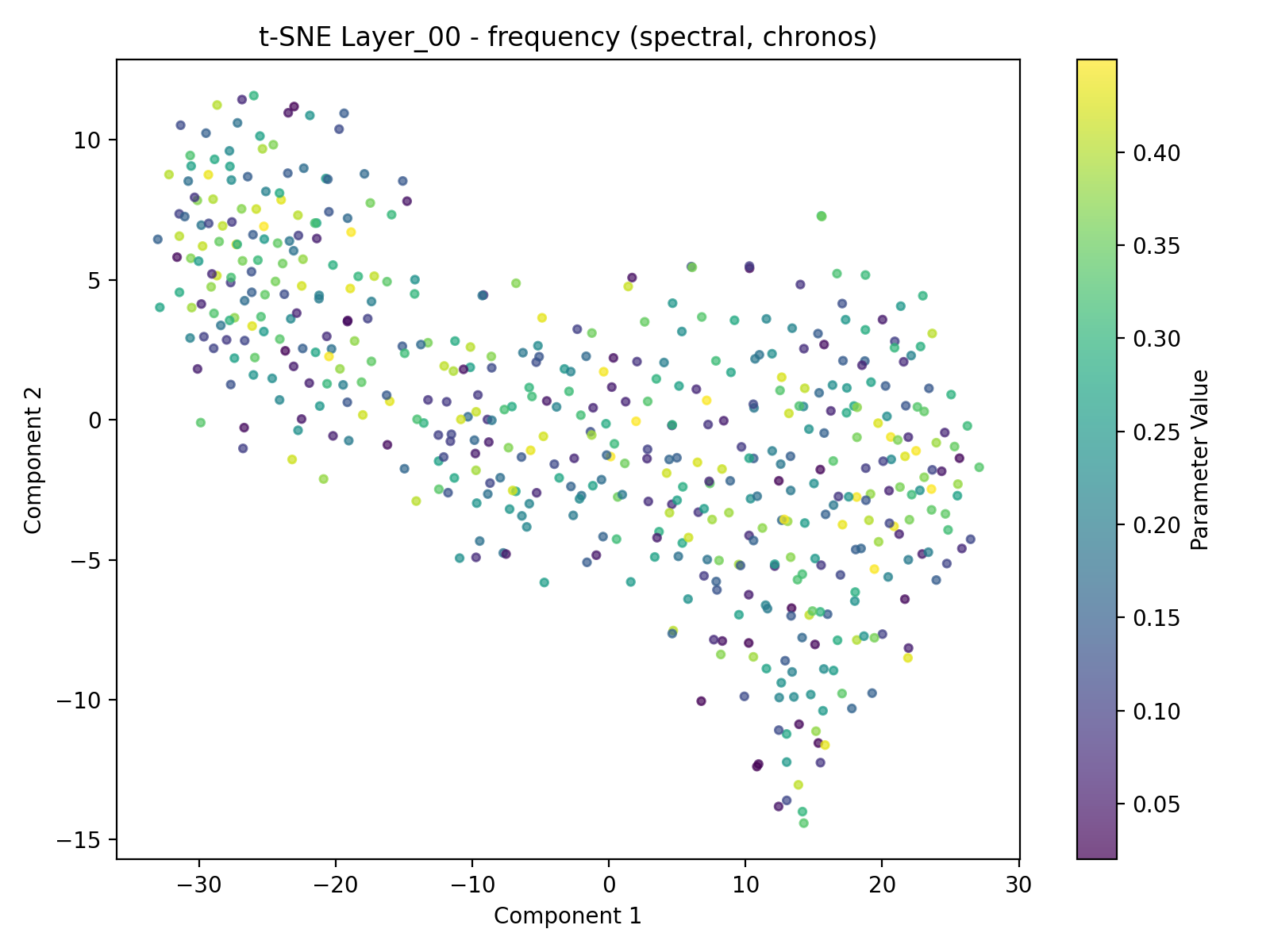}%
             {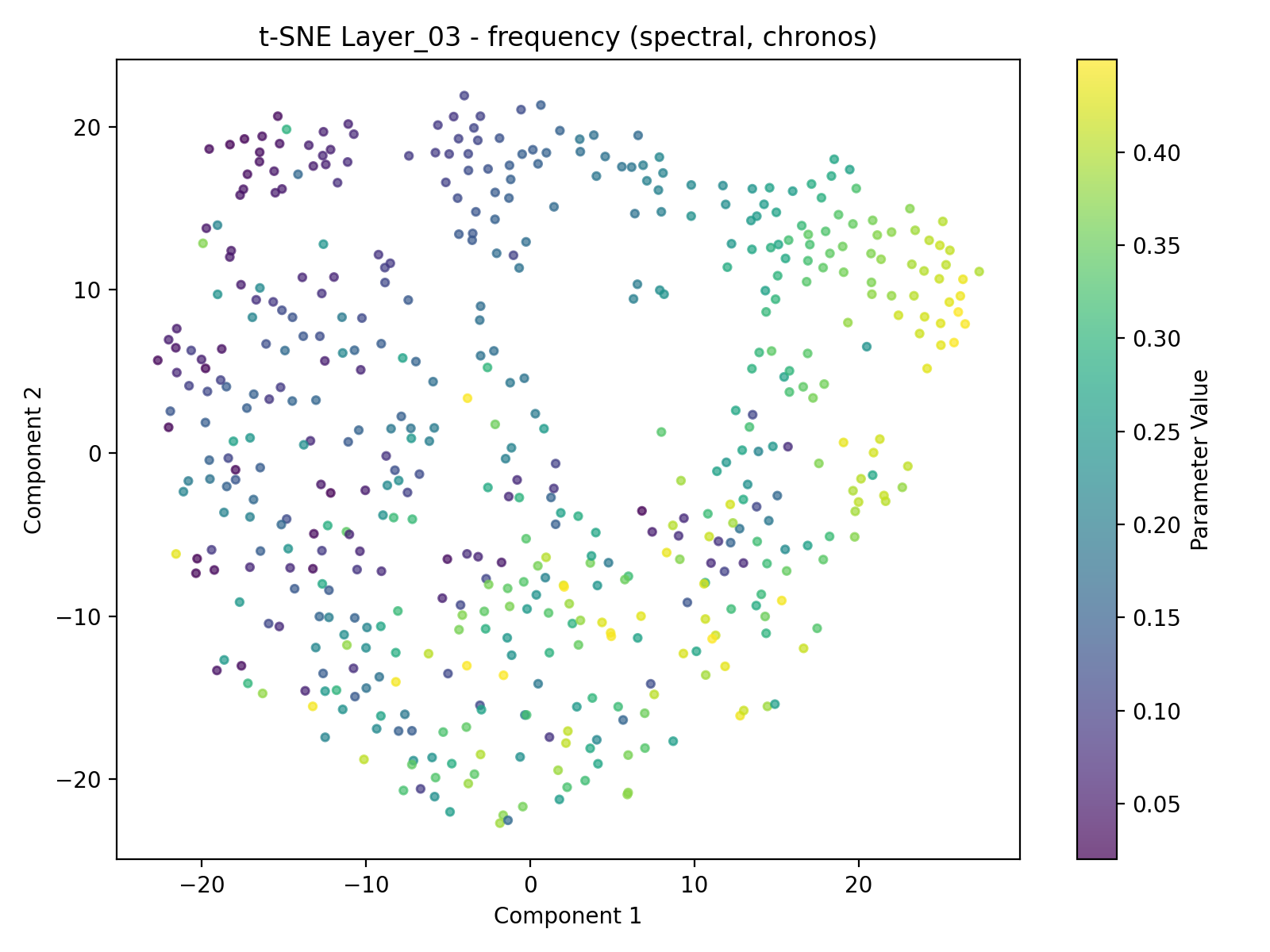}%
             {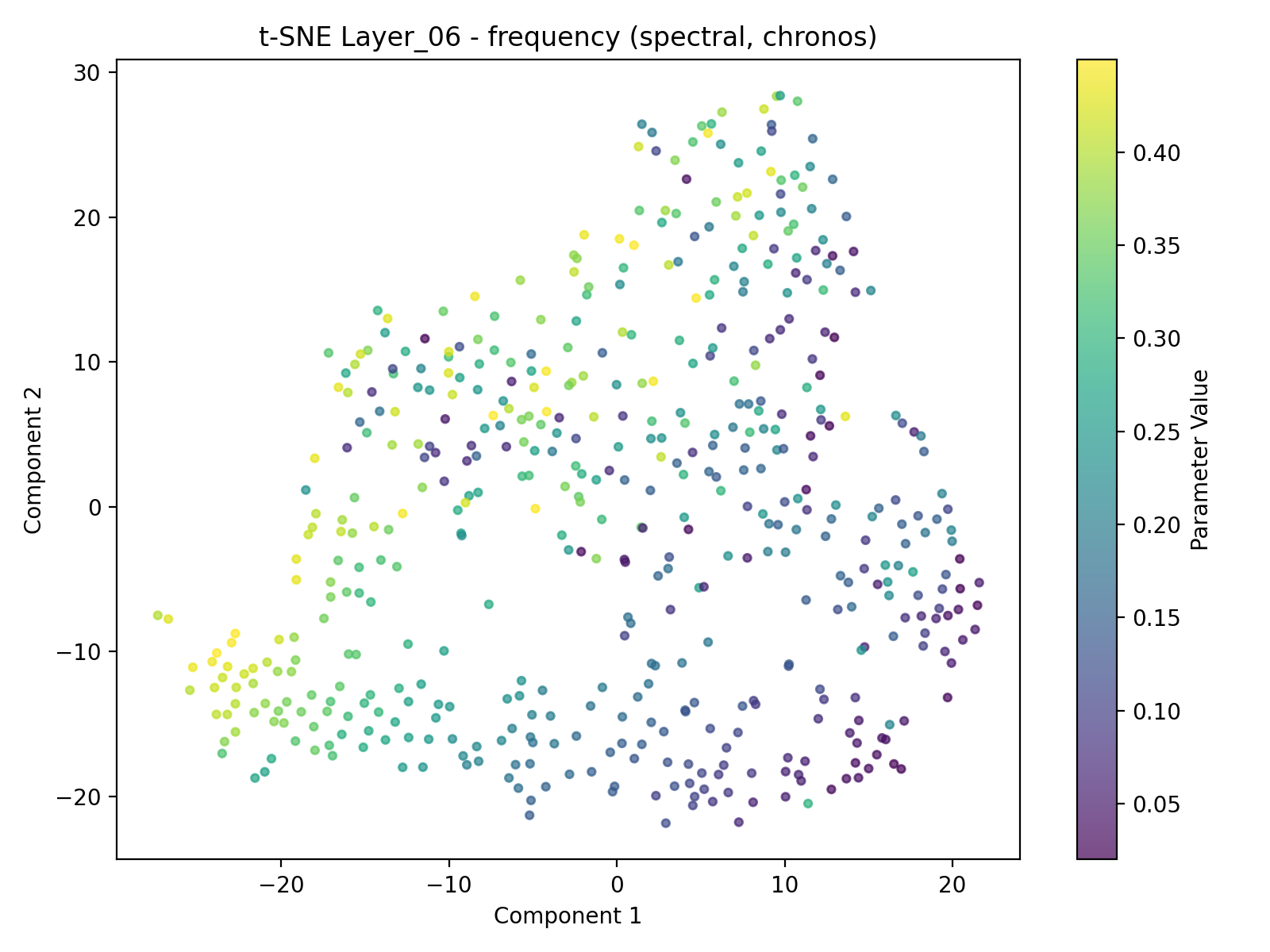}
  \caption{Spectral --- Chronos --- Frequency --- t-SNE (Layers 00/03/06)}
\end{figure}

\begin{figure}[t]
  \centering
  \threeplots{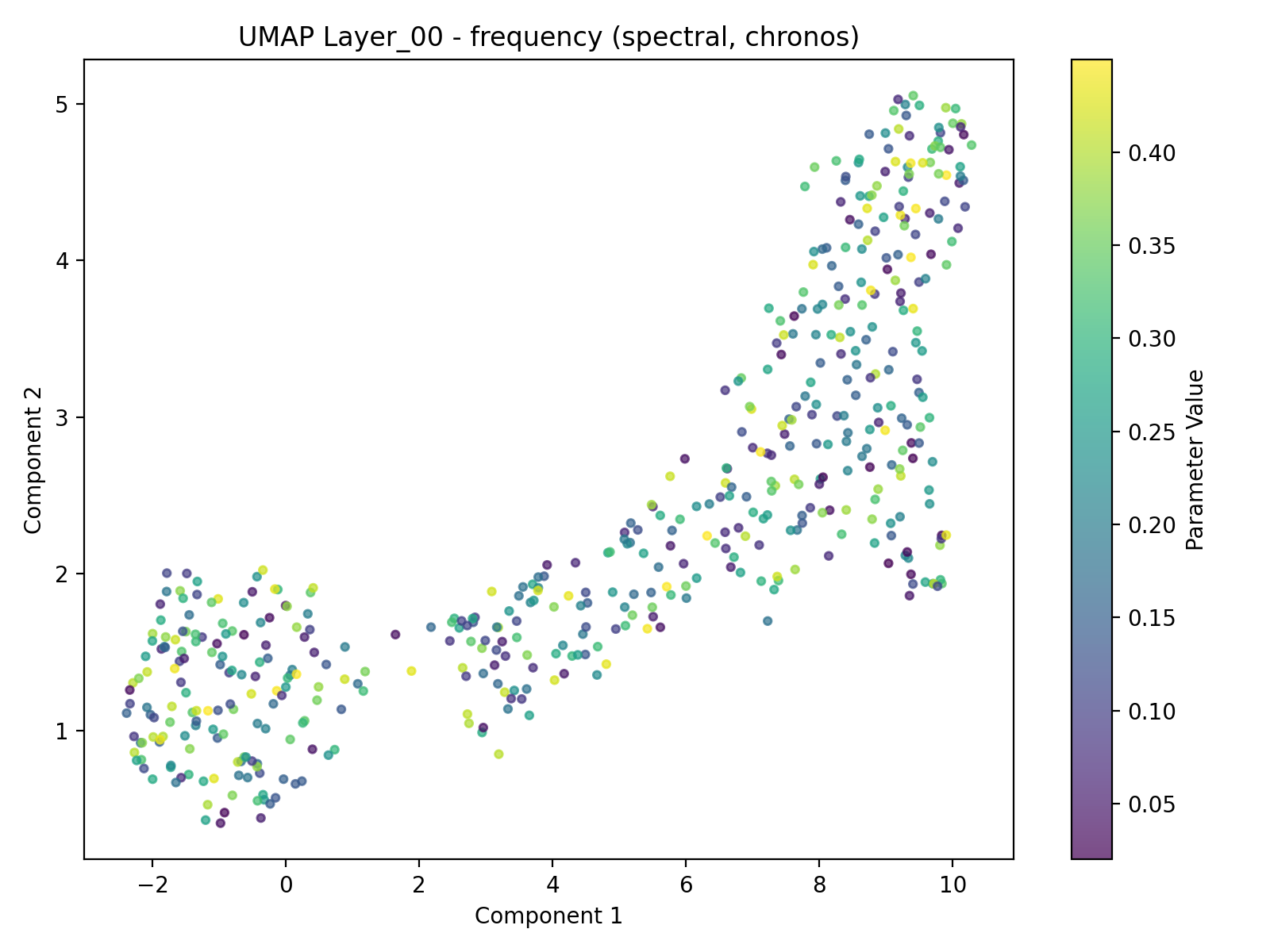}%
             {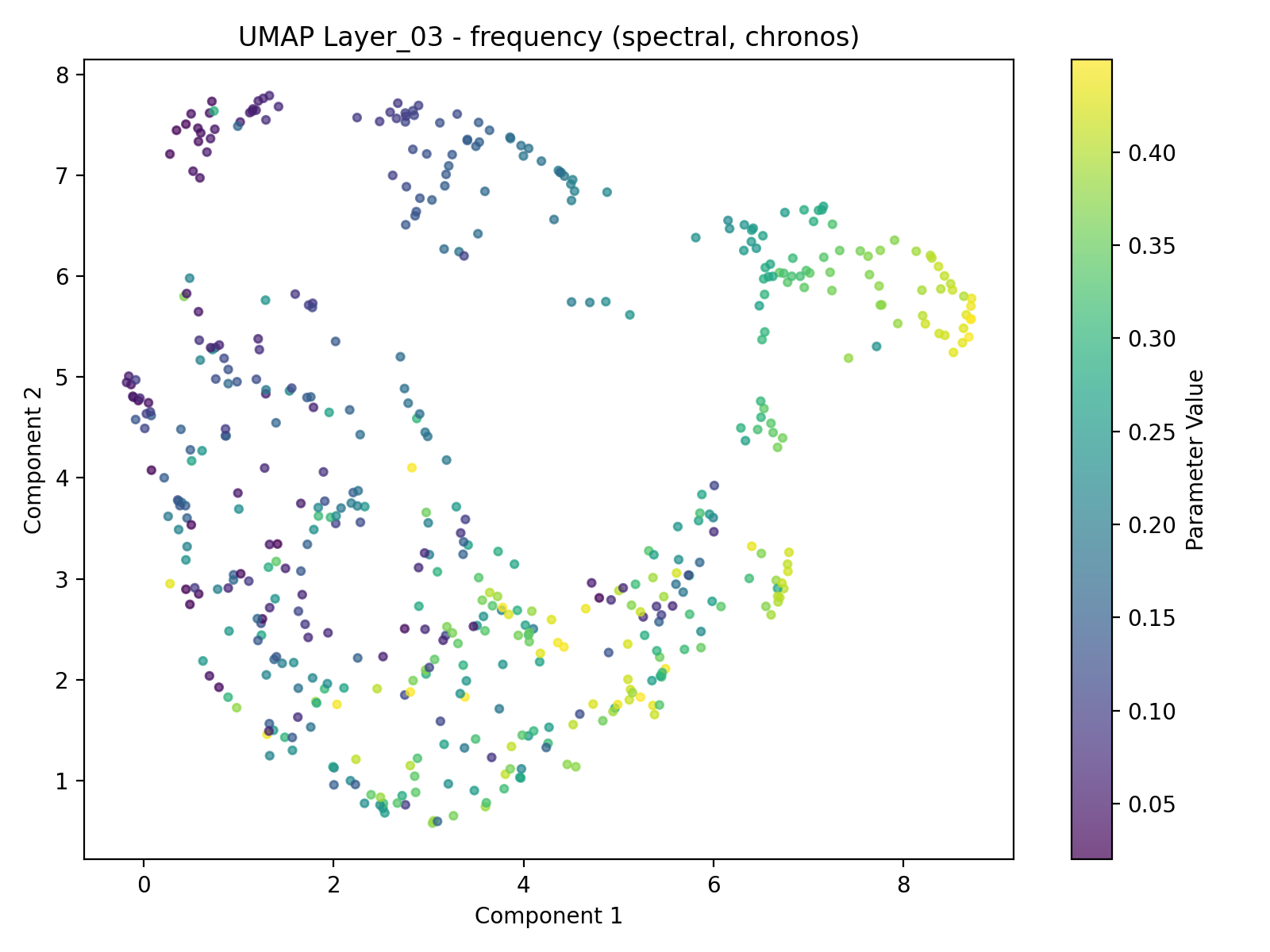}%
             {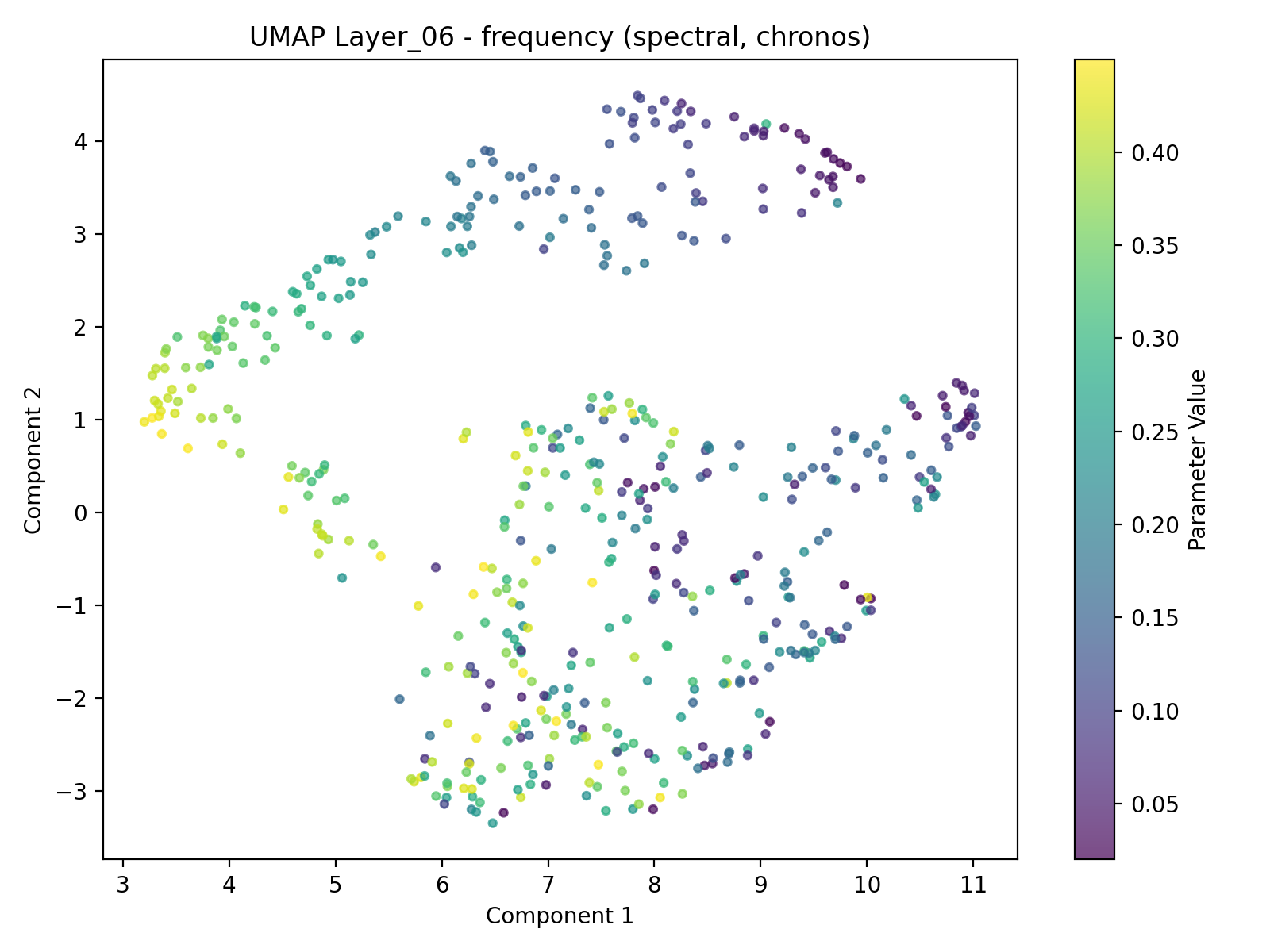}
  \caption{Spectral --- Chronos --- Frequency --- UMAP (Layers 00/03/06)}
\end{figure}

% \begin{figure}[t]
%   \centering
%   \threeplots{figs/spectral_chronos/spectral/chronos_spectral_pca_Layer_00_phase.png}%
%              {figs/spectral_chronos/spectral/chronos_spectral_pca_Layer_03_phase.png}%
%              {figs/spectral_chronos/spectral/chronos_spectral_pca_Layer_06_phase.png}
%   \caption{Spectral --- Chronos --- Phase --- PCA (Layers 00/03/06)}
% \end{figure}

% \begin{figure}[t]
%   \centering
%   \threeplots{figs/spectral_chronos/spectral/chronos_spectral_tsne_Layer_00_phase.png}%
%              {figs/spectral_chronos/spectral/chronos_spectral_tsne_Layer_03_phase.png}%
%              {figs/spectral_chronos/spectral/chronos_spectral_tsne_Layer_06_phase.png}
%   \caption{Spectral --- Chronos --- Phase --- t-SNE (Layers 00/03/06)}
% \end{figure}

% \begin{figure}[t]
%   \centering
%   \threeplots{figs/spectral_chronos/spectral/chronos_spectral_umap_Layer_00_phase.png}%
%              {figs/spectral_chronos/spectral/chronos_spectral_umap_Layer_03_phase.png}%
%              {figs/spectral_chronos/spectral/chronos_spectral_umap_Layer_06_phase.png}
%   \caption{Spectral --- Chronos --- Phase --- UMAP (Layers 00/03/06)}
% \end{figure}
\newpage
\paragraph{Moment ( frequency).}
% \begin{figure}[t]
%   \centering
%   \threeplots{figs/spectral_moment/spectral/moment_spectral_pca_Layer_00_amplitude.png}%
%              {figs/spectral_moment/spectral/moment_spectral_pca_Layer_06_amplitude.png}%
%              {figs/spectral_moment/spectral/moment_spectral_pca_Layer_12_amplitude.png}
%   \caption{Spectral --- Moment --- Amplitude --- PCA (Layers 00/06/12)}
% \end{figure}

% \begin{figure}[t]
%   \centering
%   \threeplots{figs/spectral_moment/spectral/moment_spectral_tsne_Layer_00_amplitude.png}%
%              {figs/spectral_moment/spectral/moment_spectral_tsne_Layer_06_amplitude.png}%
%              {figs/spectral_moment/spectral/moment_spectral_tsne_Layer_12_amplitude.png}
%   \caption{Spectral --- Moment --- Amplitude --- t-SNE (Layers 00/06/12)}
% \end{figure}

% \begin{figure}[t]
%   \centering
%   \threeplots{figs/spectral_moment/spectral/moment_spectral_umap_Layer_00_amplitude.png}%
%              {figs/spectral_moment/spectral/moment_spectral_umap_Layer_06_amplitude.png}%
%              {figs/spectral_moment/spectral/moment_spectral_umap_Layer_12_amplitude.png}
%   \caption{Spectral --- Moment --- Amplitude --- UMAP (Layers 00/06/12)}
% \end{figure}

\begin{figure}[t]
  \centering
  \threeplots{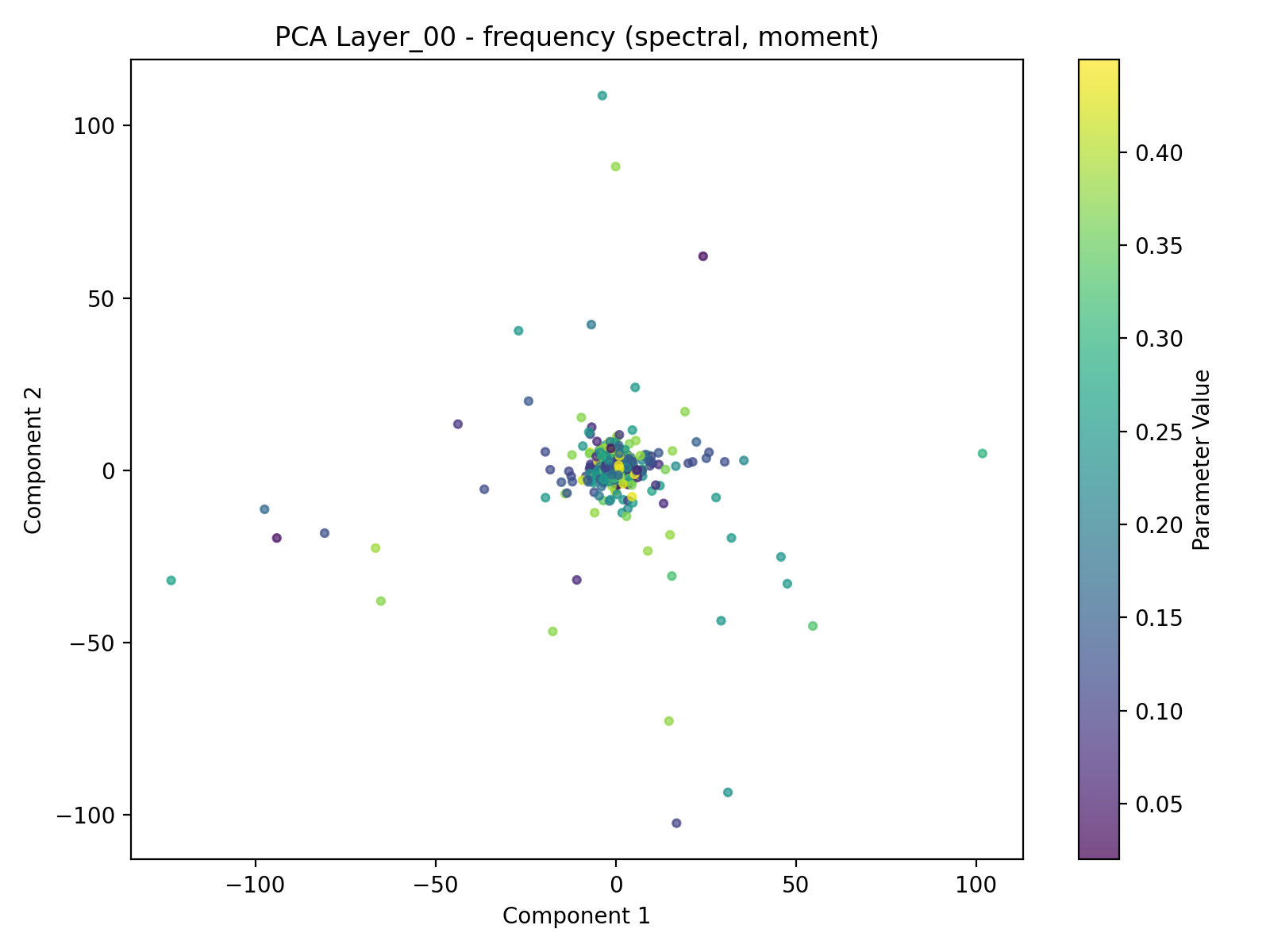}%
             {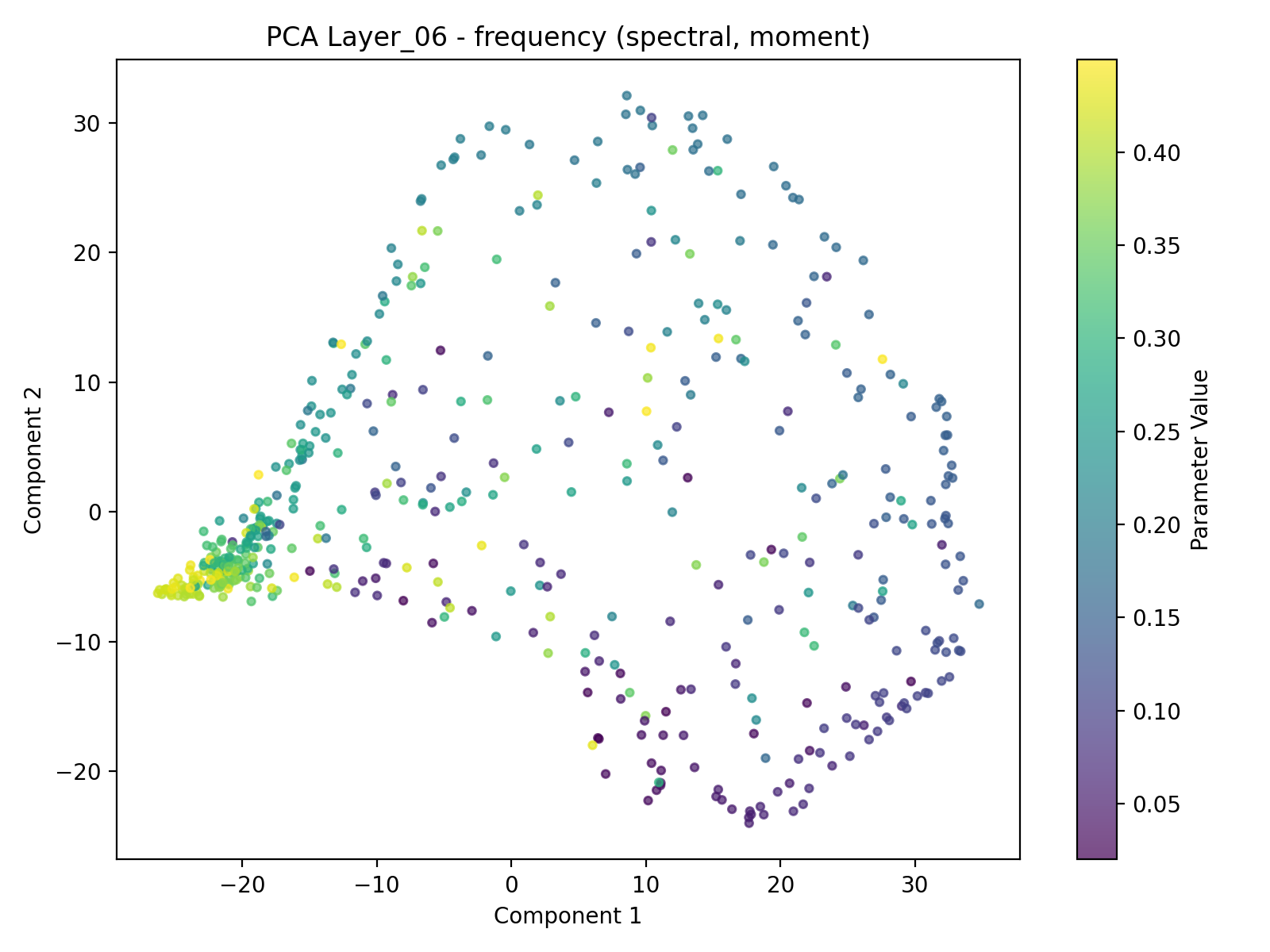}%
             {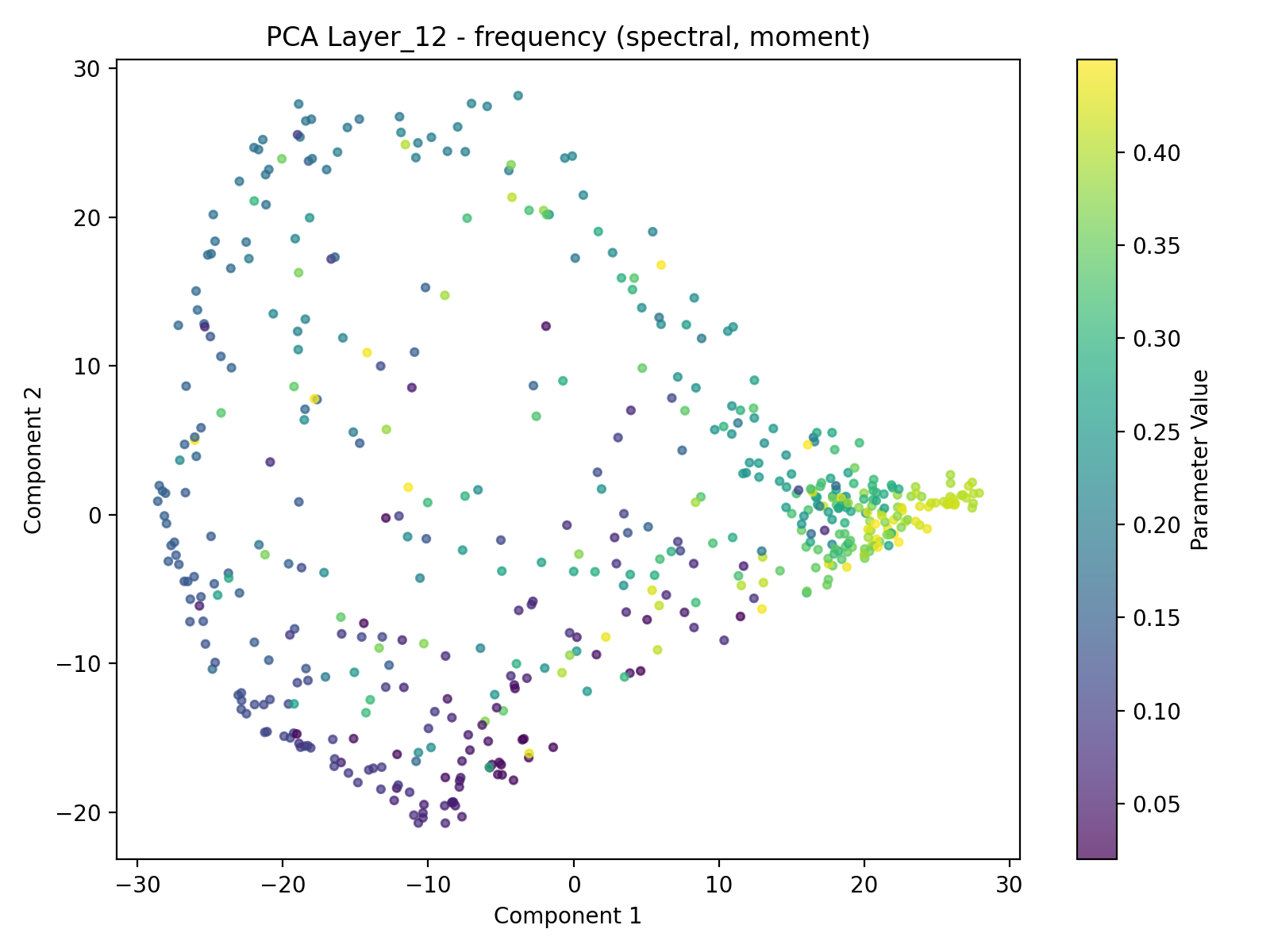}
  \caption{Spectral --- Moment --- Frequency --- PCA (Layers 00/06/12)}
\end{figure}

\begin{figure}[t]
  \centering
  \threeplots{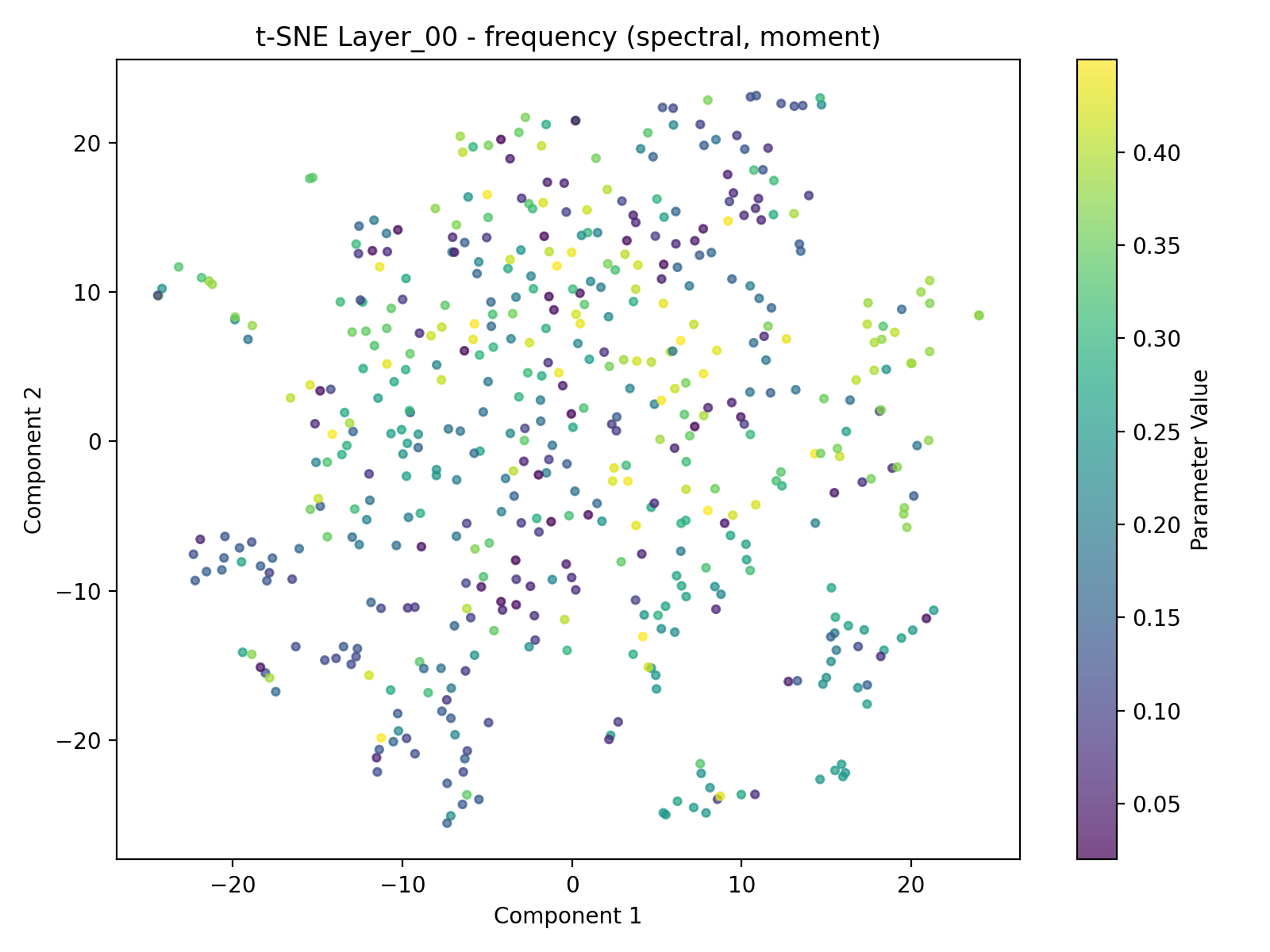}%
             {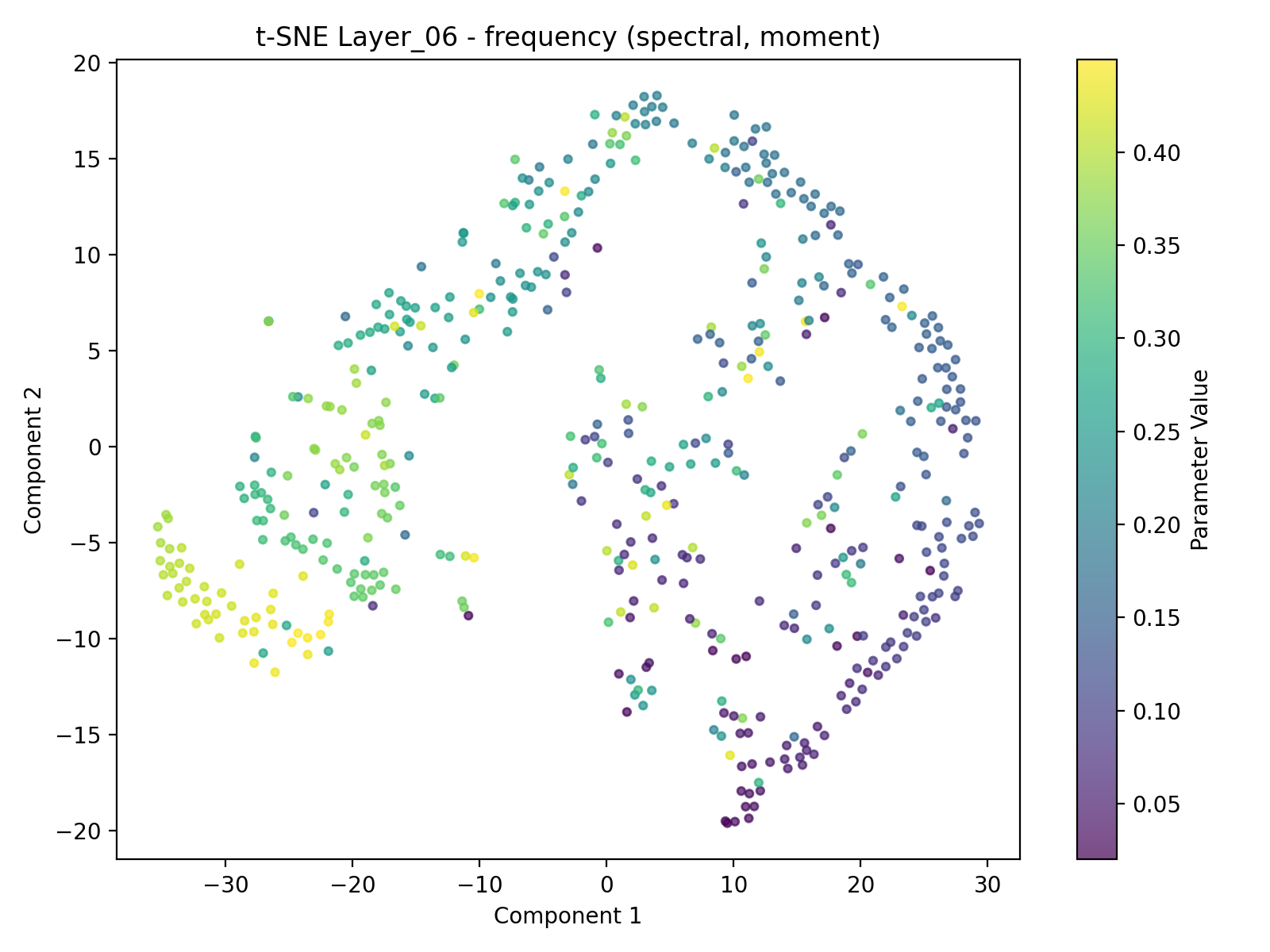}%
             {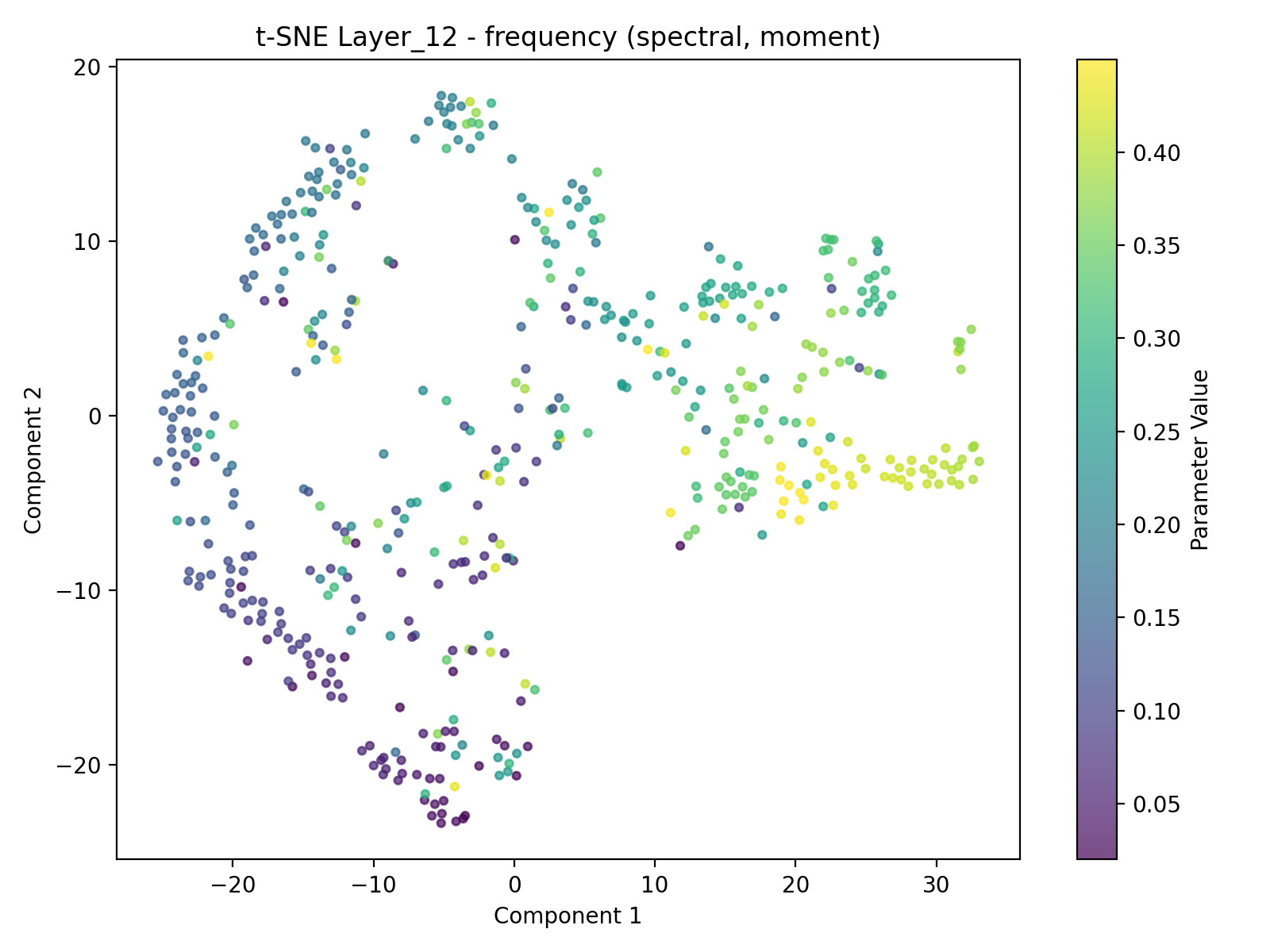}
  \caption{Spectral --- Moment --- Frequency --- t-SNE (Layers 00/06/12)}
\end{figure}

\begin{figure}[t]
  \centering
  \threeplots{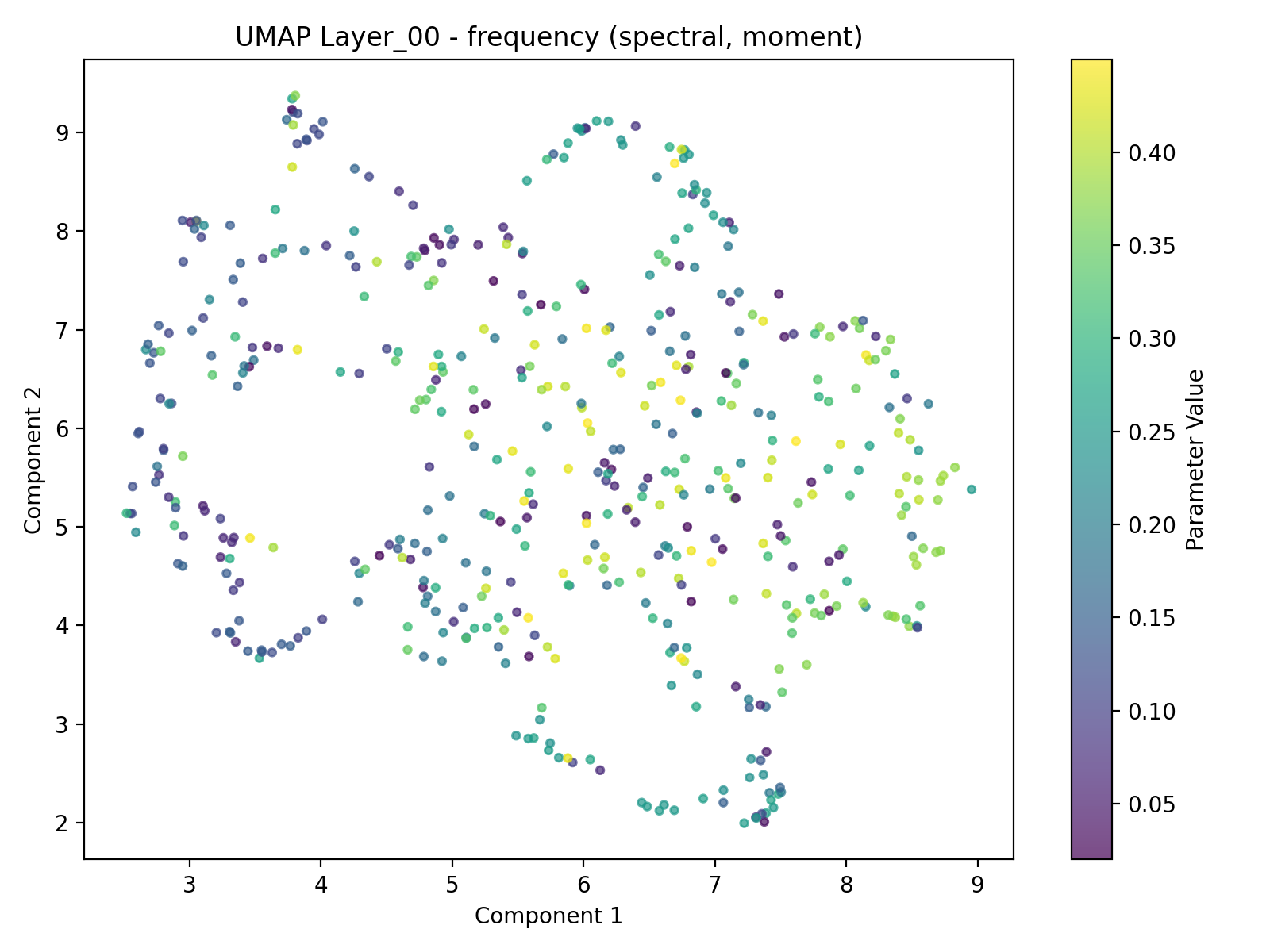}%
             {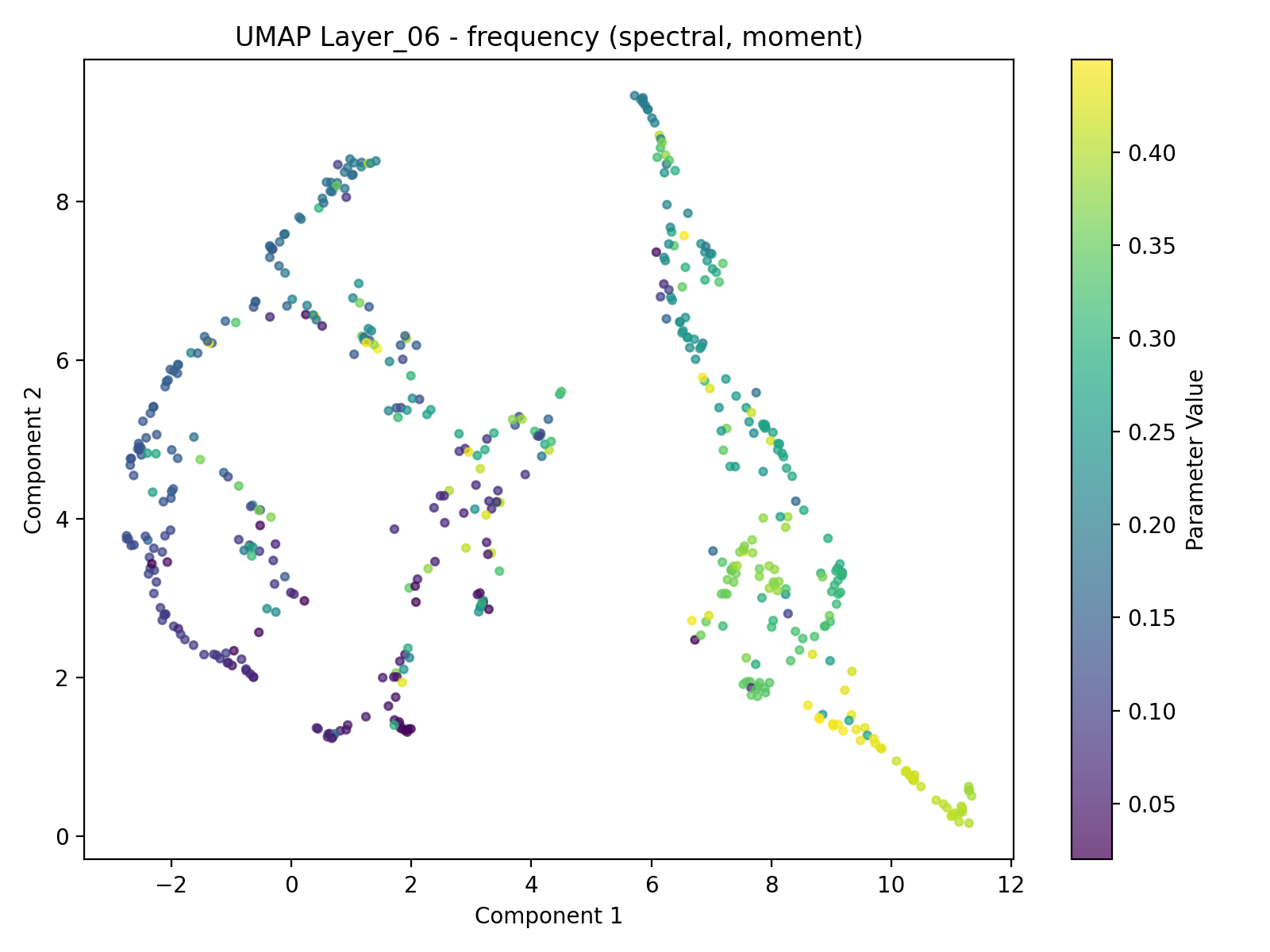}%
             {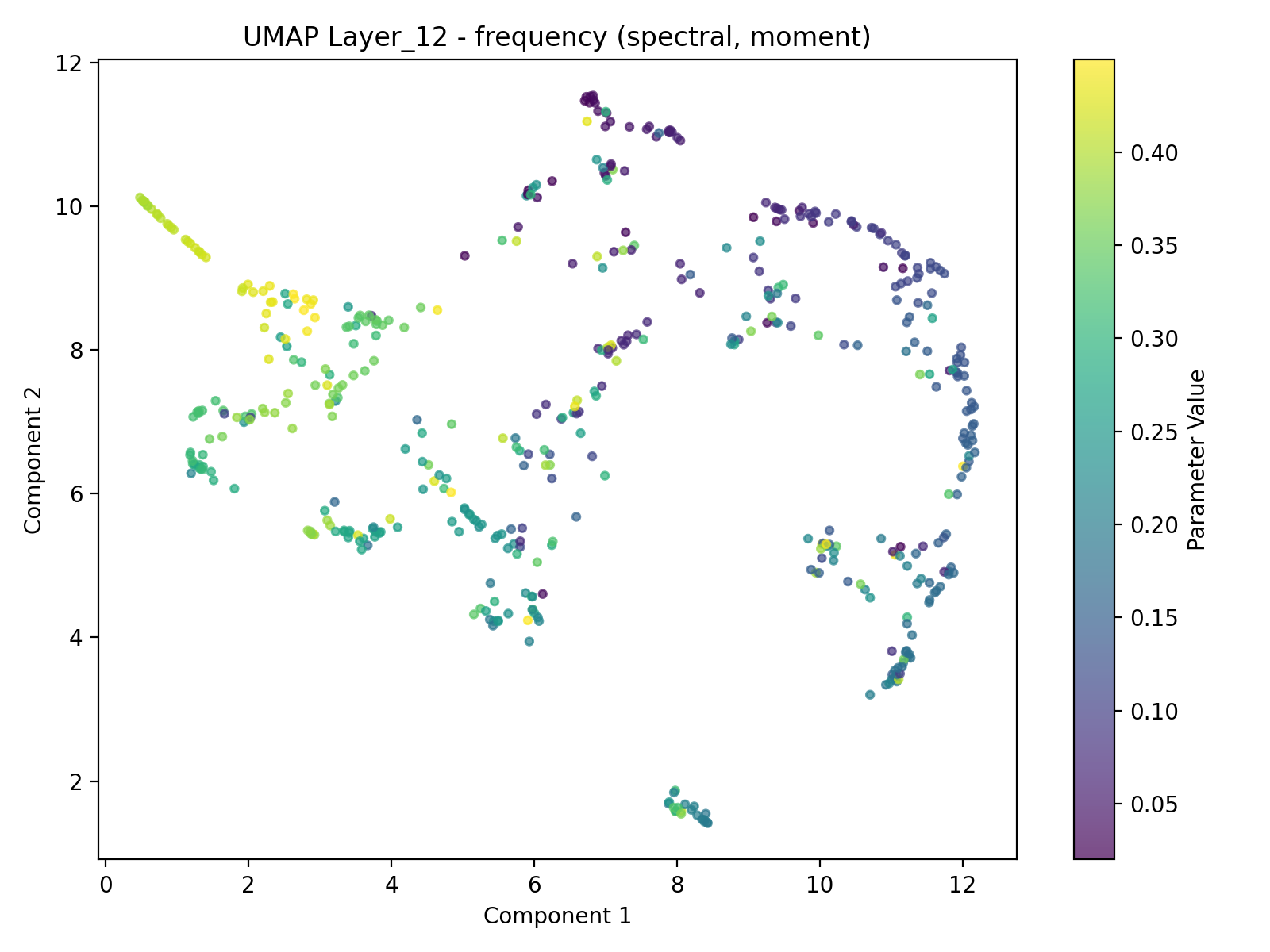}
  \caption{Spectral --- Moment --- Frequency --- UMAP (Layers 00/06/12)}
\end{figure}

% \begin{figure}[t]
%   \centering
%   \threeplots{figs/spectral_moment/spectral/moment_spectral_pca_Layer_00_phase.png}%
%              {figs/spectral_moment/spectral/moment_spectral_pca_Layer_06_phase.png}%
%              {figs/spectral_moment/spectral/moment_spectral_pca_Layer_12_phase.png}
%   \caption{Spectral --- Moment --- Phase --- PCA (Layers 00/06/12)}
% \end{figure}

% \begin{figure}[t]
%   \centering
%   \threeplots{figs/spectral_moment/spectral/moment_spectral_tsne_Layer_00_phase.png}%
%              {figs/spectral_moment/spectral/moment_spectral_tsne_Layer_06_phase.png}%
%              {figs/spectral_moment/spectral/moment_spectral_tsne_Layer_12_phase.png}
%   \caption{Spectral --- Moment --- Phase --- t-SNE (Layers 00/06/12)}
% \end{figure}

% \begin{figure}[t]
%   \centering
%   \threeplots{figs/spectral_moment/spectral/moment_spectral_umap_Layer_00_phase.png}%
%              {figs/spectral_moment/spectral/moment_spectral_umap_Layer_06_phase.png}%
%              {figs/spectral_moment/spectral/moment_spectral_umap_Layer_12_phase.png}
%   \caption{Spectral --- Moment --- Phase --- UMAP (Layers 00/06/12)}
% \end{figure}
\newpage
% ---------------- Time-Warped ----------------
\subsection{Time-Warped Sinusoid}

\paragraph{Moment (freq).}
\begin{figure}[t]
  \centering
  \threeplots{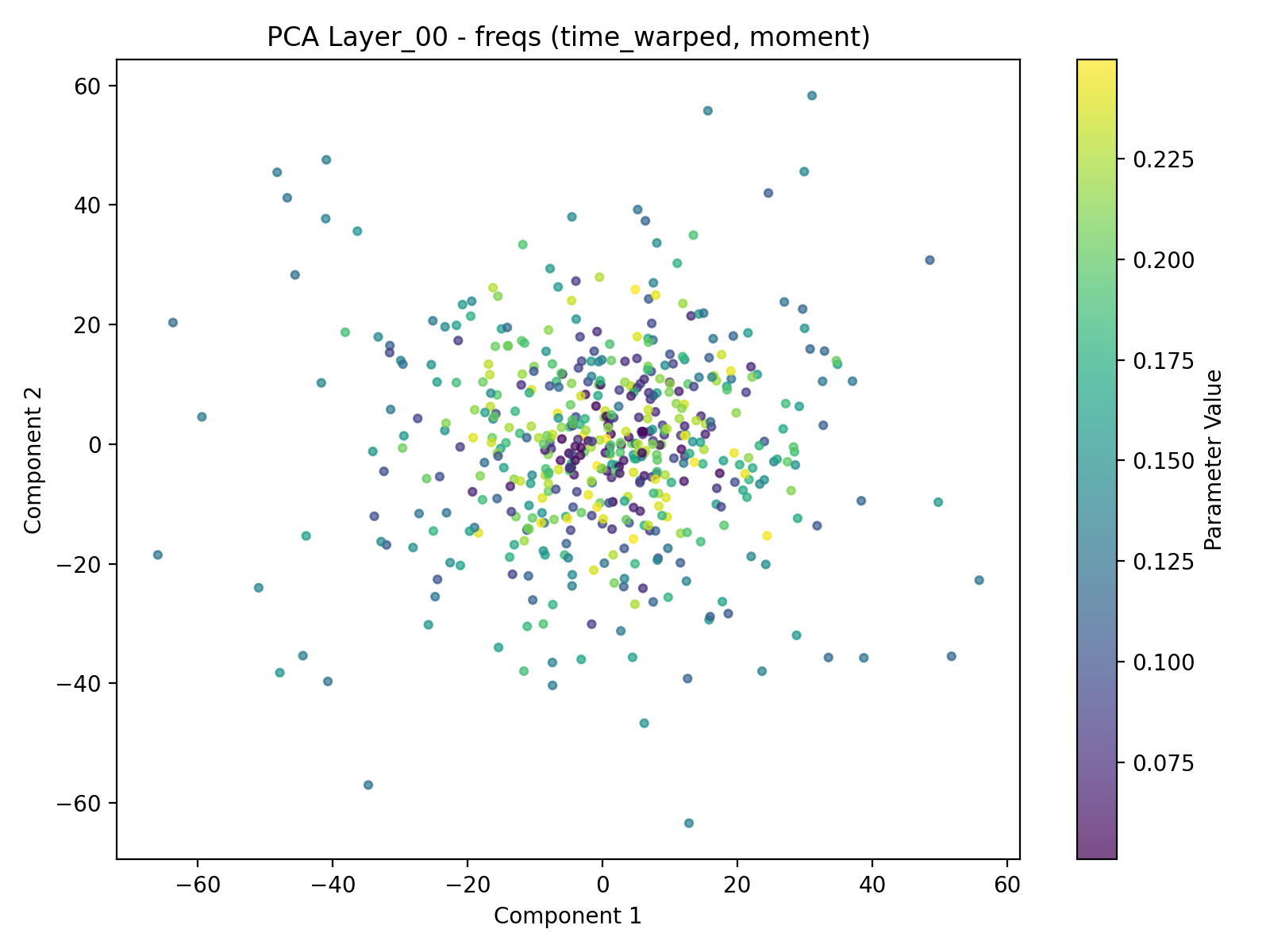}%
             {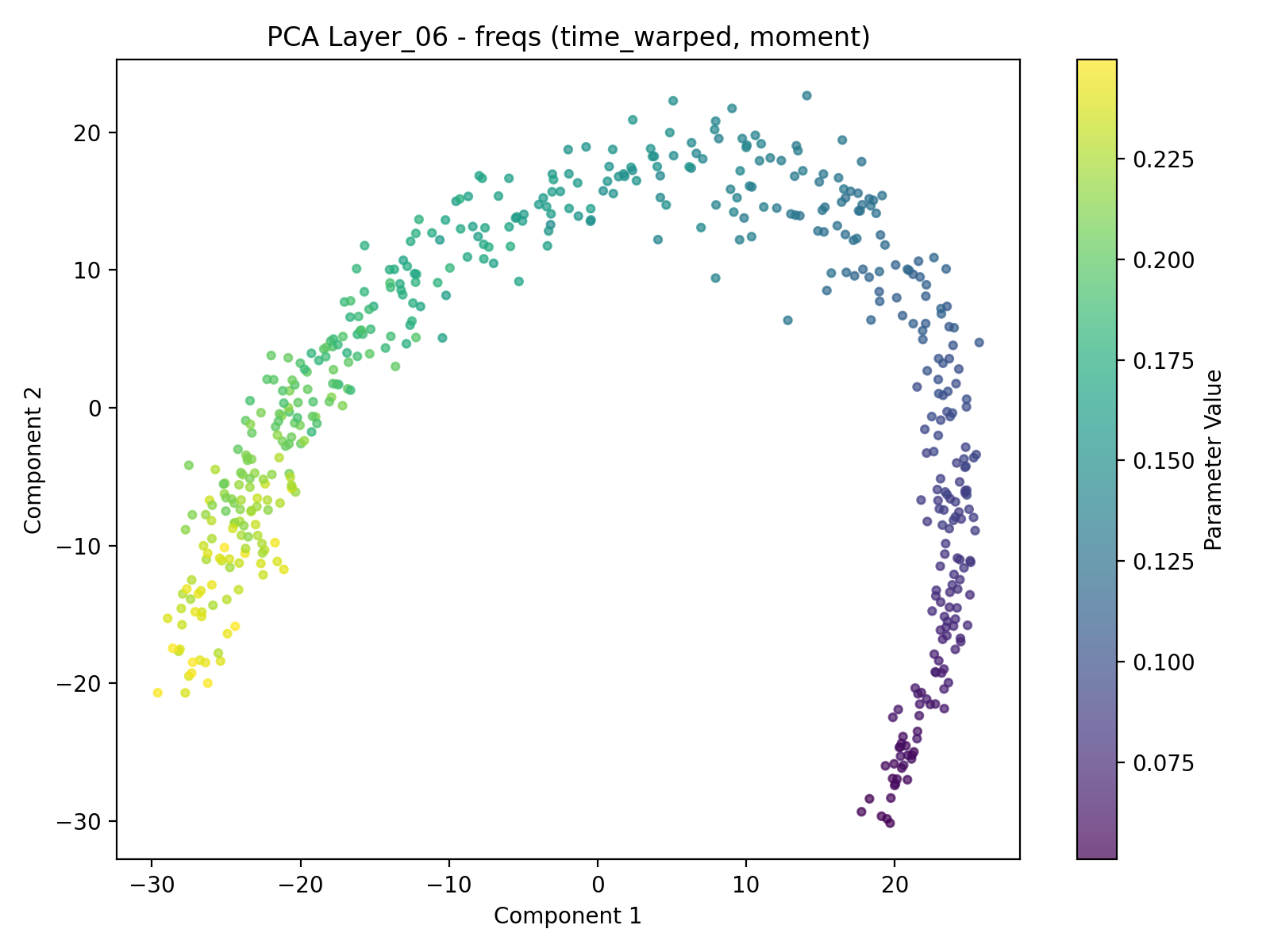}%
             {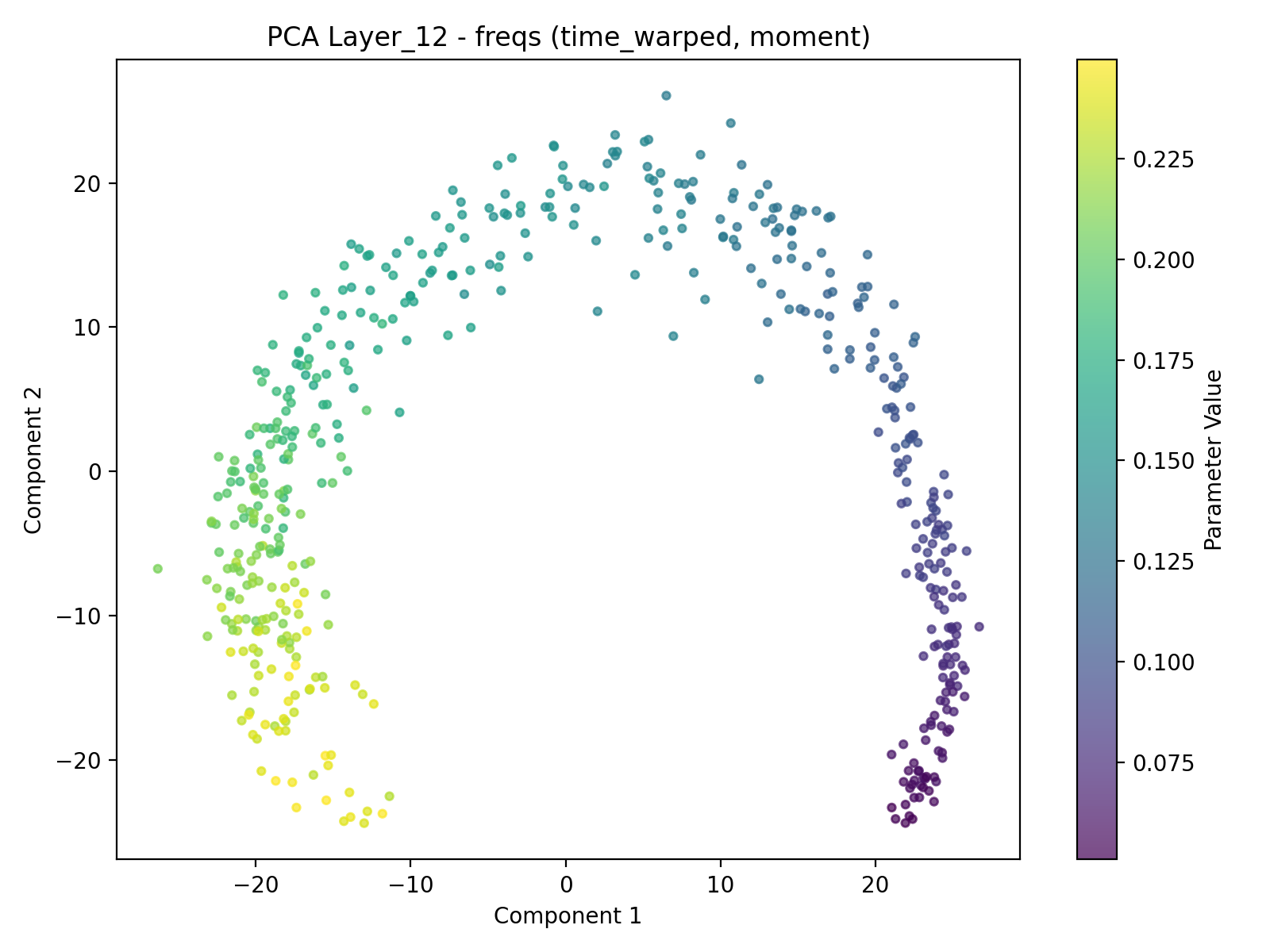}
  \caption{Time-Warped --- Moment --- Freqs --- PCA (Layers 00/06/12)}
\end{figure}

\begin{figure}[t]
  \centering
  \threeplots{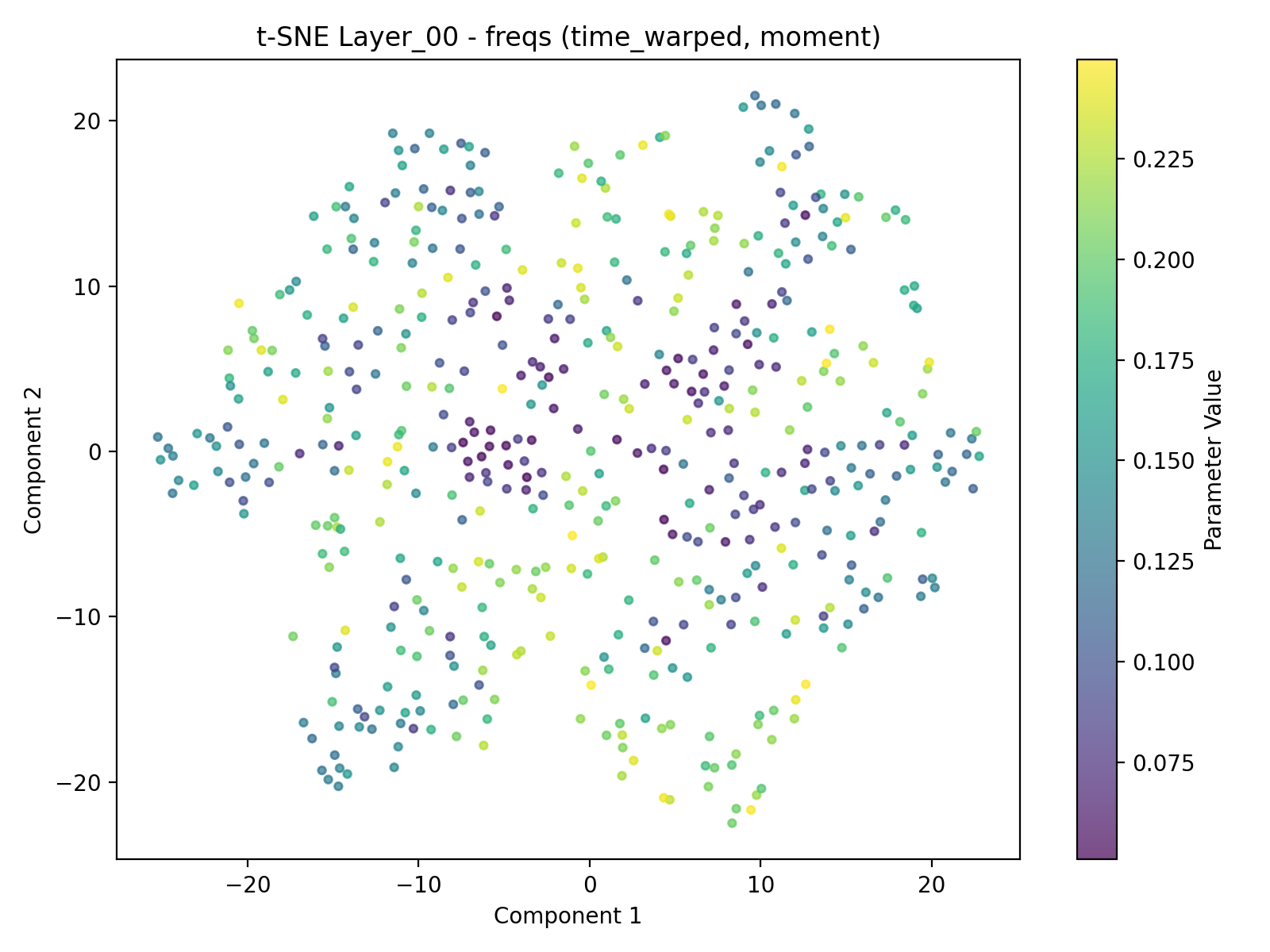}%
             {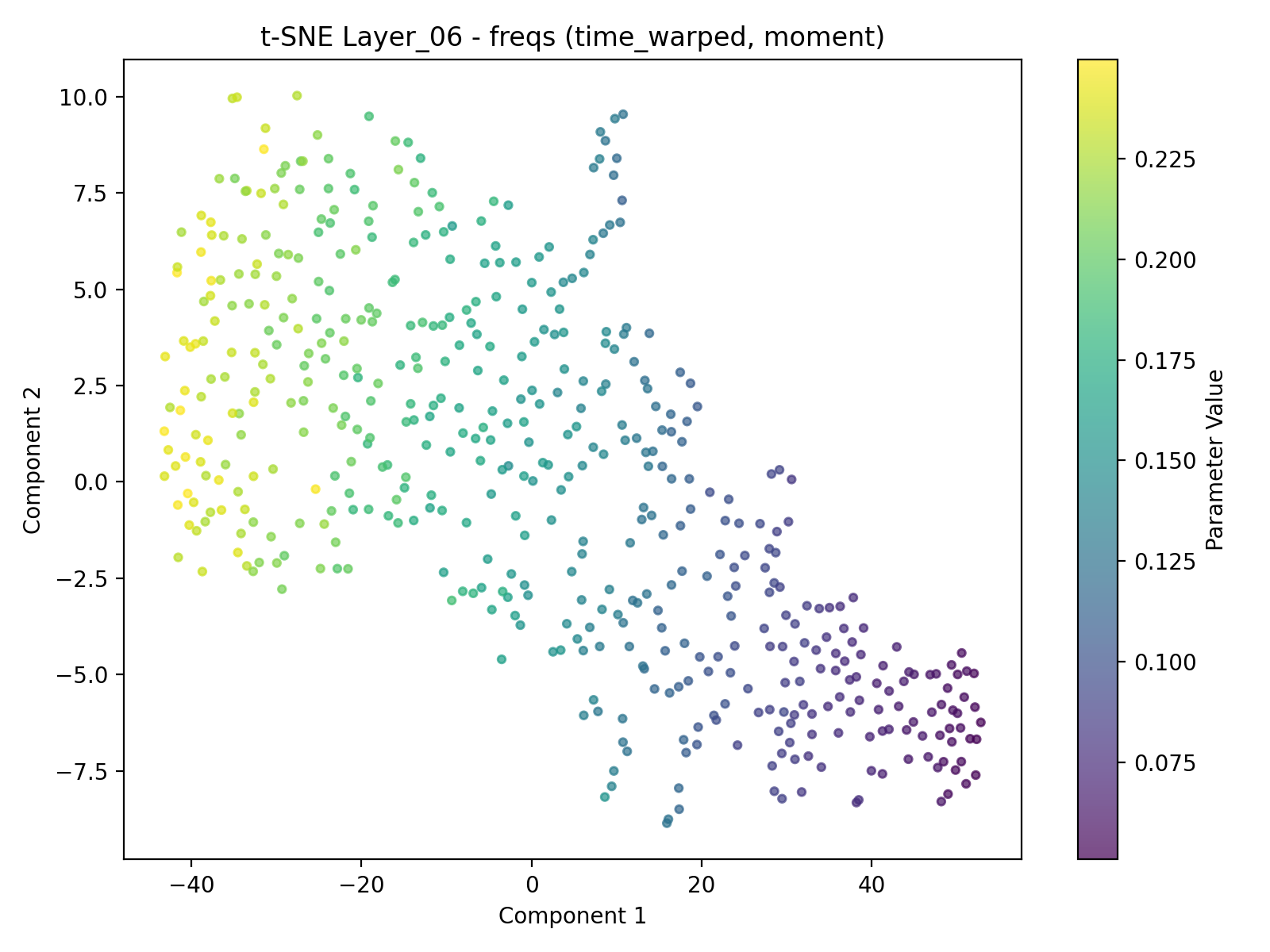}%
             {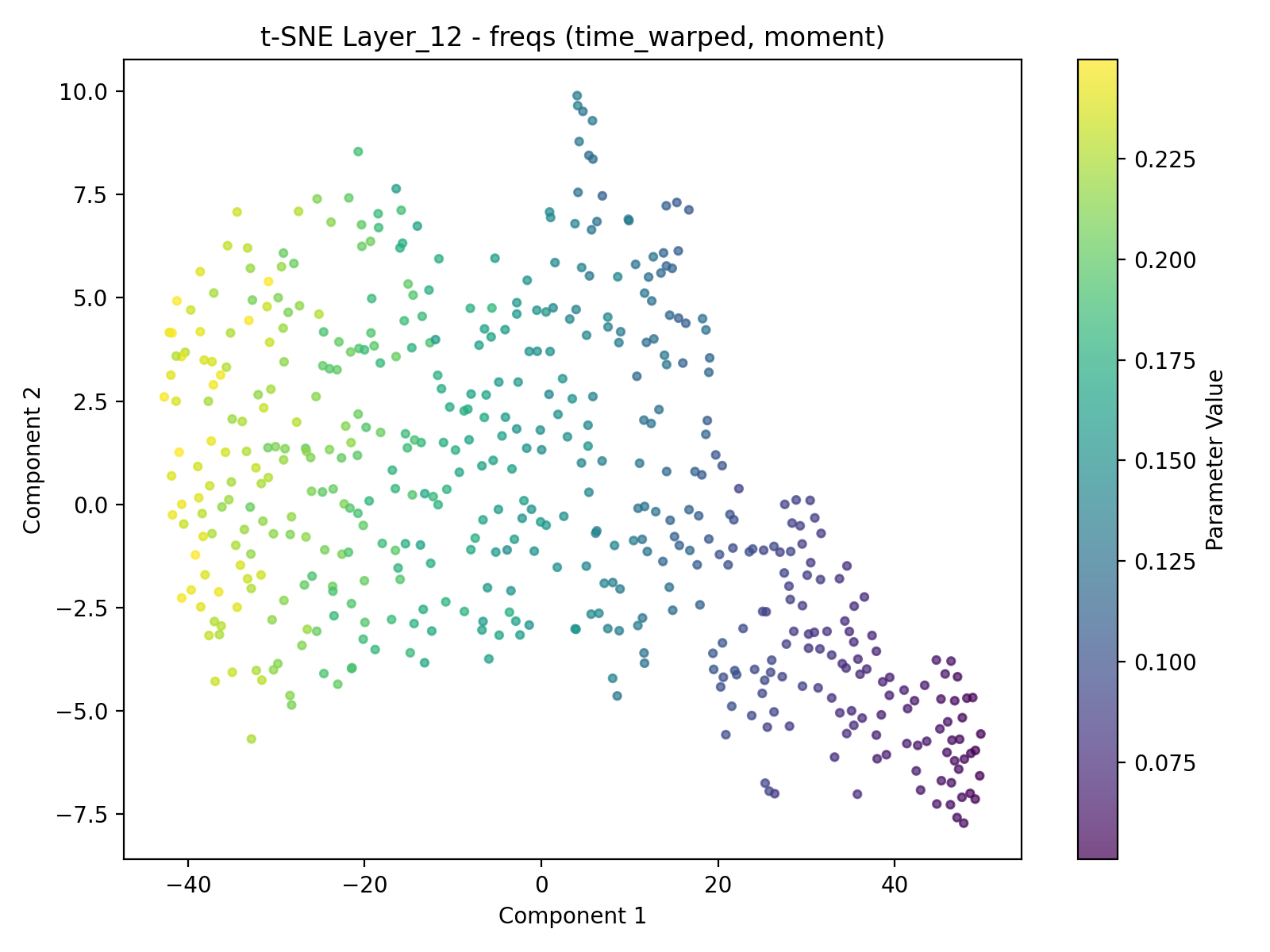}
  \caption{Time-Warped --- Moment --- Freqs --- t-SNE (Layers 00/06/12)}
\end{figure}

\begin{figure}[t]
  \centering
  \threeplots{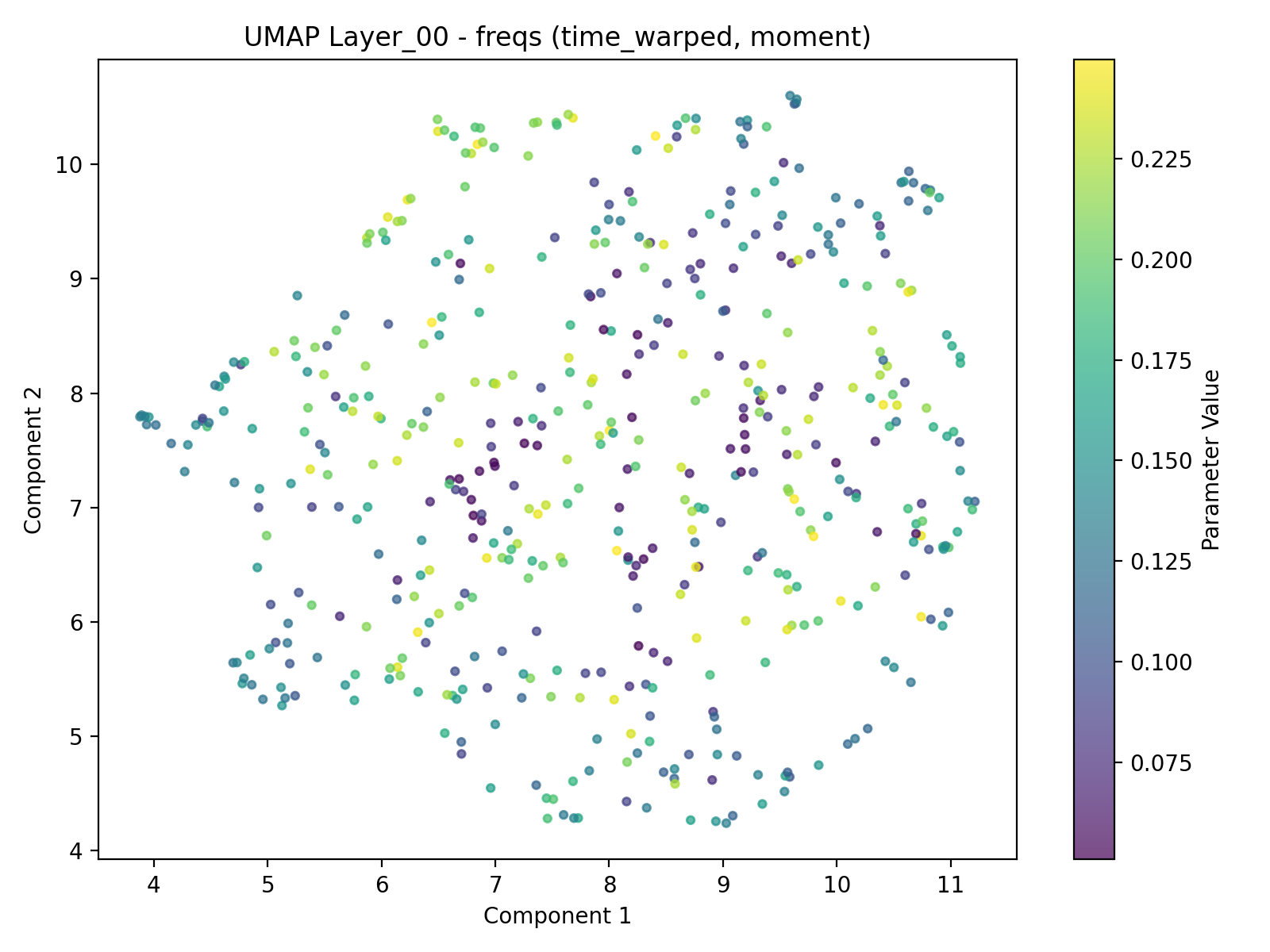}%
             {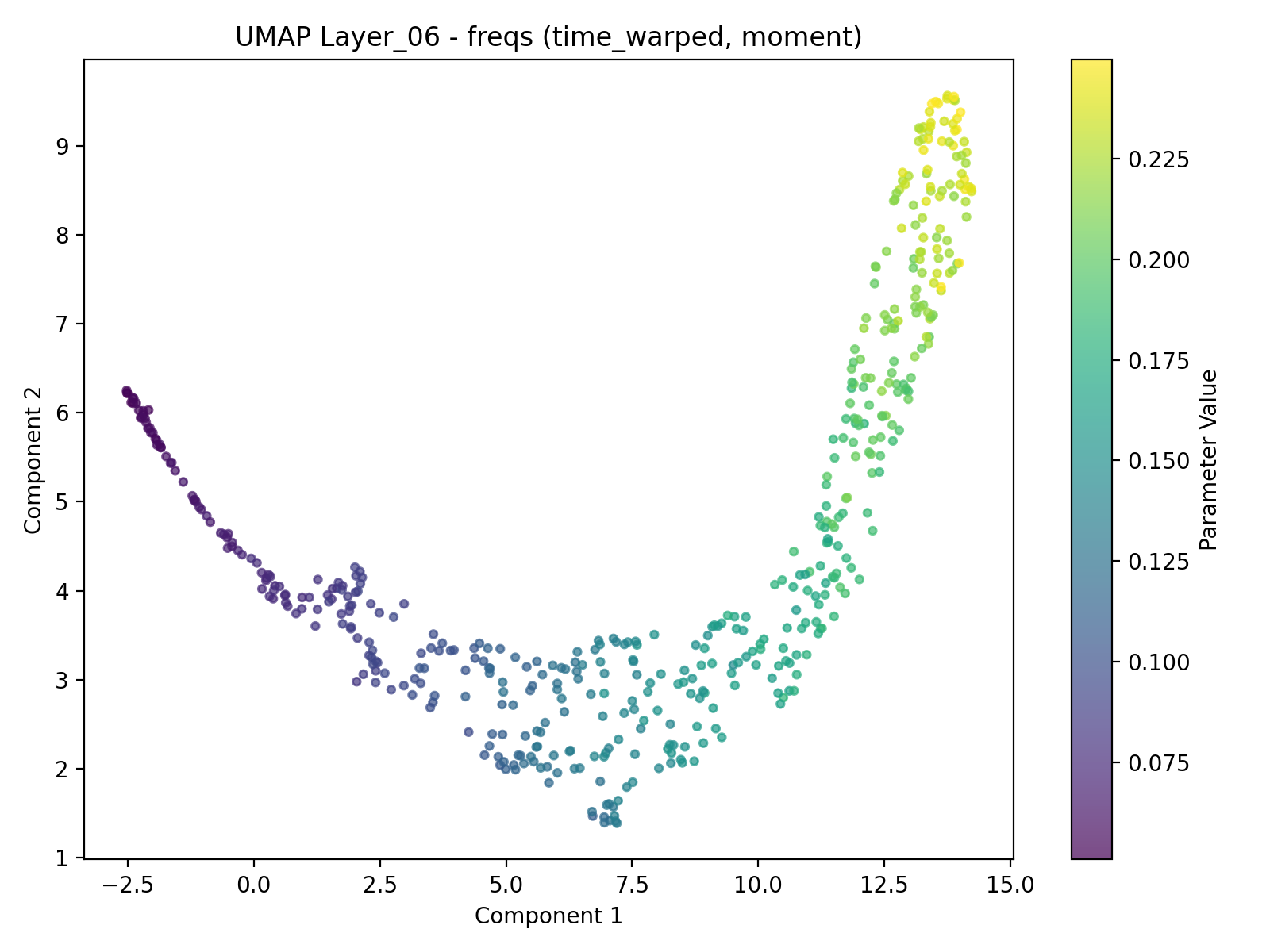}%
             {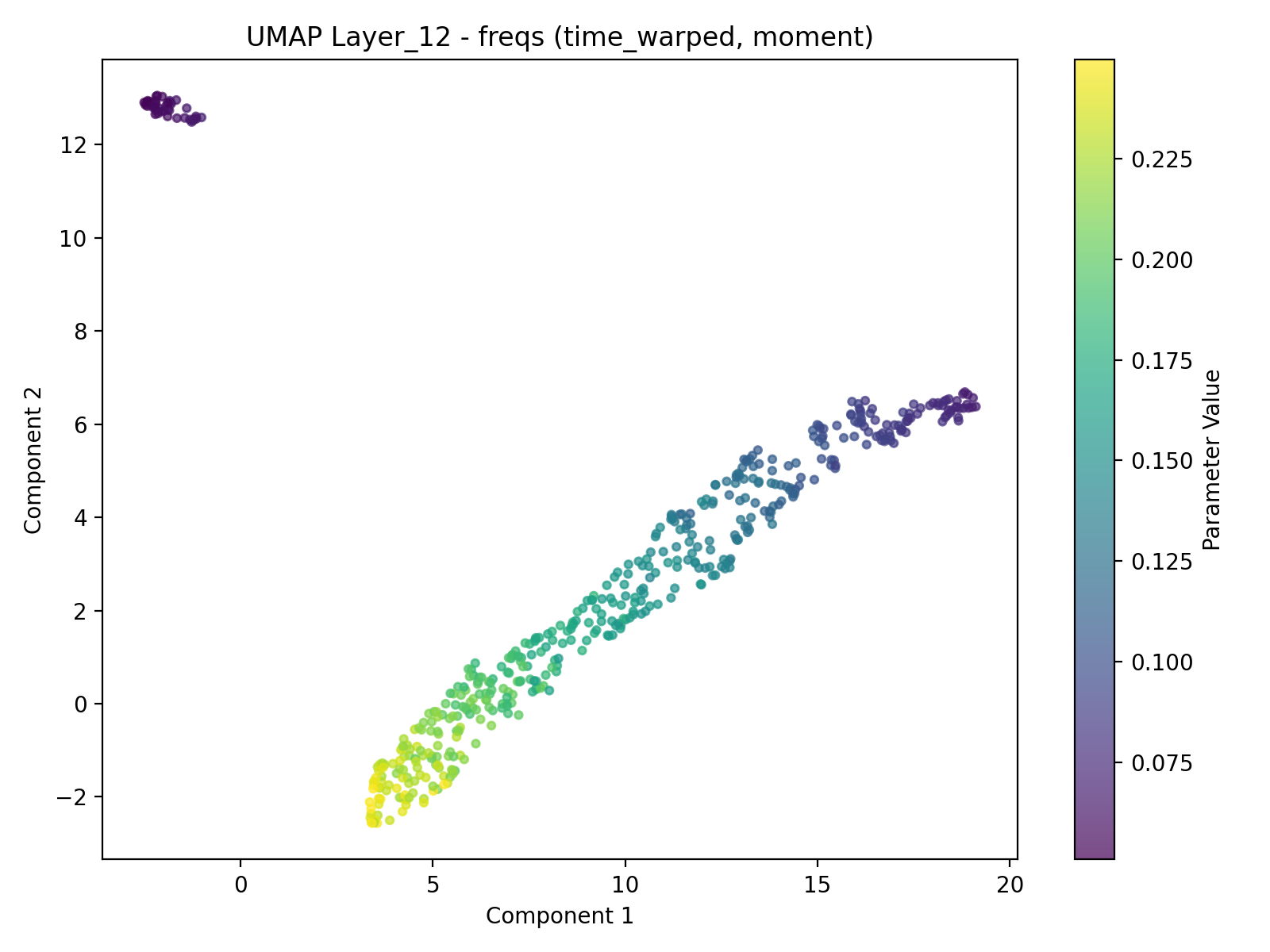}
  \caption{Time-Warped --- Moment --- Freqs --- UMAP (Layers 00/06/12)}
\end{figure}

% \begin{figure}[t]
%   \centering
%   \threeplots{figs/time_warped_moment/time_warped/moment_time_warped_pca_Layer_00_phases.png}%
%              {figs/time_warped_moment/time_warped/moment_time_warped_pca_Layer_06_phases.png}%
%              {figs/time_warped_moment/time_warped/moment_time_warped_pca_Layer_12_phases.png}
%   \caption{Time-Warped --- Moment --- Phases --- PCA (Layers 00/06/12)}
% \end{figure}

% \begin{figure}[t]
%   \centering
%   \threeplots{figs/time_warped_moment/time_warped/moment_time_warped_tsne_Layer_00_phases.png}%
%              {figs/time_warped_moment/time_warped/moment_time_warped_tsne_Layer_06_phases.png}%
%              {figs/time_warped_moment/time_warped/moment_time_warped_tsne_Layer_12_phases.png}
%   \caption{Time-Warped --- Moment --- Phases --- t-SNE (Layers 00/06/12)}
% \end{figure}

% \begin{figure}[t]
%   \centering
%   \threeplots{figs/time_warped_moment/time_warped/moment_time_warped_umap_Layer_00_phases.png}%
%              {figs/time_warped_moment/time_warped/moment_time_warped_umap_Layer_06_phases.png}%
%              {figs/time_warped_moment/time_warped/moment_time_warped_umap_Layer_12_phases.png}
%   \caption{Time-Warped --- Moment --- Phases --- UMAP (Layers 00/06/12)}
% \end{figure}
\newpage
\paragraph{Chronos (freq).}
\begin{figure}[t]
  \centering
  \threeplots{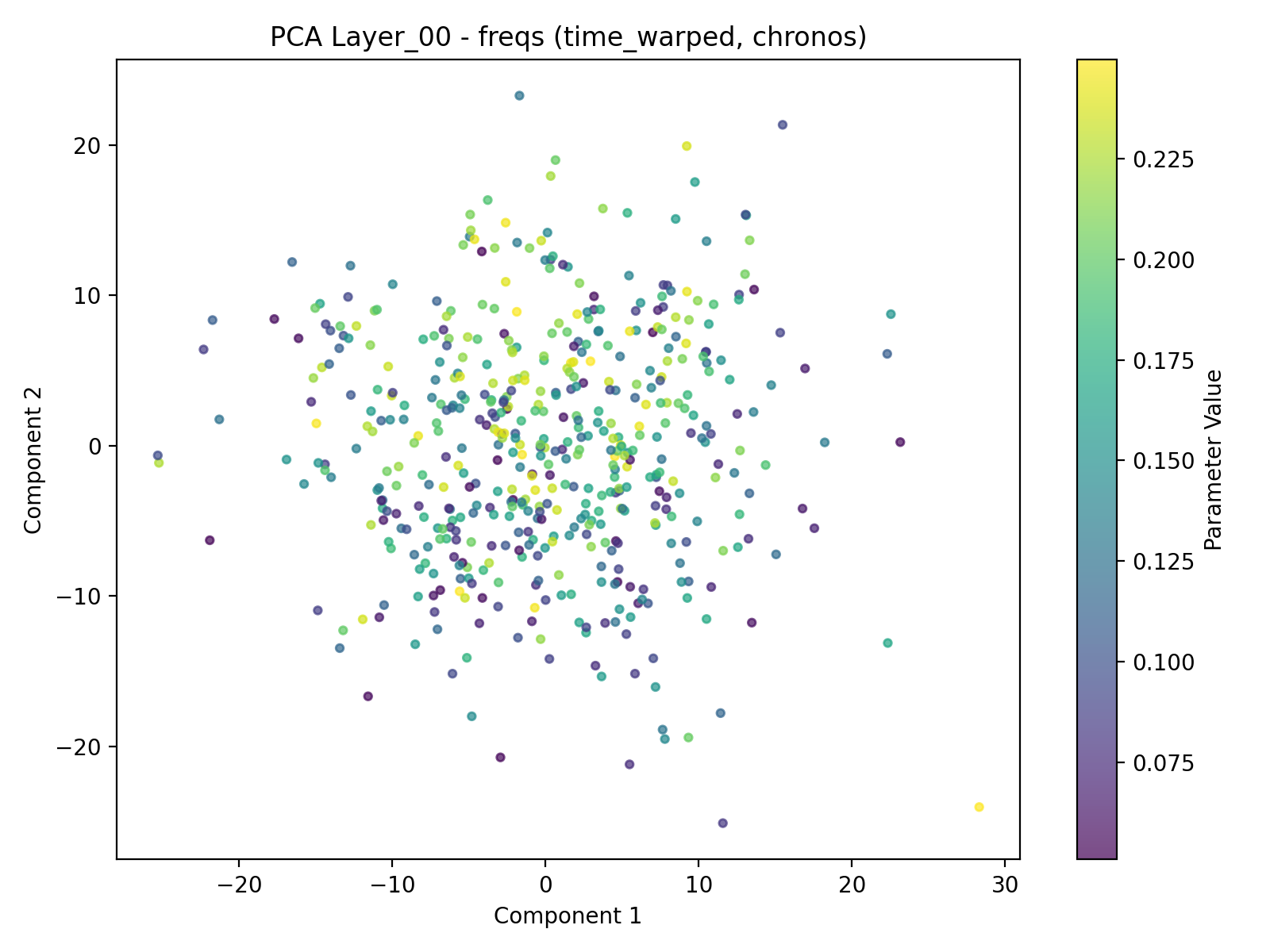}%
             {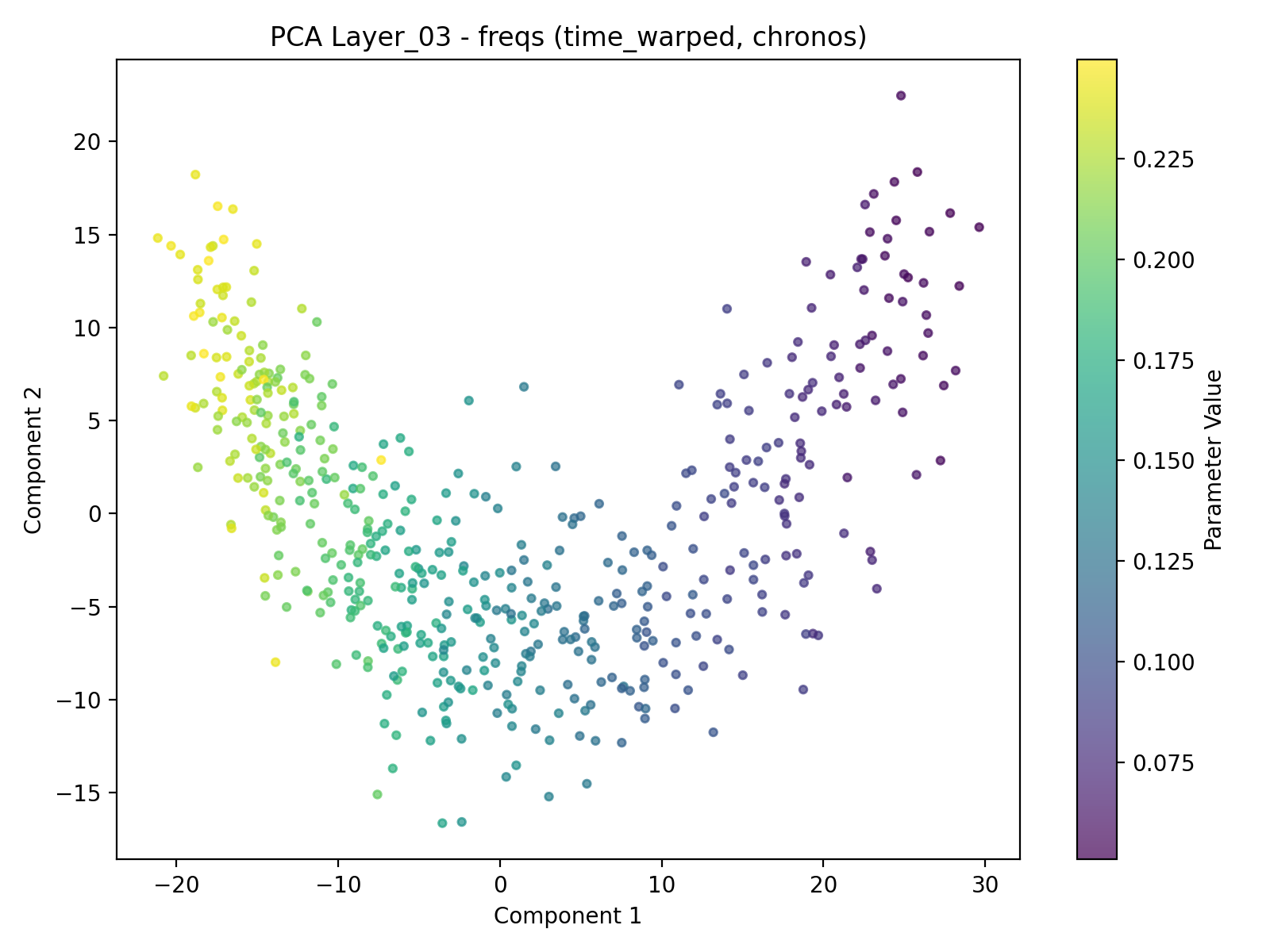}%
             {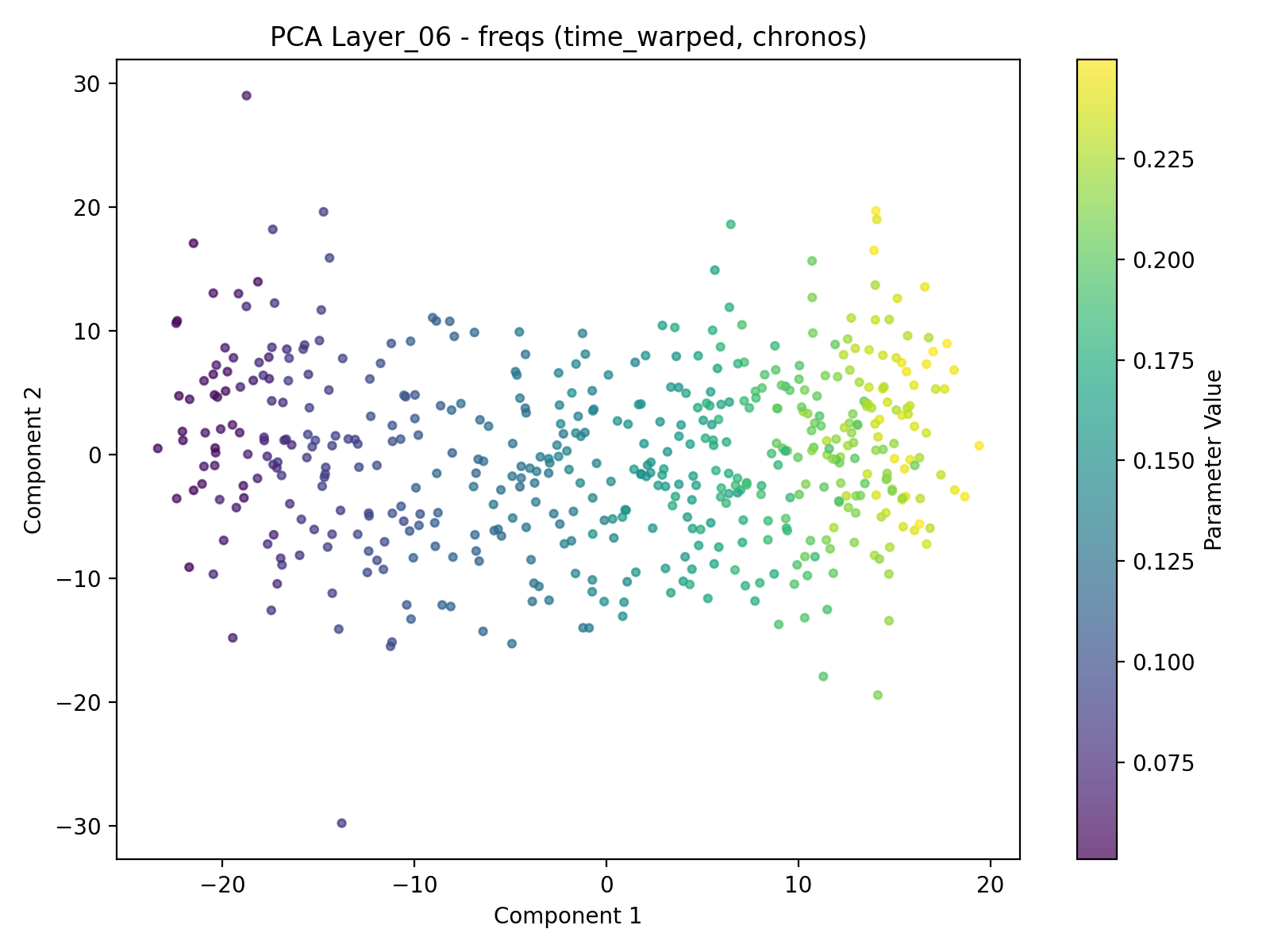}
  \caption{Time-Warped --- Chronos --- Freqs --- PCA (Layers 00/03/06)}
\end{figure}

\begin{figure}[t]
  \centering
  \threeplots{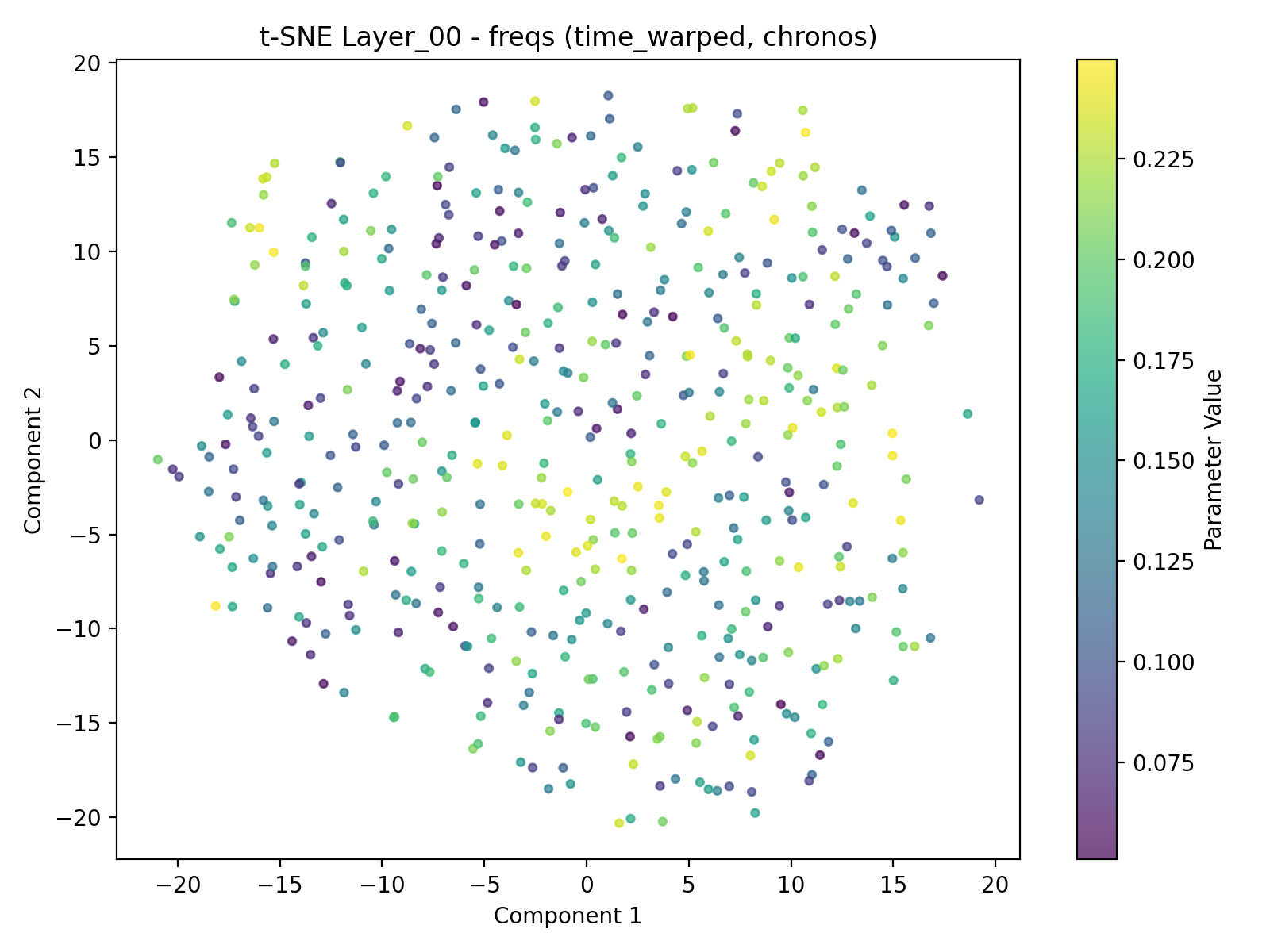}%
             {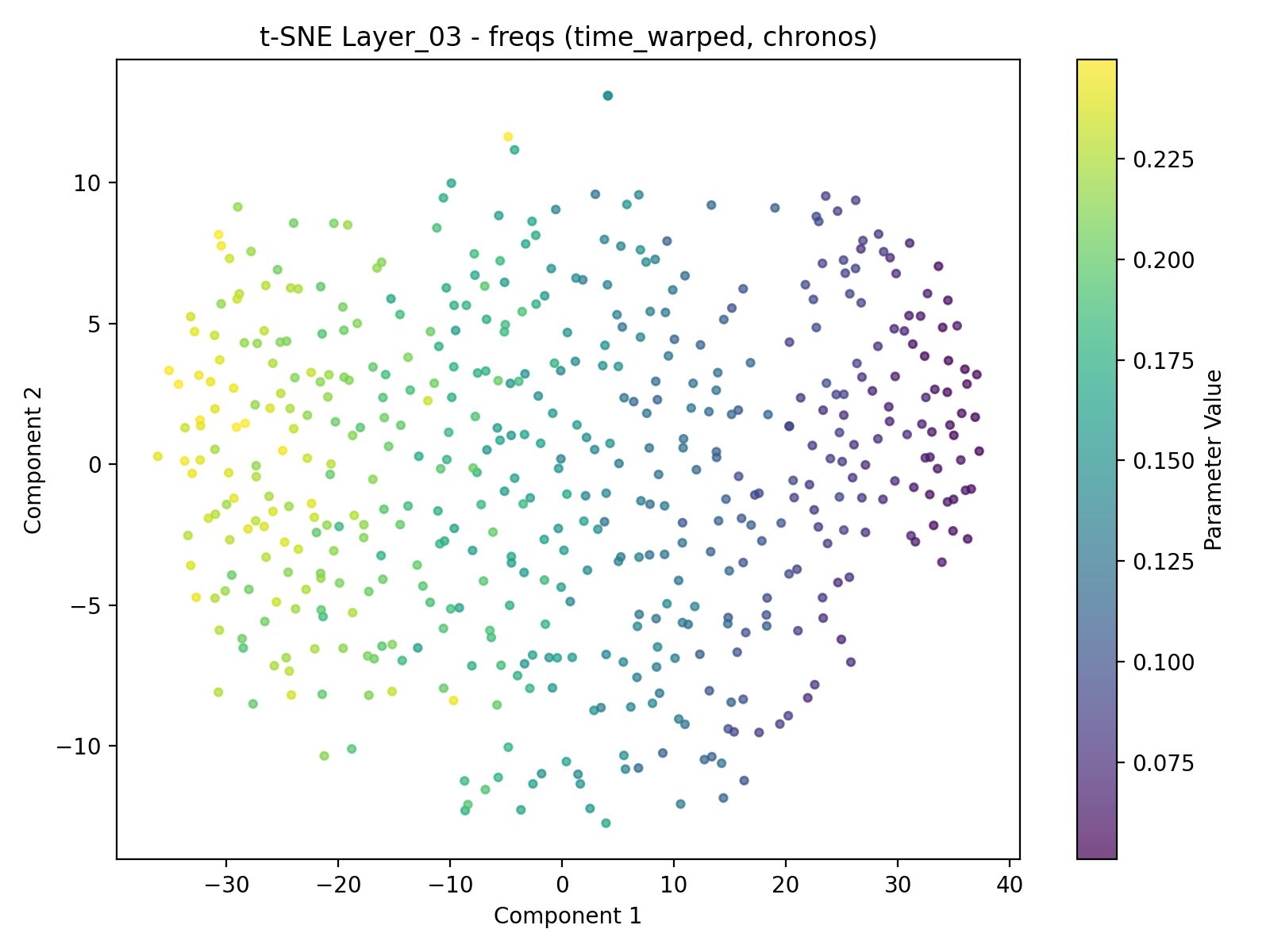}%
             {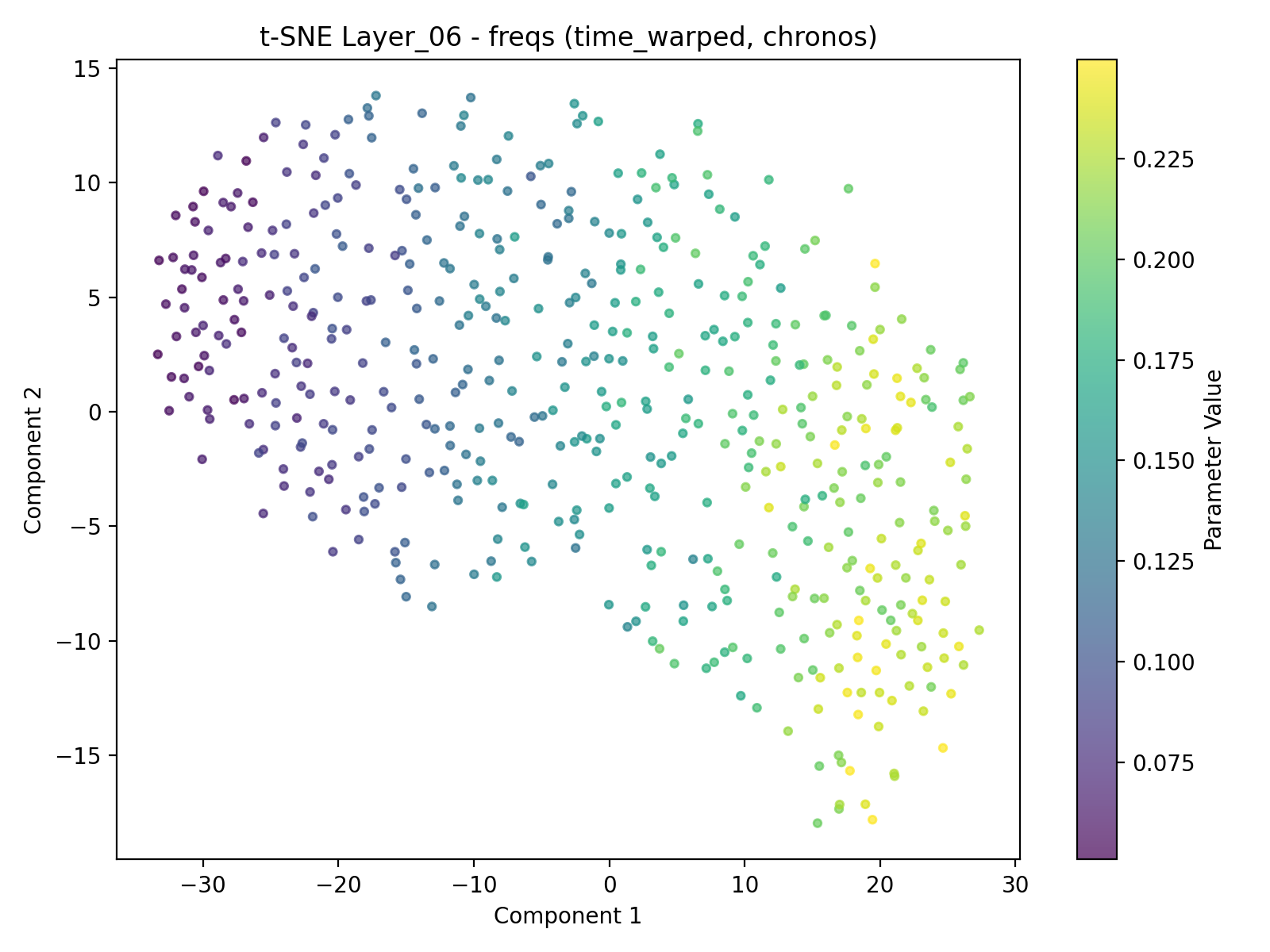}
  \caption{Time-Warped --- Chronos --- Freqs --- t-SNE (Layers 00/03/06)}
\end{figure}

\begin{figure}[t]
  \centering
  \threeplots{figs/time_warped_chronos/time_warped/chronos_time_warped_umap_Layer_00_freqs.png}%
             {figs/time_warped_chronos/time_warped/chronos_time_warped_umap_Layer_03_freqs.png}%
             {figs/time_warped_chronos/time_warped/chronos_time_warped_umap_Layer_06_freqs.png}
  \caption{Time-Warped --- Chronos --- Freqs --- UMAP (Layers 00/03/06)}
\end{figure}

% \begin{figure}[t]
%   \centering
%   \threeplots{figs/time_warped_chronos/time_warped/chronos_time_warped_pca_Layer_00_phases.png}%
%              {figs/time_warped_chronos/time_warped/chronos_time_warped_pca_Layer_03_phases.png}%
%              {figs/time_warped_chronos/time_warped/chronos_time_warped_pca_Layer_06_phases.png}
%   \caption{Time-Warped --- Chronos --- Phases --- PCA (Layers 00/03/06)}
% \end{figure}

% \begin{figure}[t]
%   \centering
%   \threeplots{figs/time_warped_chronos/time_warped/chronos_time_warped_tsne_Layer_00_phases.png}%
%              {figs/time_warped_chronos/time_warped/chronos_time_warped_tsne_Layer_03_phases.png}%
%              {figs/time_warped_chronos/time_warped/chronos_time_warped_tsne_Layer_06_phases.png}
%   \caption{Time-Warped --- Chronos --- Phases --- t-SNE (Layers 00/03/06)}
% \end{figure}

% \begin{figure}[t]
%   \centering
%   \threeplots{figs/time_warped_chronos/time_warped/chronos_time_warped_umap_Layer_00_phases.png}%
%              {figs/time_warped_chronos/time_warped/chronos_time_warped_umap_Layer_03_phases.png}%
%              {figs/time_warped_chronos/time_warped/chronos_time_warped_umap_Layer_06_phases.png}
%   \caption{Time-Warped --- Chronos --- Phases --- UMAP (Layers 00/03/06)}
% \end{figure}
\newpage
% ---------------- Trend ----------------
\subsection{Deterministic Trend}

\paragraph{Moment (slope).}
% \begin{figure}[t]
%   \centering
%   \threeplots{figs/trend_deterministic_moment/trend_deterministic/moment_trend_deterministic_pca_Layer_00_slopes.png}%
%              {figs/trend_deterministic_moment/trend_deterministic/moment_trend_deterministic_pca_Layer_06_slopes.png}%
%              {figs/trend_deterministic_moment/trend_deterministic/moment_trend_deterministic_pca_Layer_12_slopes.png}
%   \caption{Trend --- Moment --- PCA (Layers 00/06/12)}
% \end{figure}

% \begin{figure}[t]
%   \centering
%   \threeplots{figs/trend_deterministic_moment/trend_deterministic/moment_trend_deterministic_tsne_Layer_00_slopes.png}%
%              {figs/trend_deterministic_moment/trend_deterministic/moment_trend_deterministic_tsne_Layer_06_slopes.png}%
%              {figs/trend_deterministic_moment/trend_deterministic/moment_trend_deterministic_tsne_Layer_12_slopes.png}
%   \caption{Trend --- Moment --- t-SNE (Layers 00/06/12)}
% \end{figure}

\begin{figure}[t]
  \centering
  \threeplots{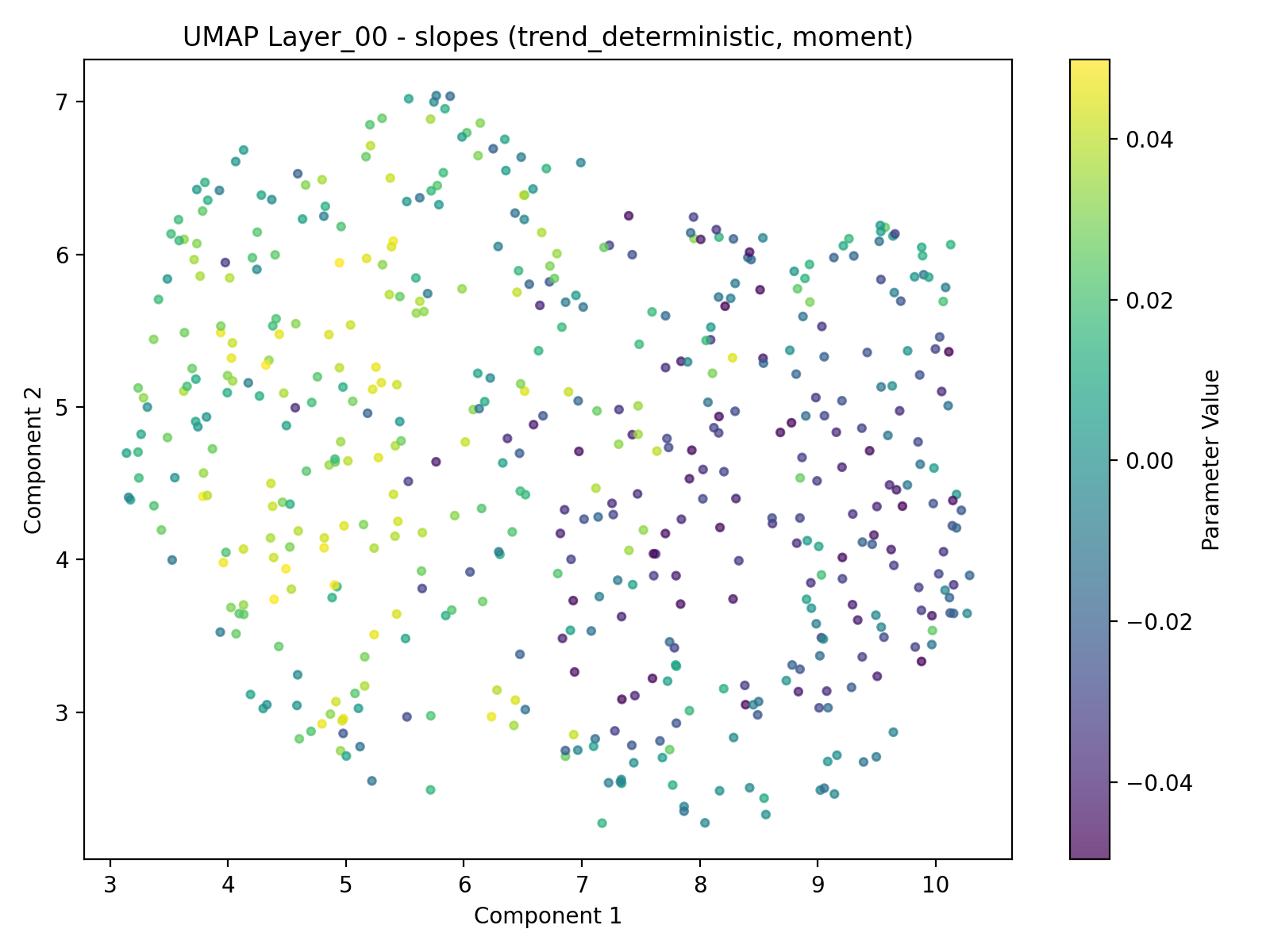}%
             {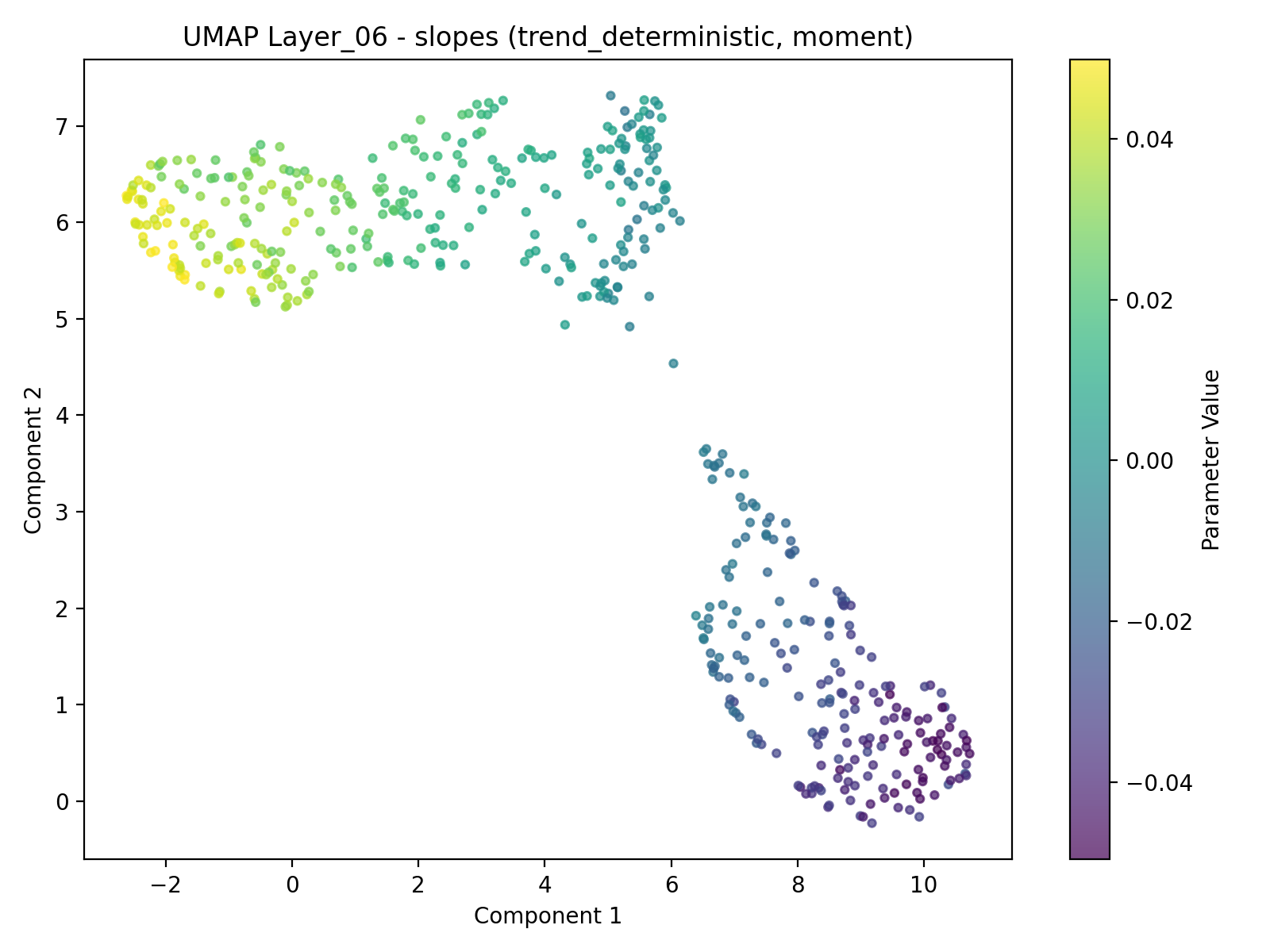}%
             {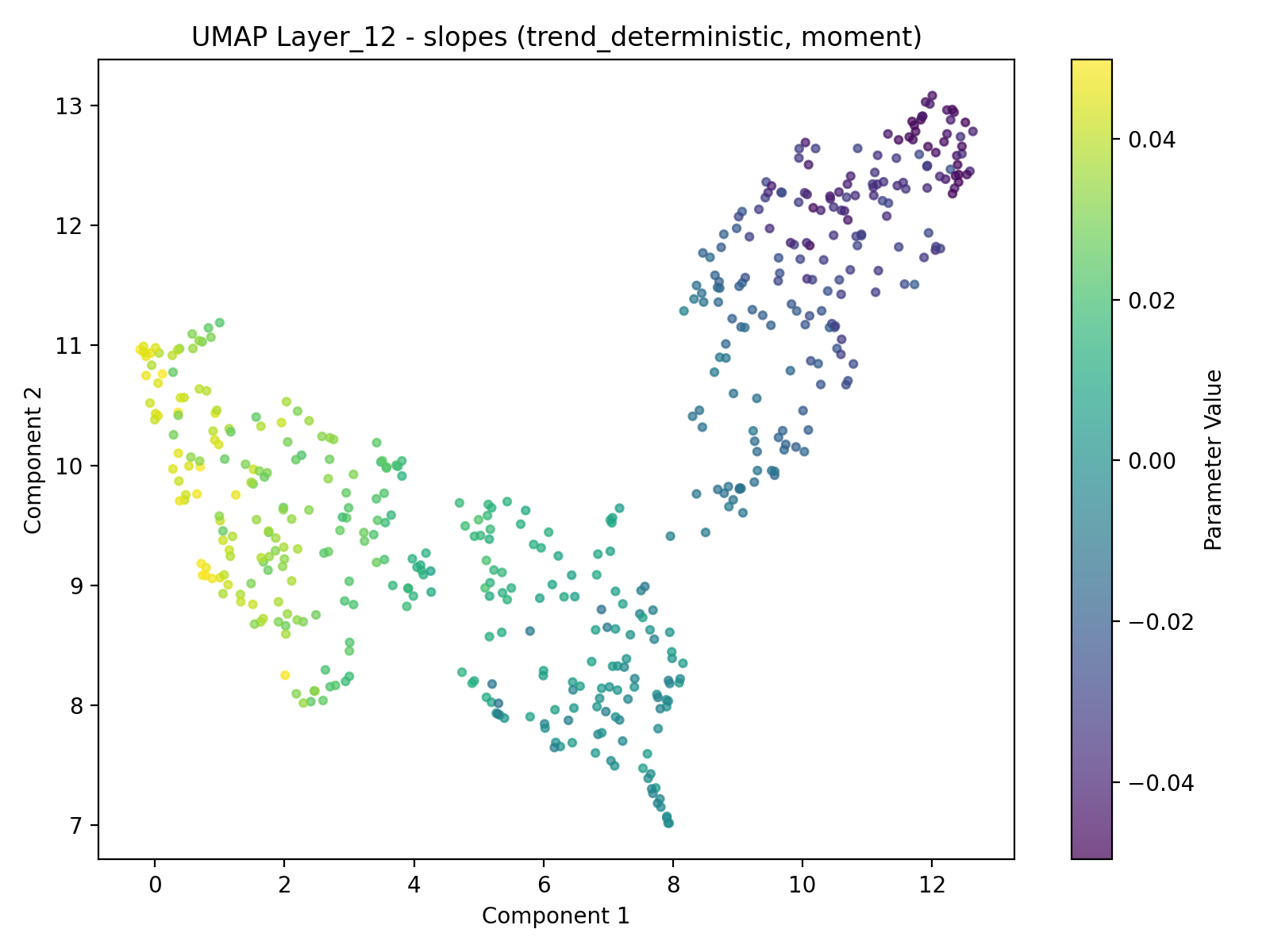}
  \caption{Trend --- Moment --- UMAP (Layers 00/06/12)}
\end{figure}
\newpage
\paragraph{Chronos (slope).}
% \begin{figure}[t]
%   \centering
%   \threeplots{figs/trend_deterministic_chronos/trend_deterministic/chronos_trend_deterministic_pca_Layer_00_slopes.png}%
%              {figs/trend_deterministic_chronos/trend_deterministic/chronos_trend_deterministic_pca_Layer_03_slopes.png}%
%              {figs/trend_deterministic_chronos/trend_deterministic/chronos_trend_deterministic_pca_Layer_06_slopes.png}
%   \caption{Trend --- Chronos --- PCA (Layers 00/03/06)}
% \end{figure}

% \begin{figure}[t]
%   \centering
%   \threeplots{figs/trend_deterministic_chronos/trend_deterministic/chronos_trend_deterministic_tsne_Layer_00_slopes.png}%
%              {figs/trend_deterministic_chronos/trend_deterministic/chronos_trend_deterministic_tsne_Layer_03_slopes.png}%
%              {figs/trend_deterministic_chronos/trend_deterministic/chronos_trend_deterministic_tsne_Layer_06_slopes.png}
%   \caption{Trend --- Chronos --- t-SNE (Layers 00/03/06)}
% \end{figure}

\begin{figure}[t]
  \centering
  \threeplots{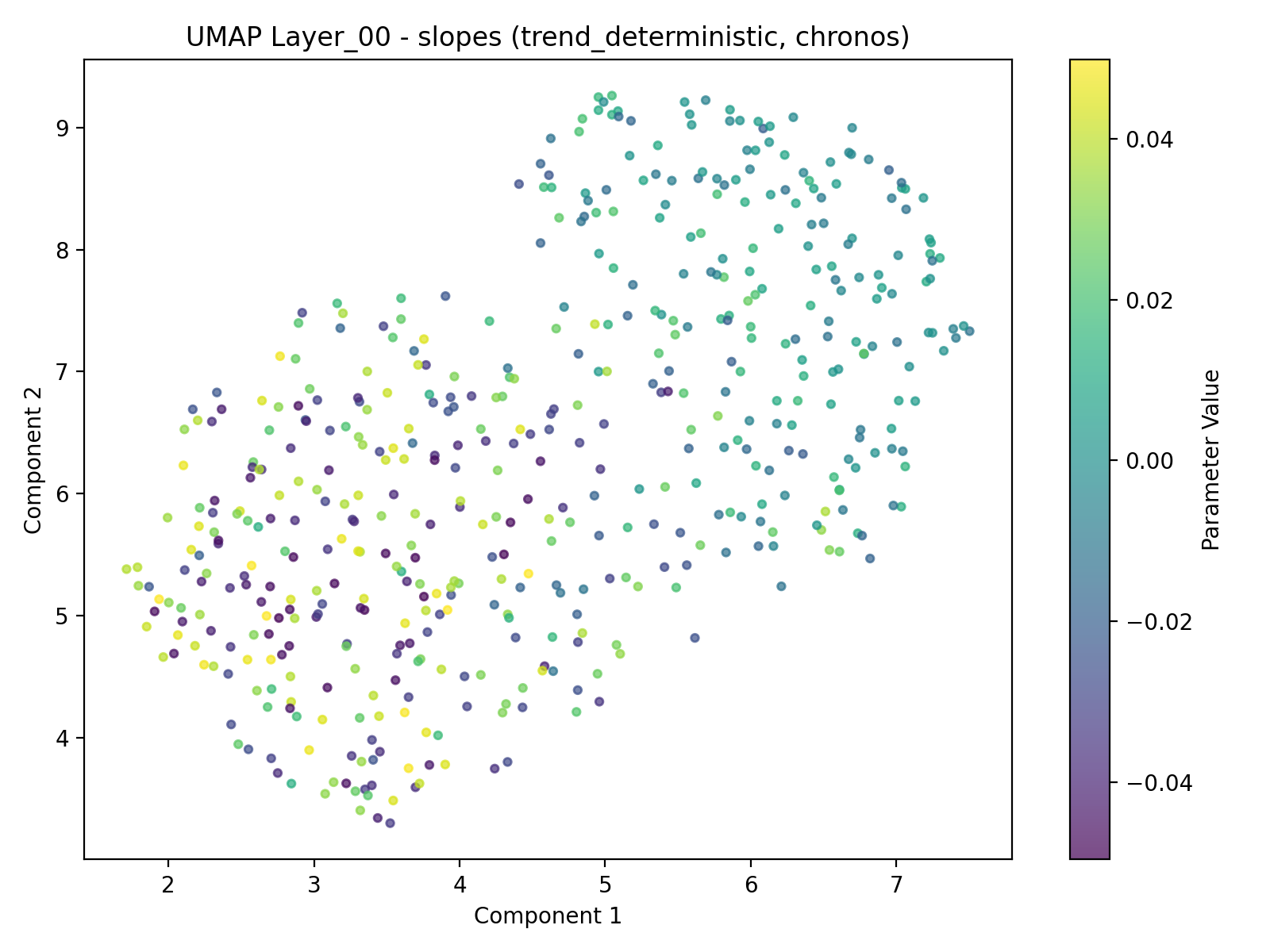}%
             {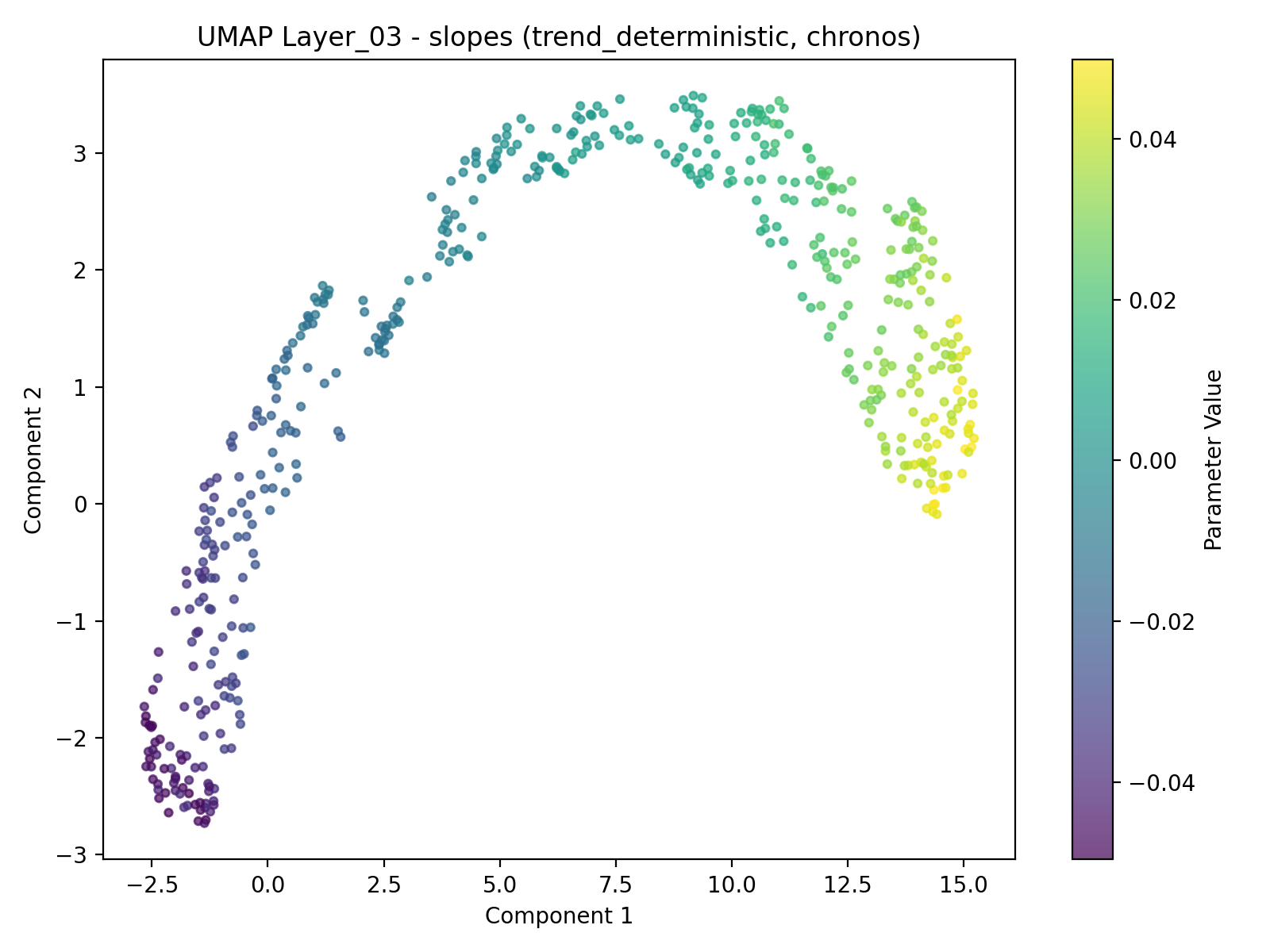}%
             {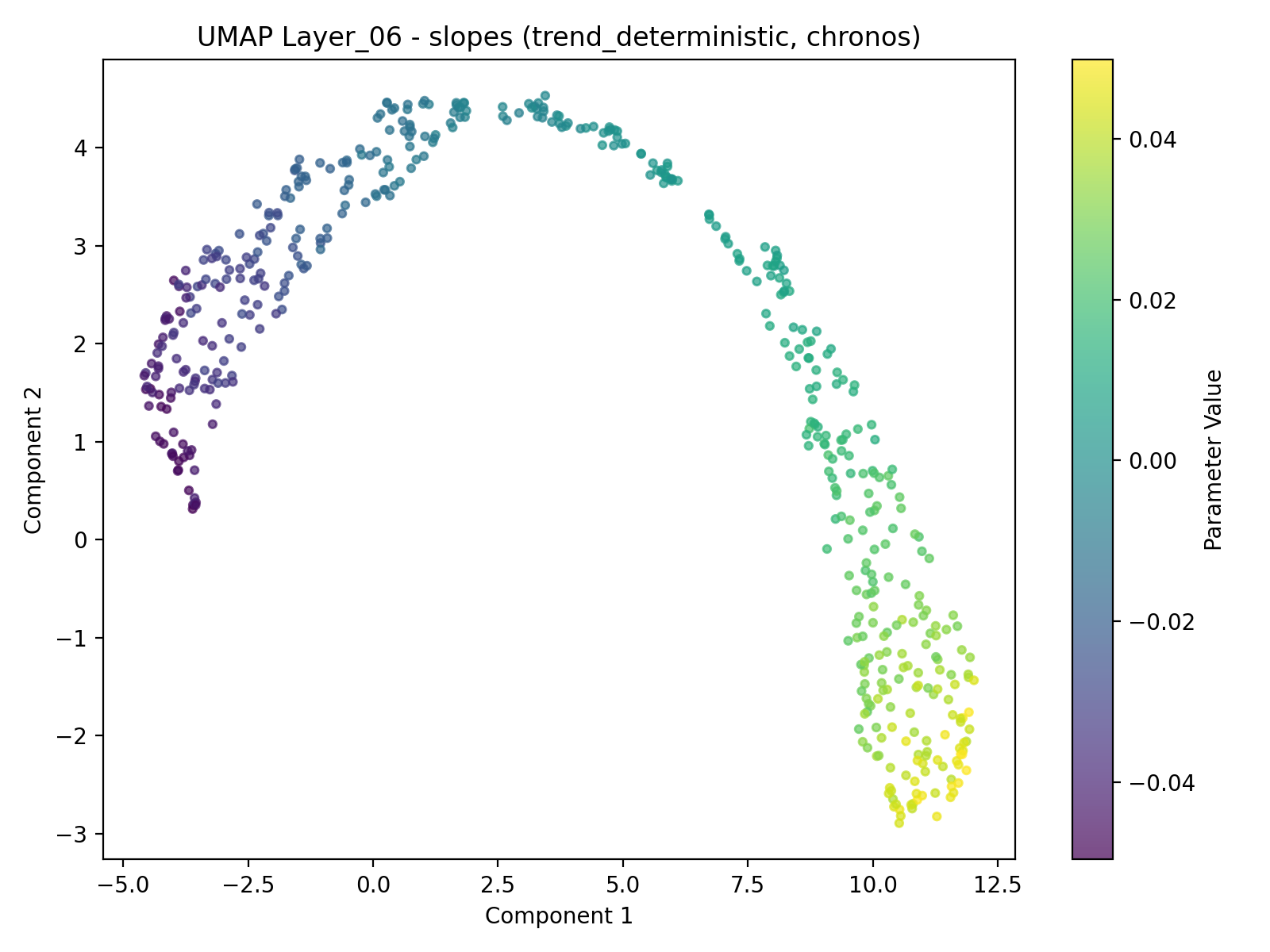}
  \caption{Trend --- Chronos --- UMAP (Layers 00/03/06)}
\end{figure}
\newpage
% ---------------- Variance Shift ----------------
\subsection{Variance Shift}

\paragraph{Chronos ($\tau$).}
\begin{figure}[t]
  \centering
  \threeplots{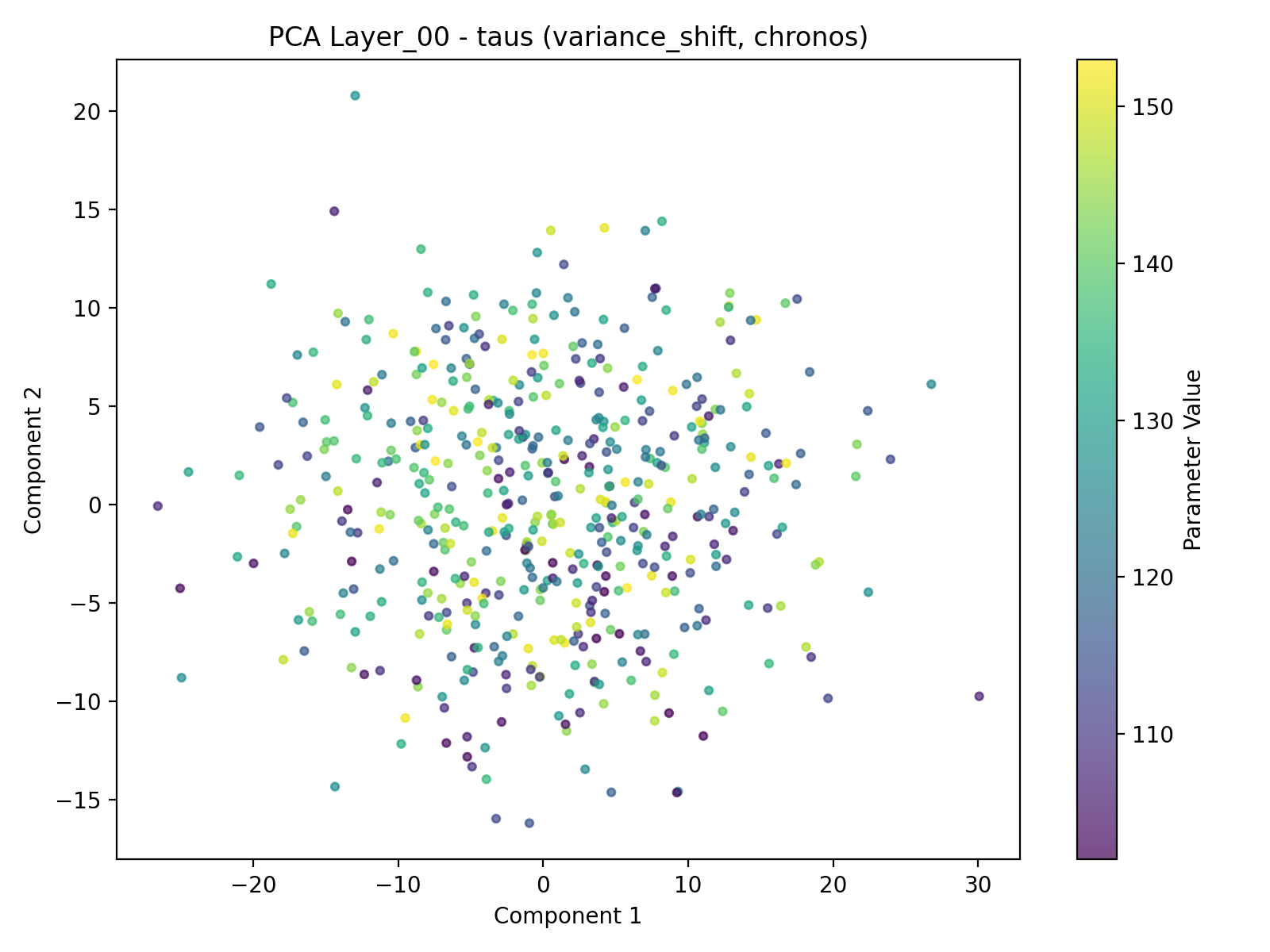}%
             {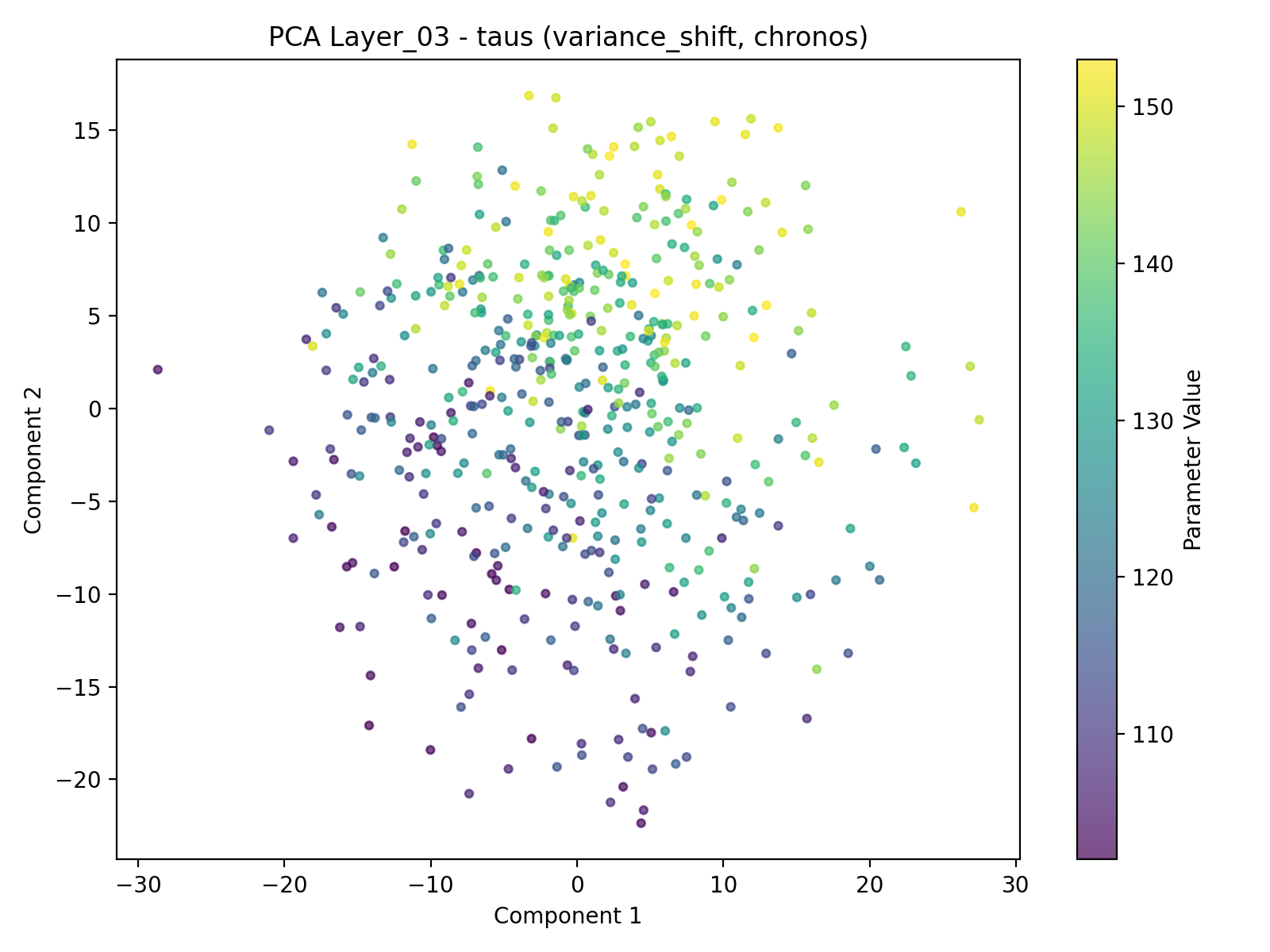}%
             {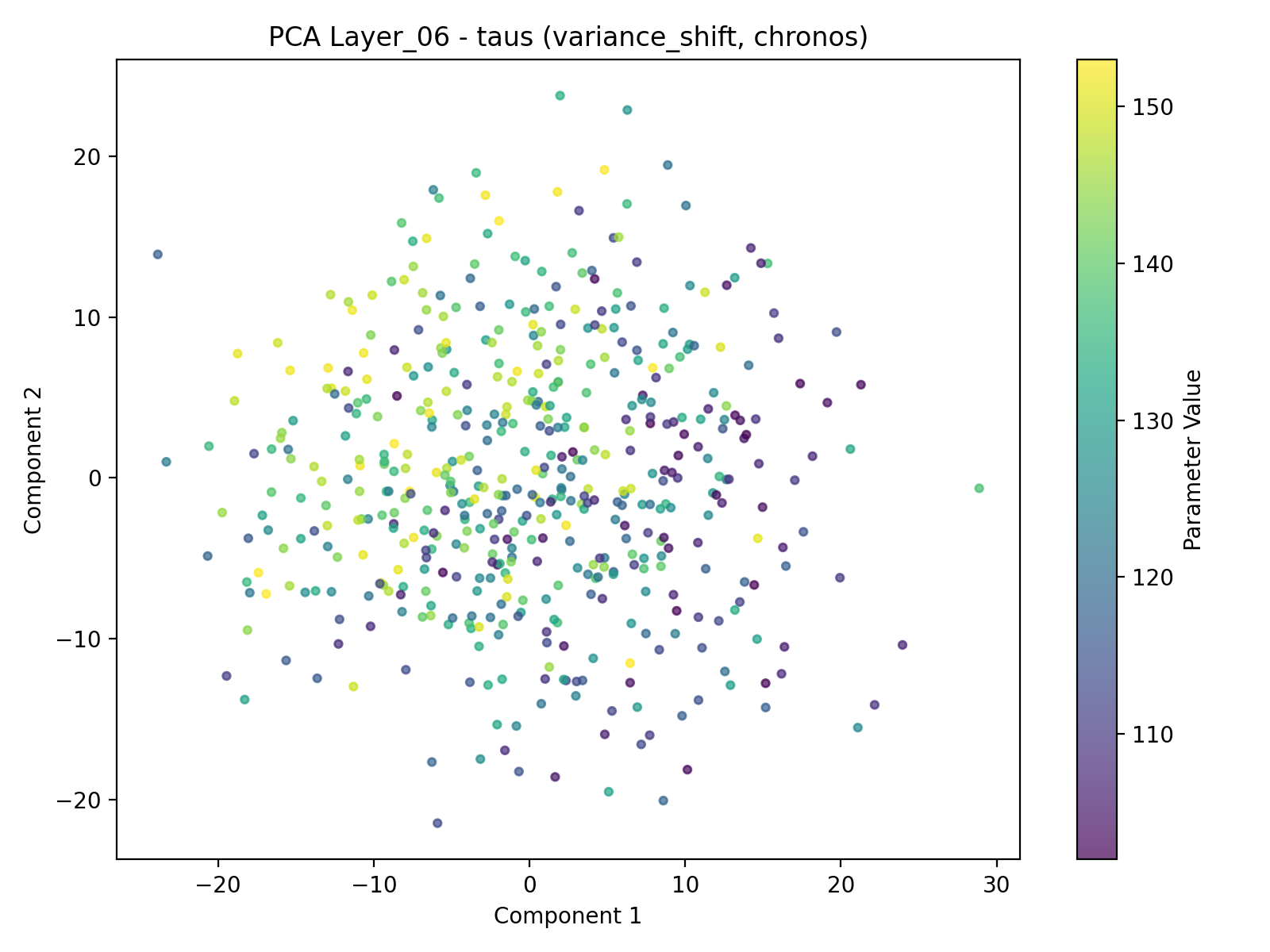}
  \caption{Variance Shift --- Chronos --- PCA (Layers 00/03/06)}
\end{figure}

\begin{figure}[t]
  \centering
  \threeplots{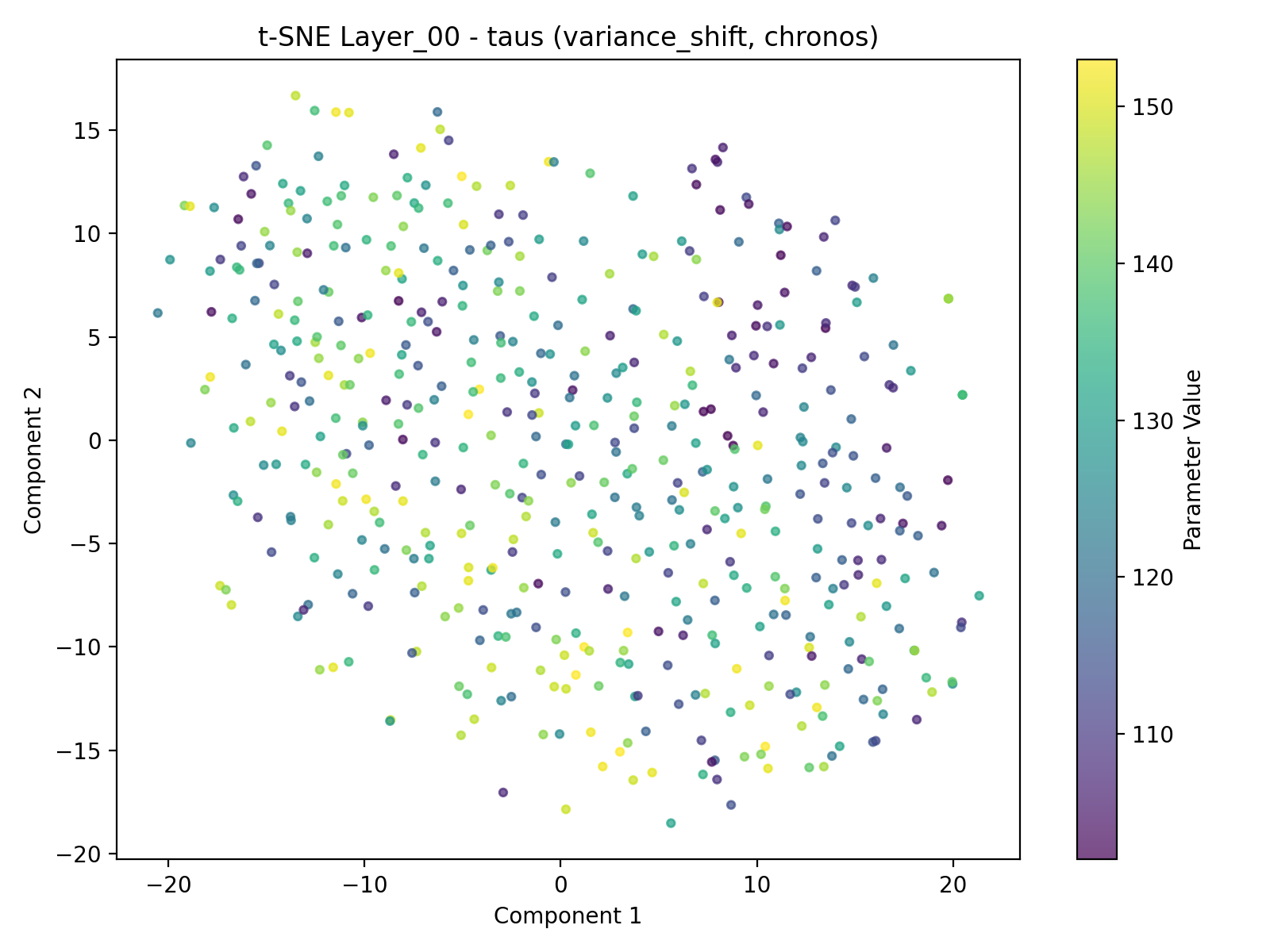}%
             {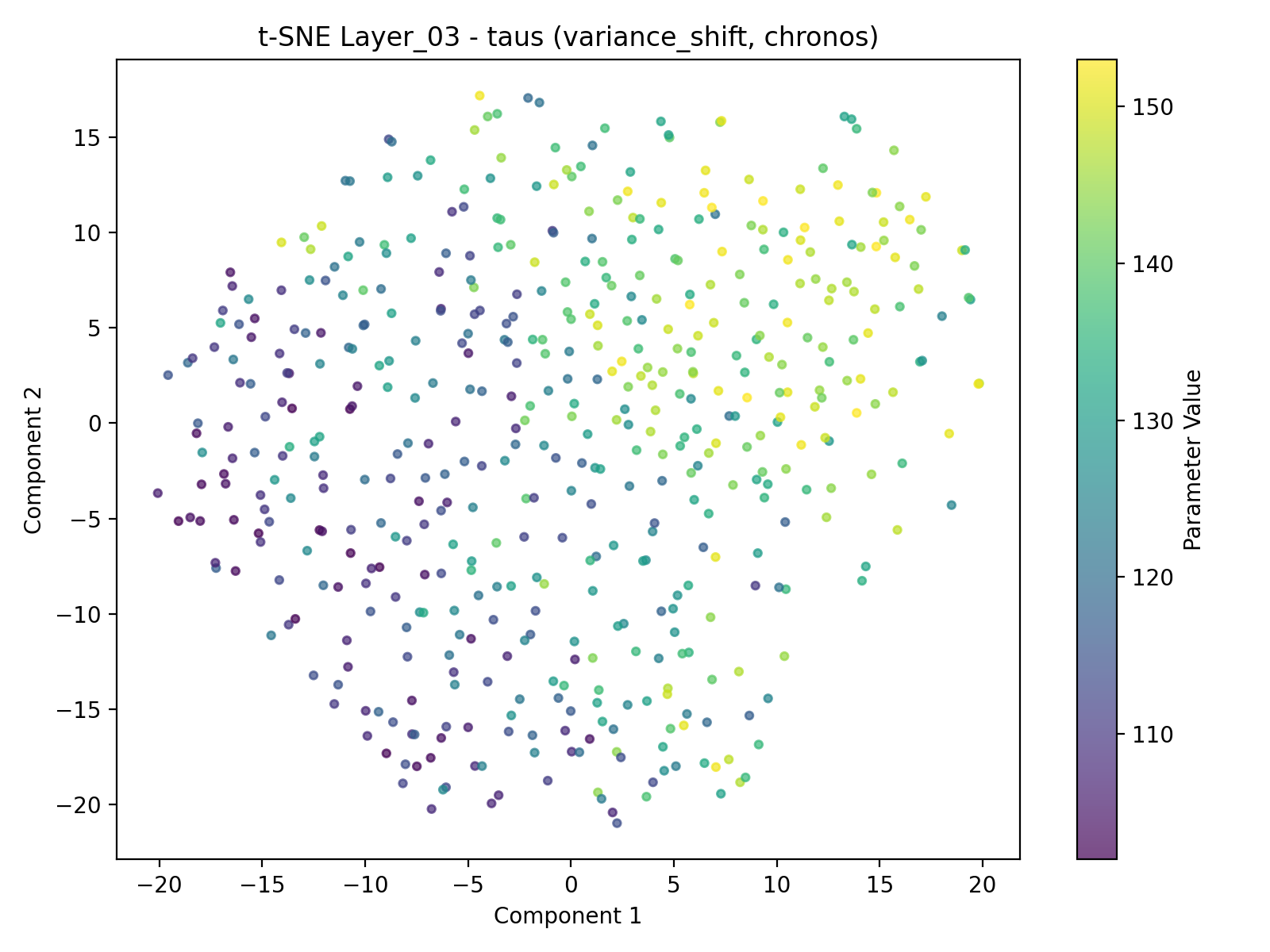}%
             {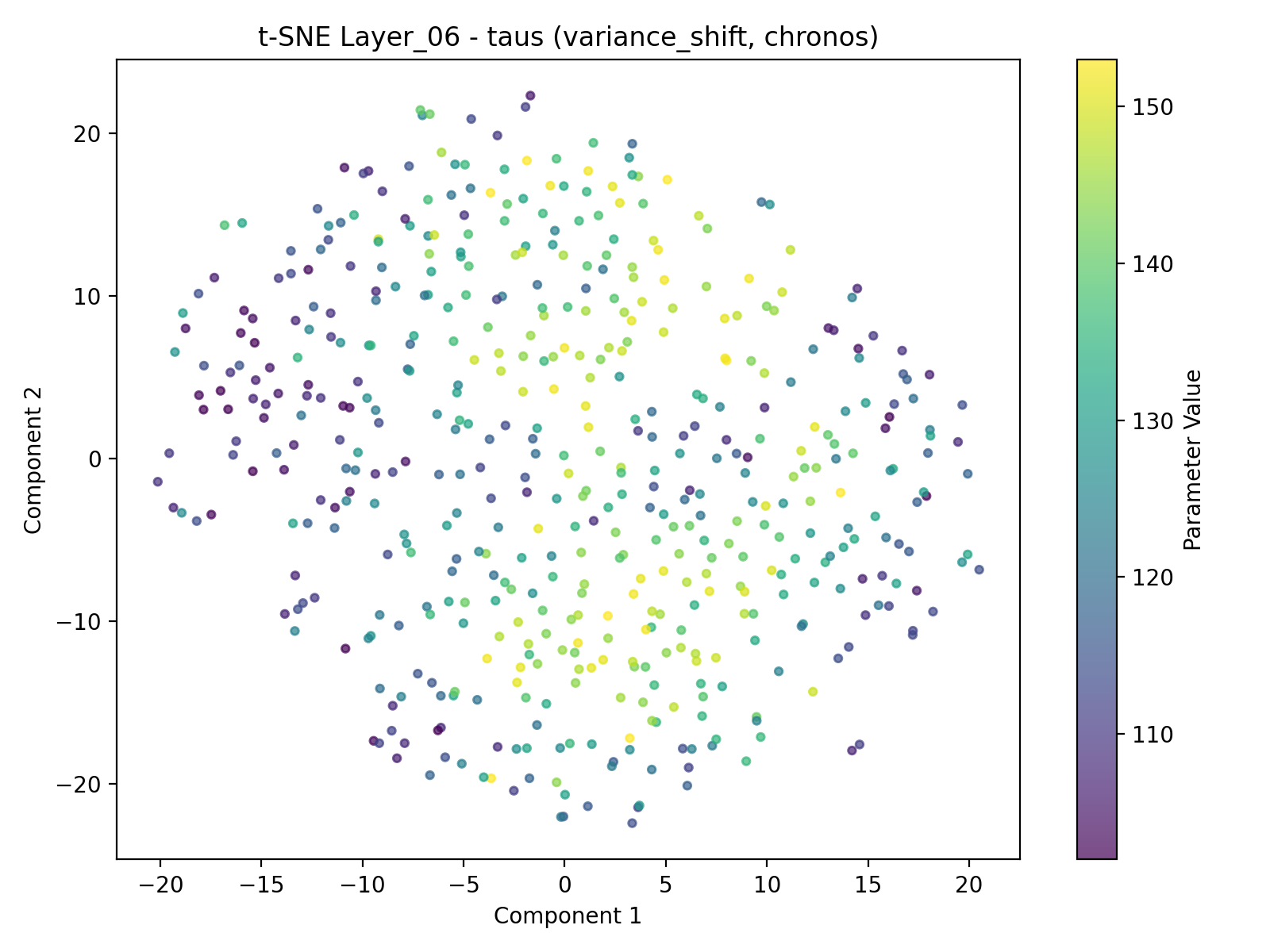}
  \caption{Variance Shift --- Chronos --- t-SNE (Layers 00/03/06)}
\end{figure}

\begin{figure}[t]
  \centering
  \threeplots{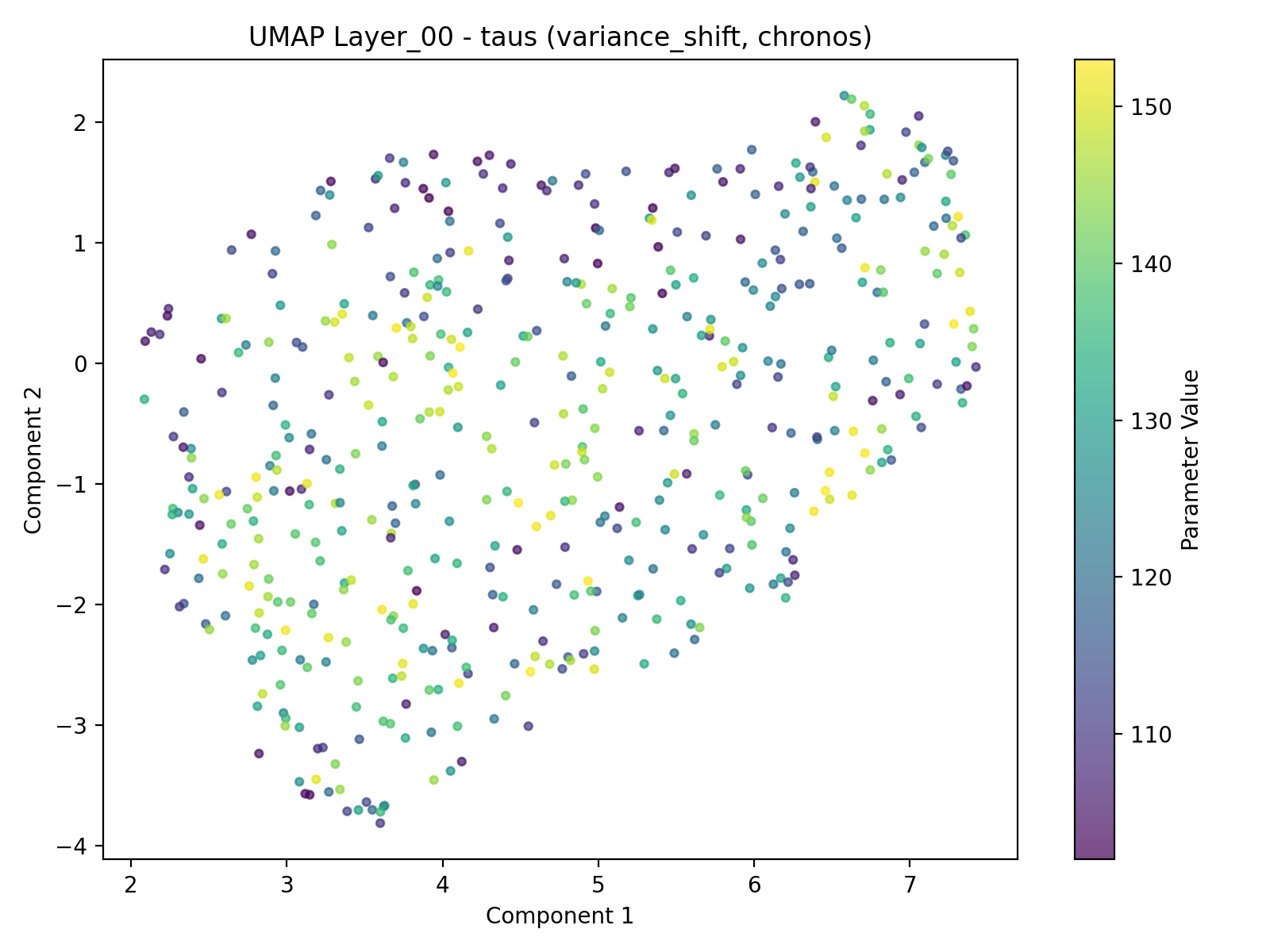}%
             {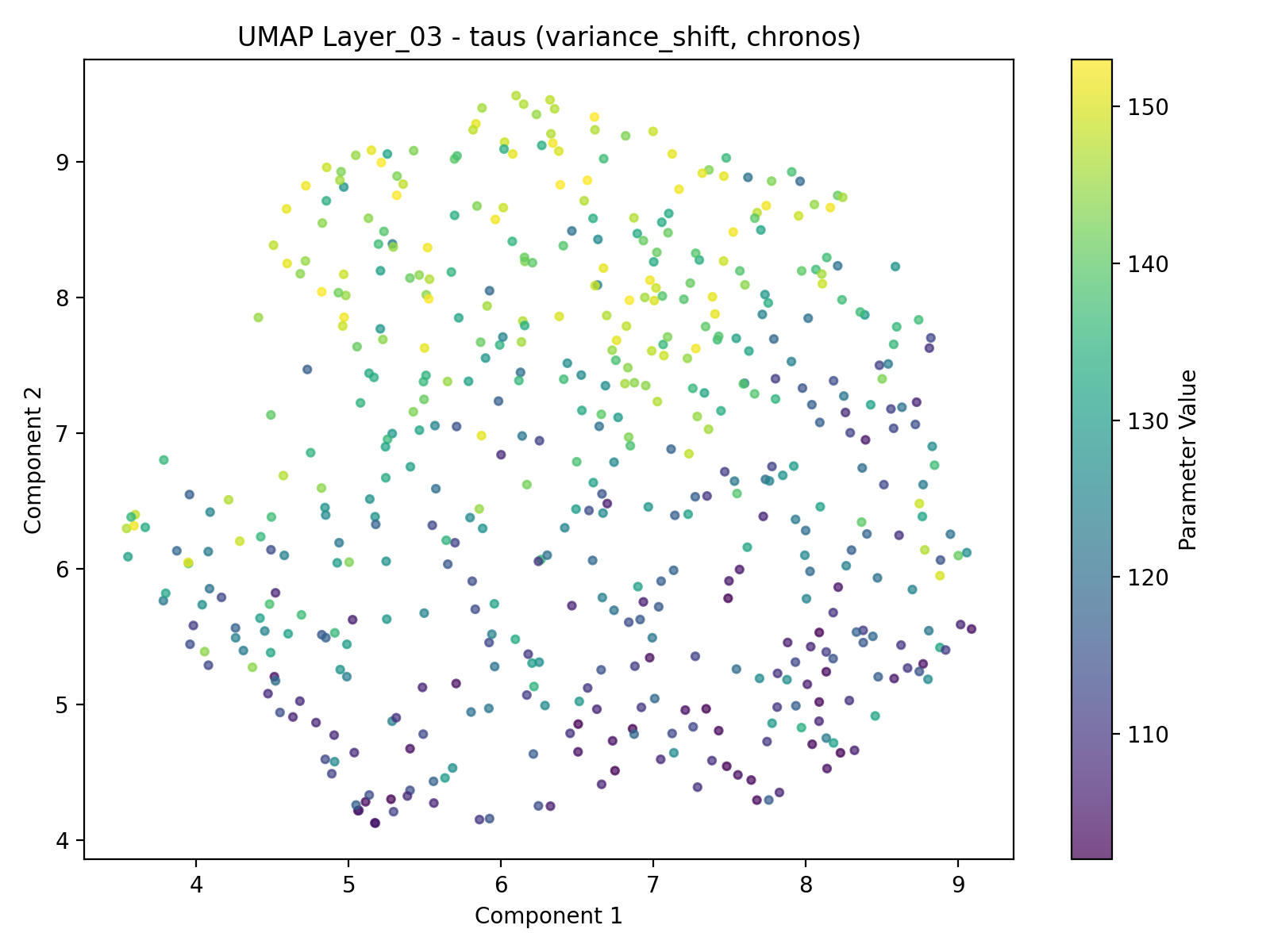}%
             {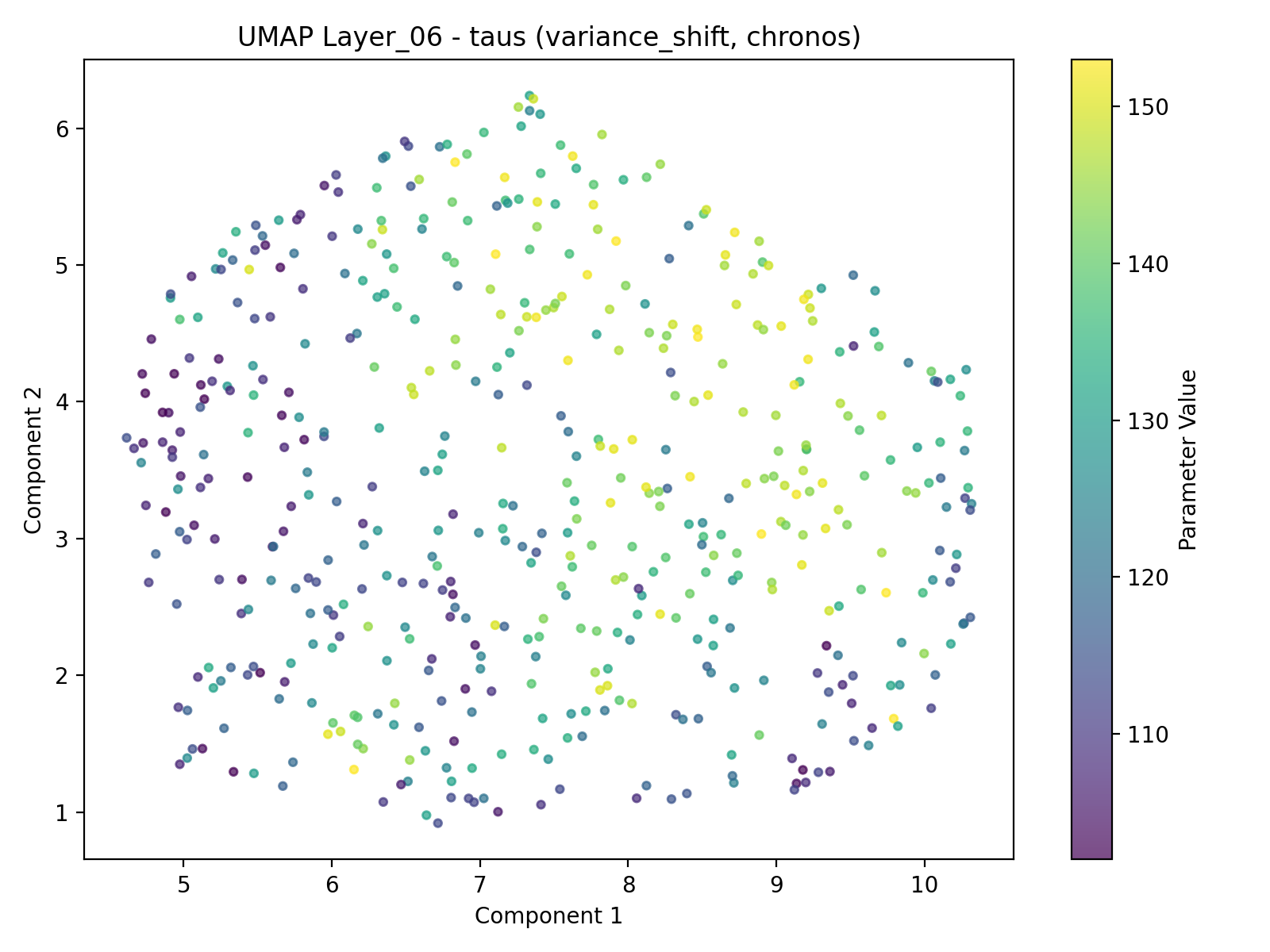}
  \caption{Variance Shift --- Chronos --- UMAP (Layers 00/03/06)}
\end{figure}
\newpage
\paragraph{Moment ($\tau$).}
\begin{figure}[t]
  \centering
  \threeplots{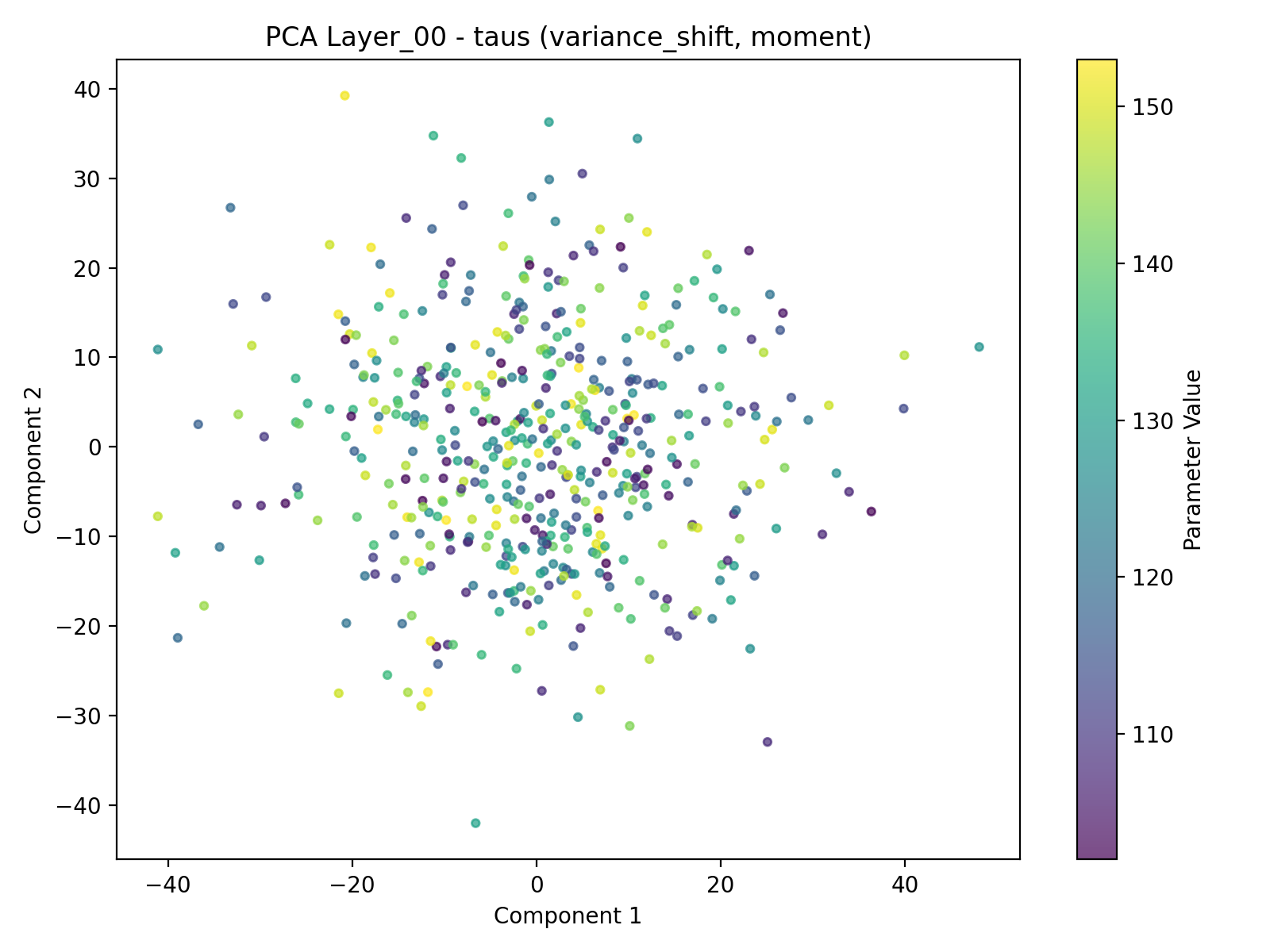}%
             {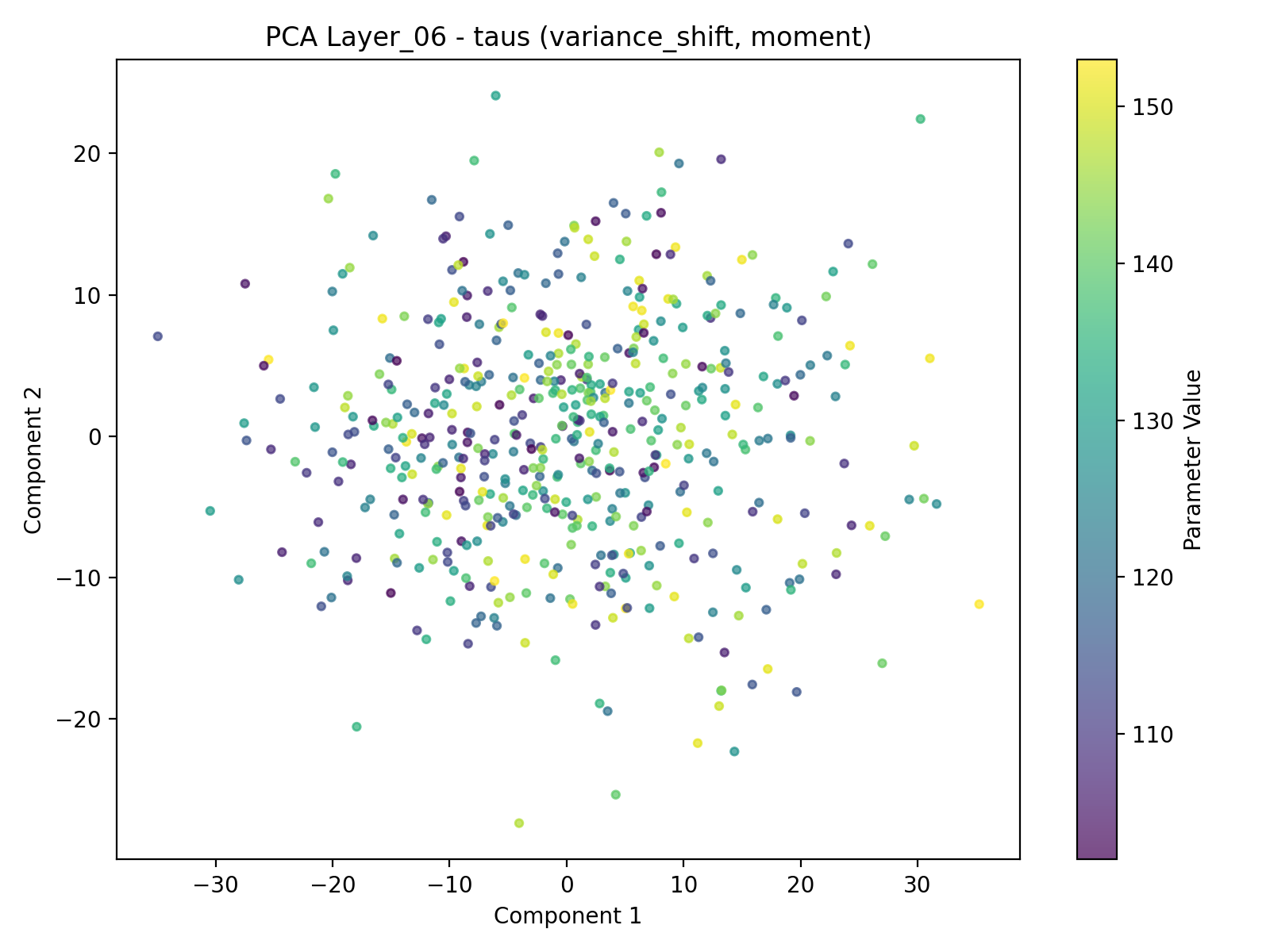}%
             {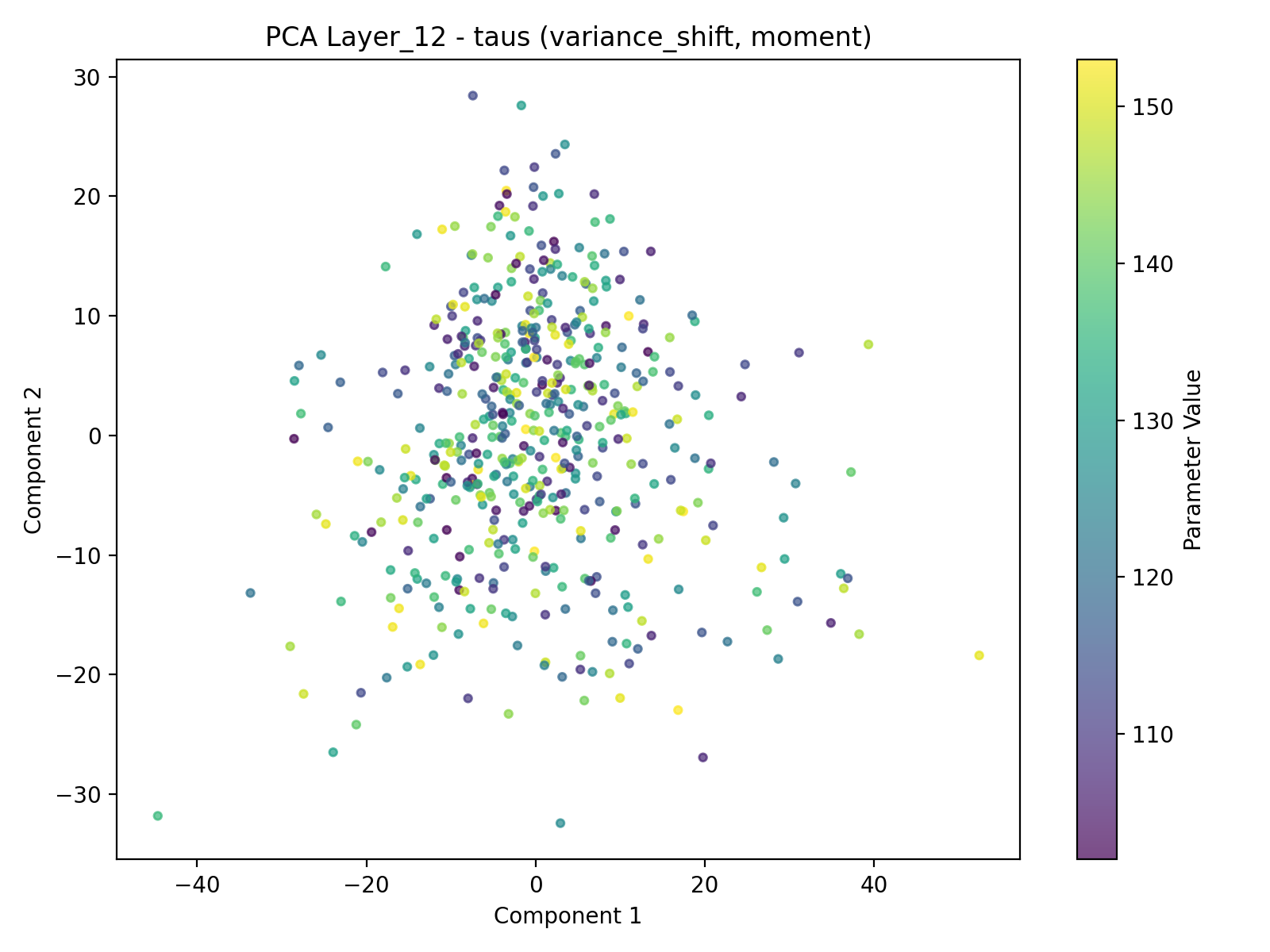}
  \caption{Variance Shift --- Moment --- PCA (Layers 00/06/12)}
\end{figure}

\begin{figure}[t]
  \centering
  \threeplots{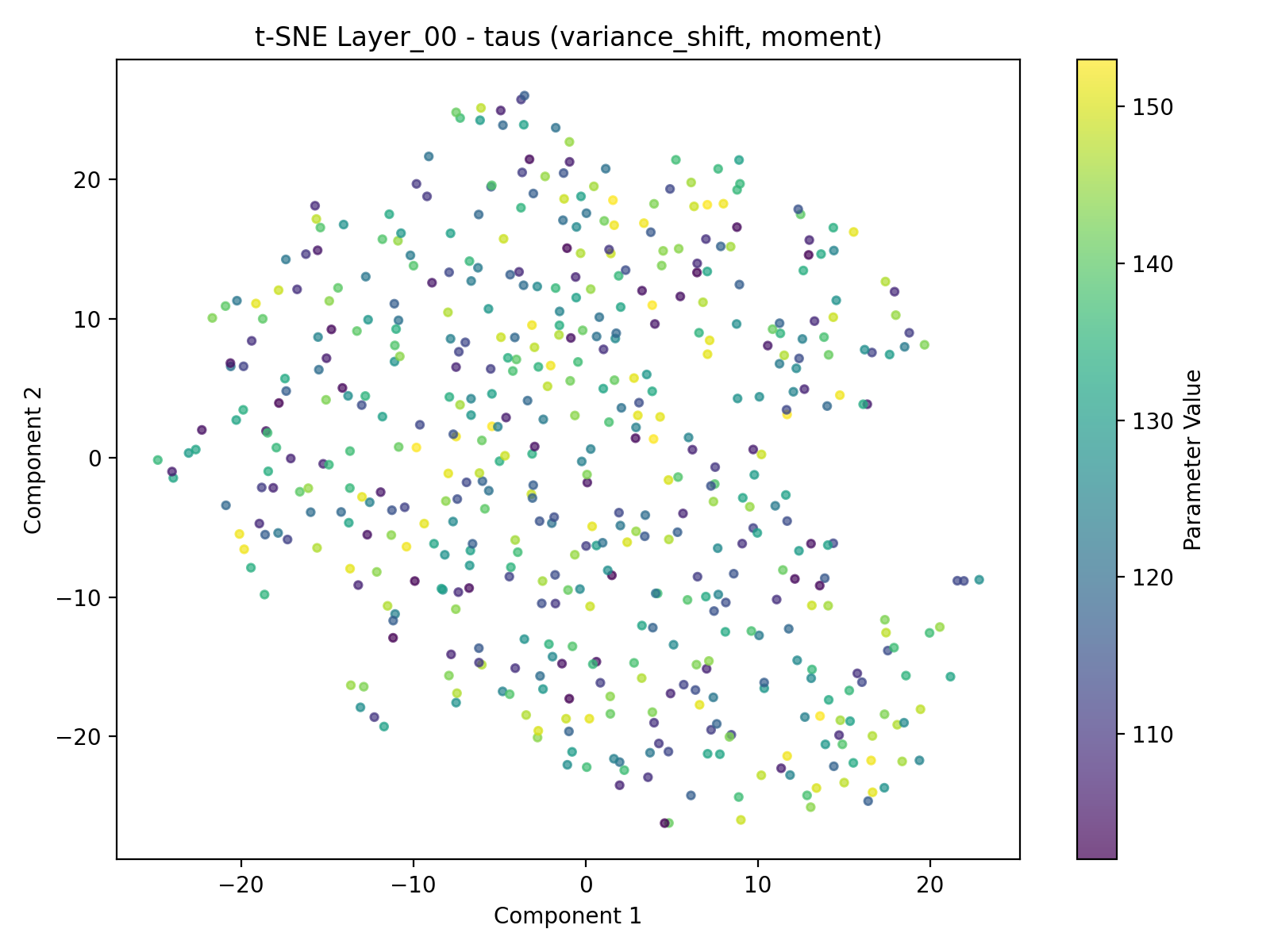}%
             {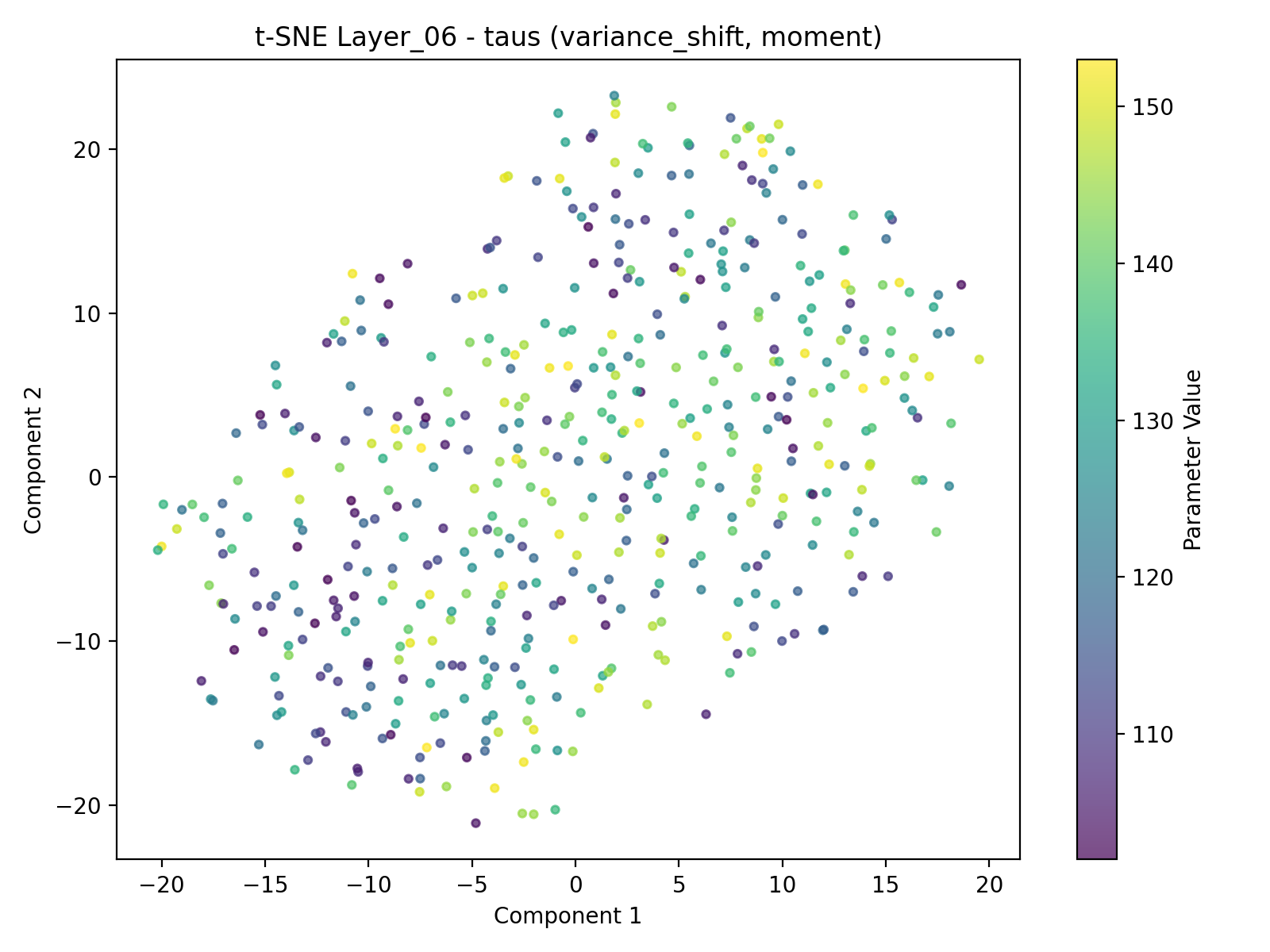}%
             {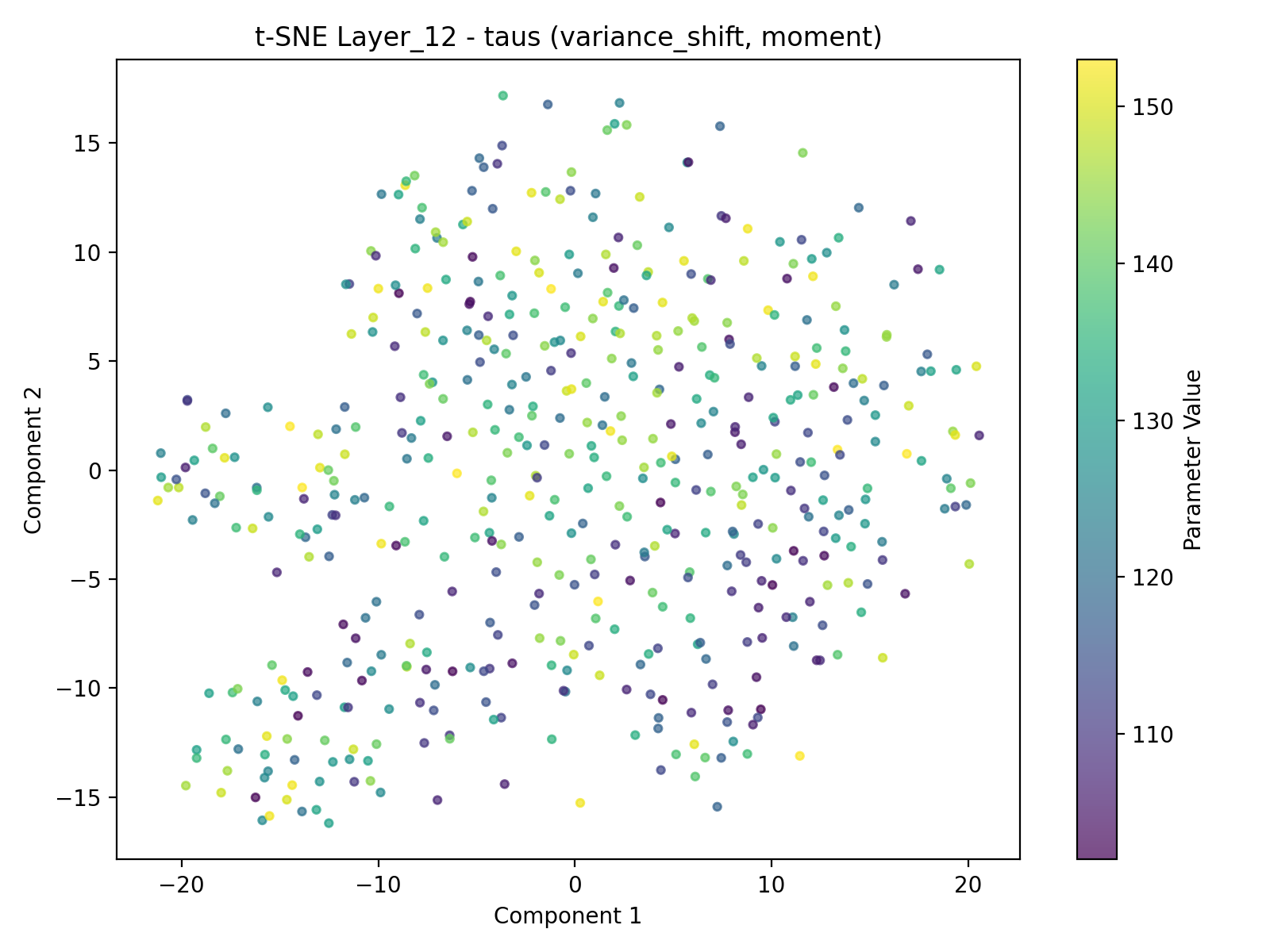}
  \caption{Variance Shift --- Moment --- t-SNE (Layers 00/06/12)}
\end{figure}

\begin{figure}[t]
  \centering
  \threeplots{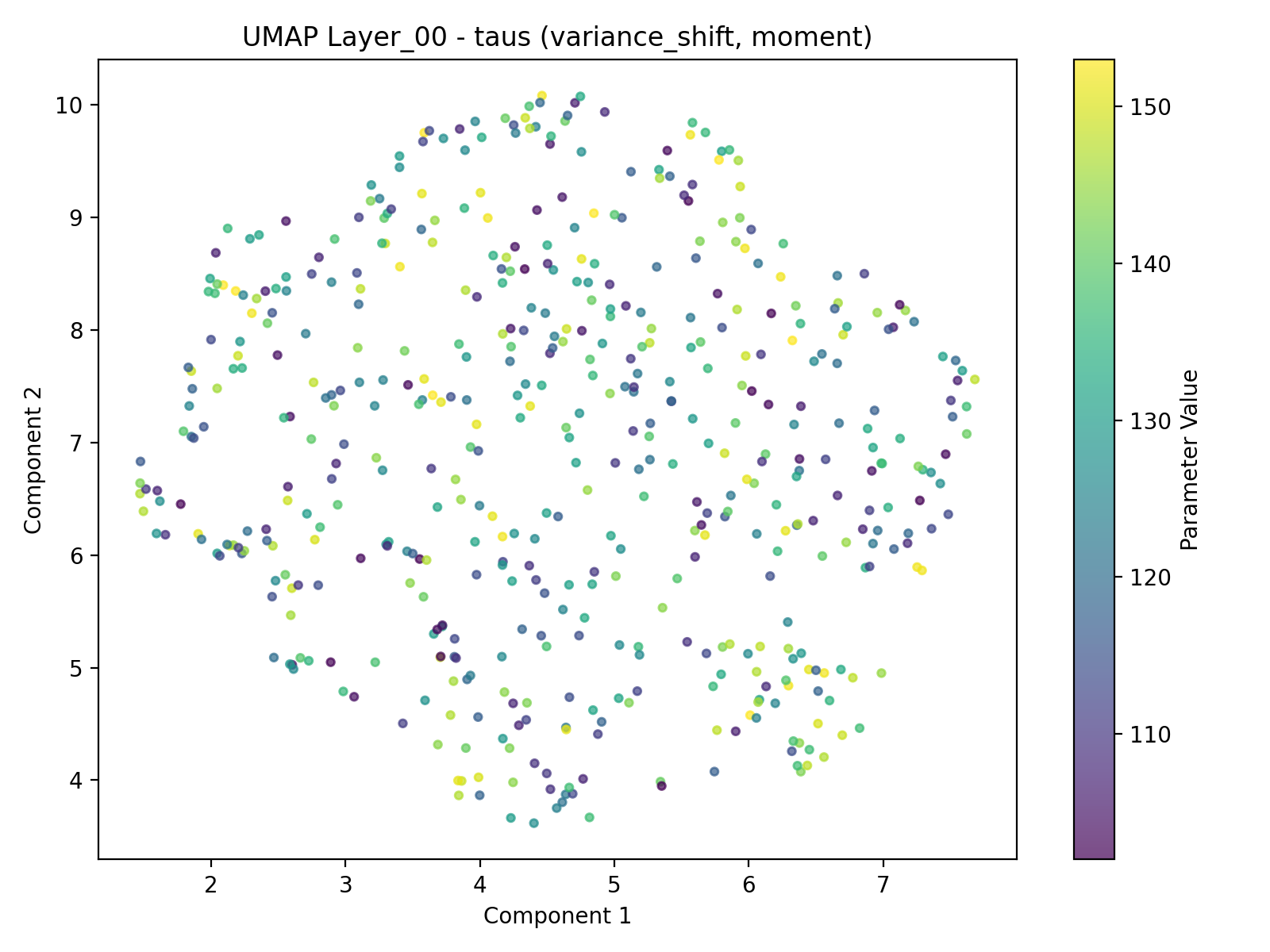}%
             {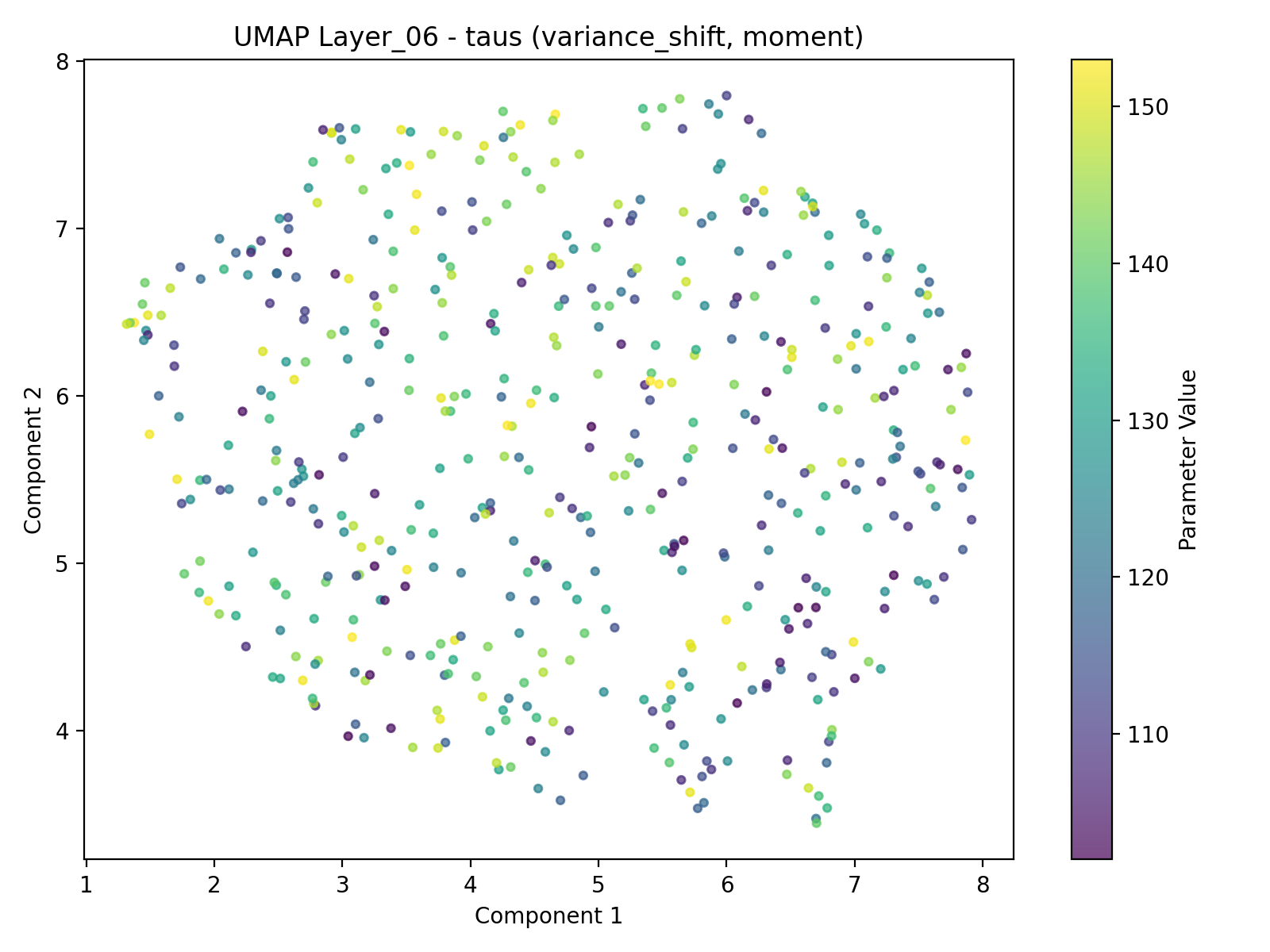}%
             {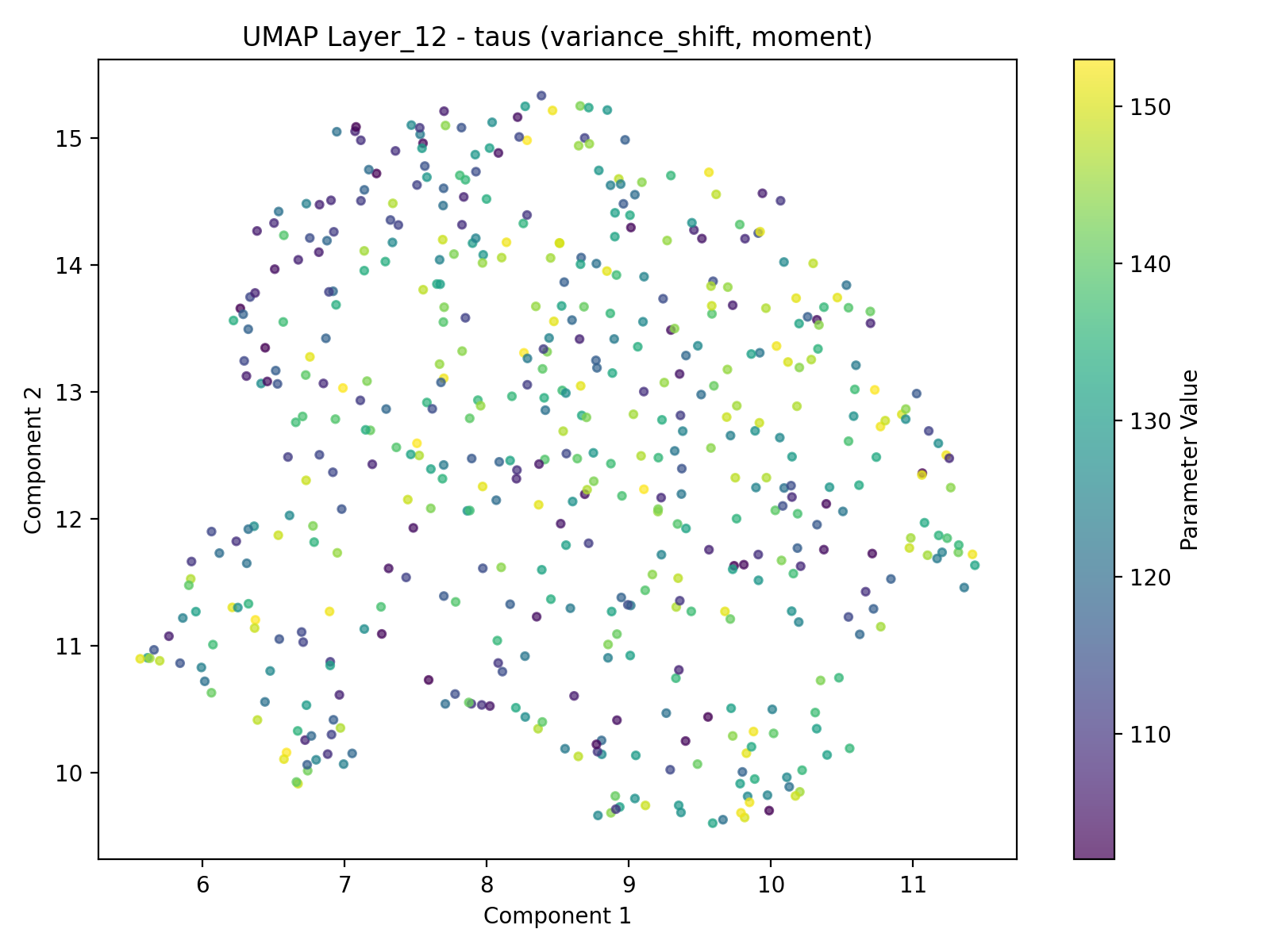}
  \caption{Variance Shift --- Moment --- UMAP (Layers 00/06/12)}
\end{figure}

\section{Compositionality results}
\label{app:compositeresults}
\begin{figure}

    % -------- Left Subfigure --------
    \begin{subfigure}[t]{0.48\textwidth}
        \centering
        \includegraphics[width=\linewidth]{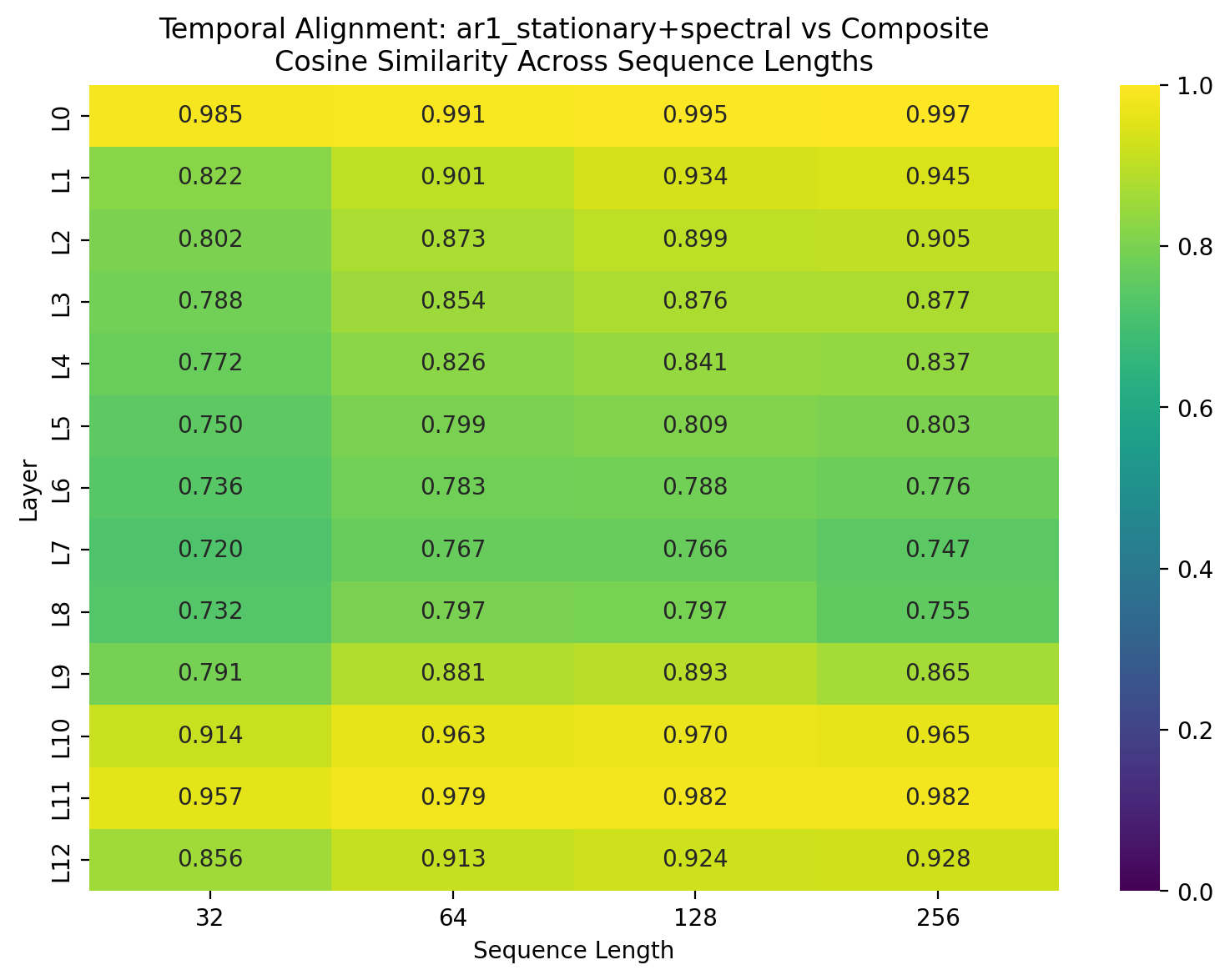} \\[4pt]
        \includegraphics[width=\linewidth]{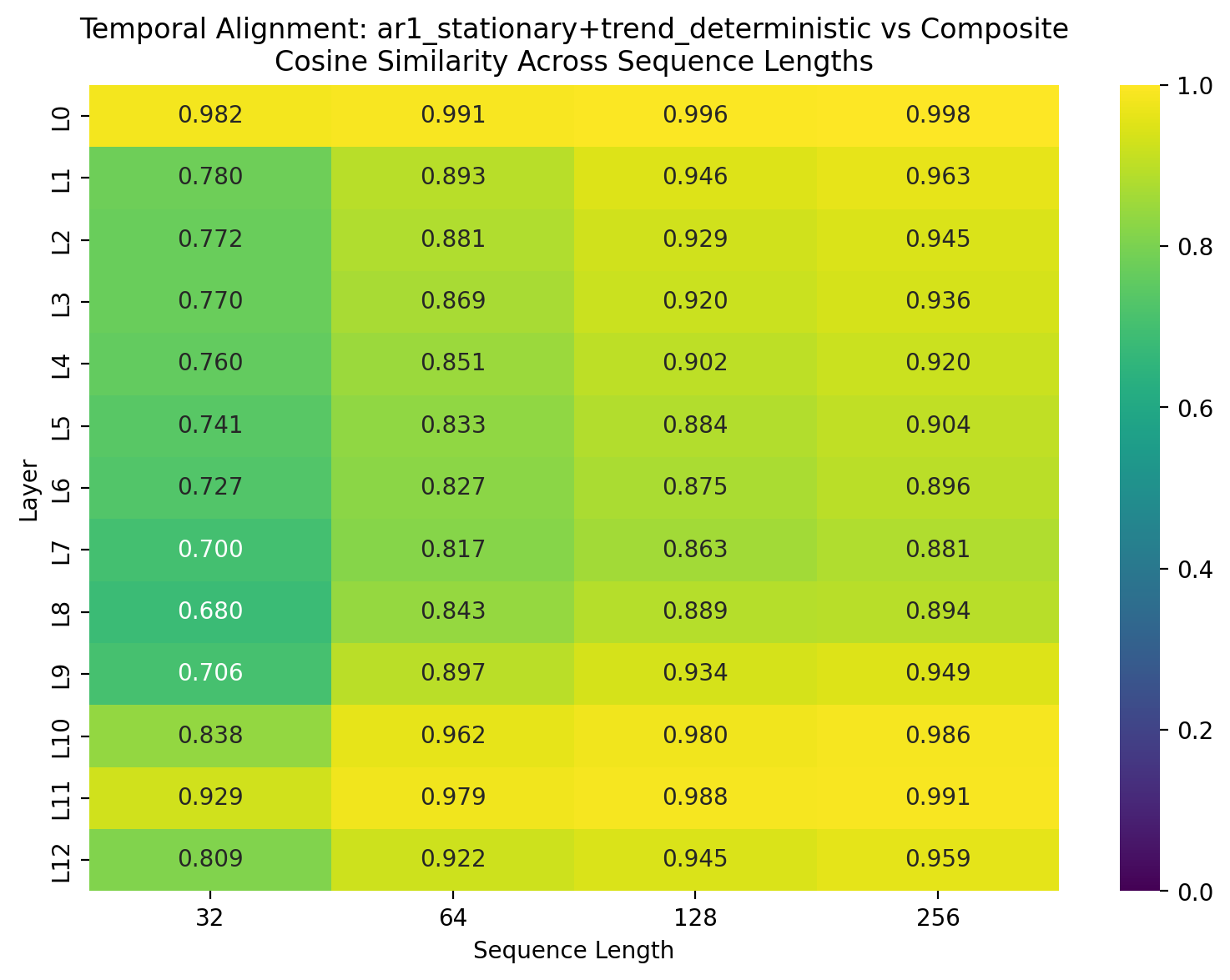} \\[4pt]
        \includegraphics[width=\linewidth]{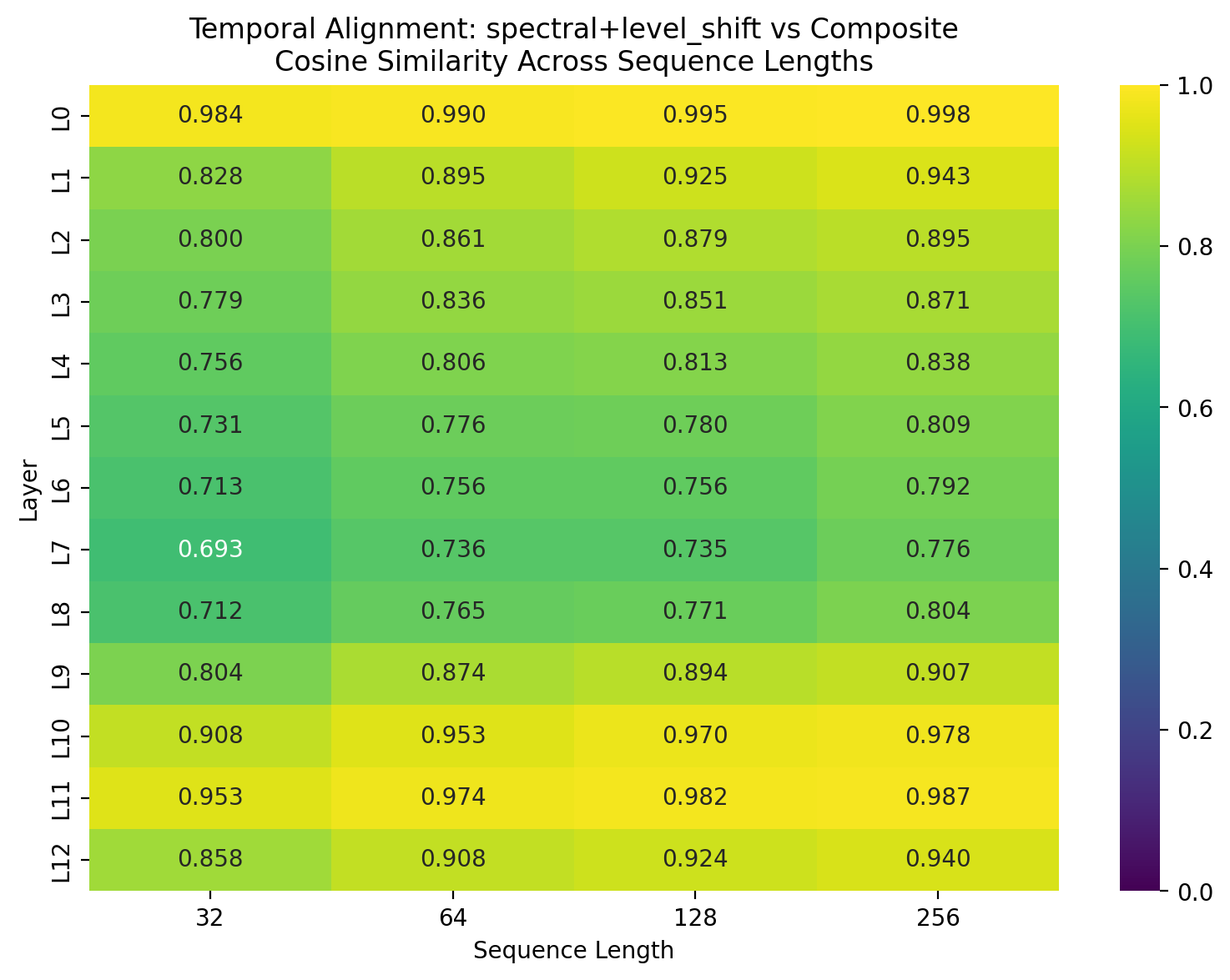}
        \caption{MOMENT - Temporal alignment experiments}
    \end{subfigure}
    \hfill
    % -------- Right Subfigure --------
    \begin{subfigure}[t]{0.48\textwidth}
        \centering
        \includegraphics[width=\linewidth]{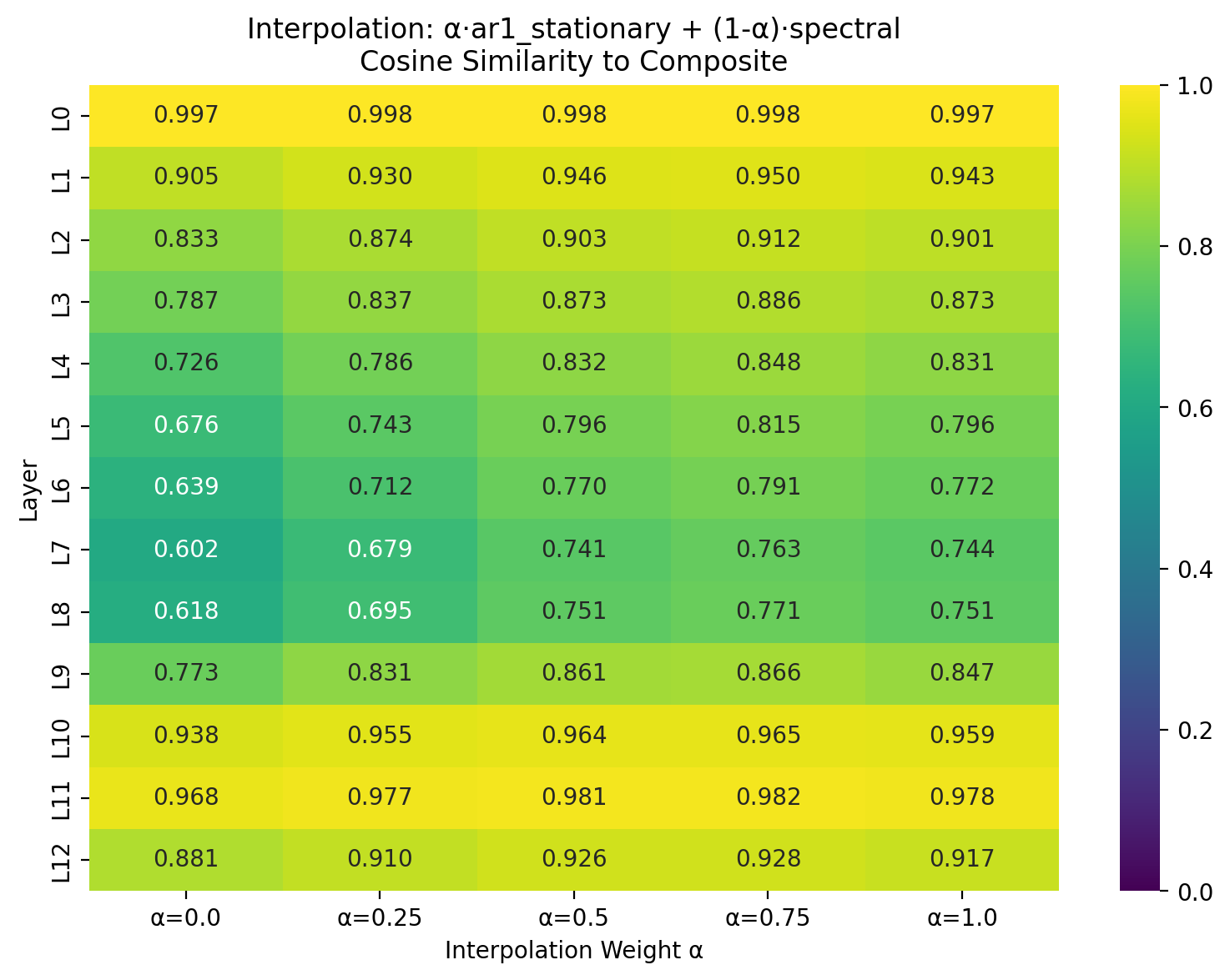} \\[4pt]
        \includegraphics[width=\linewidth]{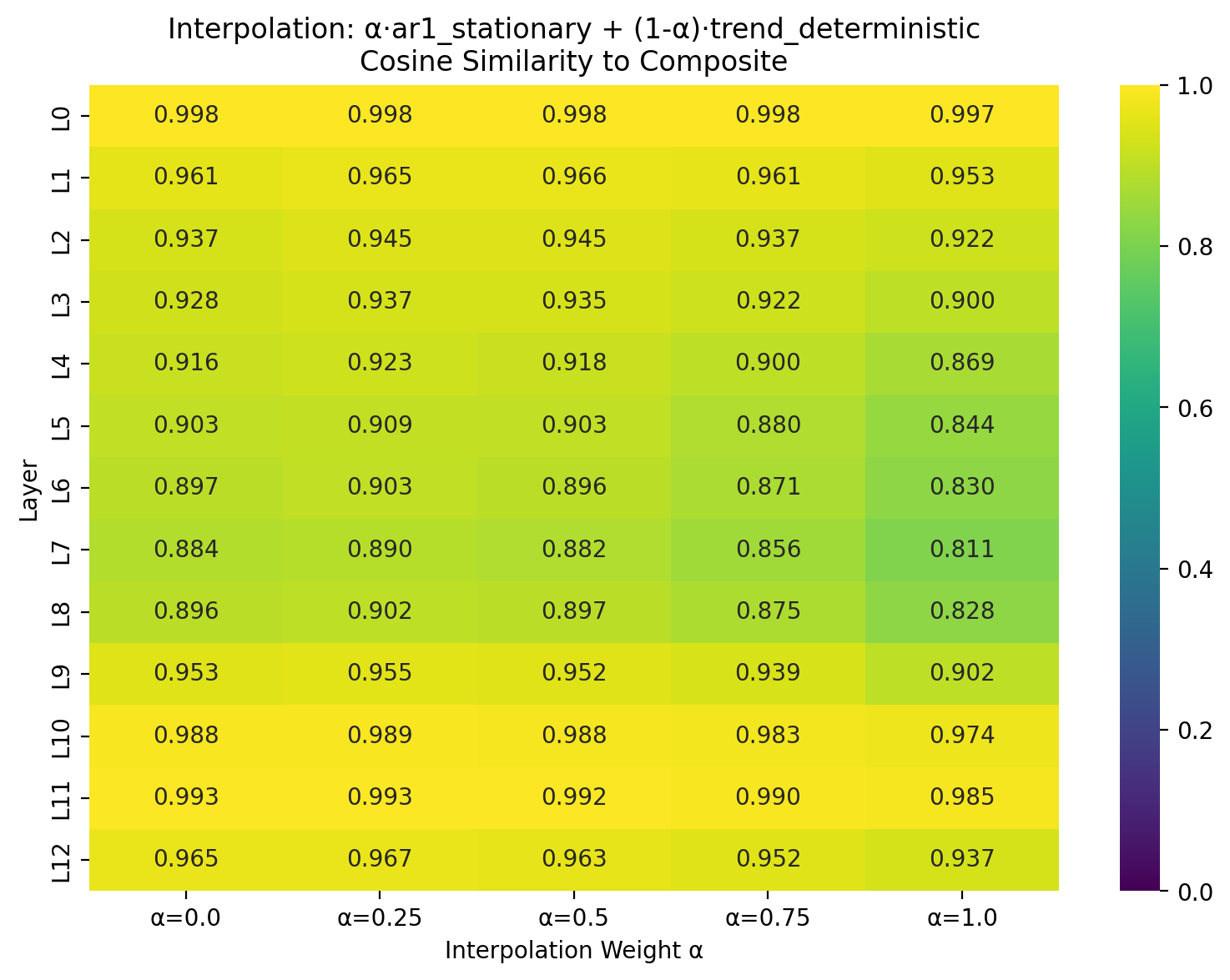} \\[4pt]
        \includegraphics[width=\linewidth]{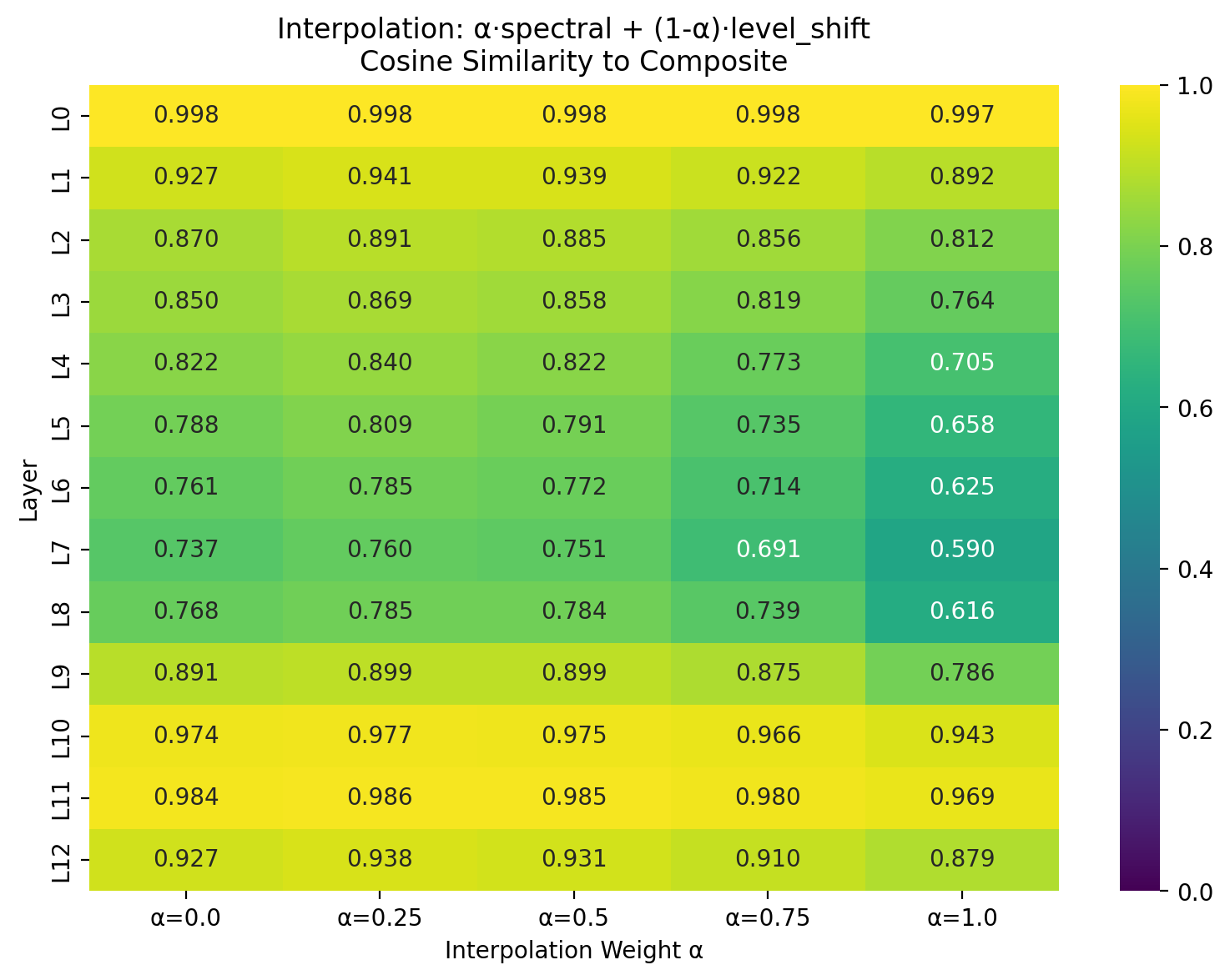}
        \caption{MOMENT - Interpolation analysis}
    \end{subfigure}
\end{figure}

\begin{figure}

    % -------- Left Subfigure --------
    \begin{subfigure}[t]{0.48\textwidth}
        \centering
        \includegraphics[width=\linewidth]{figures/chronos_temporal_ar1_stationary_spectral_mean.png} \\[4pt]
        \includegraphics[width=\linewidth]{figures/chronos_temporal_ar1_stationary_trend_deterministic_mean.png} \\[4pt]
        \includegraphics[width=\linewidth]{figures/chronos_temporal_spectral_level_shift_mean.png}
        \caption{Chronos - Temporal alignment experiments}
    \end{subfigure}
    \hfill
    % -------- Right Subfigure --------
    \begin{subfigure}[t]{0.48\textwidth}
        \centering
        \includegraphics[width=\linewidth]{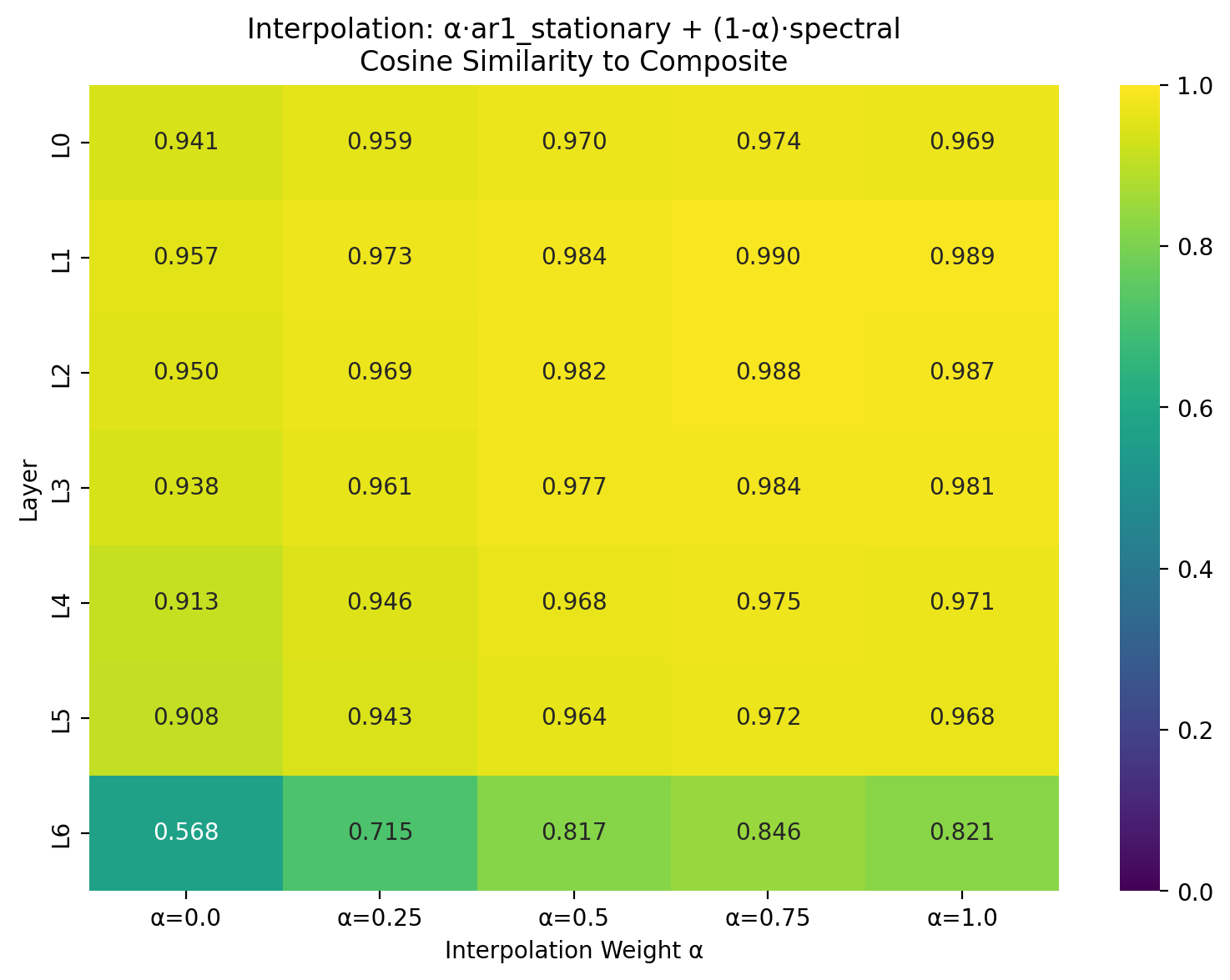} \\[4pt]
        \includegraphics[width=\linewidth]{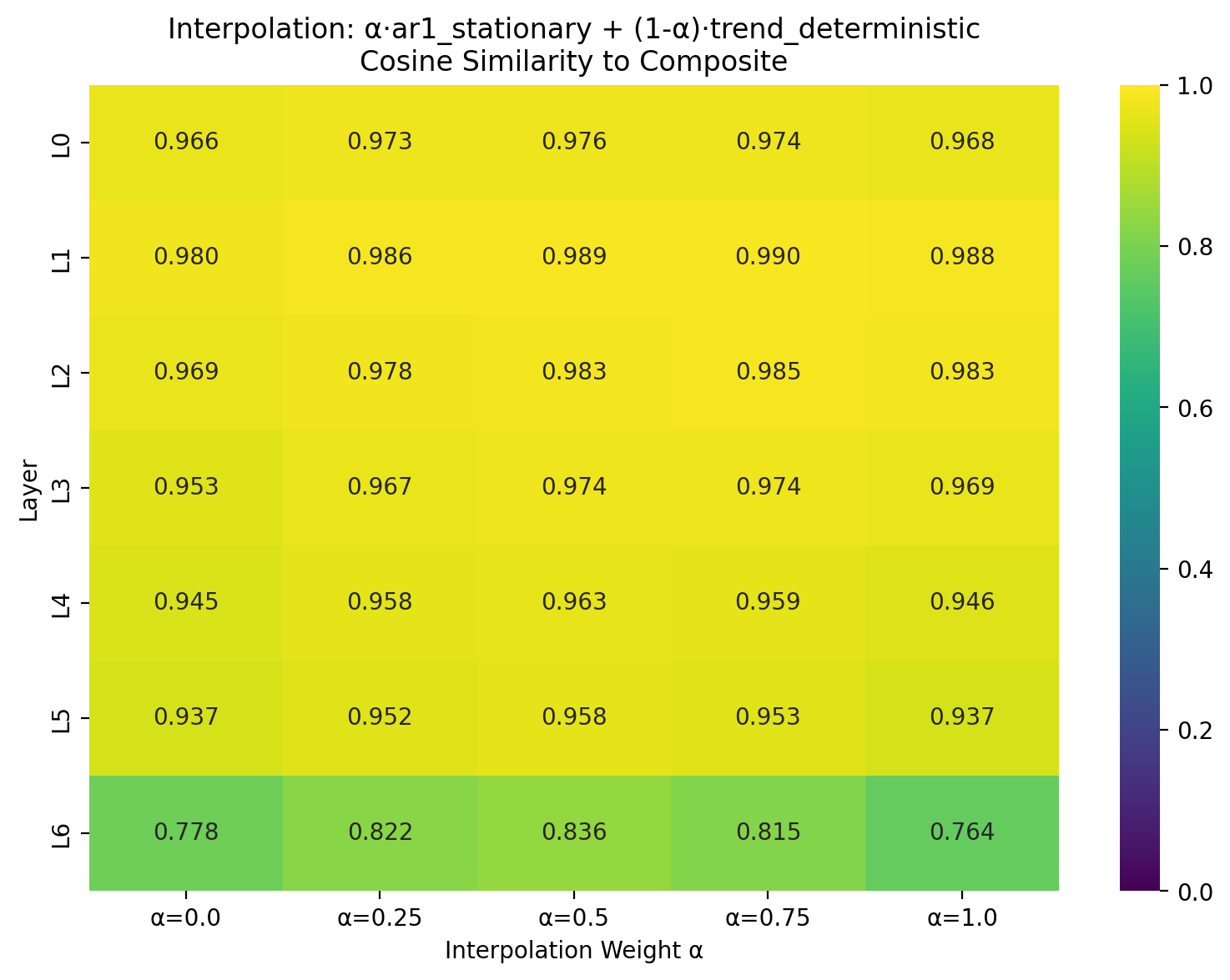} \\[4pt]
        \includegraphics[width=\linewidth]{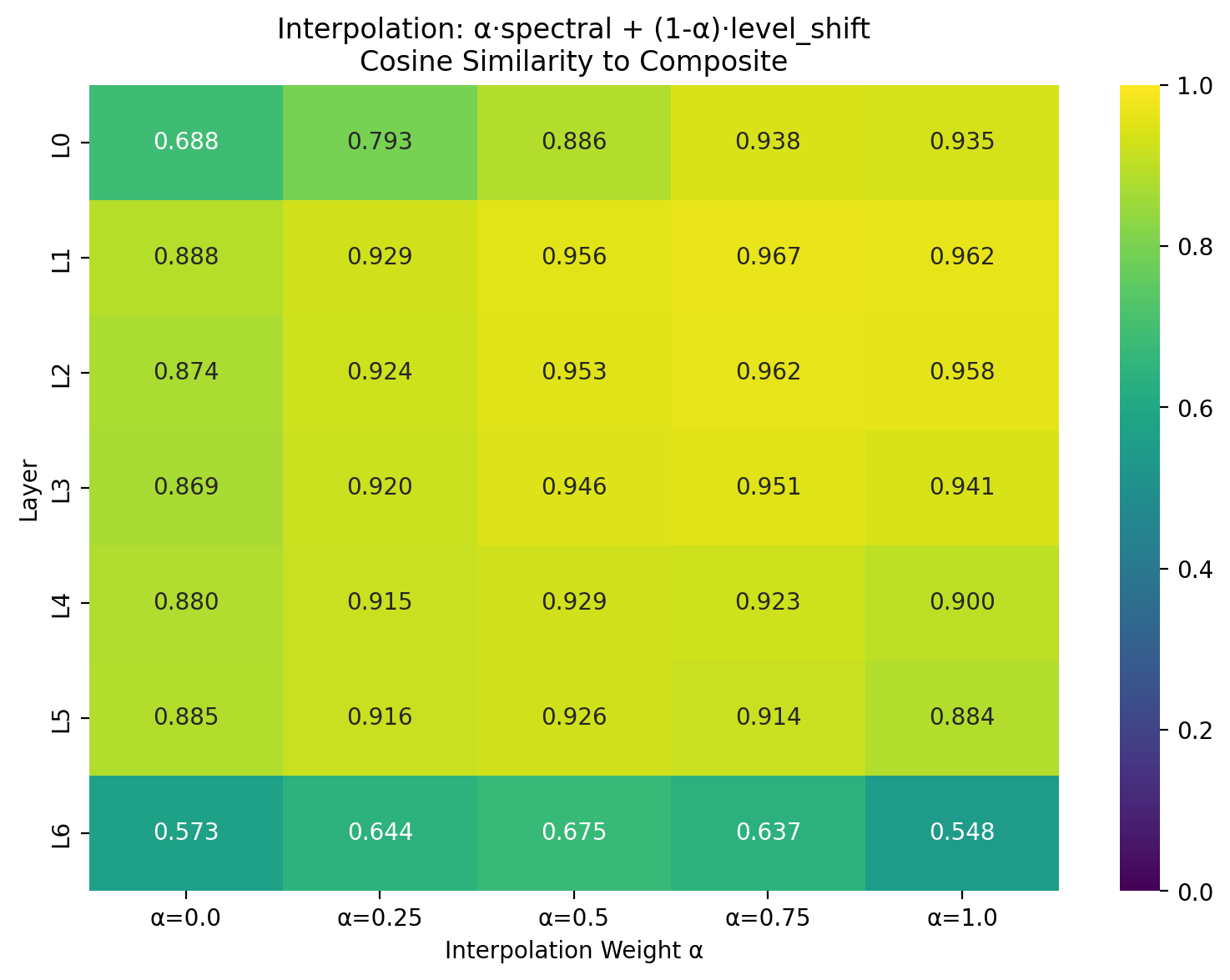}
        \caption{Chronos - Interpolation analysis}
    \end{subfigure}

\end{figure}

\begin{figure*}[htbp]
    \centering
    \begin{subfigure}{0.48\textwidth}
        \includegraphics[width=\linewidth]{figures/vector_arithmetic_chronos.png}
        \caption{Vector Arithmetic Experiments with Chronos}
        \label{fig:moment_context_length_ablation_heatmap_time_warped}
    \end{subfigure}
    \begin{subfigure}{0.48\textwidth}
        \includegraphics[width=\linewidth]{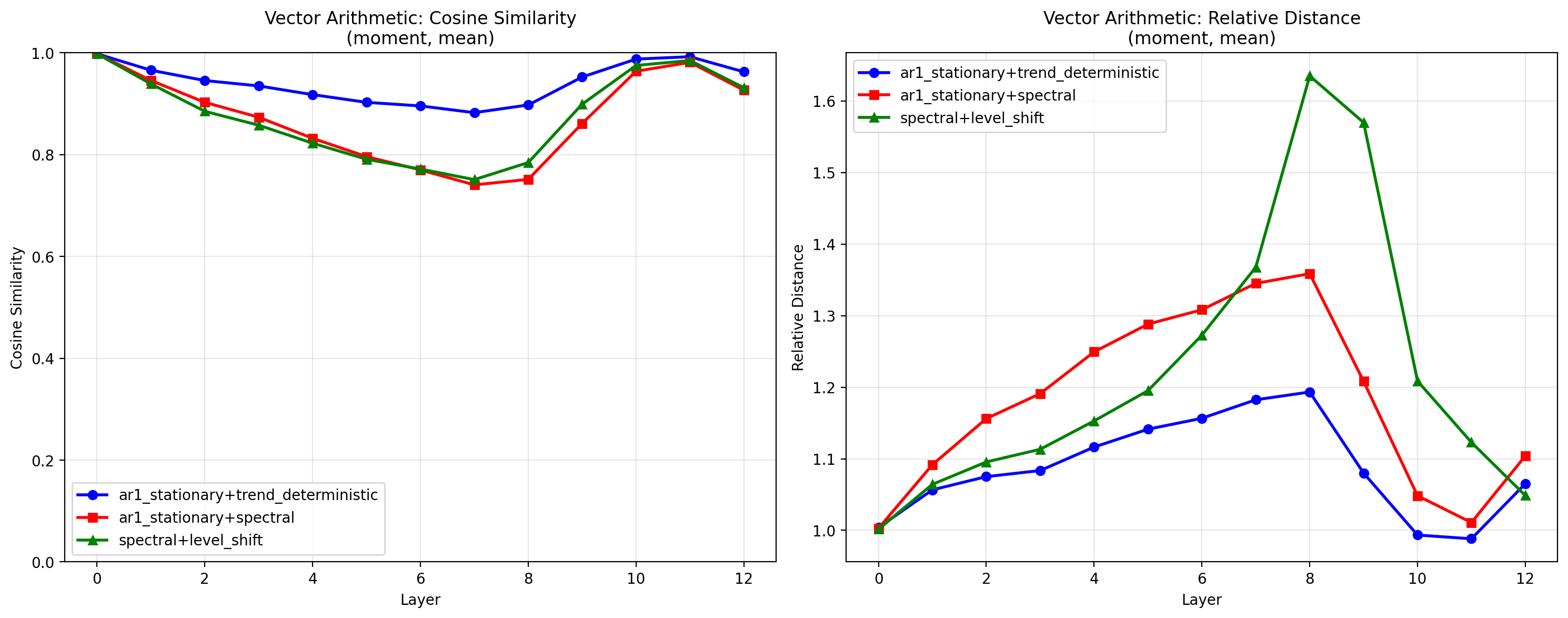}
        \caption{Vector Arithmetic Experiments with MOMENT}
        \label{fig:moment_variance_shift_context_length_ablation_heatmap}
    \end{subfigure}
    \caption{Vector Arithmetic Experiments}
\end{figure*}

\end{document}